\documentclass[lettersize,journal]{IEEEtran}
\usepackage{amsmath,amsfonts}
\usepackage{algorithmic}
\usepackage{algorithm}
\usepackage{array}
\usepackage[caption=false,font={footnotesize,rm},labelfont=rm,textfont=rm]{subfig}
\usepackage{textcomp}
\usepackage{stfloats}
\usepackage{url}
\usepackage{verbatim}
\usepackage{graphicx}
\usepackage{cite}
\usepackage{booktabs}
\usepackage{multirow}
\usepackage{cleveref}
\usepackage{makecell}
\usepackage{float} 
\usepackage{caption}

\usepackage{xcolor}
\usepackage{colortbl} 
\usepackage{hhline}  

\definecolor{l_void}{HTML}{000000}
\definecolor{l_road}{HTML}{7f417e}
\definecolor{l_sidewalk}{HTML}{f422ea}
\definecolor{l_building}{HTML}{464646}
\definecolor{l_wall}{HTML}{67669e}
\definecolor{l_fence}{HTML}{bf9a98}
\definecolor{l_pole}{HTML}{999999}
\definecolor{l_light}{HTML}{f9a920}
\definecolor{l_sign}{HTML}{dbdc03}
\definecolor{l_veg}{HTML}{6c8d25}
\definecolor{l_terrain}{HTML}{9ff89f}
\definecolor{l_sky}{HTML}{4582b6}
\definecolor{l_people}{HTML}{de133d}
\definecolor{l_rider}{HTML}{fd0102}
\definecolor{l_car}{HTML}{01008e}
\definecolor{l_truck}{HTML}{010048}
\definecolor{l_bus}{HTML}{013b66}
\definecolor{l_train}{HTML}{005164}
\definecolor{l_moto}{HTML}{0101e5}
\definecolor{l_bike}{HTML}{770c1f}

\newcommand{\legfont}{\fontsize{7pt}{8pt}\selectfont\bfseries\color{white}}
\newcommand{\legfontblk}{\fontsize{7pt}{8pt}\selectfont\bfseries\color{black}}
\newcommand{\sidefont}{\fontsize{7pt}{8pt}\selectfont\bfseries}

\begin{document}

\title{Unifying Amplitude-Phase Coupling and Reliable Endogenous Priors for Low-Light Image Enhancement}

\author{Tongshun Zhang\textsuperscript{\textdagger},
        Zixuan Zhong \textsuperscript{\textdagger},
        Pingping Liu\textsuperscript{*},
        Jin Kuang,
        Yubing Lu,
        Dong Liu and
        Qiuzhan Zhou
\thanks{Tongshun Zhang, Pingping Liu, and Yubing Lu are with the College of Computer Science and Technology, Jilin University, Changchun, Jilin, 130012, China, and the Key Laboratory of Symbolic Computation and Knowledge Engineering of Ministry of Education, Jilin University, Changchun, Jilin 130012, China.

Zixuan Zhong is with the College of Software, Jilin University, Changchun, Jilin, 130012, China.

Jin Kuang and Dong Liu are School of Computer and Artificial Intelligence, XiangnanUniversity, Chenzhou, 423300, China.

Qiuzhan Zhou is with the College of Communication Engineering, Jilin University, Changchun, Jilin, 130012, China.
}
\thanks{E-mail: \{tszhang23@mails.jlu.edu.cn, zhongzx24@mails.jlu.edu.cn, liupp@jlu.edu.cn, luyb24@mails.jlu.edu.cn, zhouqz@jlu.edu.cn\}}

\thanks{\textsuperscript{\textdagger}Equal contribution.}
\thanks{* Corresponding author: Pingping Liu (e-mail: liupp@jlu.edu.cn).}
}

\markboth{Journal of \LaTeX\ Class Files,~Vol.~14, No.~8, August~2021}%
{Shell \MakeLowercase{\textit{et al.}}: A Sample Article Using IEEEtran.cls for IEEE Journals}


\maketitle

\begin{abstract}
Frequency-domain methods have shown strong potential for low-light image enhancement (LLIE) because they can effectively regulate global illumination. However, most existing methods assume that amplitude and phase can be restored independently, while overlooking their inherent spectral coupling. This artificial decoupling often leads to information inconsistency and structural distortions. To address this issue, we propose a frequency-spatial framework that jointly models spectral interaction and reliable prior guidance. We first empirically analyze the coupling between amplitude and phase and develop an Amplitude-Phase Adaptive Compensation Fusion (APACF) module. Instead of enforcing a hard separation between the two components, APACF enables their bidirectional interaction, resulting in spectrally coherent restoration. In addition, directly extracting priors from raw low-light inputs is unreliable, since severe noise and underexposure destroy critical visual content. To alleviate this problem, we introduce an Endogenous Reliability Prior (ERP). Specifically, we treat an intermediate result, preliminarily restored in the frequency domain and already close to the ground-truth illumination, as a high-fidelity source for extracting complementary pseudo-infrared and gradient cues. Based on ERP, we further design a Dual-Modal Guided Refinement Network (DGRN) to inject these purified structural cues into the decoder, thereby recovering missing fine textures more accurately. Extensive experiments on benchmark datasets show that our method achieves state-of-the-art performance and consistently surpasses existing methods in both quantitative metrics and visual quality. Our code is available at \url{https://github.com/bywlzts/Amplitude-Phase-Coupling-Net}.
\end{abstract}

\begin{IEEEkeywords}
Low-light image enhancement, frequency domain, amplitude-phase collaboration
\end{IEEEkeywords}

\section{Introduction}

\begin{figure}[!t]
    \centering
    \includegraphics[width=1.0\linewidth]{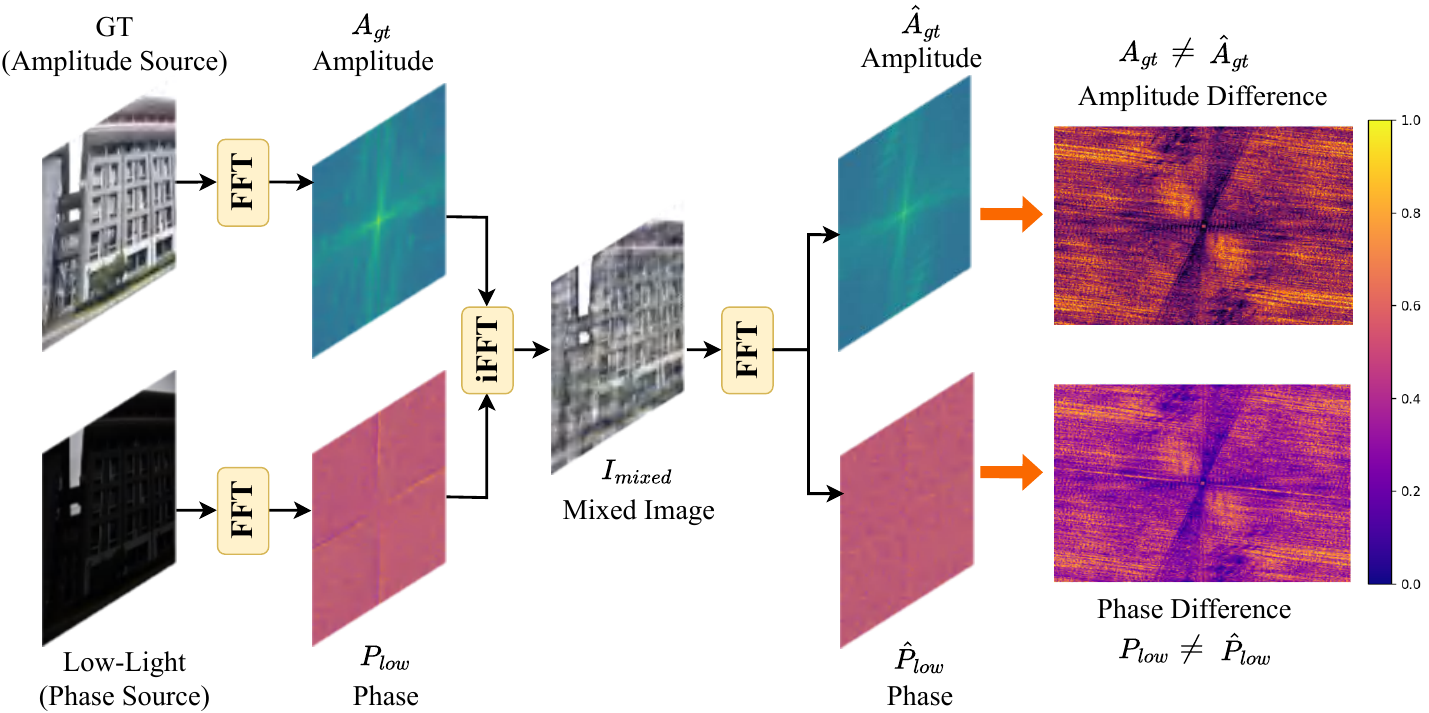}
    \caption{Reprojection consistency test. Combining the clean amplitude $A_{gt}$ with the degraded phase $P_{low}$ produces a mixed image $I_{mixed}$ with severe artifacts. Reprojecting $I_{mixed}$ to the frequency domain yields clear discrepancies between the original and reconstructed components, e.g., $A_{gt} \neq \hat{A}_{gt}$, revealing strong coupling between amplitude and phase and challenging the assumption that they can be restored independently.}
    \vspace{-0.3cm}
    \label{fig:exchange}
\end{figure}

\IEEEPARstart{I}{n} real-world imaging scenarios, low-light images often suffer from insufficient brightness, low contrast, amplified noise, and noticeable color distortion. These degradations not only impair visual perception but also undermine the reliability of downstream vision tasks~\cite{sun2022shift, ye2023llod,lu2025infrared, xiao2025disentangled}. Low-Light Image Enhancement (LLIE) aims to restore visually clear and semantically faithful images from underexposed inputs, thereby improving both scene visibility and perceptual quality.

Early LLIE methods mainly relied on histogram-based adjustment~\cite{pisano1998contrast} and Retinex-inspired physical modeling~\cite{land1977retinex}, but their performance was often limited in real scenes with spatially varying illumination and heavy noise. With the rise of deep learning, learning-based methods~\cite{cai2023retinexformer,jiang2023low,yan2025hvi, cciftcci2025low} have substantially advanced LLIE by learning direct mappings from low-light inputs to normal-light targets. More recently, Fourier-domain methods have attracted increasing attention because they are well suited to modeling global illumination and can suppress noise with relatively high efficiency. Representative works, including FECNet~\cite{Huang2022DeepFE} and FourLLIE~\cite{Wang2023FourLLIEBL}, explored the complementary roles of amplitude and phase in Fourier representations; UHDFour~\cite{li2023embedding} extended this line to ultra-high-definition restoration; DMFourLLIE~\cite{Zhang2024DMFourLLIEDA} highlighted the importance of phase modeling in multimodal enhancement; and FDN~\cite{Tu2025FourierbasedDN} further exploited spectral representations for joint low-light deblurring.

Despite their progress, existing Fourier-domain LLIE methods are still largely built on a hard-decoupling assumption: amplitude is treated as a carrier of illumination-dominant low-frequency content, whereas phase is regarded as the main source of structure-dominant high-frequency details. This simplified view neglects the fact that amplitude and phase are inherently coupled in real images. To examine this issue, we conduct a reprojection consistency test in Fig.~\ref{fig:exchange}. Specifically, we combine the clean amplitude $A_{gt}$ from the reference image with the degraded phase $P_{low}$ from the low-light input to construct a mixed image $I_{mixed}$. Although $I_{mixed}$ inherits the reference amplitude, it still exhibits obvious structural artifacts and color distortion, indicating that plausible illumination cannot be recovered by manipulating amplitude alone. We then reproject $I_{mixed}$ to the frequency domain and obtain $(\hat{A}_{gt}, \hat{P}_{low})$. The resulting difference maps show clear discrepancies between the original and re-extracted components, e.g., $A_{gt} \neq \hat{A}_{gt}$, providing direct evidence that degradation in one component propagates to the other. This observation suggests that forcibly separating amplitude and phase can lead to feature misassignment and ultimately hinder faithful detail restoration.

A second issue lies in the reliability of prior construction. Existing prior-guided methods~\cite{Zhang2024DMFourLLIEDA, Qian2025LLEFuseLI} often extract structural cues directly from raw low-light inputs. In practice, however, these inputs are heavily corrupted by shot noise and underexposure, so the resulting edges and gradients are often incomplete or spurious. Such priors are therefore unreliable and provide limited guidance for recovering fine structures. To address this issue, we introduce the Endogenous Reliability Prior (ERP), whose core idea is to extract priors from a rectified intermediate representation rather than from severely degraded observations.

\begin{figure}[!t]
    \centering
    \includegraphics[width=1.0\linewidth]{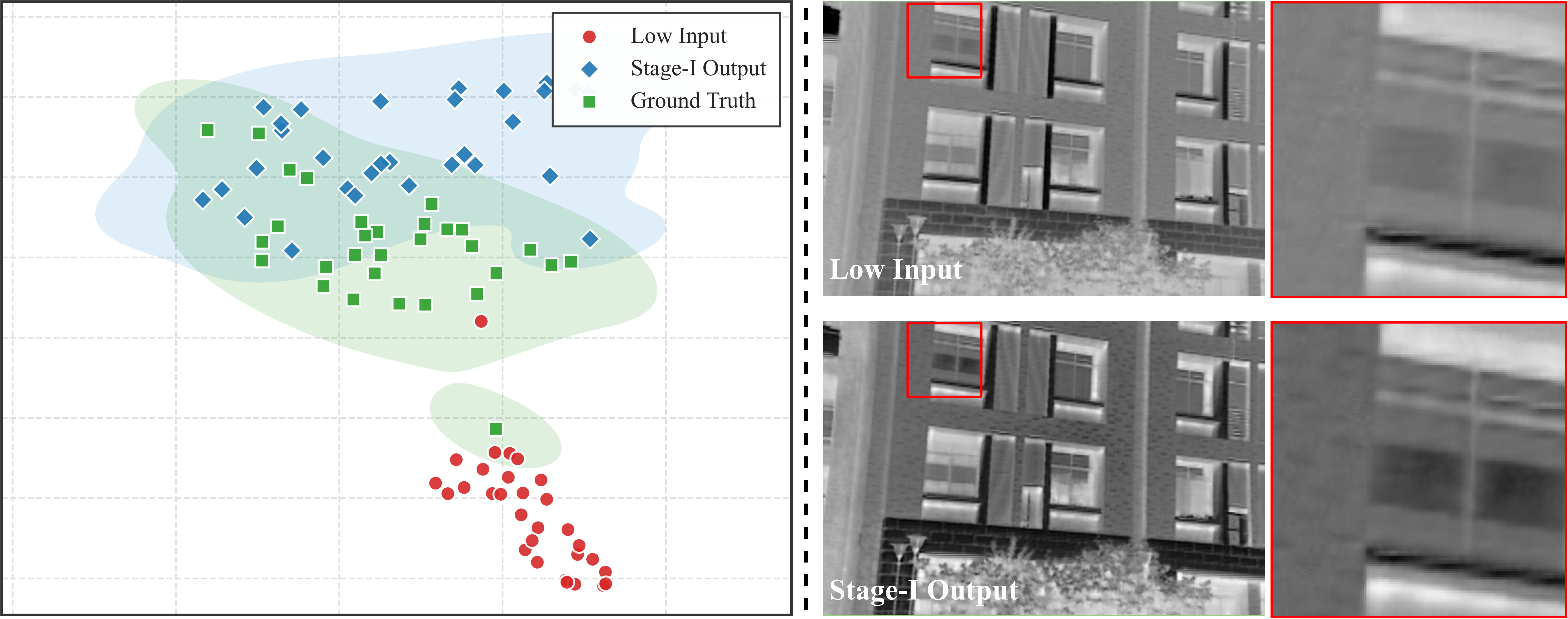}
    \caption{Validation of ERP. Left: $t$-SNE visualization of feature distributions for the low-light input (red), the Stage I output (blue), and the ground truth (green). Right: Comparison of Pseudo-Infrared (PI) features extracted from the low-light input and the Stage I output.}
    \vspace{-0.3cm}
    \label{fig:prior_reliability_combined}
\end{figure}

Existing Fourier-based methods~\cite{Huang2022DeepFE,Wang2023FourLLIEBL,Zhang2024DMFourLLIEDA} generally follow a two-stage pipeline: global restoration in the frequency domain followed by spatial detail refinement. In line with this design, our Stage I performs frequency-domain restoration to produce an intermediate result. As shown in the left panel of Fig.~\ref{fig:prior_reliability_combined}, its feature distribution is much closer to the ground-truth distribution than that of the raw low-light input. We therefore use the Stage I output as a reliable source for extracting Pseudo-Infrared (PI) and gradient priors. The right panel further shows that the endogenous priors derived from the Stage I output preserve clearer boundaries and finer structures than those extracted directly from the raw input, making them more suitable for subsequent detail refinement.

Motivated by the above observations, we propose a two-stage frequency-spatial framework that integrates spectral coupling modeling and ERP-guided refinement. In Stage I, we introduce an Amplitude-Phase Adaptive Compensation Fusion (APACF) module to explicitly model the bidirectional interaction between amplitude and phase, thereby alleviating the artifacts caused by hard decoupling and producing a cleaner intermediate representation. In Stage II, we build a Dual-Modal Guided Refinement Network (DGRN) to exploit ERP for detail recovery. Specifically, DGRN first adopts a Prior-Injected Guided Encoder (PIGE) to inject PI and multi-scale gradient priors into feature encoding while suppressing redundant responses. It then employs a Prior-Enhanced Spatial-Frequency Block (PESFB) to jointly refine features in the spatial and frequency domains. Finally, a reconstruction head aggregates the refined features to generate the enhanced image.

The main contributions are summarized as follows:
\begin{itemize}
    \item We revisit the hard-decoupling assumption in Fourier-domain LLIE and show, through a reprojection consistency test, that amplitude and phase are strongly coupled. To model this dependency, we propose APACF for bidirectional amplitude-phase interaction.

    \item We propose ERP, a ``rectify-then-extract'' strategy that derives PI and gradient priors from the rectified Stage I output instead of the corrupted raw input, providing more reliable guidance for structure restoration.

    \item We develop a two-stage frequency-spatial framework in which DGRN leverages ERP for refinement. Within DGRN, PIGE injects dual-modal priors into feature encoding, and PESFB performs joint spatial-frequency refinement for accurate detail recovery.

    \item Extensive experiments on multiple benchmarks show that our method achieves highly competitive performance in both quantitative evaluation and visual quality, and ranks among the top-performing methods on several datasets and metrics.
\end{itemize}

\section{Related Work}

\subsection{Deep Learning-based LLIE}

Deep learning has substantially advanced LLIE, with methods evolving from early convolutional models to more sophisticated prior-guided frameworks. Early studies mainly focused on directly learning image-to-image mappings. For example, LLNet~\cite{lore2017llnet} employed stacked autoencoders for joint enhancement and denoising, while Retinex-Net~\cite{wei2018deep} incorporated Retinex theory~\cite{land1977retinex} into a learnable decomposition-and-enhancement pipeline, improving the interpretability of the restoration process.

To further enhance representation capability, subsequent works introduced stronger network architectures. LPNet~\cite{li2020luminance} exploited multi-scale feature pyramids to better model illumination variations, whereas ABSGNet~\cite{Chen2022AttentionbasedBS} and HCSAMNet~\cite{Chen2024HCSAMNetMN} incorporated attention mechanisms to improve feature selection and contextual aggregation. More recently, Transformer-based methods have shown strong capability in modeling long-range dependencies, with representative examples including RetinexFormer~\cite{cai2023retinexformer} and LLFormer~\cite{wang2023ultra}.

Despite these advances, spatial-domain LLIE methods still face challenges in maintaining global illumination consistency while preserving fine details under extreme darkness. To alleviate the ill-posedness of LLIE, introducing additional priors has become an important direction. One line of work relies on external priors from auxiliary modalities, especially infrared signals, to provide structural guidance~\cite{Wen2025InfraredguidedGN,Jenifa2025MultimodalLI}. Although such methods can offer reliable complementary cues, they usually require additional hardware and accurate cross-modal registration. To avoid these limitations, recent studies have turned to internal priors derived from the input image itself, such as semantic cues~\cite{Wang2024SBCNetSB,Hou2024SemanticAG}, edge information~\cite{Huang2024EdgeguidedLI}, and histogram statistics~\cite{zhang2021unsupervised}. However, when the input is severely degraded, directly extracted priors are often unstable and may provide limited support for faithful structure recovery.

\subsection{Frequency-Domain Learning in LLIE}

Frequency-domain learning has attracted growing attention in low-level vision because of its advantages in modeling global image statistics and separating degradations~\cite{Wang2023FourLLIEBL,Zhang2024DMFourLLIEDA,Wang2024LearningFR,li2023embedding}. A common strategy is to transform images into the Fourier domain and restore them by manipulating spectral components. Representative methods such as FourLLIE~\cite{Wang2023FourLLIEBL}, DMFourLLIE~\cite{Zhang2024DMFourLLIEDA}, UHDFour~\cite{li2023embedding}, FDN~\cite{Tu2025FourierbasedDN}, and SFDiff~\cite{Wan2024SFDiffDM} have demonstrated the effectiveness of this paradigm for illumination correction and related restoration tasks.

Most existing Fourier-domain LLIE methods loosely associate amplitude with illumination-dominant or low-frequency information and phase with structure-dominant or high-frequency information. Based on this view, they typically process the two components separately, for example by enhancing amplitude while preserving, exchanging, or conditionally modulating phase. Although effective to some extent, this design is still rooted in a simplified decoupling assumption. In practice, amplitude and phase are not strictly independent, and degradation in one component can influence the other. Ignoring this coupling may lead to spectral inconsistency and structural artifacts. Moreover, while some methods further incorporate prior guidance, they rarely examine whether the extracted priors are themselves reliable under severe low-light degradation. Therefore, jointly modeling amplitude-phase coupling and reliable prior guidance remains insufficiently explored in current LLIE research.

\begin{figure*}[t]
    \centering
    \includegraphics[width=1.0\linewidth]{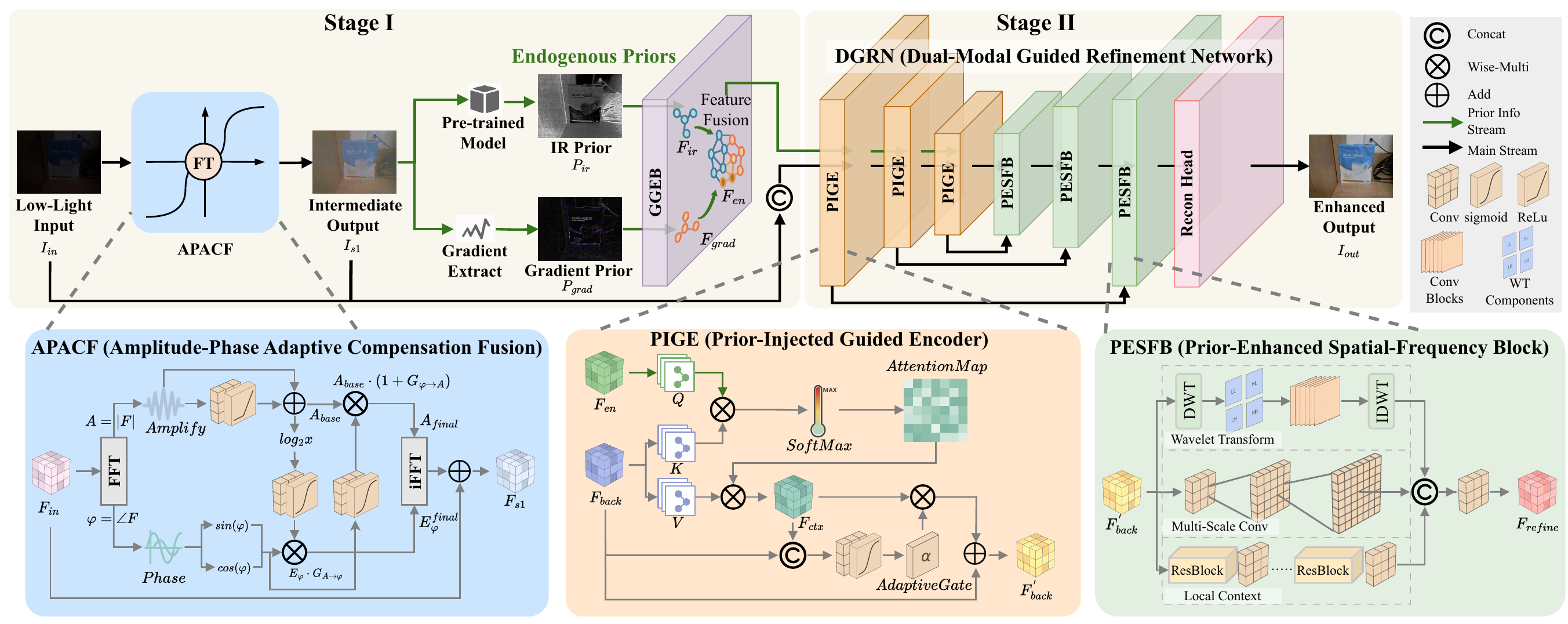}
    \vspace{-0.3cm}
    \caption{
    Overview of the proposed framework.
    Stage I: APACF performs coupled amplitude-phase restoration to produce $I_{s1}$, from which ERP is extracted.
    Stage II: DGRN refines details under ERP guidance to generate the enhanced output $I_{out}$.
    }
    \label{fig:pipeline}
    \vspace{-0.3cm}
\end{figure*}

\section{Methods}

The overall pipeline of our method is illustrated in Fig.~\ref{fig:pipeline}. 
Given a low-light input image $I_{in} \in \mathbb{R}^{H \times W \times 3}$, our goal is to learn a mapping $\Phi$ that restores a high-quality image $I_{out}$ with normal exposure, natural color, and faithful structural details. 
To this end, we adopt a two-stage frequency-spatial framework that progressively performs global rectification and local refinement.

\noindent \textbf{Stage I: Global Frequency Rectification.} 
Existing Fourier-domain LLIE methods often process amplitude and phase separately, ignoring their intrinsic coupling and potentially introducing spectral inconsistency. 
To address this, we design the Amplitude-Phase Adaptive Compensation Fusion (APACF) module to explicitly model their bidirectional interaction. 
Given $I_{in}$, APACF produces a rectified feature representation $F_{s1}$, from which the intermediate image $I_{s1}$ is reconstructed. 
This intermediate result restores global illumination, color distribution, and coarse structures, providing a more reliable foundation for subsequent prior extraction.

\noindent \textbf{ERP.} 
Instead of extracting priors directly from the degraded input, we construct the Endogenous Reliability Prior from $I_{s1}$. 
Specifically, a pretrained estimator derives the Pseudo-Infrared (PI) prior $P_{ir}$, while a gradient extractor produces the gradient prior $P_{grad}$. 
These two cues are then fused by the Gradient-Guided Enhancement Block (GGEB), where the gradient prior modulates the PI prior to produce an endogenous prior feature $F_{en}$ with enhanced structural awareness.

\noindent \textbf{Stage II: ERP-Guided Detail Refinement.} 
We employ the Dual-Modal Guided Refinement Network (DGRN) for detail recovery. 
DGRN first uses the Prior-Injected Guided Encoder (PIGE) to inject $F_{en}$ into feature encoding while suppressing redundant responses. 
It then applies the Prior-Enhanced Spatial-Frequency Block (PESFB) to jointly refine features in the spatial and wavelet domains. 
Finally, a reconstruction head aggregates the refined features to produce the enhanced image $I_{out}$. 
Benefiting from ERP-guided refinement, DGRN recovers fine textures more accurately while suppressing residual noise and artifacts.

Together, the two-stage design decouples the restoration task into a coarse-to-fine process: Stage I handles global illumination and color recovery via coupled frequency-domain modeling, while Stage II focuses on local detail refinement under reliable prior guidance.

\subsection{Stage I: Frequency Restoration Network}

Stage I aims to rectify degraded illumination and structural statistics in the frequency domain. 
As shown in Fig.~\ref{fig:pipeline}, this stage is implemented by APACF. 
Unlike conventional methods~\cite{Huang2022DeepFE, Wang2023FourLLIEBL, Zhang2024DMFourLLIEDA, li2023embedding} that process amplitude and phase independently, APACF is designed around the inherent spectral coupling between the two components.

\noindent \textbf{Motivation for Coupled Processing.} 
This coupling can be understood from the signal reconstruction perspective. 
For a 2D spatial signal $x(m,n)$ of size $M \times N$, the inverse discrete Fourier transform (IDFT) gives:
\begin{equation}
    x(m,n) = \frac{1}{MN} \sum_{u=0}^{M-1} \sum_{v=0}^{N-1} \mathcal{A}(u,v) \cos\big(\theta_{m,n}(u,v) - \boldsymbol{\varphi}(u,v)\big),
\end{equation}
where $\mathcal{A}(u,v)$ and $\boldsymbol{\varphi}(u,v)$ denote the amplitude and phase spectra, and $\theta_{m,n}(u,v) = 2\pi(\frac{um}{M} + \frac{vn}{N})$ is the spatial phase offset. 
At each frequency, the amplitude and phase are multiplicatively coupled through the cosine function, meaning that errors in one component inevitably distort the pixel-level reconstruction.
This has two important consequences for low-light restoration. 
First, when phase is degraded by noise, the reconstructed signal at each pixel is distorted even if amplitude is perfectly restored, because the erroneous phase shifts the sinusoidal bases incorrectly. 
Second, and more critically, if we reconstruct an image using clean amplitude $\mathcal{A}_{\text{gt}}$ but degraded phase $\boldsymbol{\varphi}_{\text{low}}$, the resulting spatial-domain image $I_{\text{mixed}} = \mathcal{F}^{-1}(\mathcal{A}_{\text{gt}}, \boldsymbol{\varphi}_{\text{low}})$ often contains pixel values outside the valid intensity range (e.g., negative values or values exceeding the maximum). 
To visualize or further process such an image, it is typically clipped or normalized to the admissible range $[0,1]$ (or $[0,255]$), which is a nonlinear operation. 
Consequently, the Fourier transform of this postprocessed image,
\begin{equation}
    \hat{X}(u,v) = \mathcal{F}(I_{\text{mixed}}) = \hat{\mathcal{A}}(u,v) e^{j\hat{\boldsymbol{\varphi}}(u,v)},
\end{equation}
no longer matches the original components. 
Formally, we have $\hat{\mathcal{A}} \neq \mathcal{A}_{\text{gt}}$ and $\hat{\boldsymbol{\varphi}} \neq \boldsymbol{\varphi}_{\text{low}}$, demonstrating that amplitude and phase errors propagate into each other under realistic image-domain constraints. 
This explains the reprojection inconsistency observed in Fig.~\ref{fig:exchange} and motivates our bidirectional interaction design.

\noindent \textbf{APACF Architecture.} 
Given an input feature $F_{in}$, we first transform it into the Fourier domain:
\begin{equation}
\mathcal{A}(u,v), \boldsymbol{\varphi}(u,v) = |\mathcal{F}(F_{in})|, \angle \mathcal{F}(F_{in}).
\end{equation}
To avoid phase wrapping discontinuities around $\pm \pi$, we embed the phase into a trigonometric space:
\begin{equation}
E_{\boldsymbol{\varphi}} = [\sin(\boldsymbol{\varphi}), \cos(\boldsymbol{\varphi})].
\end{equation}

We then model the bidirectional amplitude-phase interaction. 
A residual transformation first enhances the amplitude branch: $\mathcal{A}_{base} = \mathrm{Conv}(\mathcal{A}) + \mathcal{A}$.

\textit{Phase-to-Amplitude Path:} 
A phase-guided gate $G_{\boldsymbol{\varphi}\rightarrow\mathcal{A}} = \sigma(\mathrm{Conv}(E_{\boldsymbol{\varphi}}))$ is generated to modulate amplitude enhancement:
\begin{equation}
\mathcal{A}_{final} = \mathcal{A}_{base} \odot (1 + G_{\boldsymbol{\varphi}\rightarrow\mathcal{A}}).
\end{equation}
This allows structural phase cues to selectively boost informative amplitude responses.

\textit{Amplitude-to-Phase Path:} 
To suppress unreliable phase in dark regions, we apply a logarithmic transform (following Weber's law~\cite{Kim2014WeightedHE}) to the amplitude before generating a reliability gate:
\begin{equation}
G_{\mathcal{A}\rightarrow\boldsymbol{\varphi}} = \sigma(\mathrm{Conv}(\log(\mathcal{A}_{base} + \epsilon))),
\end{equation}
where $\epsilon$ is a small constant. The phase embedding is refined as:
\begin{equation}
E_{\boldsymbol{\varphi}}^{final} = E_{\boldsymbol{\varphi}} \odot G_{\mathcal{A}\rightarrow\boldsymbol{\varphi}}.
\end{equation}
This gate attenuates noisy phase responses in low-energy regions while preserving informative structures.

Finally, we convert the refined embedding back to the phase domain using $\mathrm{atan2}$, and reconstruct the spatial feature via the inverse Fourier transform with a residual connection:
\begin{equation}
F_{s1} = \mathcal{F}^{-1}(\mathcal{A}_{final}, \boldsymbol{\varphi}_{final}) + F_{in}.
\end{equation}
The resulting feature $F_{s1}$ serves as the frequency-restored representation, from which the intermediate image $I_{s1}$ is reconstructed for subsequent ERP construction and Stage II refinement.

\subsection{Endogenous Reliability Prior (ERP) Construction}

\begin{figure}[!t]
    \centering
    \includegraphics[width=1.0\linewidth]{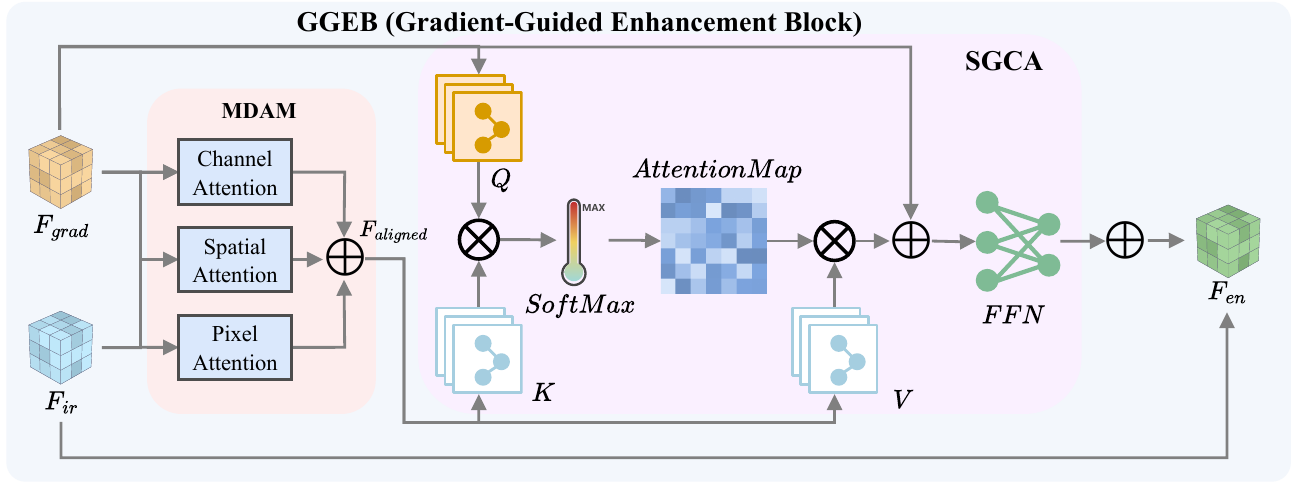}
    \caption{Gradient-Guided Enhancement Block (GGEB). MDAM first aligns PI and gradient features across channel, spatial, and pixel dimensions. SGCA then performs gradient-guided cross-attention to produce the endogenous prior feature $F_{en}$.}
    \vspace{-0.3cm}
    \label{fig:GGEB}
\end{figure}

As discussed above, directly extracting priors from severely degraded low-light inputs is often unreliable, since noise and underexposure can corrupt both semantic cues and structural boundaries. 
To obtain more trustworthy guidance, we construct the ERP from the Stage~I intermediate image $I_{s1}$, whose illumination and coarse structures have already been rectified. 
Specifically, we extract two complementary priors and fuse them through the GGEB.

The first is the PI prior $P_{ir}$, generated by a pretrained estimator~\cite{lee2023edge}. 
Since PI representations are relatively insensitive to illumination variations, they provide object-aware semantic cues even in dark environments. 
The second is the gradient prior $P_{grad}$, extracted by a multi-scale Sobel operator to emphasize local edges and texture transitions. 
We denote their corresponding feature representations as $F_{ir}$ and $F_{grad}$, respectively.

A straightforward fusion of $F_{ir}$ and $F_{grad}$ is insufficient because the two modalities emphasize different properties: PI features focus on region-level semantic responses, whereas gradient features highlight local structural details. 
To bridge this modality gap, GGEB comprises two sub-modules: the Multi-Dimensional Alignment Module (MDAM) and the Structure-Guided Cross-Attention (SGCA), as shown in Fig.~\ref{fig:GGEB}.

MDAM first aligns the PI and gradient features across multiple dimensions. Specifically, it performs channel, spatial, and pixel-wise interactions to recalibrate the two inputs:
\begin{equation}
\begin{aligned}
F_{c} &= \mathcal{M}_{\mathrm{channel}}(F_{ir}, F_{grad}), \\
F_{s} &= \mathcal{M}_{\mathrm{spatial}}(F_{ir}, F_{grad}), \\
F_{p} &= \mathcal{M}_{\mathrm{pixel}}(F_{ir}, F_{grad}),
\end{aligned}
\end{equation}
where $\mathcal{M}_{\mathrm{channel}}(\cdot)$, $\mathcal{M}_{\mathrm{spatial}}(\cdot)$, and $\mathcal{M}_{\mathrm{pixel}}(\cdot)$ denote the corresponding attention operations. The aligned feature is then obtained by
\begin{equation}
F_{aligned}=F_{c}+F_{s}+F_{p}.
\end{equation}
Through this process, MDAM reduces the discrepancy between semantic and structural representations, providing a more consistent basis for subsequent fusion.

Building on the aligned feature, SGCA injects structure-aware guidance via cross-attention. 
Since gradient cues offer more reliable localization of boundaries and fine details, we use $F_{grad}$ to generate the query, while $F_{aligned}$ generates the key and value:
\begin{equation}
\begin{aligned}
Q &= \mathrm{DWConv}(F_{grad}), \\
K &= \mathrm{DWConv}(F_{aligned}), \\
V &= \mathrm{DWConv}(F_{aligned}),
\end{aligned}
\end{equation}
where $\mathrm{DWConv}(\cdot)$ denotes depth-wise convolution. The cross-attention is formulated as
\begin{equation}
\mathrm{Attention}(Q,K,V)=\mathrm{Softmax}\left(\frac{QK^{T}}{\sqrt{d_k}}\right)V,
\end{equation}
where $d_k$ is the channel dimension of the key. In this way, structural cues serve as an anchor to retrieve semantically consistent contexts from the aligned PI feature.

As illustrated in Fig.~\ref{fig:GGEB}, the attended feature is first combined with the gradient feature through a residual connection and then passed through a feed-forward network:
\begin{equation}
F_{att}=\mathrm{Attention}(Q,K,V)+F_{grad},
\end{equation}
\begin{equation}
F_{en}=\mathrm{FFN}(F_{att})+F_{ir}.
\end{equation}
The resulting endogenous prior feature $F_{en}$ preserves the semantic robustness of the PI prior while incorporating precise structural constraints from the gradient prior. 
It is then forwarded to Stage~II to guide fine-grained detail refinement.

\begin{table*}[t]
\centering
\caption{Quantitative comparison on LOL-v1~\cite{lol}, LOL-v2-Syn~\cite{lol}, LSRW-Huawei~\cite{lsrw} and LSRW-Nikon~\cite{lsrw}. \textbf{Bold} and \underline{underline} indicate the best and second-best results, respectively.}
\vspace{-0.2cm}
\label{tab:main_results}
\resizebox{\textwidth}{!}{
\begin{tabular}{l|ccc|ccc|ccc|ccc}
\toprule
\multirow{2}{*}{\textbf{Methods}} & \multicolumn{3}{c|}{\textbf{LOL-v1}} & \multicolumn{3}{c|}{\textbf{LOL-v2-Syn}} & \multicolumn{3}{c|}{\textbf{LSRW-Huawei}} & \multicolumn{3}{c}{\textbf{LSRW-Nikon}}  \\
 & PSNR $\uparrow$ & SSIM $\uparrow$ & LPIPS $\downarrow$ & PSNR $\uparrow$ & SSIM $\uparrow$ & LPIPS $\downarrow$ & PSNR $\uparrow$ & SSIM $\uparrow$ & LPIPS $\downarrow$ & PSNR $\uparrow$ & SSIM $\uparrow$ & LPIPS $\downarrow$ \\ \midrule
MIRNet~\cite{lowlight9} & \textbf{24.14} & 0.8305 & 0.2502 & 22.52 & 0.8997 & 0.0568 & 19.98 & 0.6085 & 0.2154 & 17.10 & 0.5022 & 0.2170  \\
Kind++~\cite{kind++} & 18.97 & 0.8042 & 0.1756 & 21.17 & 0.8814 & 0.0678 & 15.43 & 0.5695 & 0.2366 & 14.79 & 0.4749 & \textbf{0.2111}  \\ 
FourLLIE~\cite{Wang2023FourLLIEBL} & 20.99 & 0.8071 & 0.0952 & 24.65 & 0.9192 & 0.0389 & 21.11 & 0.6256 & 0.1825 & 17.82 & 0.5036 & 0.2150  \\
UHDFour~\cite{li2023embedding} & 22.89 & 0.8147 & 0.0934 & 23.64 & 0.8998 & 0.0341 & 19.39 & 0.6006 & 0.2466 & \underline{17.94} & 0.5195 & 0.2546  \\
DMFourLLIE\cite{Zhang2024DMFourLLIEDA} & 22.98 & 0.8273 & 0.0792 & 25.74 & 0.9308 & 0.0251 & 21.09 & 0.6328 & 0.1804 & 17.04 & \underline{0.5274} & 0.2178  \\ 
Retinexformer~\cite{cai2023retinexformer} & 22.71 & 0.8177 & 0.0922 & 25.67 & 0.9295 & 0.0273 & 21.23 & 0.6309 & 0.1699 & 17.64 & 0.5082 & 0.2784  \\
Wave-Mamba~\cite{Zou2024WaveMambaWS} & 22.76 & 0.8419 & 0.0791 & 24.69 & 0.9271 & 0.0584 & 21.19 & 0.6391 & 0.1818 & 17.34 & 0.5192 & 0.2294 \\
RetinexMamba~\cite{Bai2024RetinexmambaRM} & 23.15 & 0.8210 & 0.0876 & \underline{25.89} & 0.9346 & 0.0389 & 20.88 & 0.6298 & 0.1689 & 17.59 & 0.5133 & 0.2534 \\ 

CIDNet~\cite{yan2025hvi} & 23.81 & \textbf{0.8574} & 0.0856 & 24.69 & \textbf{0.9419} &  0.0437 & 20.30 & 0.6054 & 0.3500 & 17.16 & 0.4975 & 0.2816 \\

CWNet~\cite{Zhang2025CWNetCW} & \underline{23.60} & \underline{0.8496} & \textbf{0.0648} & 25.50 & 0.9362 & \underline{0.0241} & \textbf{21.50} & \underline{0.6397} & \textbf{0.1562} & 17.38 & 0.5119 & 0.2575\\ \midrule \rowcolor{gray!15}

\textbf{Ours} & \underline{23.81} & \underline{0.8547} & \underline{0.0715} & \textbf{26.42} & \underline{0.9371} & \textbf{0.0222}  & \underline{21.32} & \textbf{0.6416} & \underline{0.1663} & \textbf{17.98} & \textbf{0.5319} & \underline{0.2153}  \\ \bottomrule
\end{tabular}
}
\vspace{-0.3cm}
\end{table*}

\begin{figure*}[t]
    \centering
    \setlength{\fboxsep}{0pt} 
    \subfloat[Input]{%
        \includegraphics[width=0.195\linewidth]{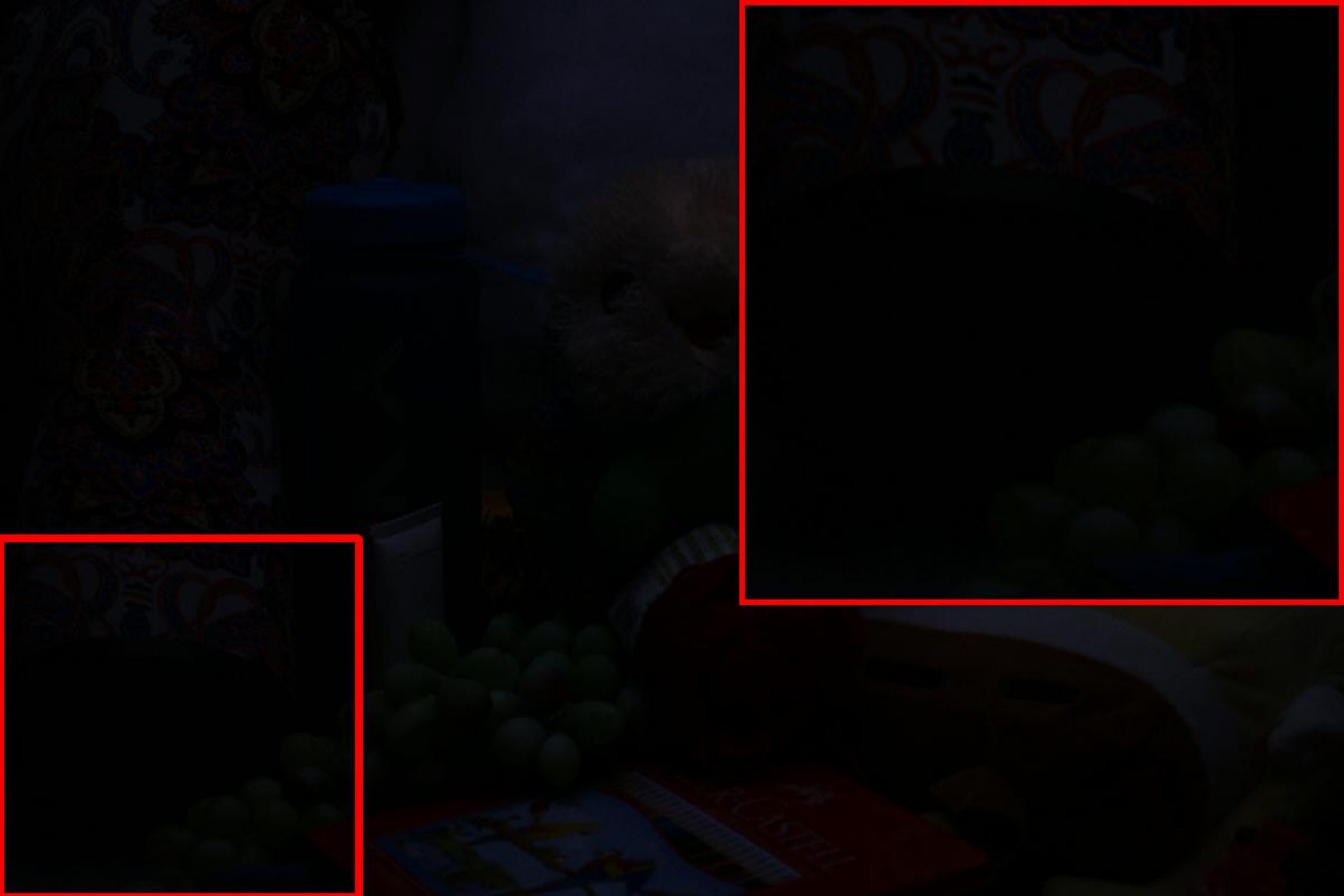}%
    }\hspace{0.5mm}%
    \subfloat[Reference]{%
        \includegraphics[width=0.195\linewidth]{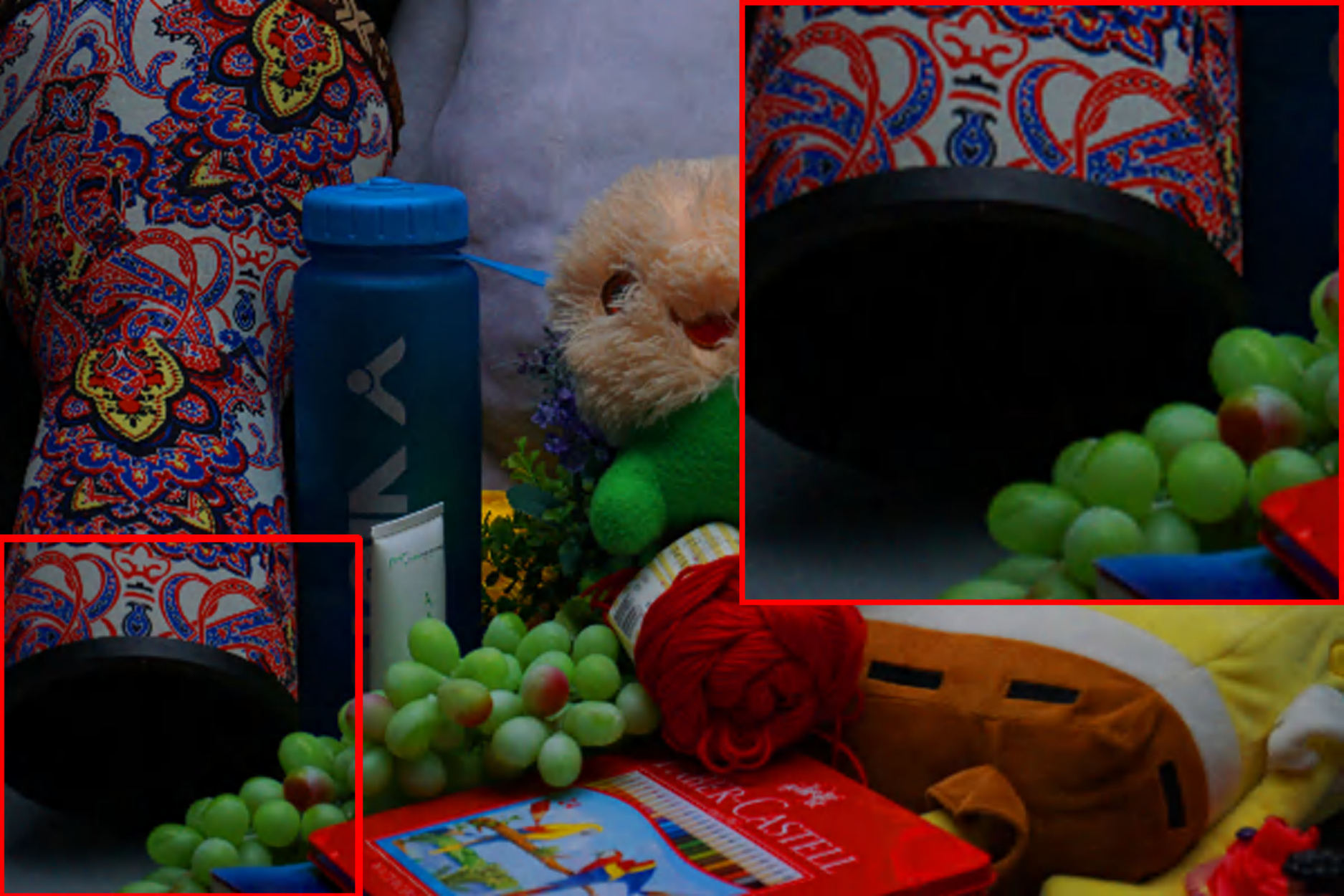}%
    }\hspace{0.5mm}%
    \subfloat[FourLLIE]{%
        \includegraphics[width=0.195\linewidth]{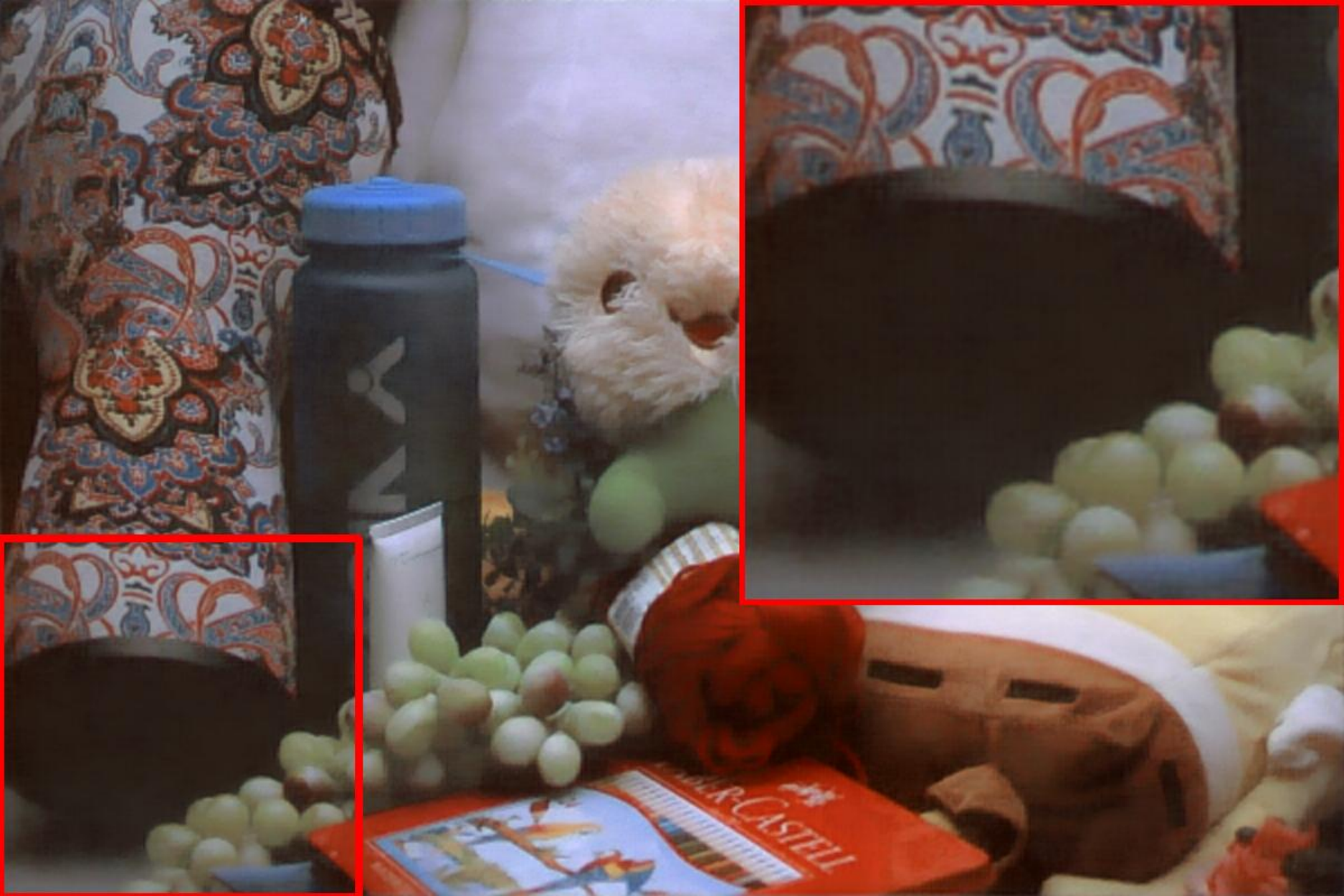}%
    }\hspace{0.5mm}%
    \subfloat[DMFourLLIE]{%
        \includegraphics[width=0.195\linewidth]{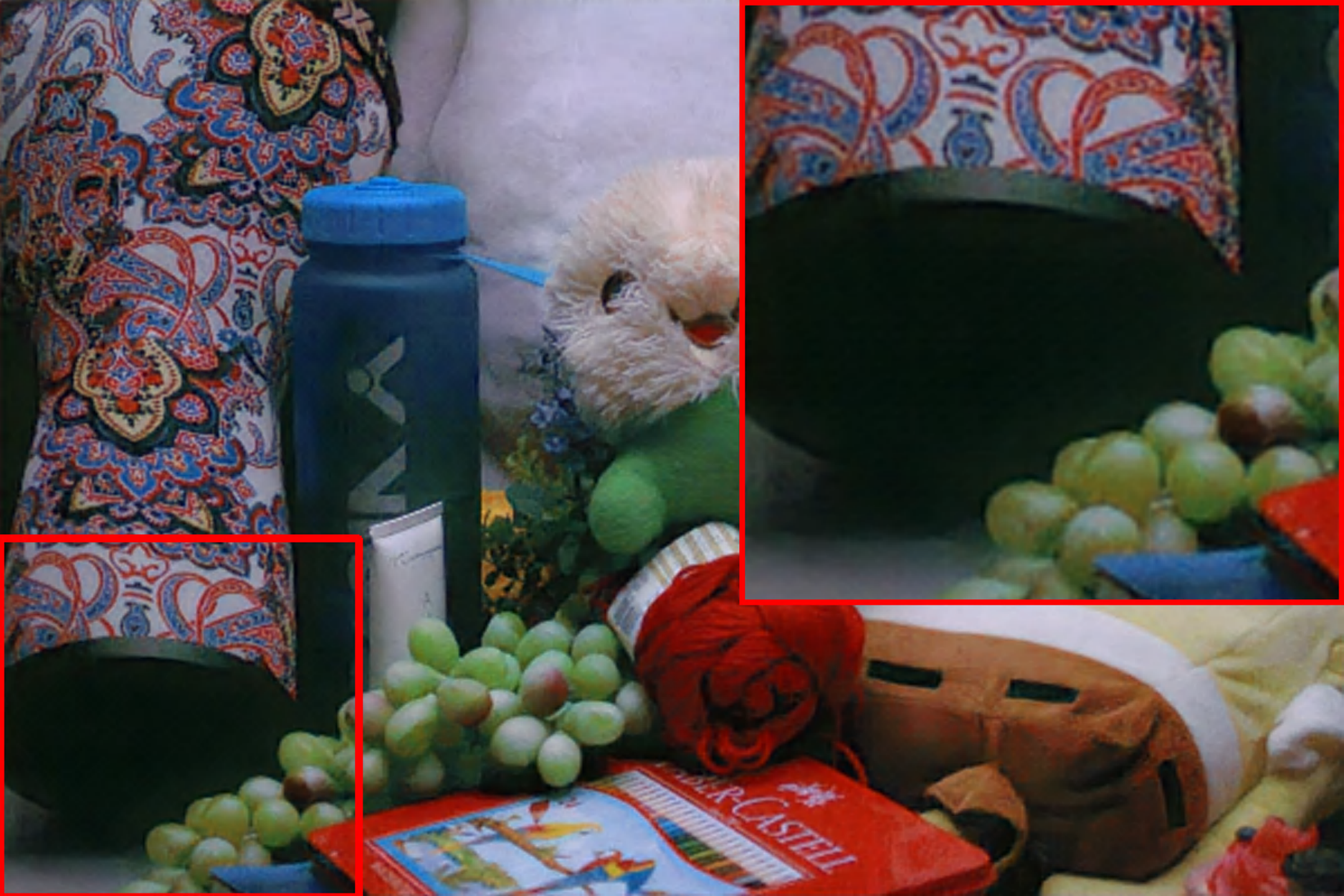}%
    }\hspace{0.5mm}%
    \subfloat[CWNet]{%
        \includegraphics[width=0.195\linewidth]{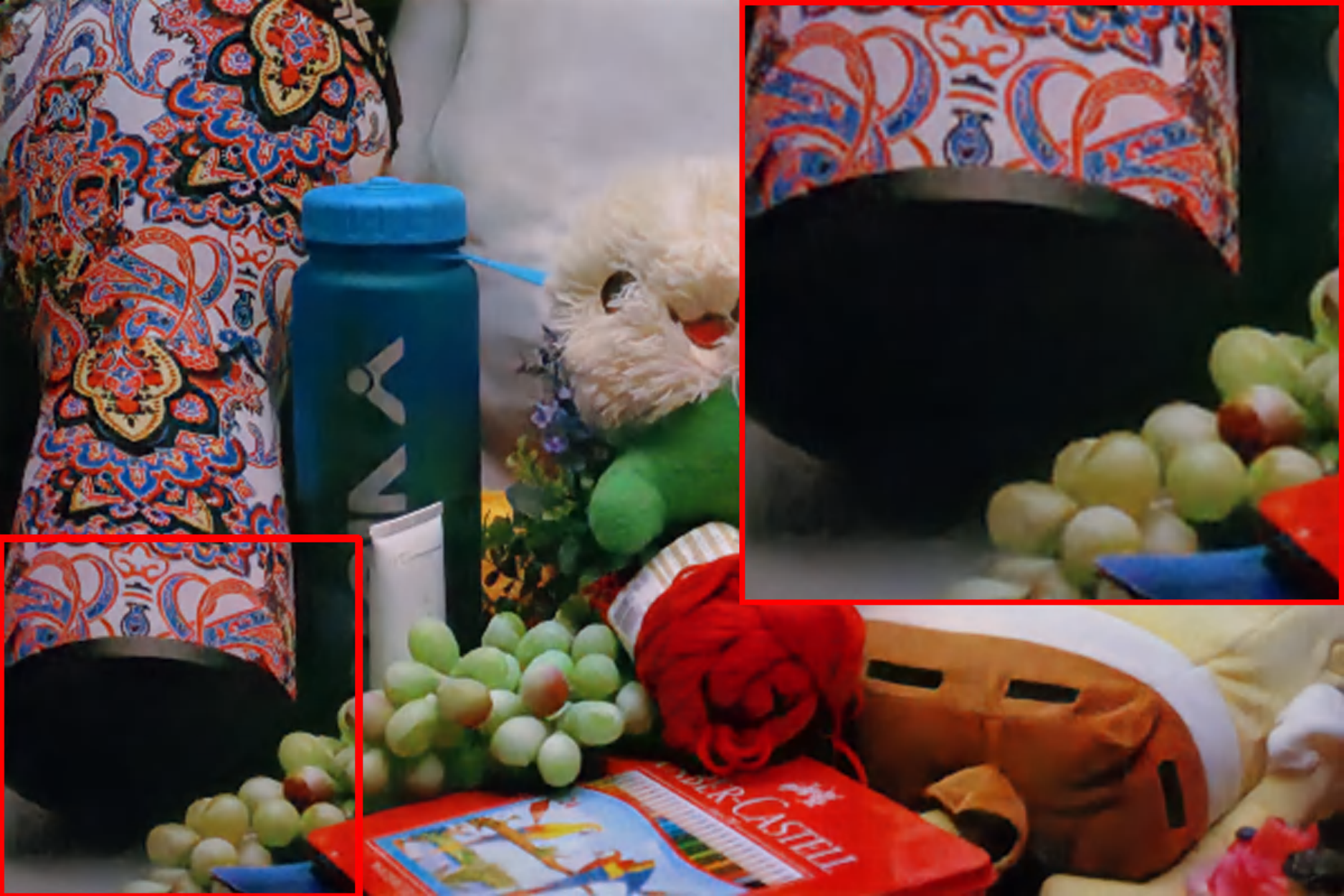}%
    }
    \vspace{-0.01mm}
    \subfloat[CIDNet]{%
        \includegraphics[width=0.195\linewidth]{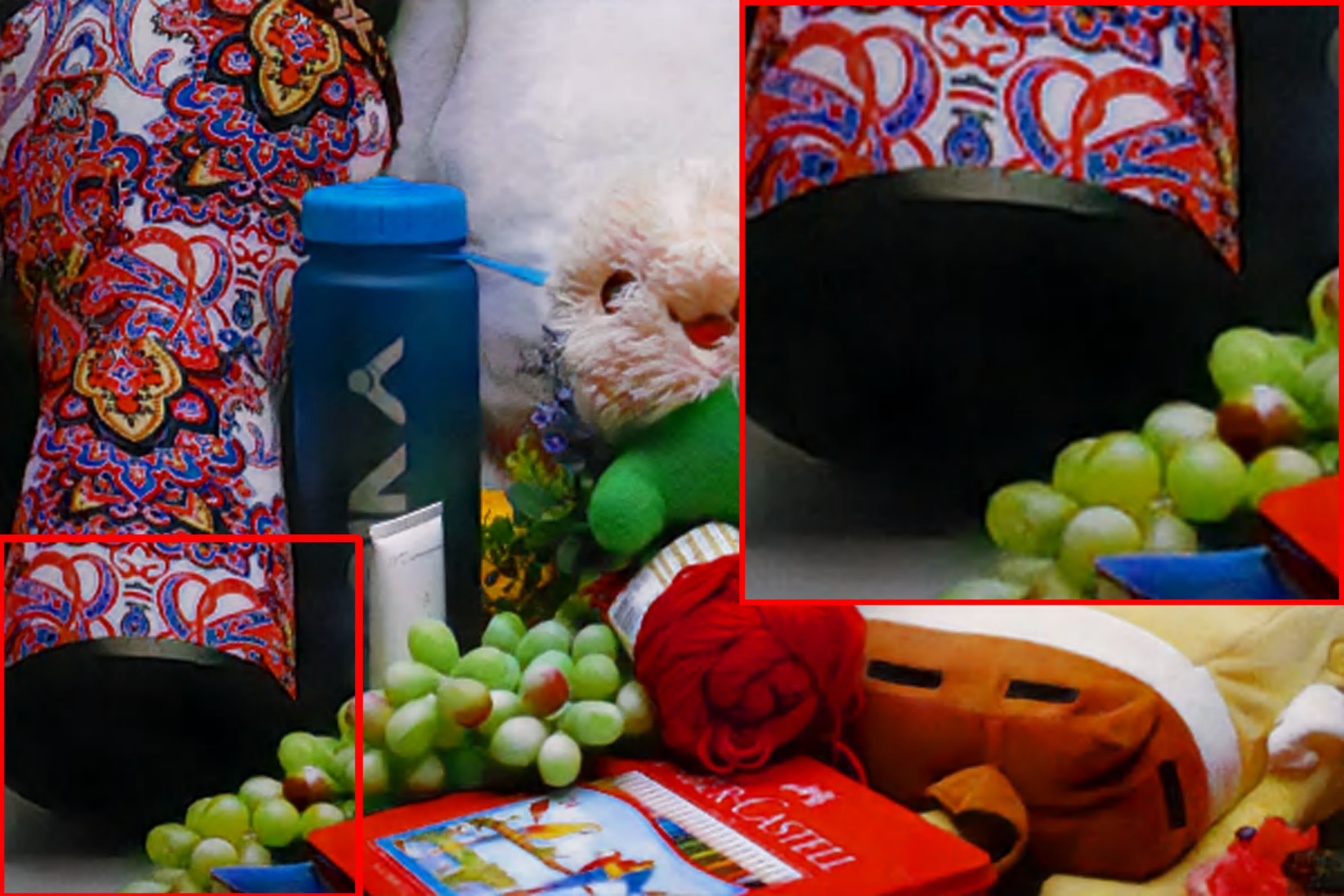}%
    }\hspace{0.5mm}%
    \subfloat[UHDFour]{%
        \includegraphics[width=0.195\linewidth]{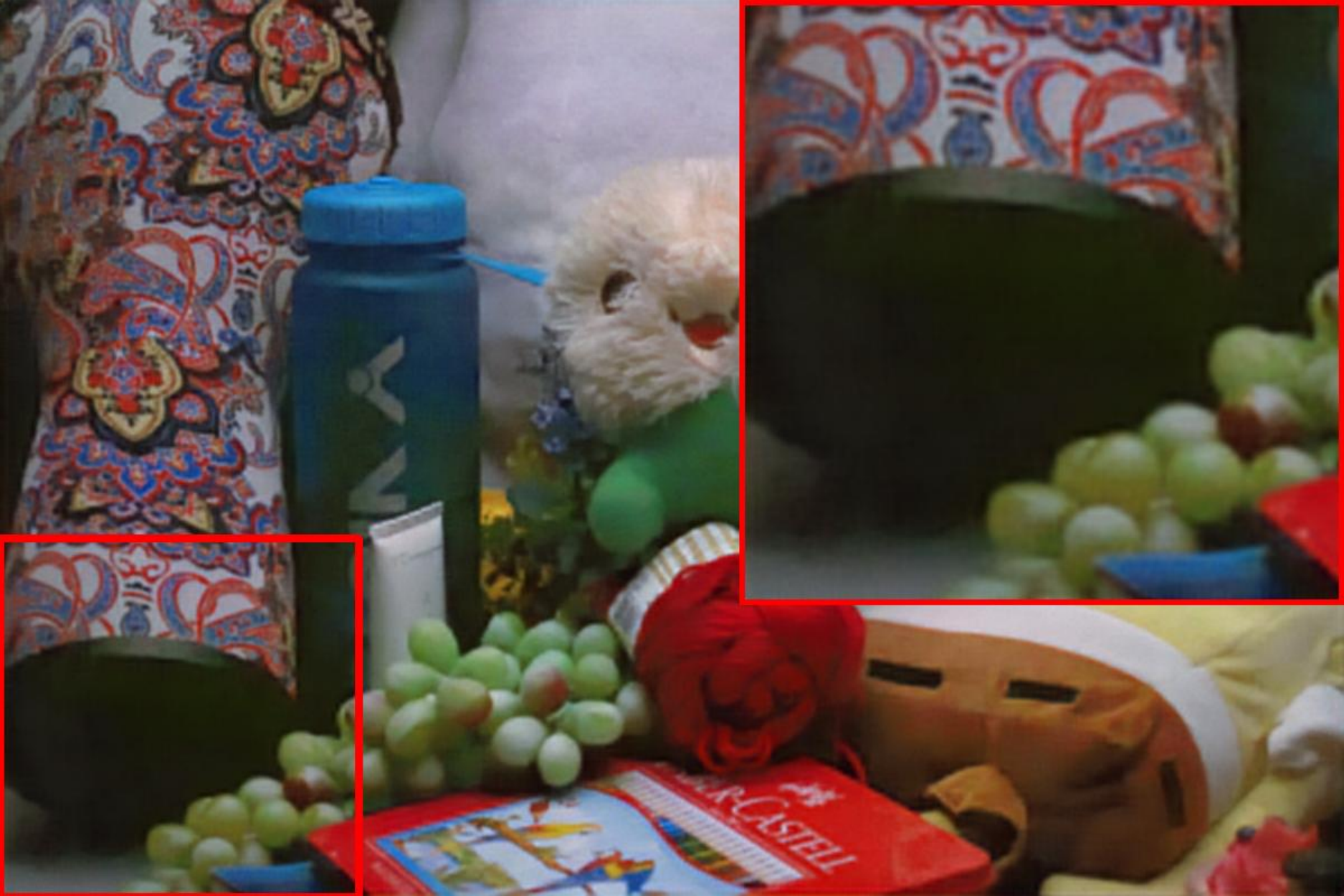}%
    }\hspace{0.5mm}%
    \subfloat[Wave-Mamba]{%
        \includegraphics[width=0.195\linewidth]{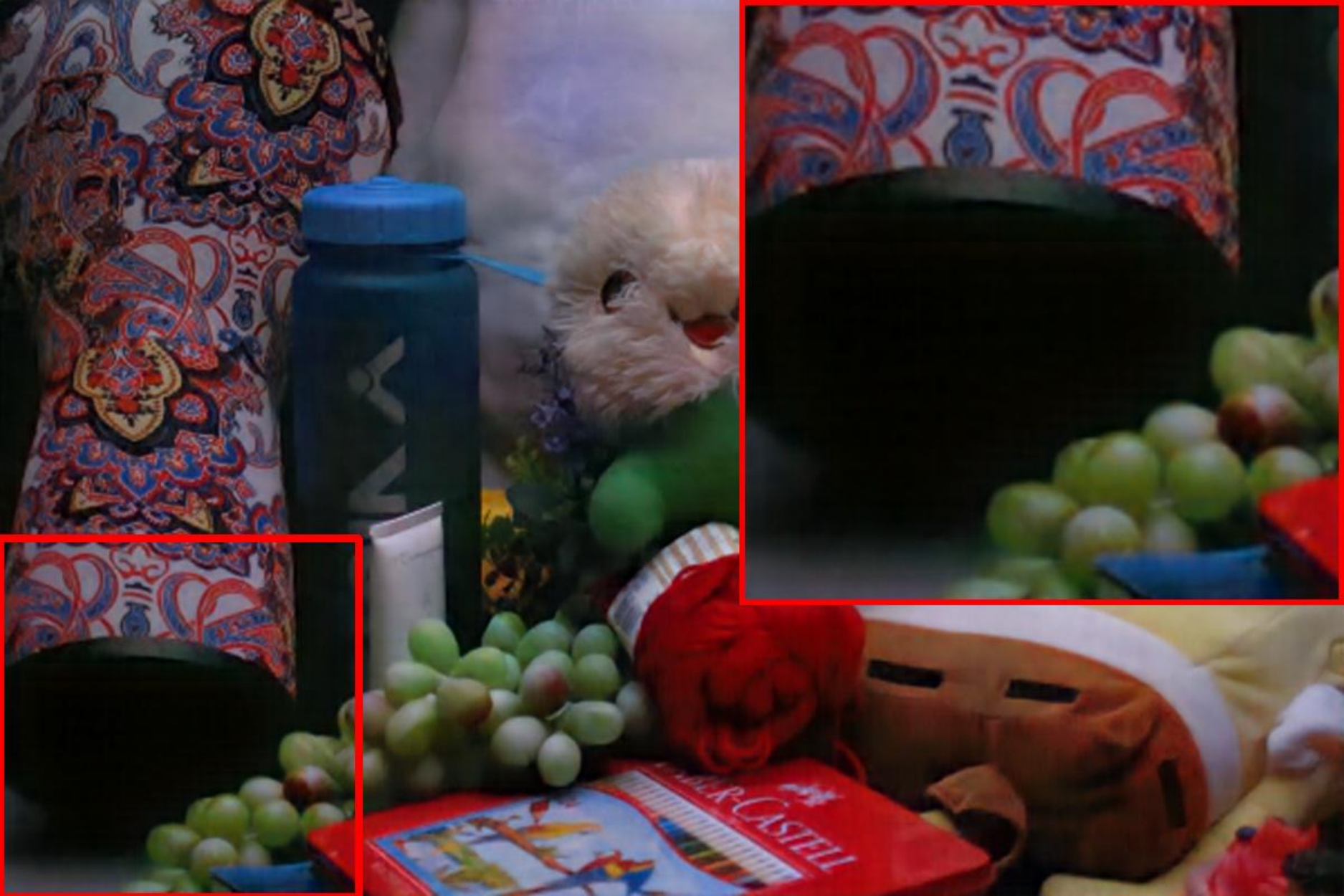}%
    }\hspace{0.5mm}%
    \subfloat[Retinexformer]{%
        \includegraphics[width=0.195\linewidth]{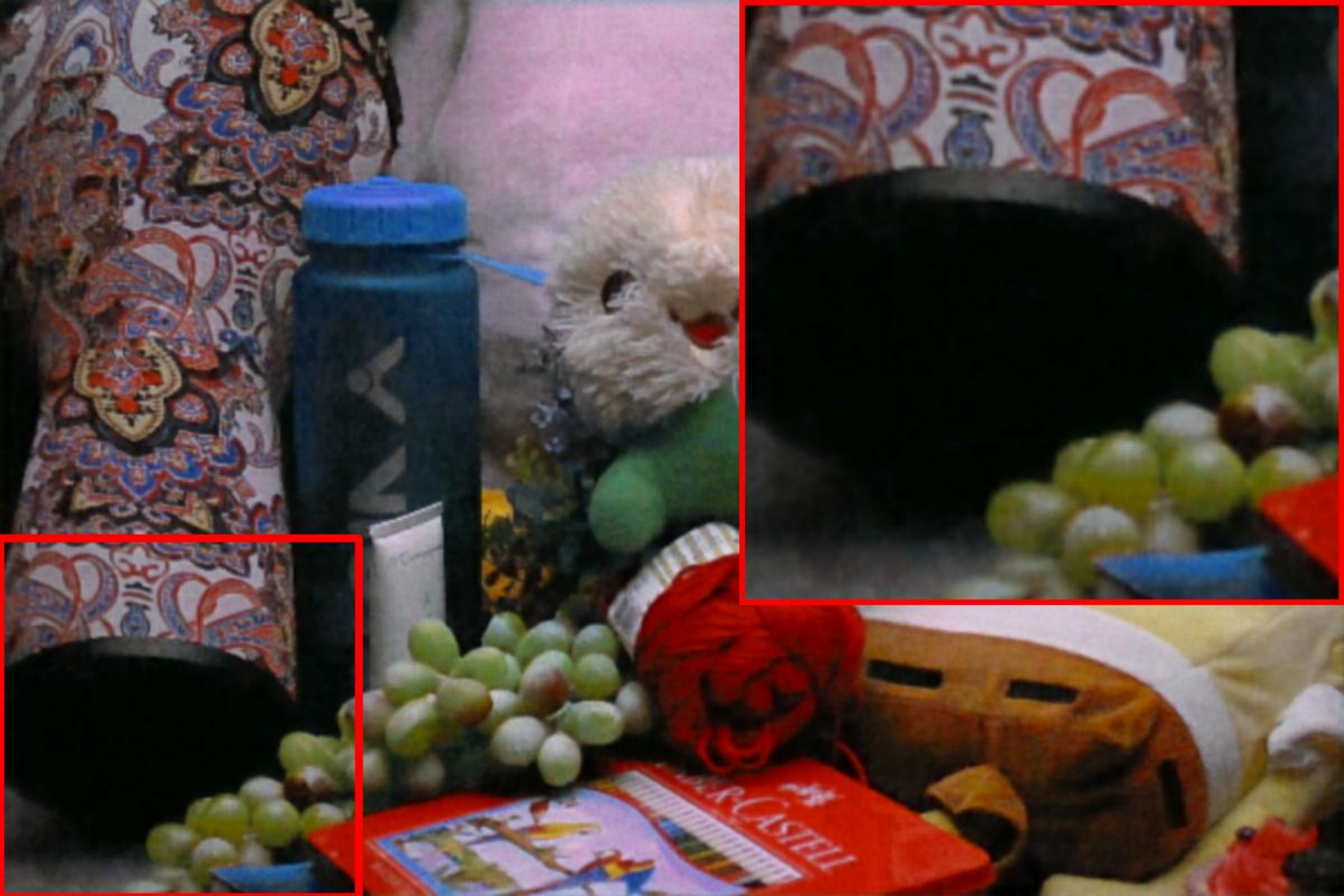}%
    }\hspace{0.5mm}%
    \subfloat[\textbf{Ours}]{%
        \includegraphics[width=0.195\linewidth]{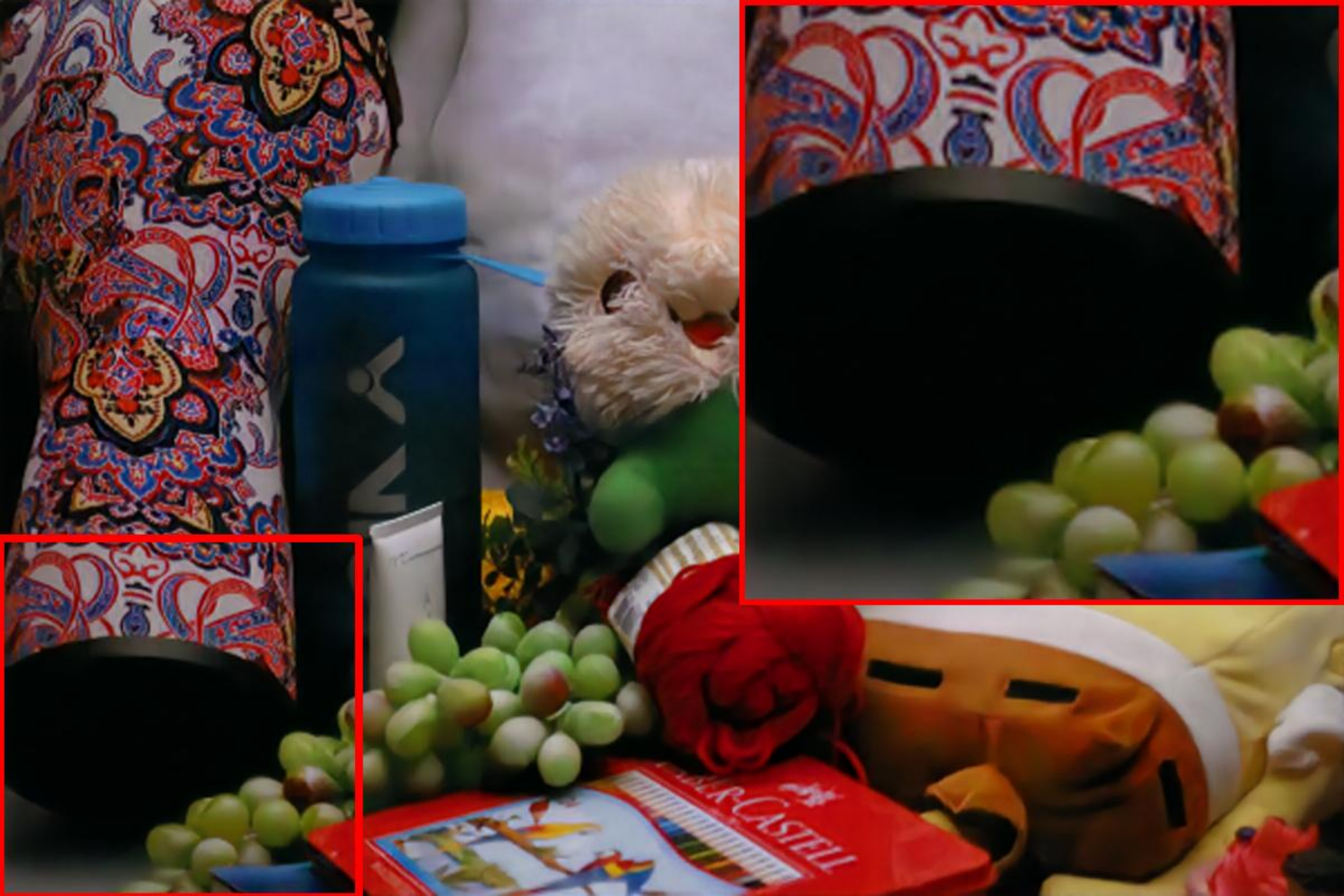}%
    }
    \vspace{-0.2cm}
    \caption{Visual comparison with state-of-the-art methods on the LOL-v1 dataset.}
    \vspace{-0.4cm}
    \label{fig:visual_compare_V1}
\end{figure*}

\begin{figure*}[!t]
    \centering
    \setlength{\fboxsep}{0pt} 
    \subfloat[Input]{%
        \includegraphics[width=0.195\linewidth]{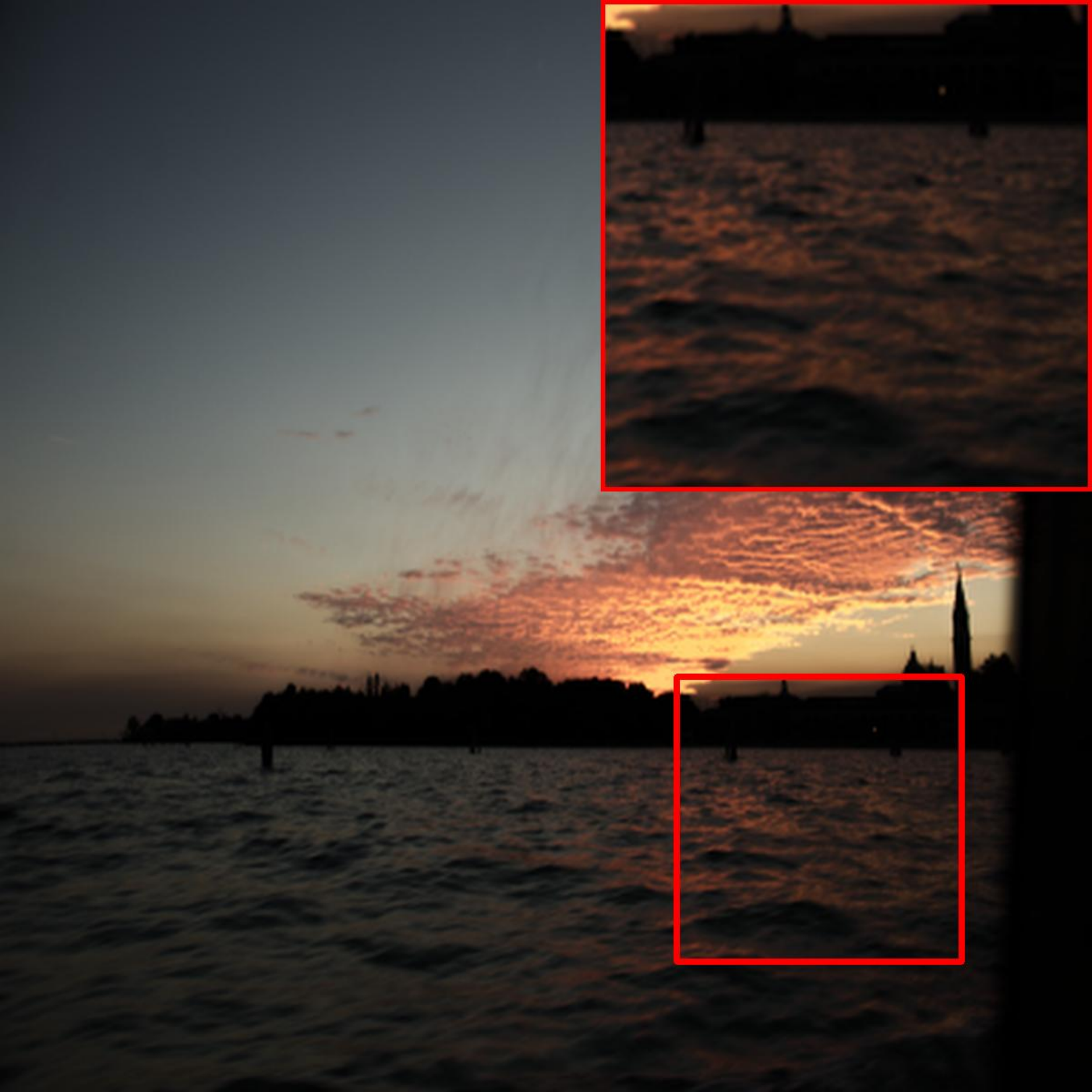}%
    }\hspace{0.5mm}%
    \subfloat[Reference]{%
        \includegraphics[width=0.195\linewidth]{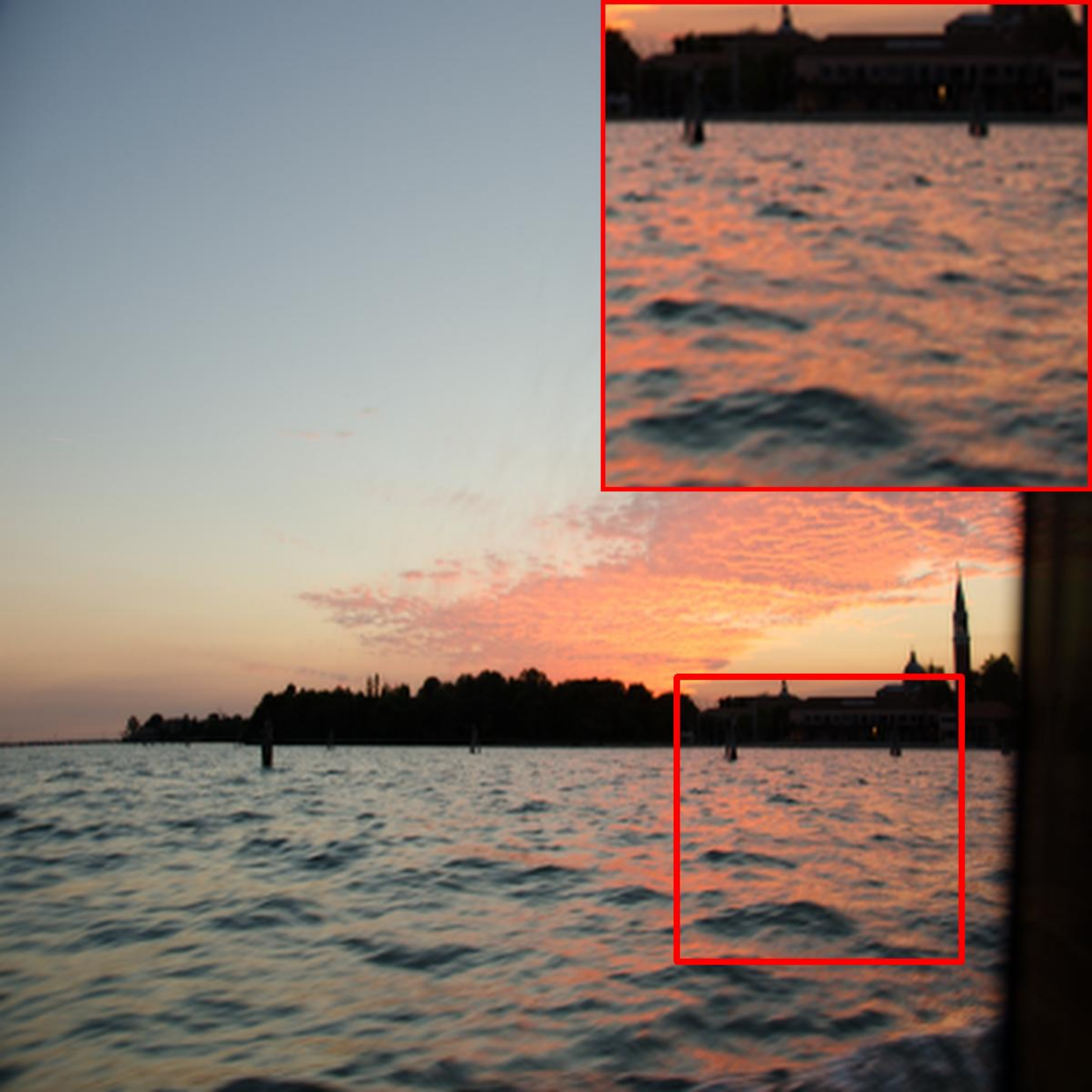}%
    }\hspace{0.5mm}%
    \subfloat[FourLLIE]{%
        \includegraphics[width=0.195\linewidth]{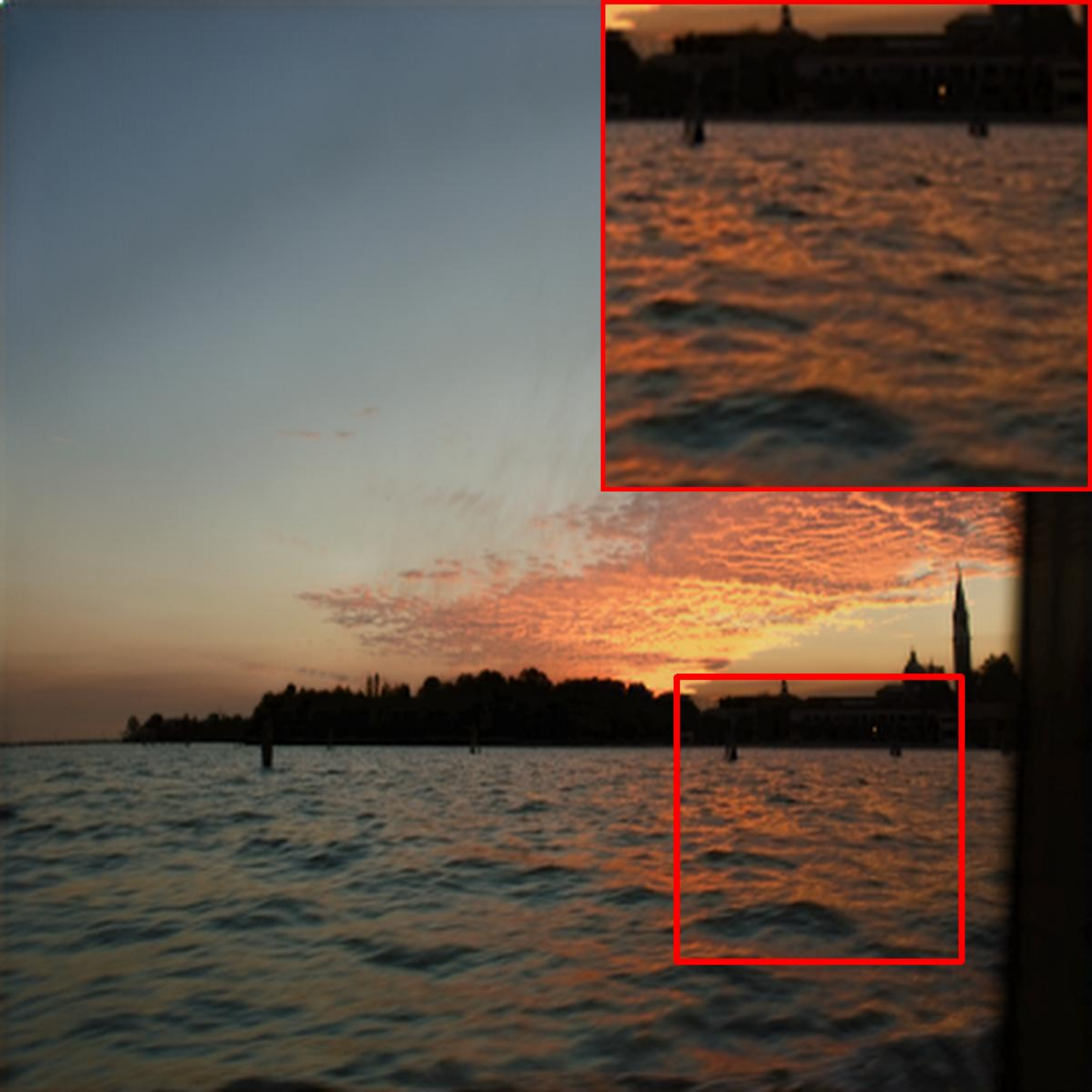}%
    }\hspace{0.5mm}%
    \subfloat[DMFourLLIE]{%
        \includegraphics[width=0.195\linewidth]{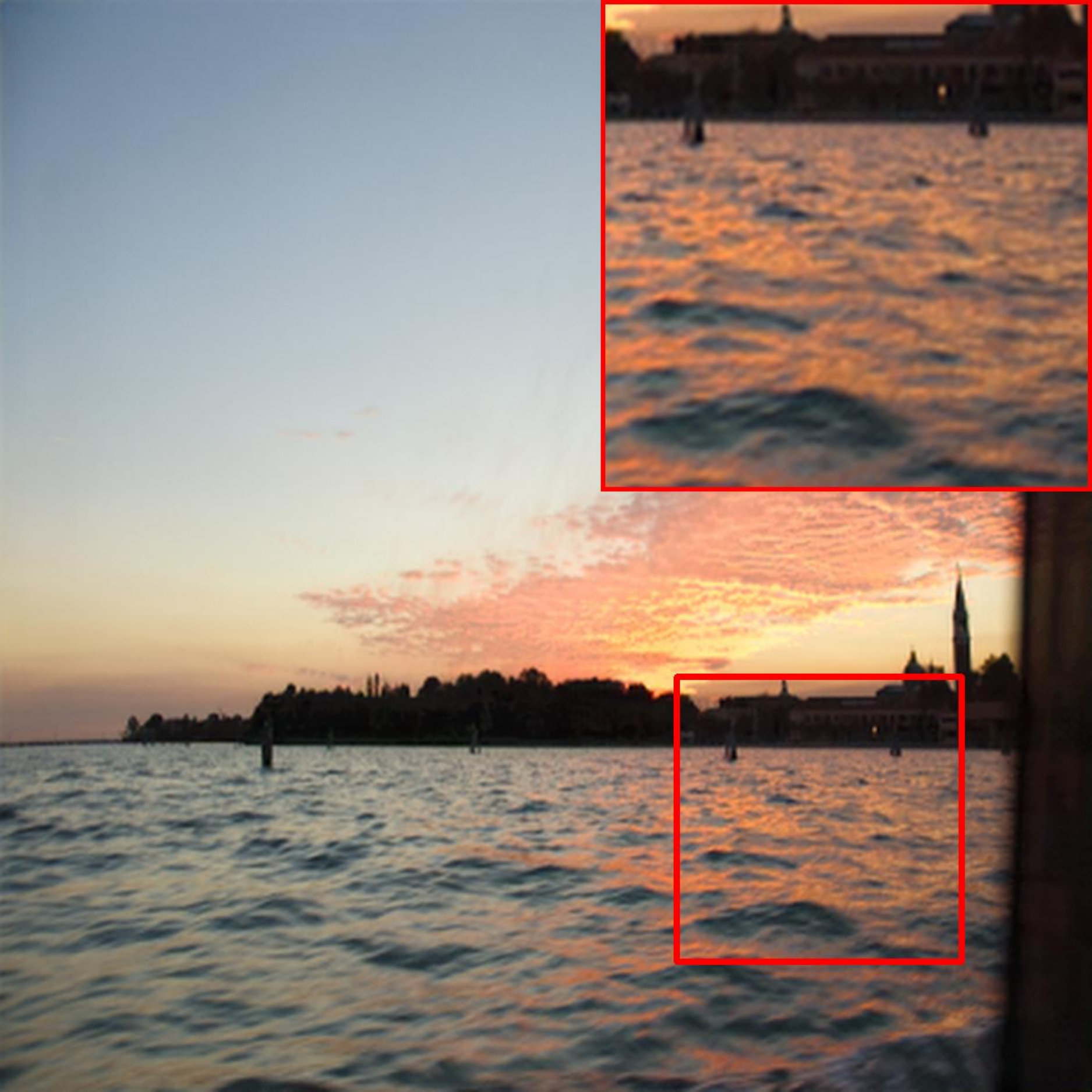}%
    }\hspace{0.5mm}%
    \subfloat[CWNet]{%
        \includegraphics[width=0.195\linewidth]{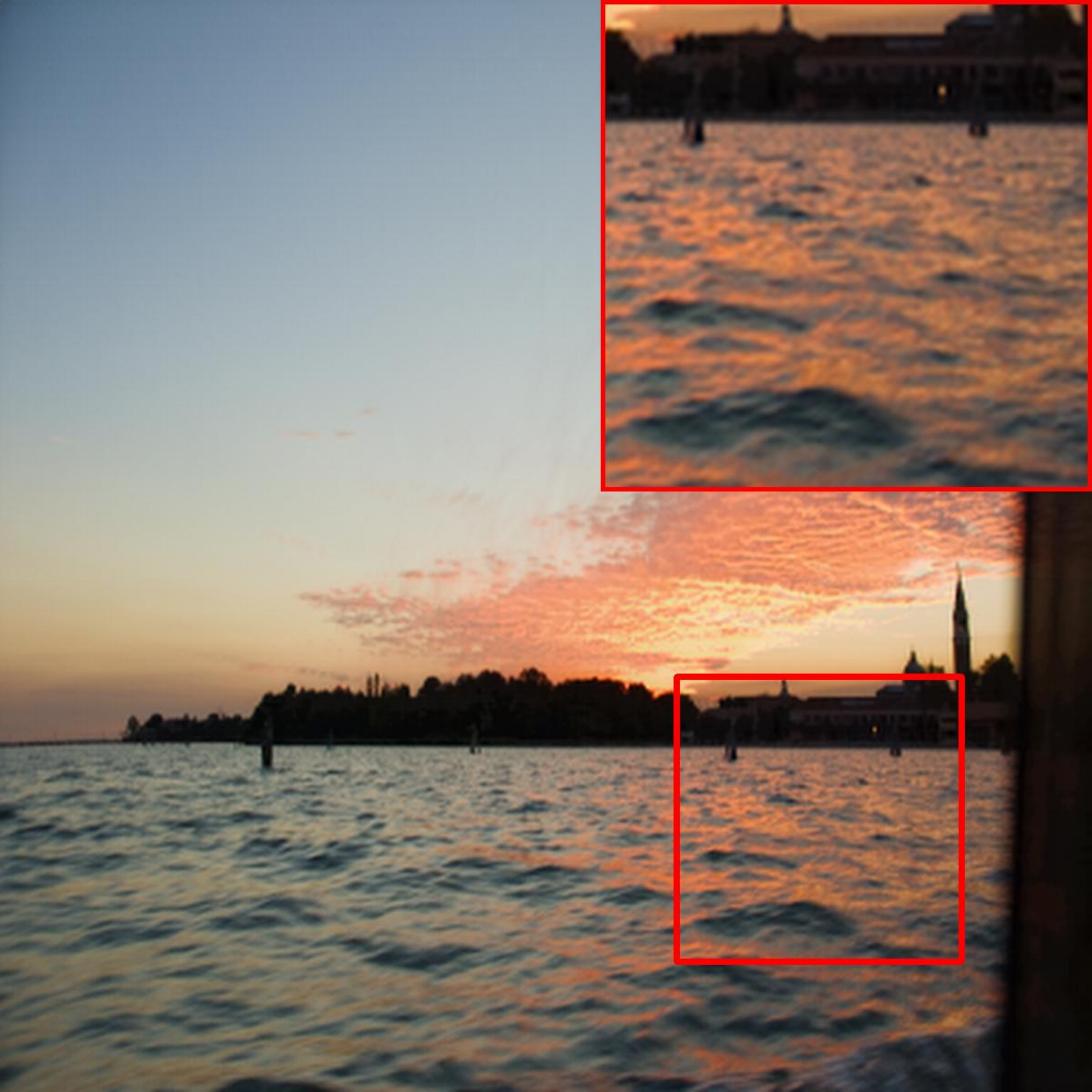}%
    }
    \vspace{-0.01mm}
    \subfloat[CIDNet]{%
        \includegraphics[width=0.195\linewidth]{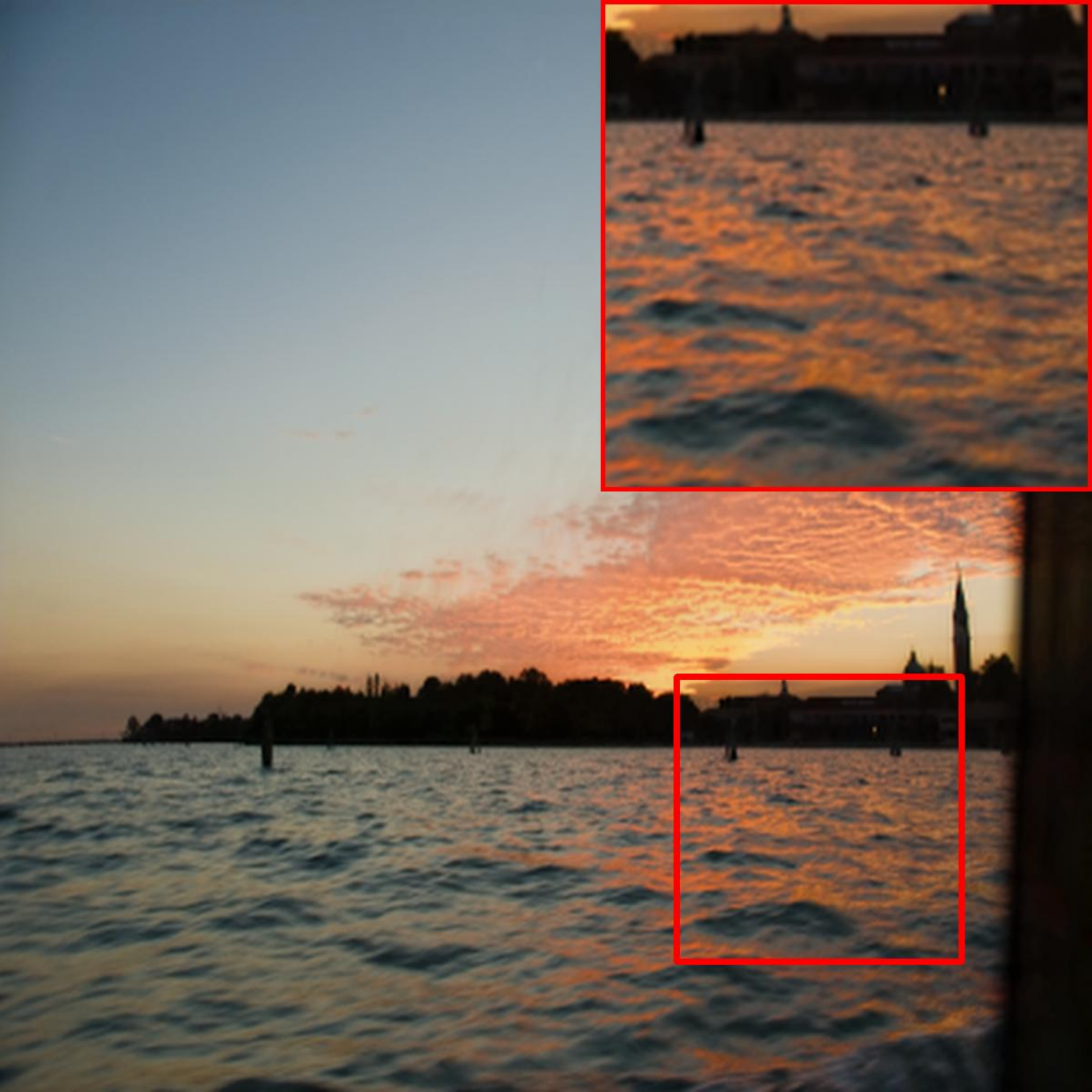}%
    }\hspace{0.5mm}%
    \subfloat[UHDFour]{%
        \includegraphics[width=0.195\linewidth]{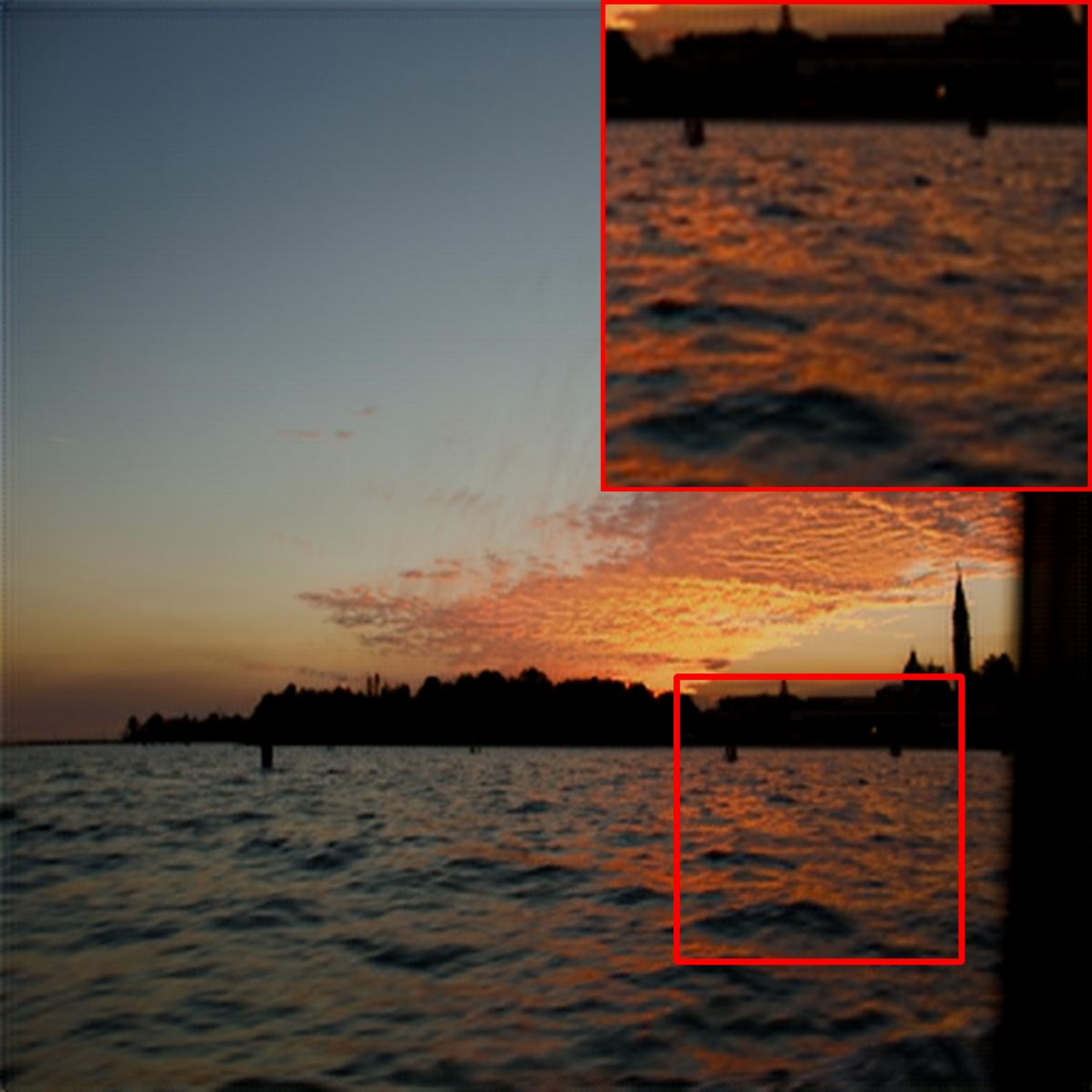}%
    }\hspace{0.5mm}%
    \subfloat[Wave-Mamba]{%
        \includegraphics[width=0.195\linewidth]{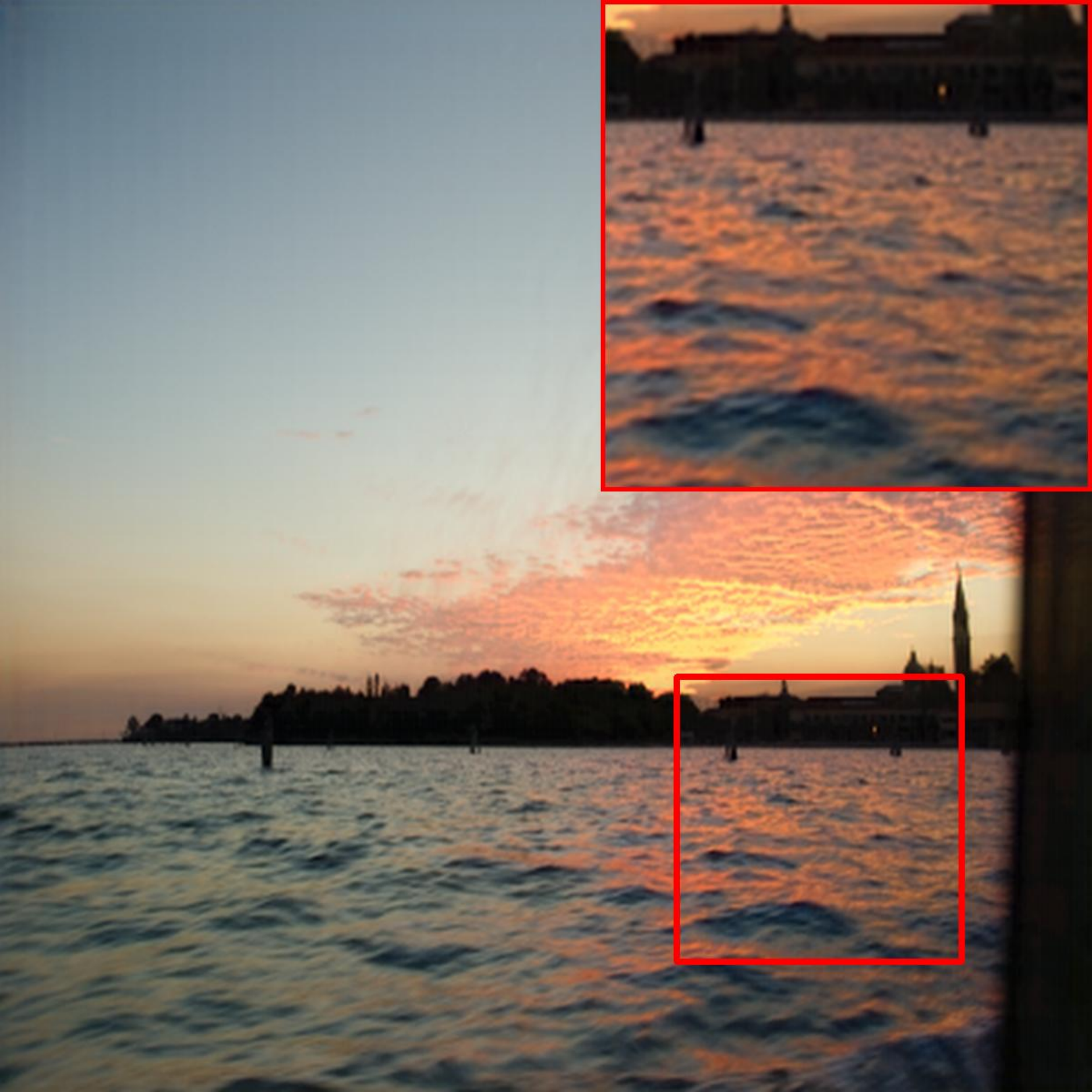}%
    }\hspace{0.5mm}%
    \subfloat[Retinexformer]{%
        \includegraphics[width=0.195\linewidth]{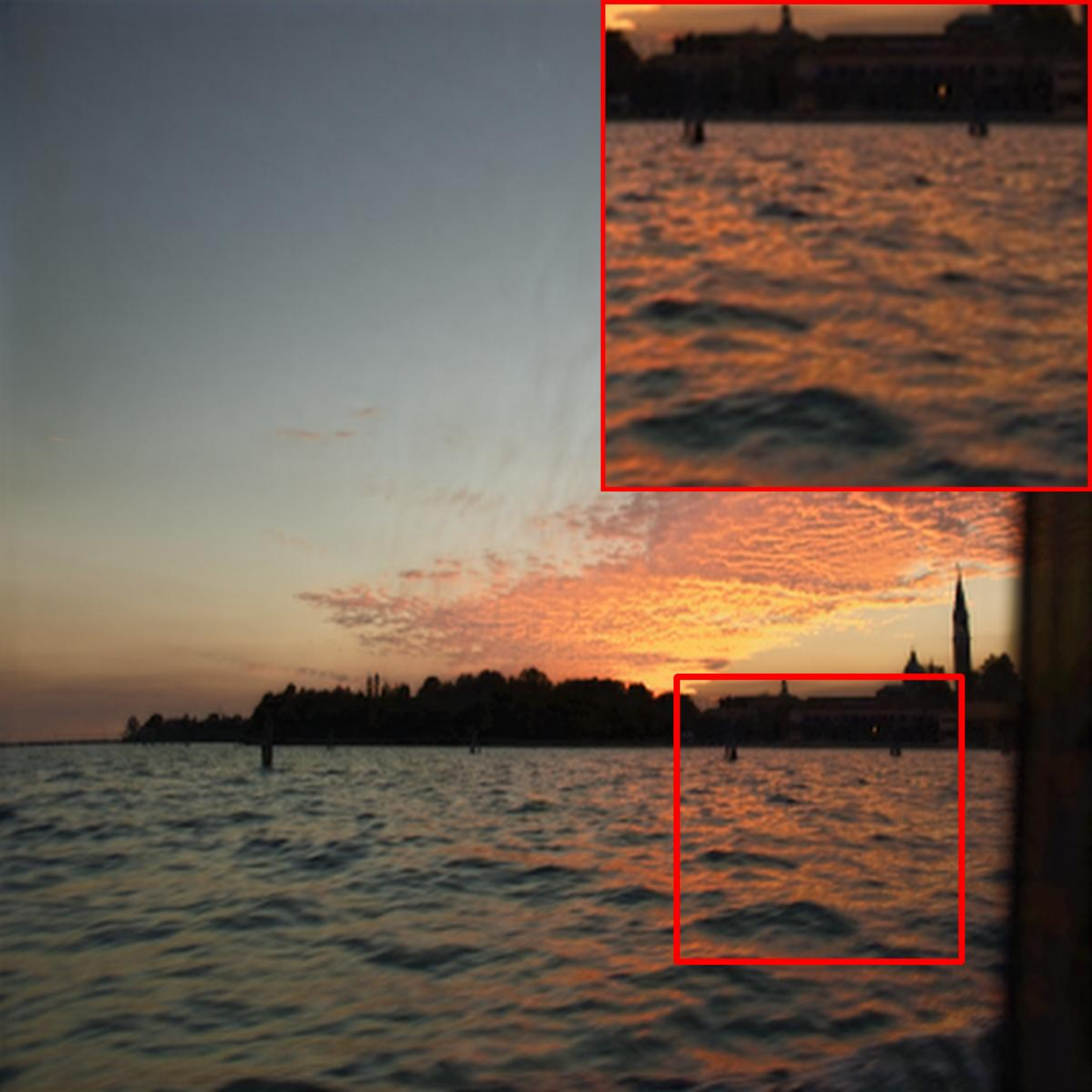}%
    }\hspace{0.5mm}%
    \subfloat[\textbf{Ours}]{%
        \includegraphics[width=0.195\linewidth]{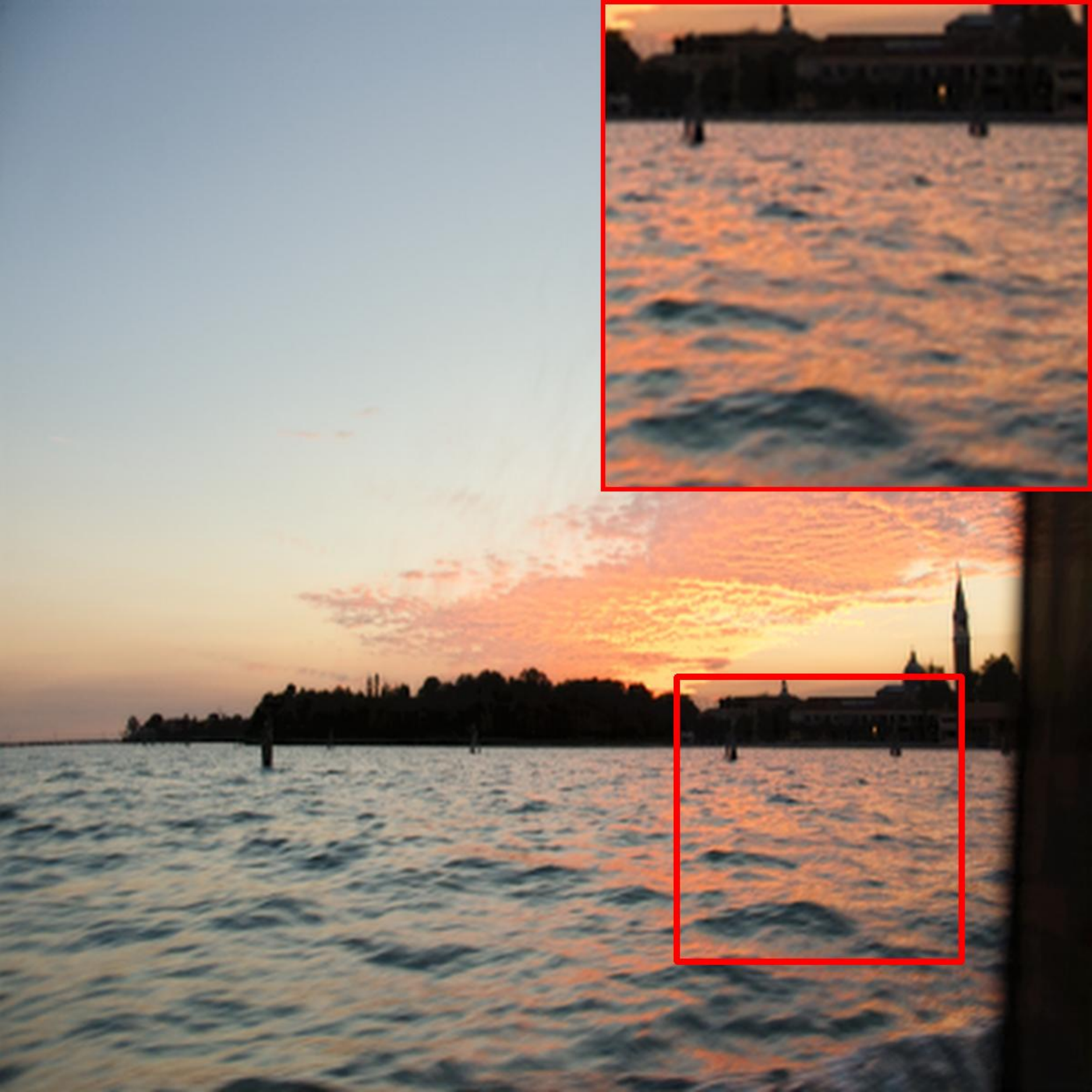}%
    }
    \vspace{-0.2cm}
    \caption{Visual comparison with state-of-the-art methods on the LOL-v2-Syn dataset.}
    \vspace{-0.3cm}
    \label{fig:visual_compare_Syn}
\end{figure*}

\subsection{Stage II: Dual-Modal Guided Refinement Network}

Stage II focuses on recovering fine-grained textures and residual high-frequency details from the globally rectified result. 
Although the endogenous prior feature $F_{en}$ provides informative structural guidance, directly injecting it into the restoration backbone may introduce feature redundancy or misalignment. 
To address this, we design the Dual-Modal Guided Refinement Network (DGRN), which consists of a Prior-Injected Guided Encoder (PIGE) and a Prior-Enhanced Spatial-Frequency Block (PESFB).

We first construct the backbone feature $F_{back}$ by concatenating the original input $I_{in}$ and the reconstructed Stage~I result $I_{s1}$, followed by a convolutional projection:
\begin{equation}
F_{back} = \mathrm{Conv}([I_{in}, I_{s1}]),
\end{equation}
where $[\cdot,\cdot]$ denotes channel-wise concatenation.

\noindent \textbf{PIGE.}
As shown in Fig.~\ref{fig:pipeline}, PIGE serves as the entry module for prior-guided refinement. 
Instead of directly concatenating prior and backbone features, we treat the endogenous prior as a query to retrieve prior-relevant information from the backbone. 
Specifically, given $F_{en}$ and $F_{back}$, we project them into query, key, and value embeddings:
\begin{equation}
Q = W_q(F_{en}), \quad K = W_k(F_{back}), \quad V = W_v(F_{back}),
\end{equation}
where $W_q$, $W_k$, and $W_v$ are learnable linear projections. 
The cross-attention map is computed as
\begin{equation}
M = \mathrm{Softmax}\left(\frac{QK^{\top}}{\sqrt{d}}\right),
\end{equation}
where $d$ is the channel dimension used for normalization. 
The retrieved context feature is then obtained by
\begin{equation}
F_{ctx} = \mathrm{Conv}(MV).
\end{equation}

To prevent excessive prior injection from overwhelming the backbone representation, we further introduce an adaptive gated fusion mechanism:
\begin{equation}
\begin{aligned}
\alpha &= \sigma\big(\mathrm{Conv}([F_{ctx}, F_{back}])\big), \\
F_{back}^{'} &= F_{back} + \alpha \odot F_{ctx},
\end{aligned}
\end{equation}
where $\sigma(\cdot)$ denotes the Sigmoid function and $\odot$ denotes element-wise multiplication. 
In this way, PIGE selectively injects prior-guided context into the backbone, enhancing informative regions while suppressing redundant responses.

\noindent \textbf{PESFB.}
After prior injection, we further refine features using PESFB, which focuses on localized spatial-frequency restoration complementary to the global Fourier rectification in Stage~I.

Multi-scale modeling has been widely adopted in image restoration to capture structures at different spatial extents, with representative methods such as CGD-TraP~\cite{liu2025condition}, MB-TaylorFormer v2~\cite{jin2025mb}, ESTINet~\cite{zhang2022enhanced}, DDMSNet~\cite{zhang2021deep}, and DBLRNet~\cite{zhang2018adversarial} demonstrating the effectiveness of coarse-to-fine or multi-resolution interactions.
Inspired by these advances, we incorporate a multi-scale convolution branch into PESFB. 
However, rather than serving as the primary reconstruction engine, PESFB is designed as a lightweight residual refinement module that explicitly operates under the guidance of our ERP.
As illustrated in Fig.~\ref{fig:pipeline}, it consists of three parallel branches:

\noindent \textit{Wavelet branch.} 
We employ the discrete wavelet transform (DWT)~\cite{jiang2023low} to decompose $F_{back}^{'}$ into one low-frequency sub-band and three high-frequency sub-bands:
\begin{equation}
\{X_{LL}, X_{LH}, X_{HL}, X_{HH}\} = \mathrm{DWT}(F_{back}^{'}).
\end{equation}
Each sub-band is refined by a dedicated convolution:
\begin{equation}
\widetilde{X}_{b} = \mathrm{Conv}(X_b), \quad b \in \{LL, LH, HL, HH\},
\end{equation}
and then reconstructed through the inverse discrete wavelet transform:
\begin{equation}
F_{wave} = \mathrm{IDWT}(\widetilde{X}_{LL}, \widetilde{X}_{LH}, \widetilde{X}_{HL}, \widetilde{X}_{HH}).
\end{equation}
This branch explicitly enhances localized edge and texture details in the wavelet domain.

\noindent \textit{Multi-scale convolution branch.} 
To capture structures of different spatial extents, this branch applies parallel convolutions with different kernel sizes:
\begin{equation}
\begin{aligned}
F_{scale} = \mathrm{Concat}\big(
\mathrm{Conv}_{1\times1}(F_{back}^{'}),
\mathrm{Conv}_{3\times3}(F_{back}^{'}), \\
\mathrm{Conv}_{5\times5}(F_{back}^{'})
\big).
\end{aligned}
\end{equation}
This design enlarges the receptive field diversity and improves the modeling of objects under non-uniform low-light conditions. 
Crucially, this branch is conditioned on the reliable structural cues from ERP to recover fine textures that are unreliable in the raw low-light input, rather than independently reconstructing the image.

\noindent \textit{Local-context branch.} 
In parallel, the local-context branch preserves spatial continuity and local identity information through residual modeling:
\begin{equation}
F_{local} = \mathrm{ResBlock}_2(\mathrm{ResBlock}_1(F_{back}^{'})).
\end{equation}

The outputs of the three branches are aggregated to obtain the refined feature:
\begin{equation}
F_{refine} = \mathrm{Conv}([F_{wave}, F_{scale}, F_{local}]).
\end{equation}
This prior-conditioned complementary design distinguishes PESFB from generic multi-scale restoration blocks and from the simple spatial compensation branches commonly used in previous frequency-domain LLIE methods~\cite{Huang2022DeepFE, Wang2023FourLLIEBL, Zhang2024DMFourLLIEDA, li2023embedding}. 
By jointly combining prior-guided attention, localized wavelet refinement, and multi-scale spatial modeling, Stage~II effectively complements the global restoration performed in Stage~I and produces features well-suited for final image reconstruction.

\begin{figure*}[!t]
    \centering
    \setlength{\fboxsep}{0pt} 
    \subfloat[Input]{%
        \includegraphics[width=0.195\linewidth]{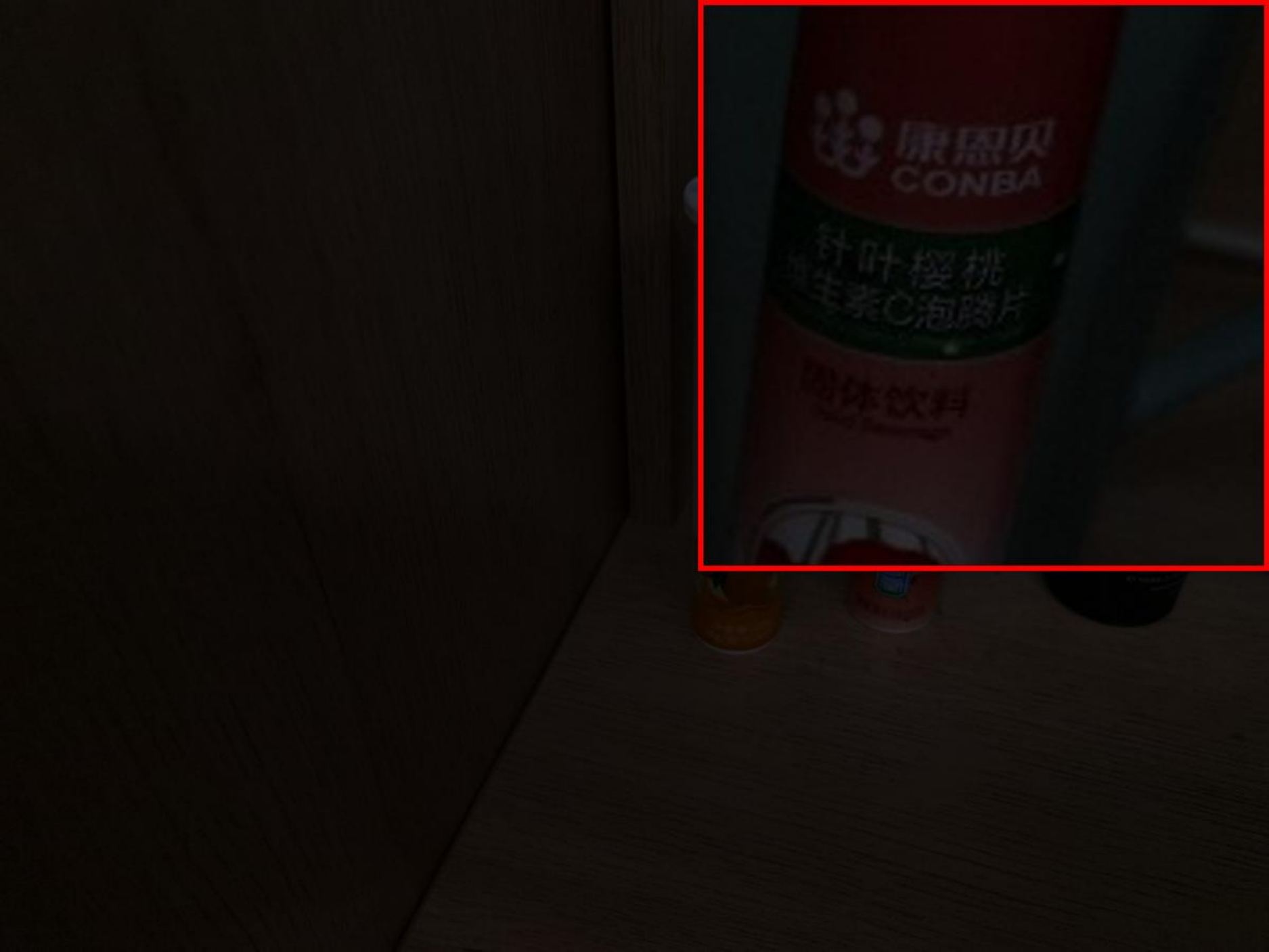}%
    }\hspace{0.5mm}%
    \subfloat[Reference]{%
        \includegraphics[width=0.195\linewidth]{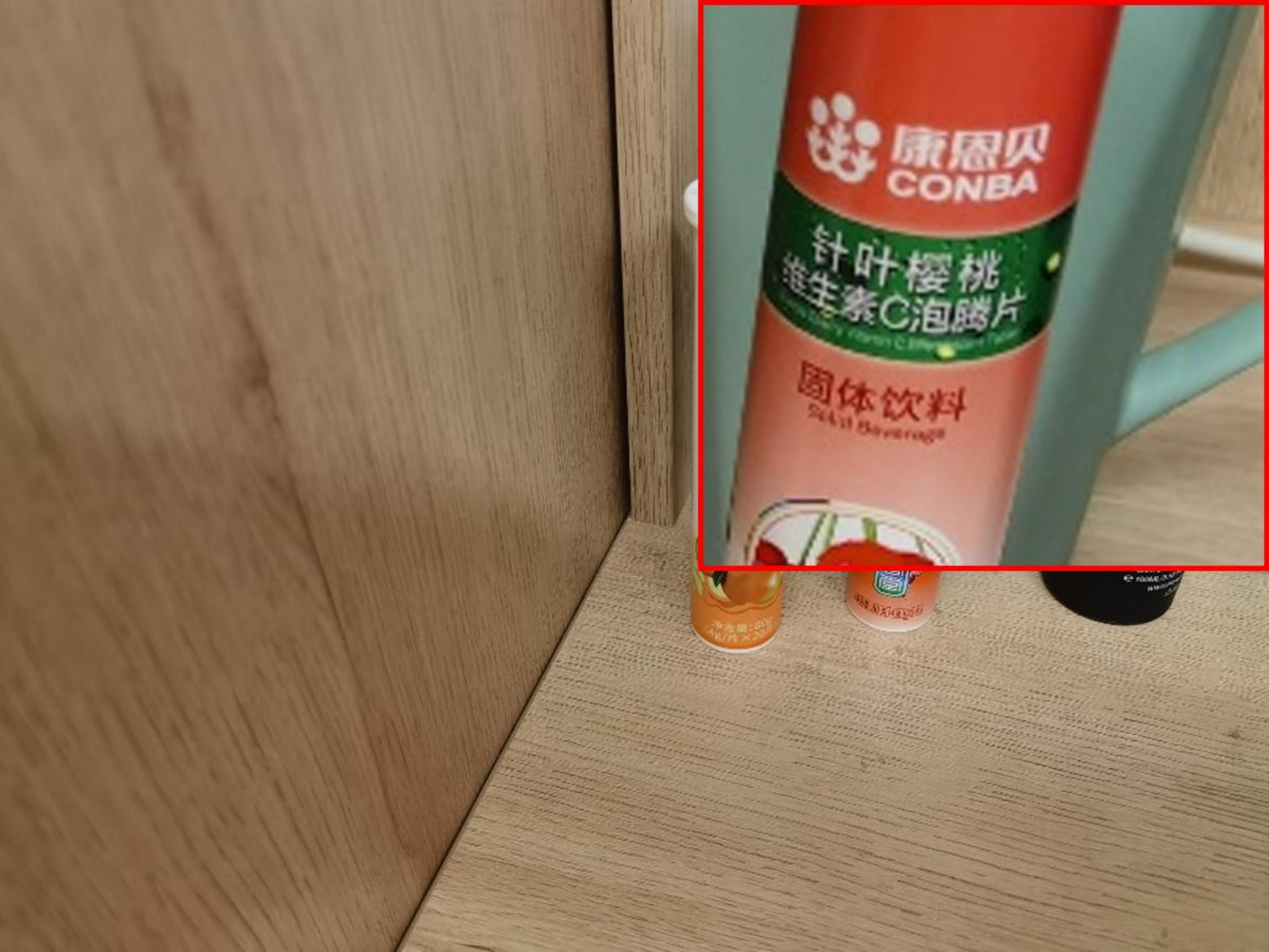}%
    }\hspace{0.5mm}%
    \subfloat[FourLLIE]{%
        \includegraphics[width=0.195\linewidth]{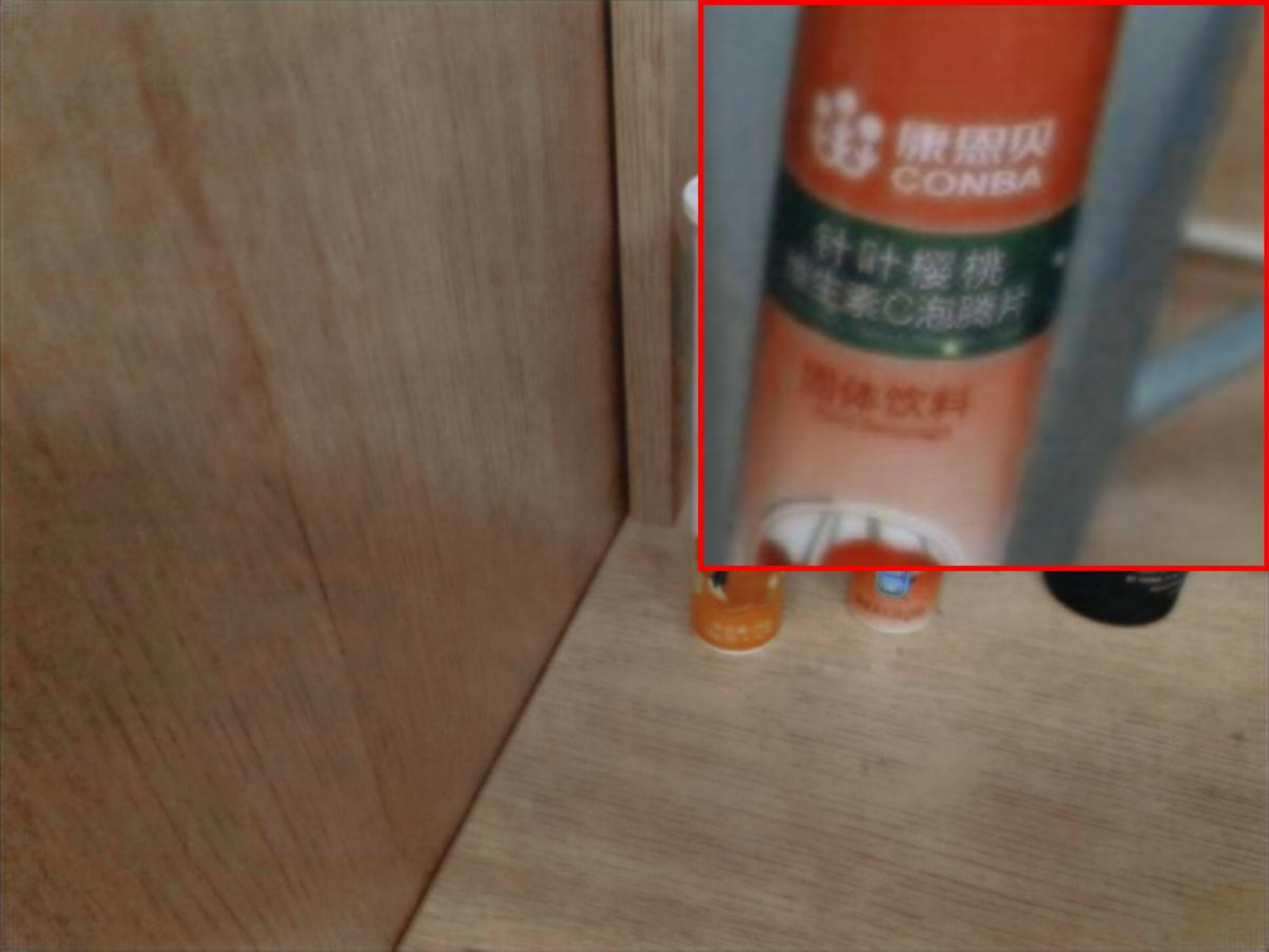}%
    }\hspace{0.5mm}%
    \subfloat[DMFourLLIE]{%
        \includegraphics[width=0.195\linewidth]{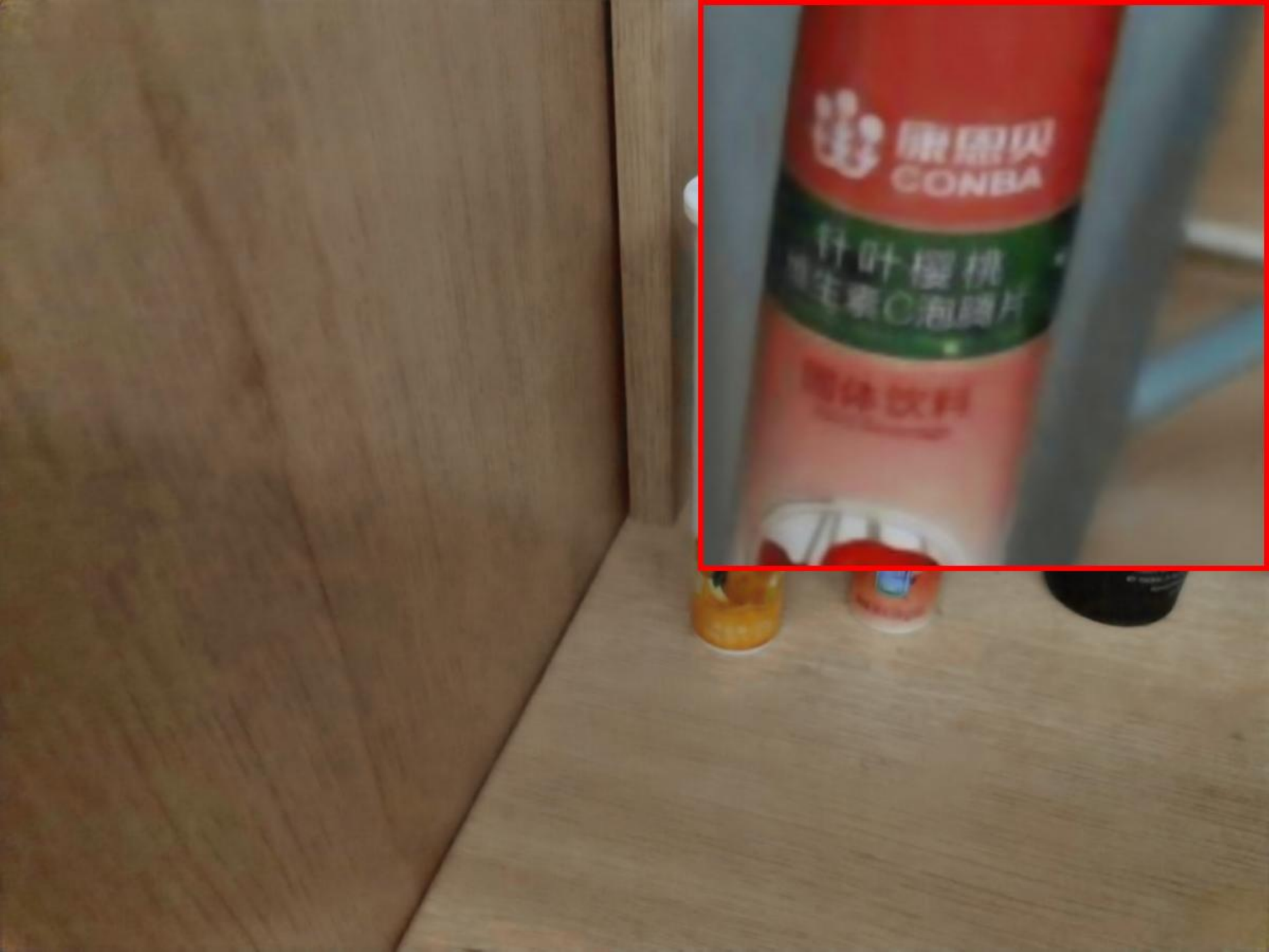}%
    }\hspace{0.5mm}%
    \subfloat[CWNet]{%
        \includegraphics[width=0.195\linewidth]{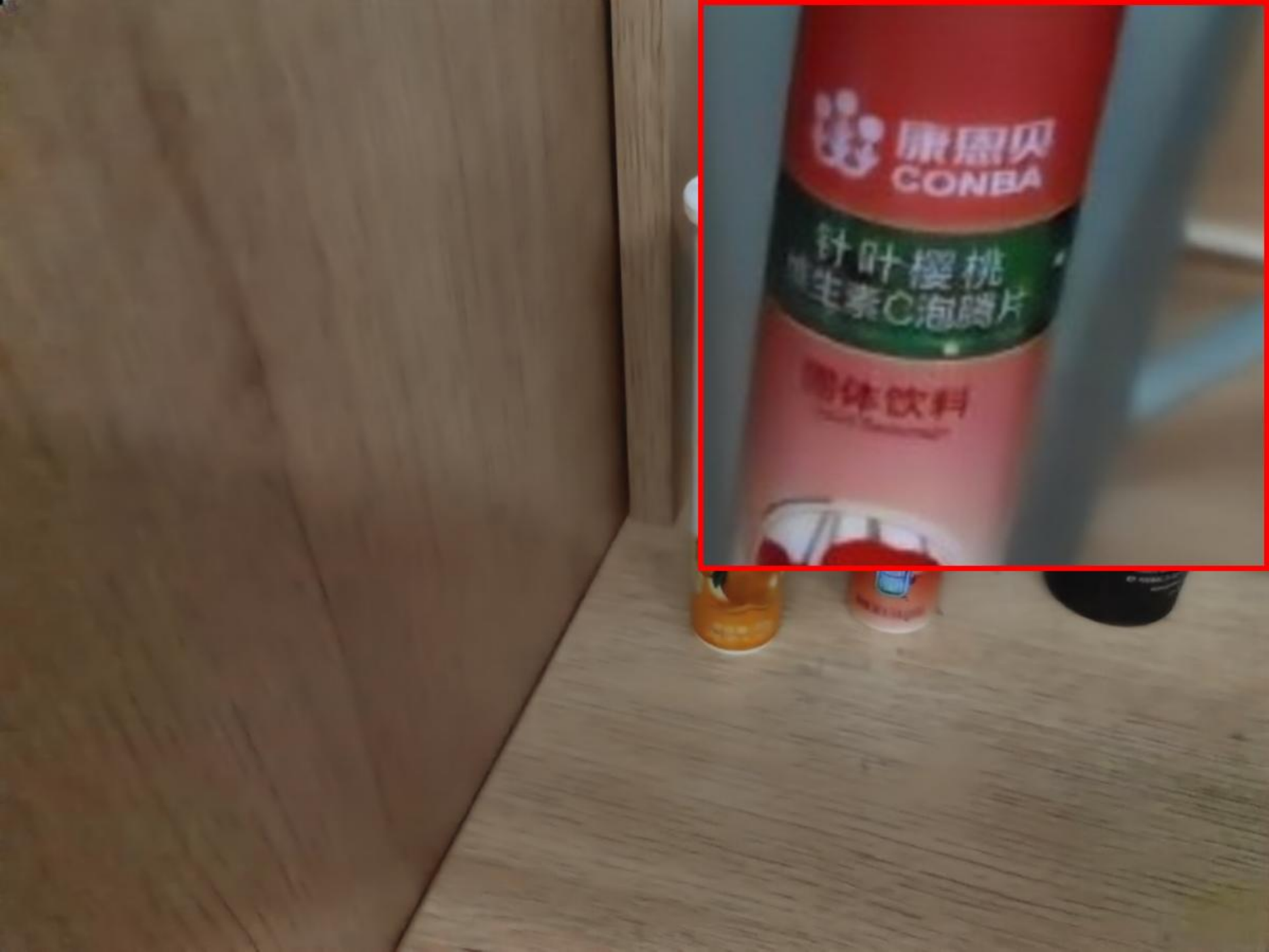}%
    }
    \vspace{-3mm}
    \subfloat[CIDNet]{%
        \includegraphics[width=0.195\linewidth]{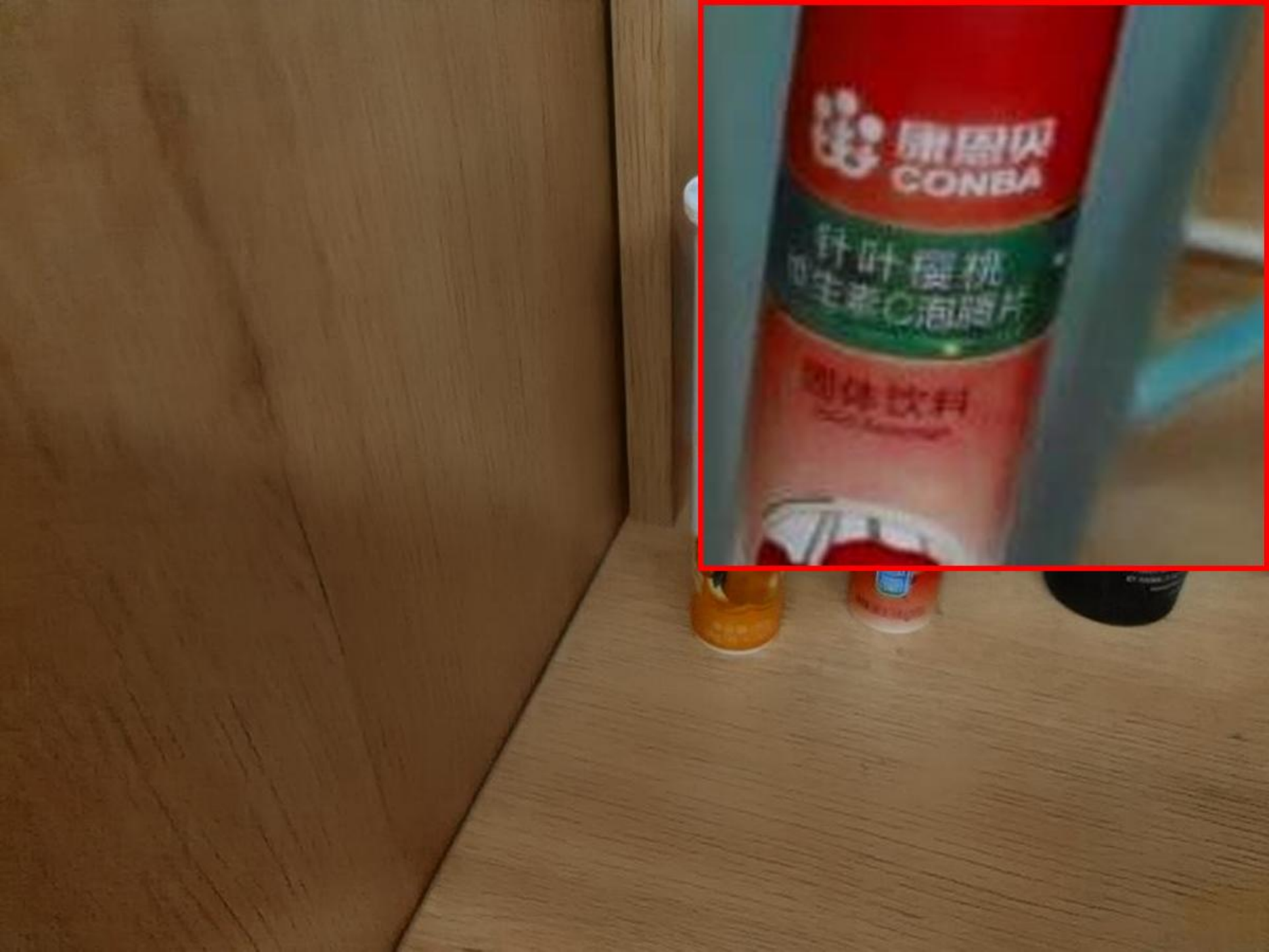}%
    }\hspace{0.5mm}%
    \subfloat[UHDFour]{%
        \includegraphics[width=0.195\linewidth]{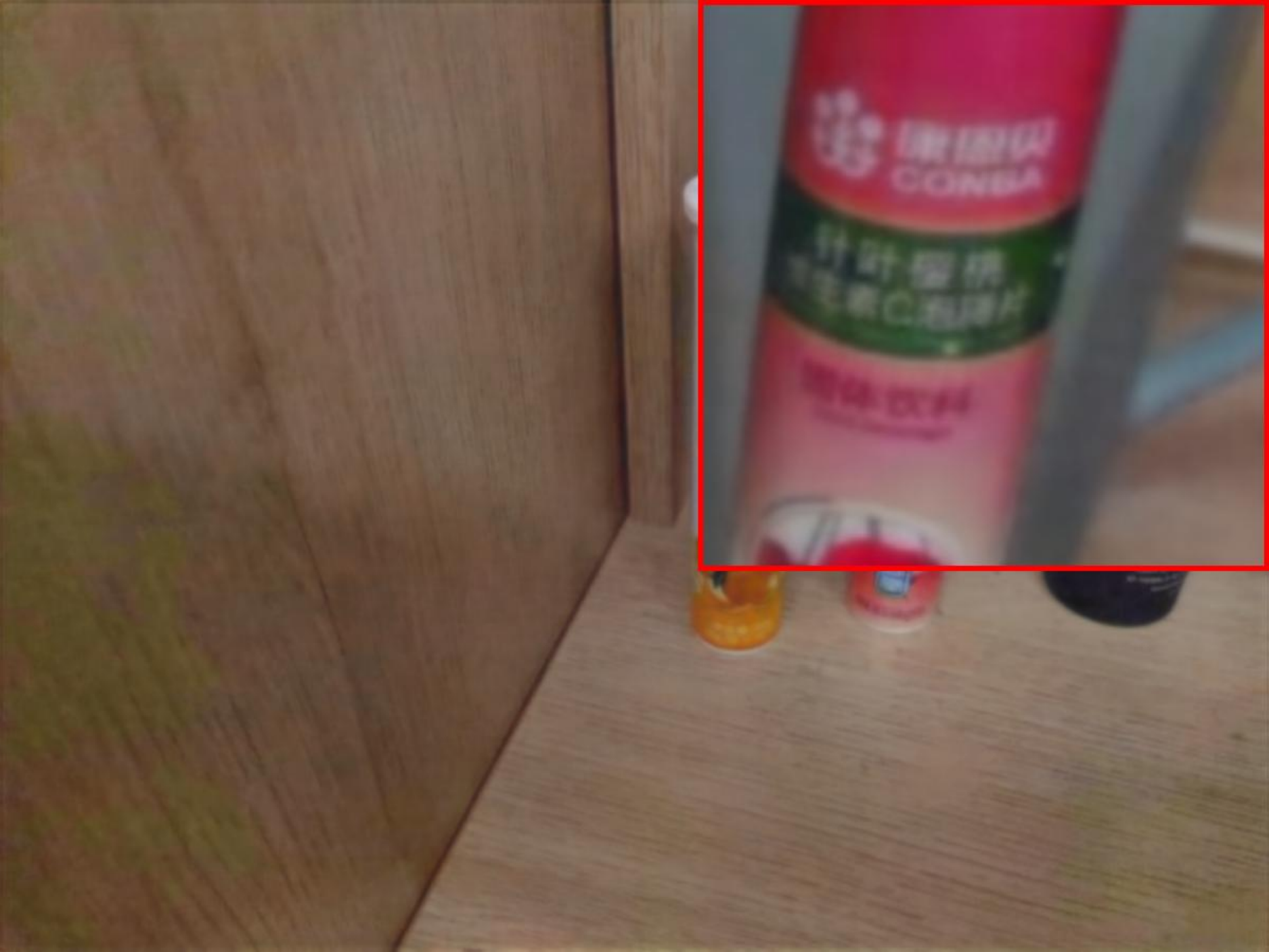}%
    }\hspace{0.5mm}%
    \subfloat[Wave-Mamba]{%
        \includegraphics[width=0.195\linewidth]{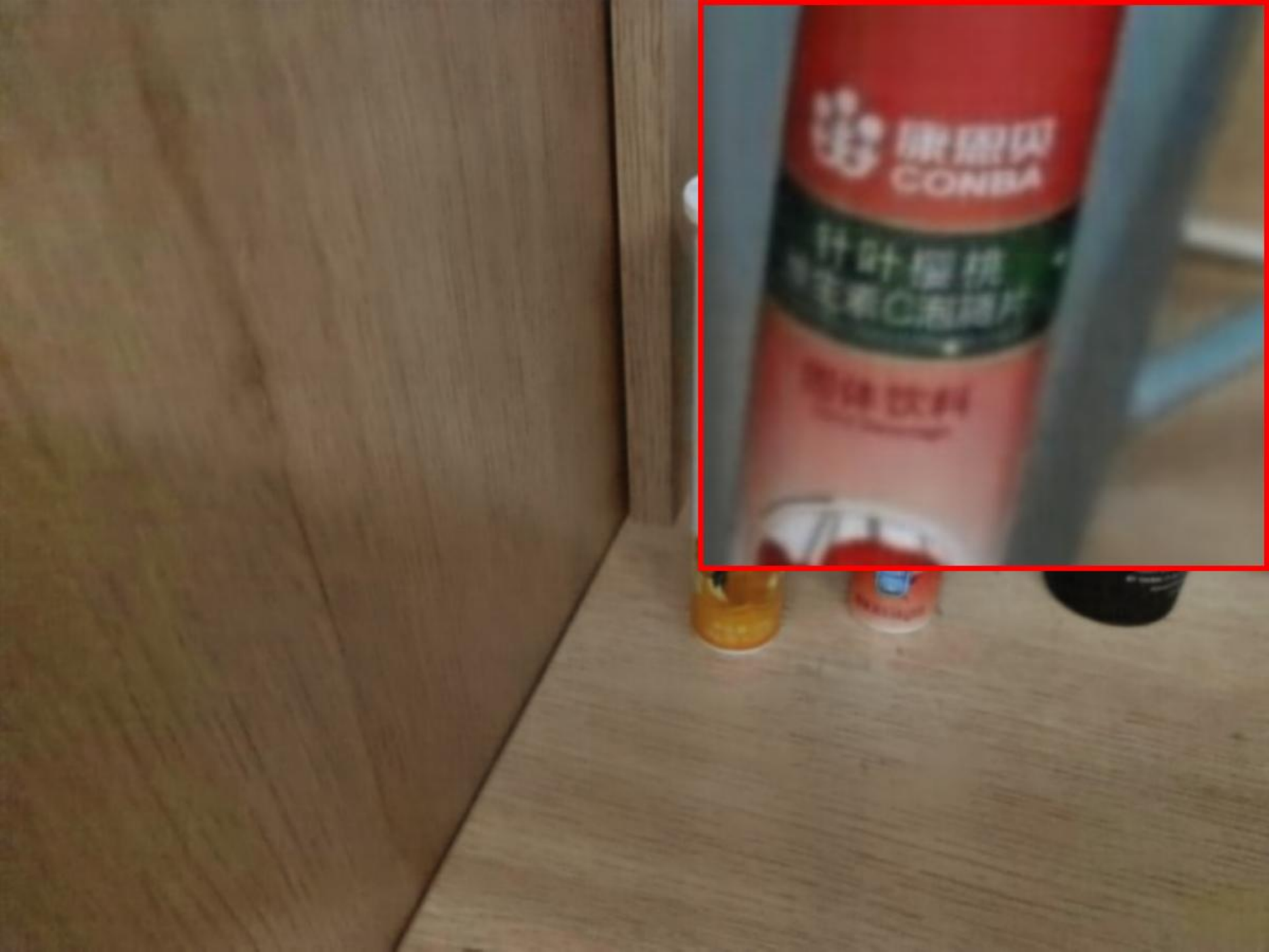}%
    }\hspace{0.5mm}%
    \subfloat[Retinexformer]{%
        \includegraphics[width=0.195\linewidth]{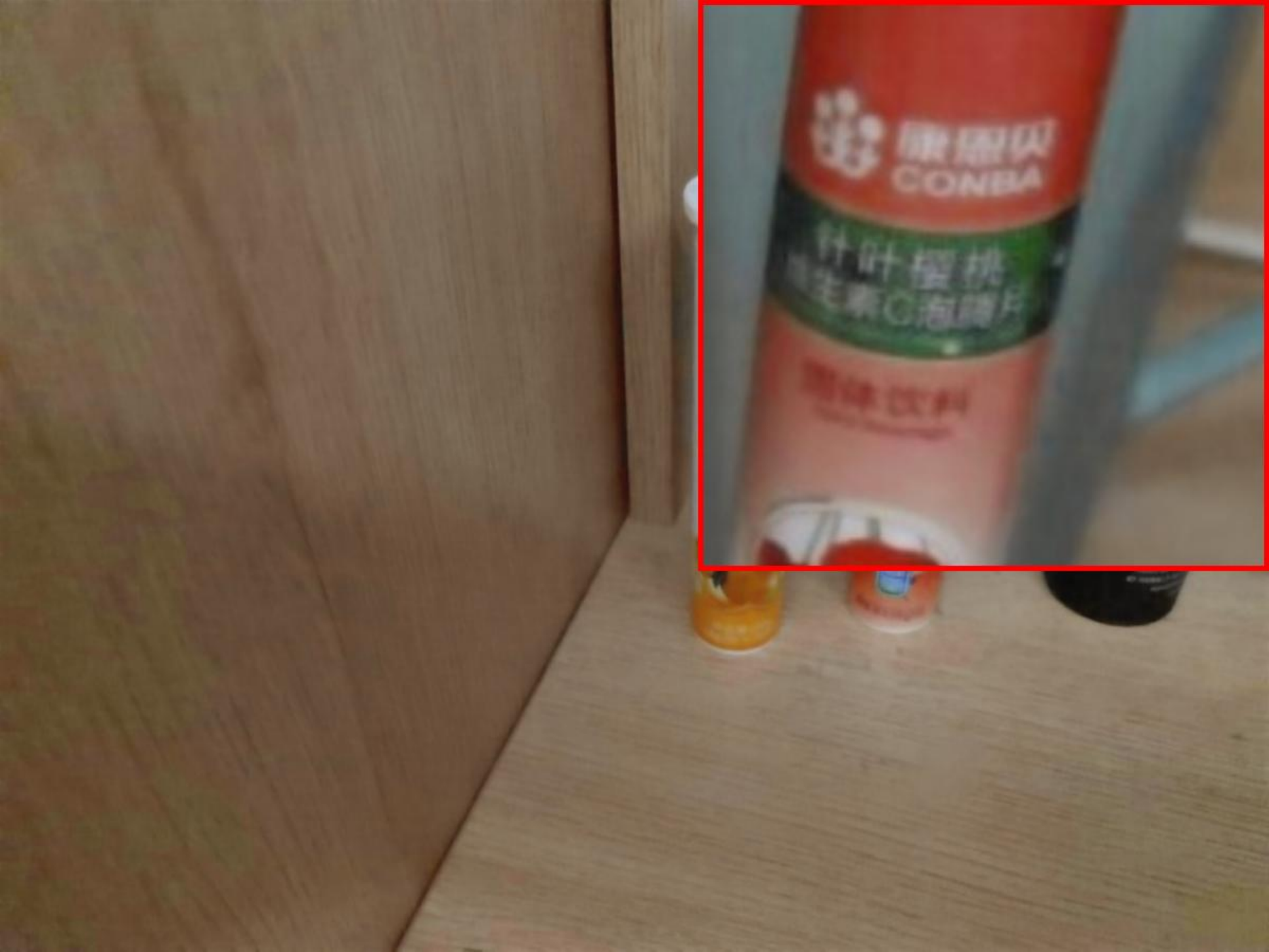}%
    }\hspace{0.5mm}%
    \subfloat[\textbf{Ours}]{%
        \includegraphics[width=0.195\linewidth]{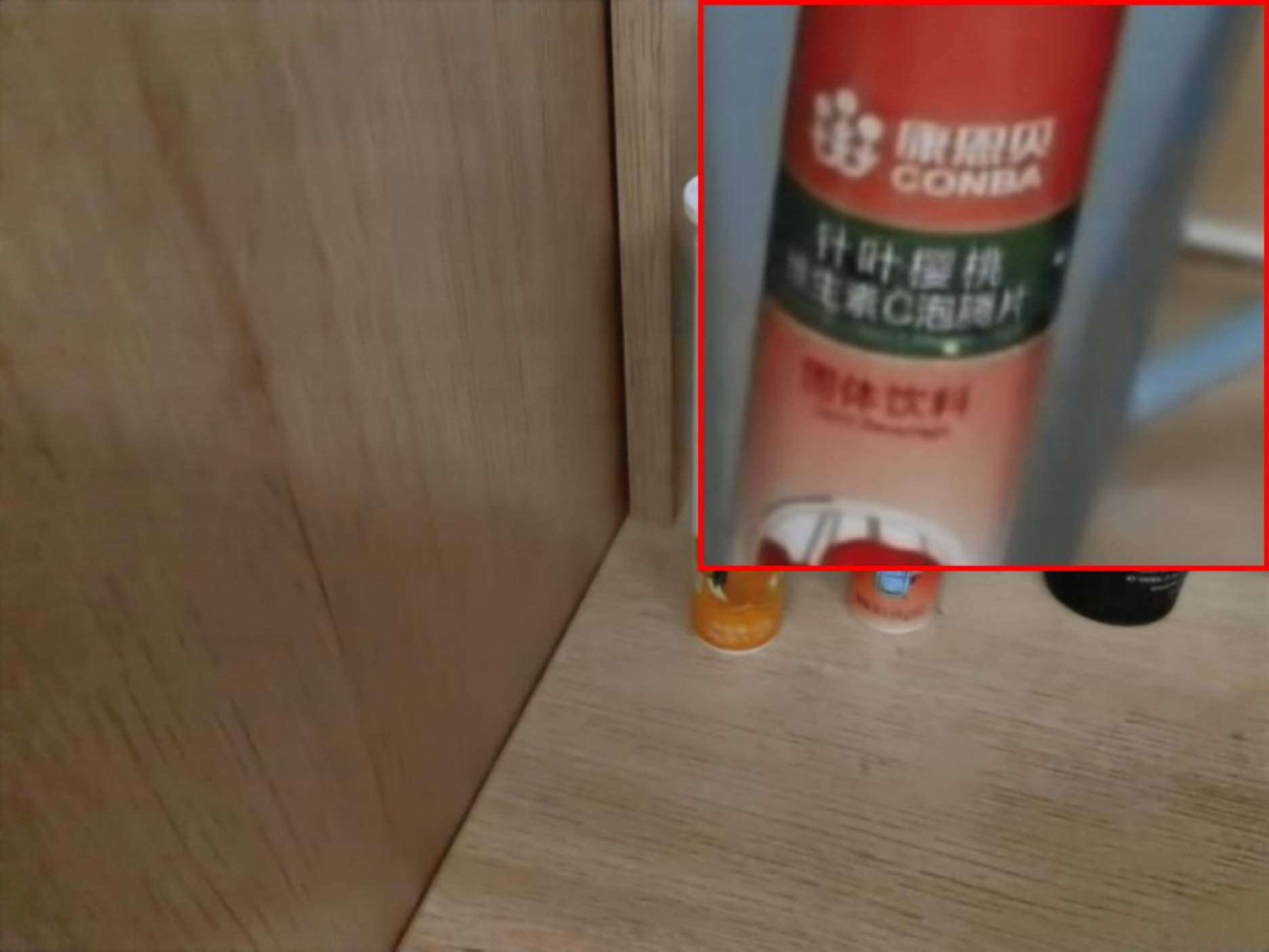}%
    }
    \vspace{-0.2cm}
    \caption{Visual comparison with state-of-the-art methods on the LSRW-Huawei dataset.}
    \vspace{-0.3cm}
    \label{fig:visual_compare_Huawei}
\end{figure*}

\begin{figure*}[t]
    \centering

    \setlength{\fboxsep}{0pt} 

    \subfloat[Input]{%
        \includegraphics[width=0.195\linewidth]{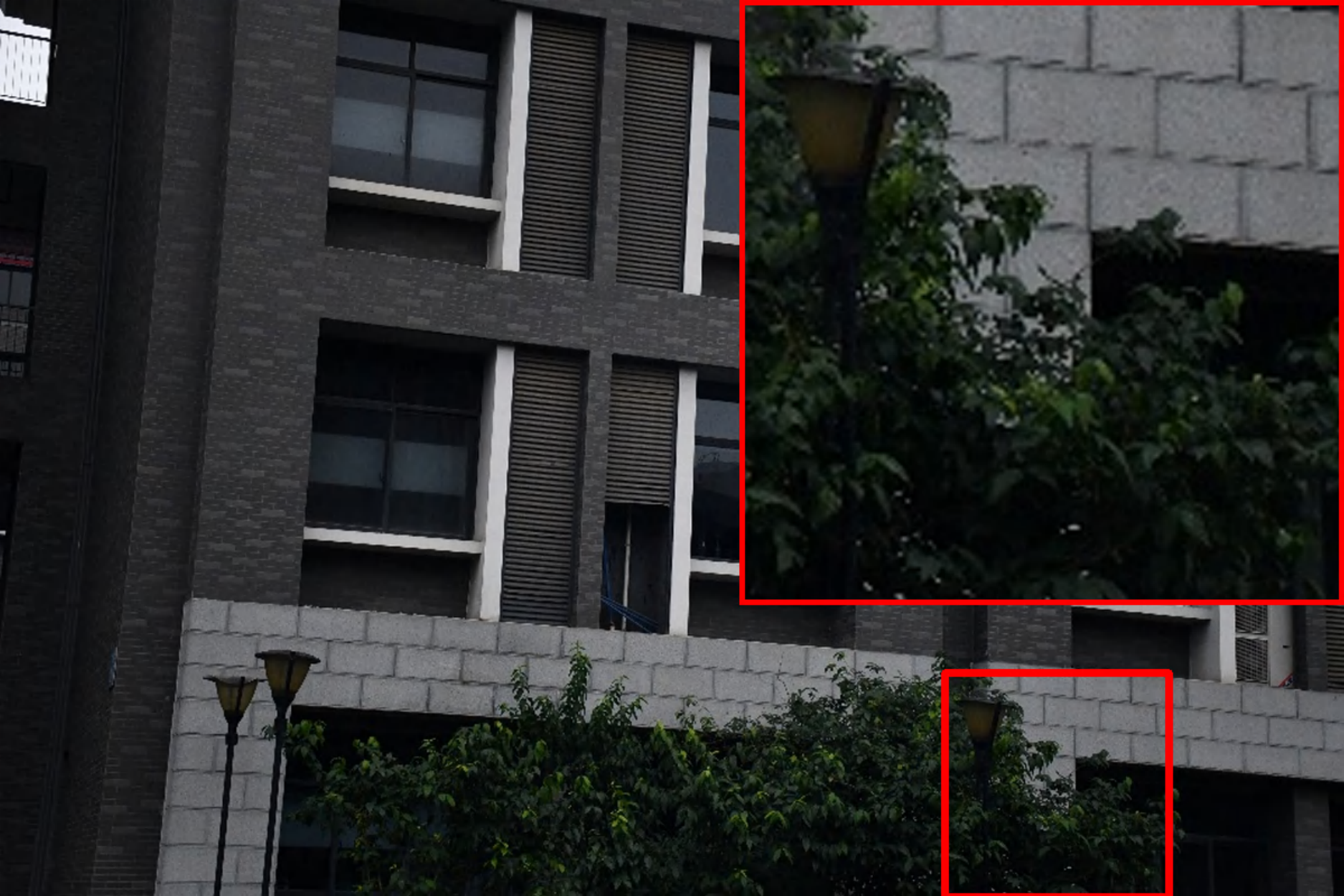}%
        \label{fig:input}%
    }\hspace{0.5mm}%
    \subfloat[Reference]{%
        \includegraphics[width=0.195\linewidth]{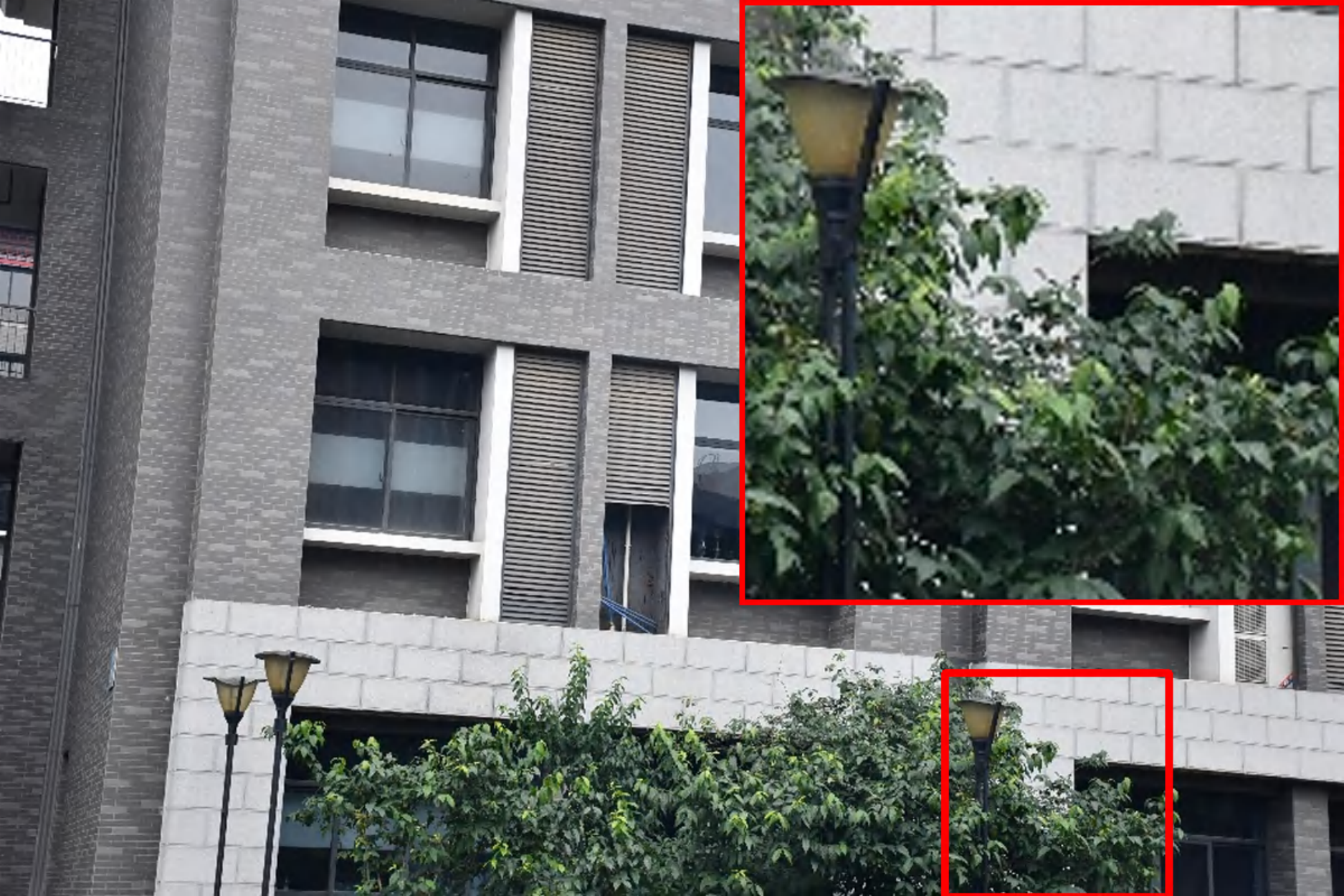}%
        \label{fig:ref}%
    }\hspace{0.5mm}%
    \subfloat[FourLLIE]{%
        \includegraphics[width=0.195\linewidth]{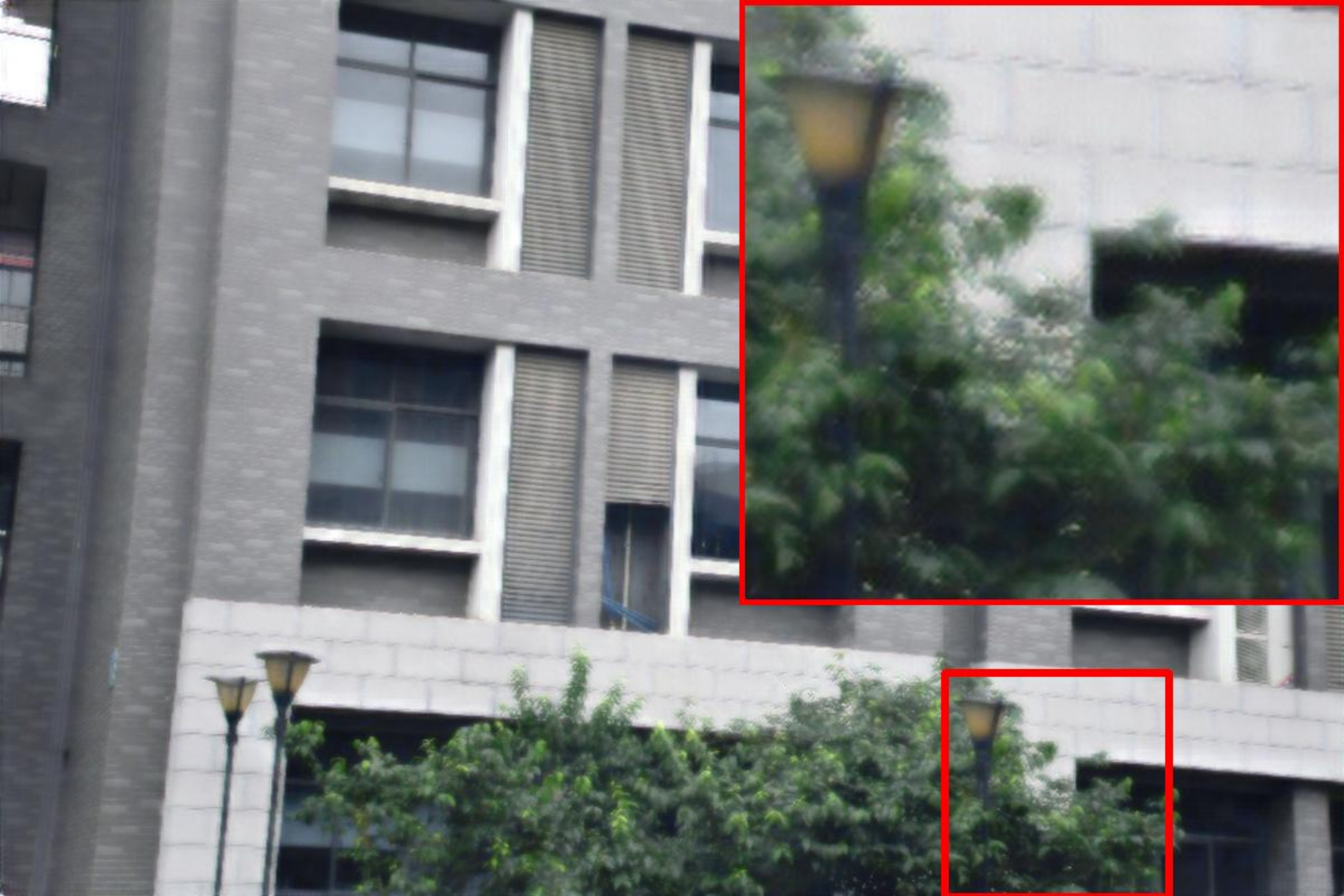}%
        \label{fig:fourllie}%
    }\hspace{0.5mm}%
    \subfloat[DMFourLLIE]{%
        \includegraphics[width=0.195\linewidth]{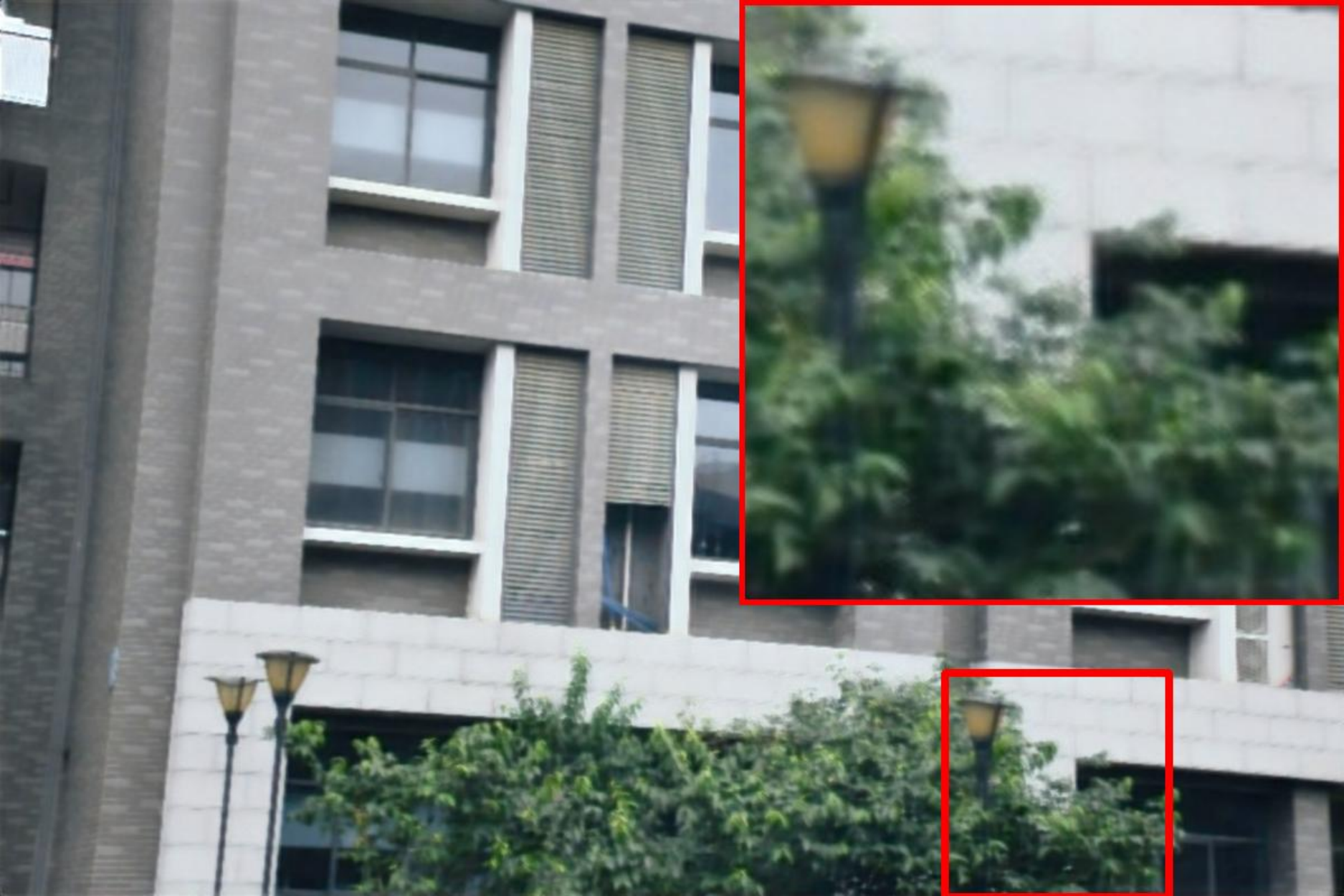}%
        \label{fig:dmfour}%
    }\hspace{0.5mm}%
    \subfloat[CWNet]{%
        \includegraphics[width=0.195\linewidth]{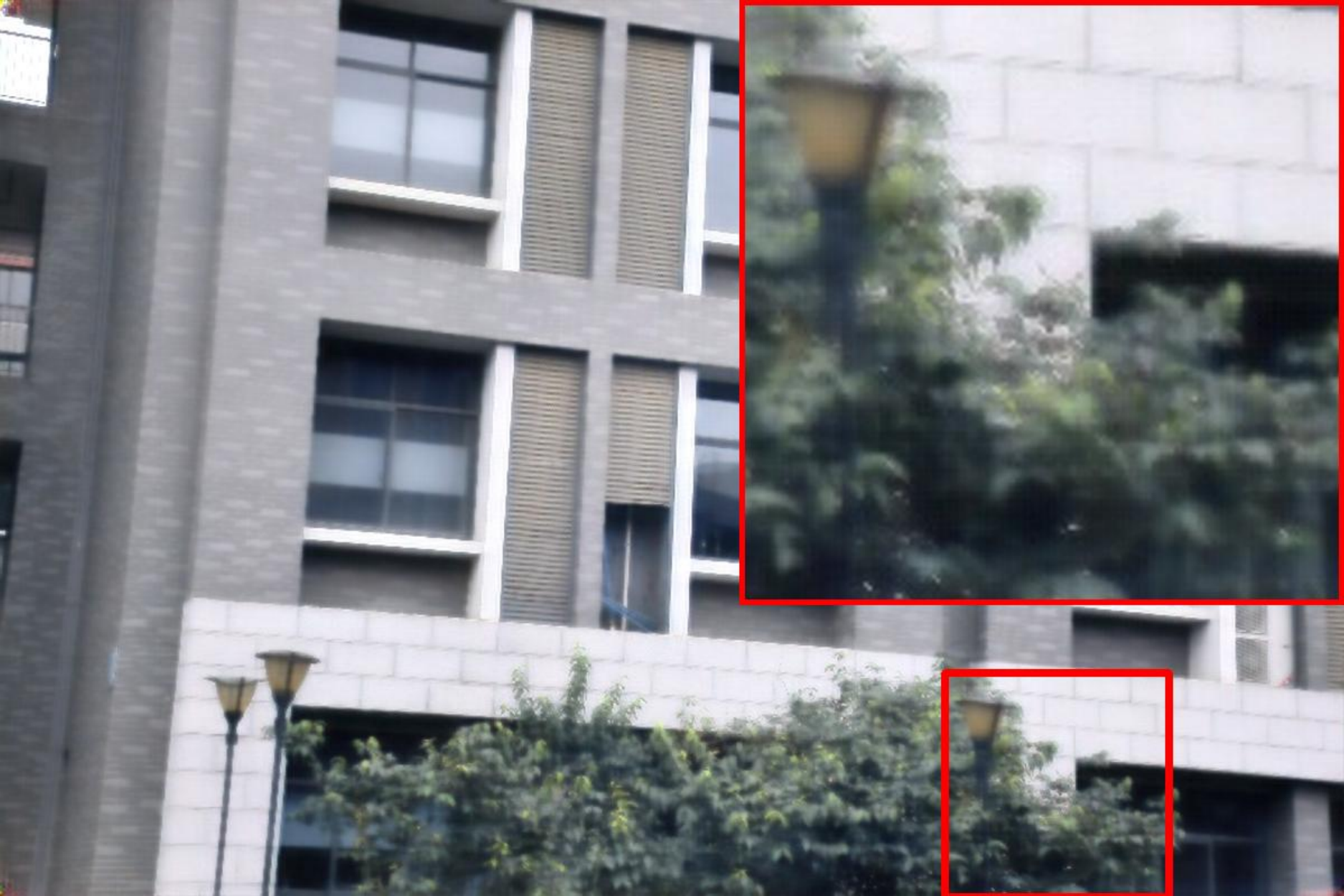}%
        \label{fig:cwnet}%
    }
    \vspace{-3mm} 
    \subfloat[CIDNet]{%
        \includegraphics[width=0.195\linewidth]{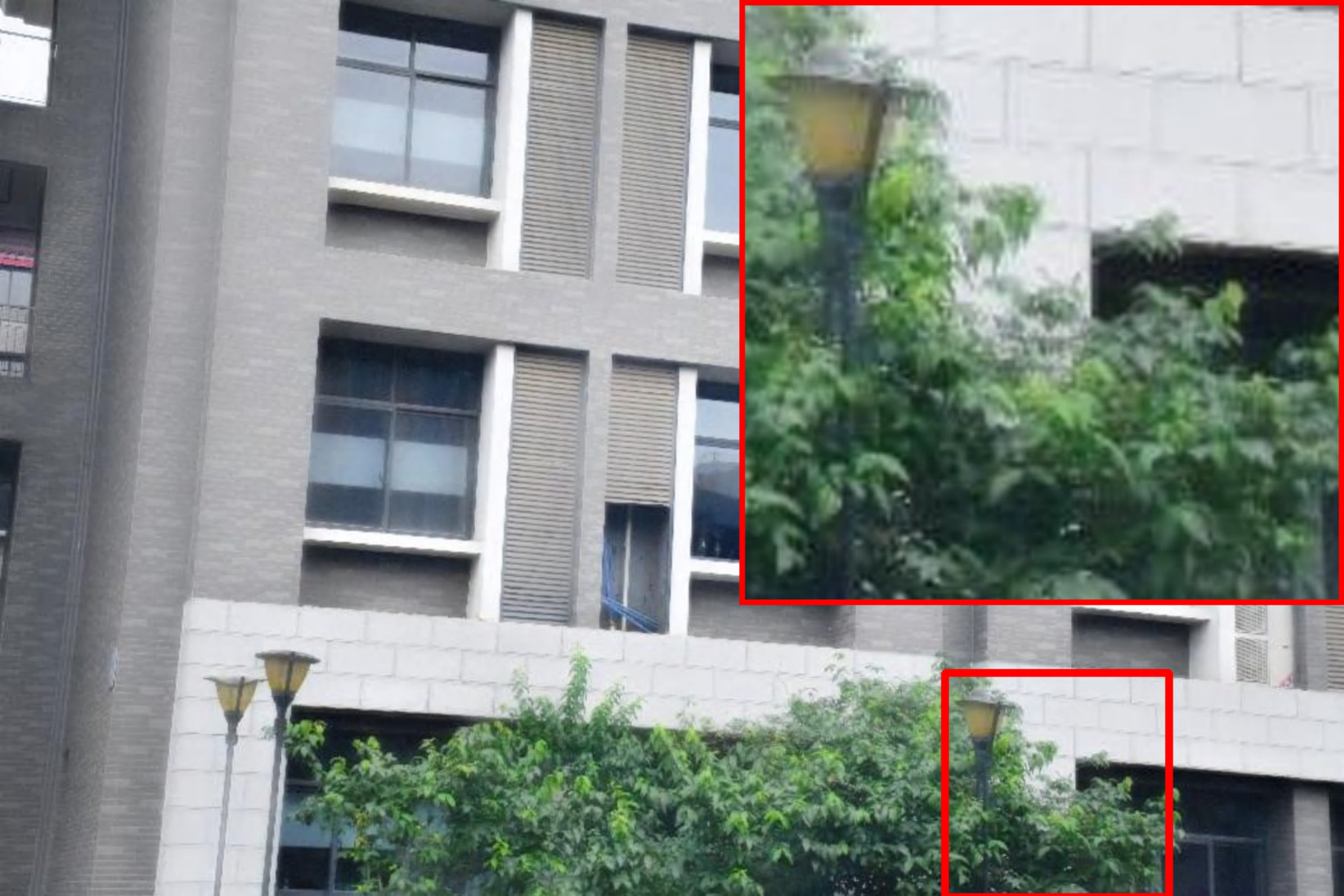}%
        \label{fig:cidnet}%
    }\hspace{0.5mm}%
    \subfloat[UHDFour]{%
        \includegraphics[width=0.195\linewidth]{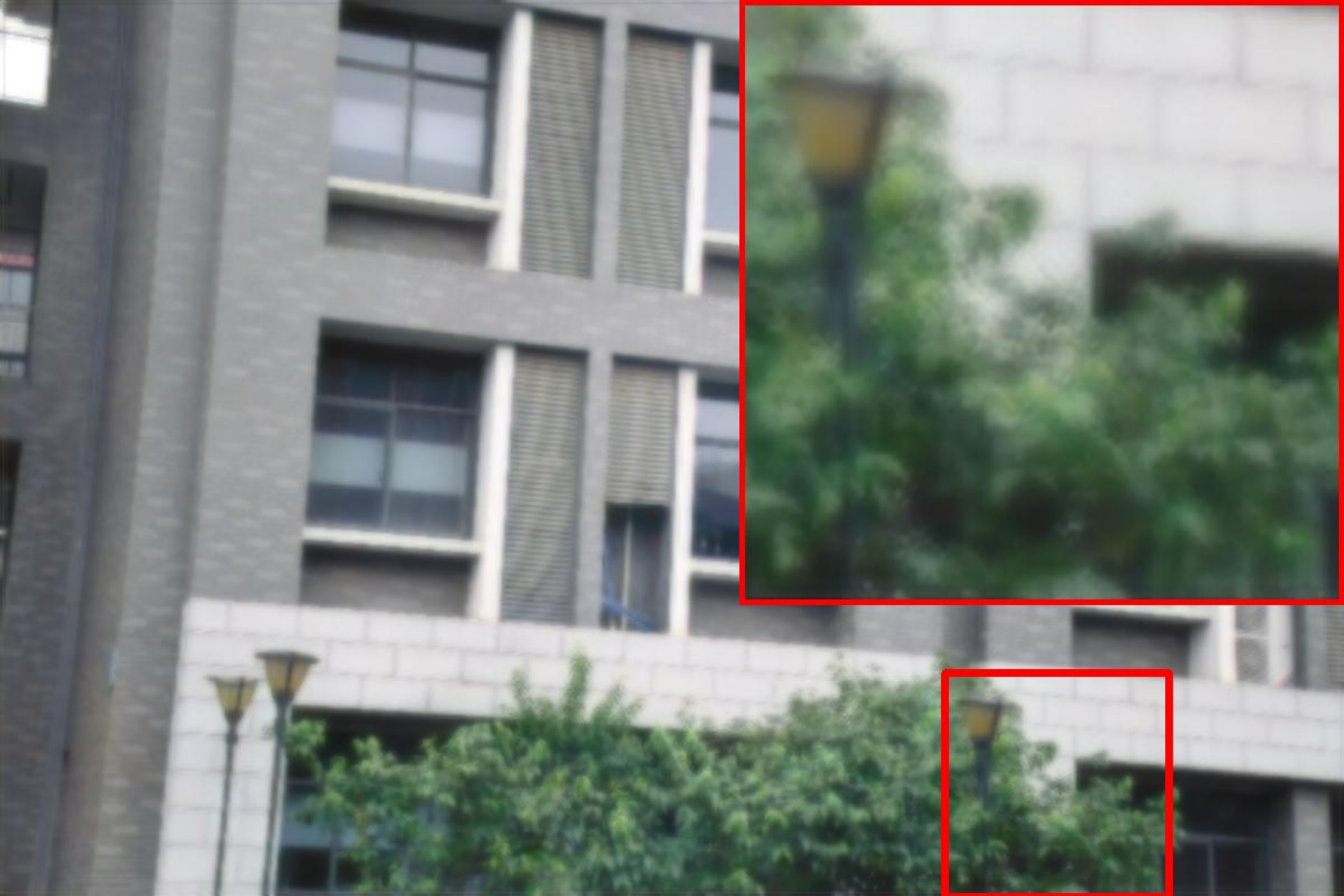}%
        \label{fig:uhd}%
    }\hspace{0.5mm}%
    \subfloat[Wave-Mamba]{%
        \includegraphics[width=0.195\linewidth]{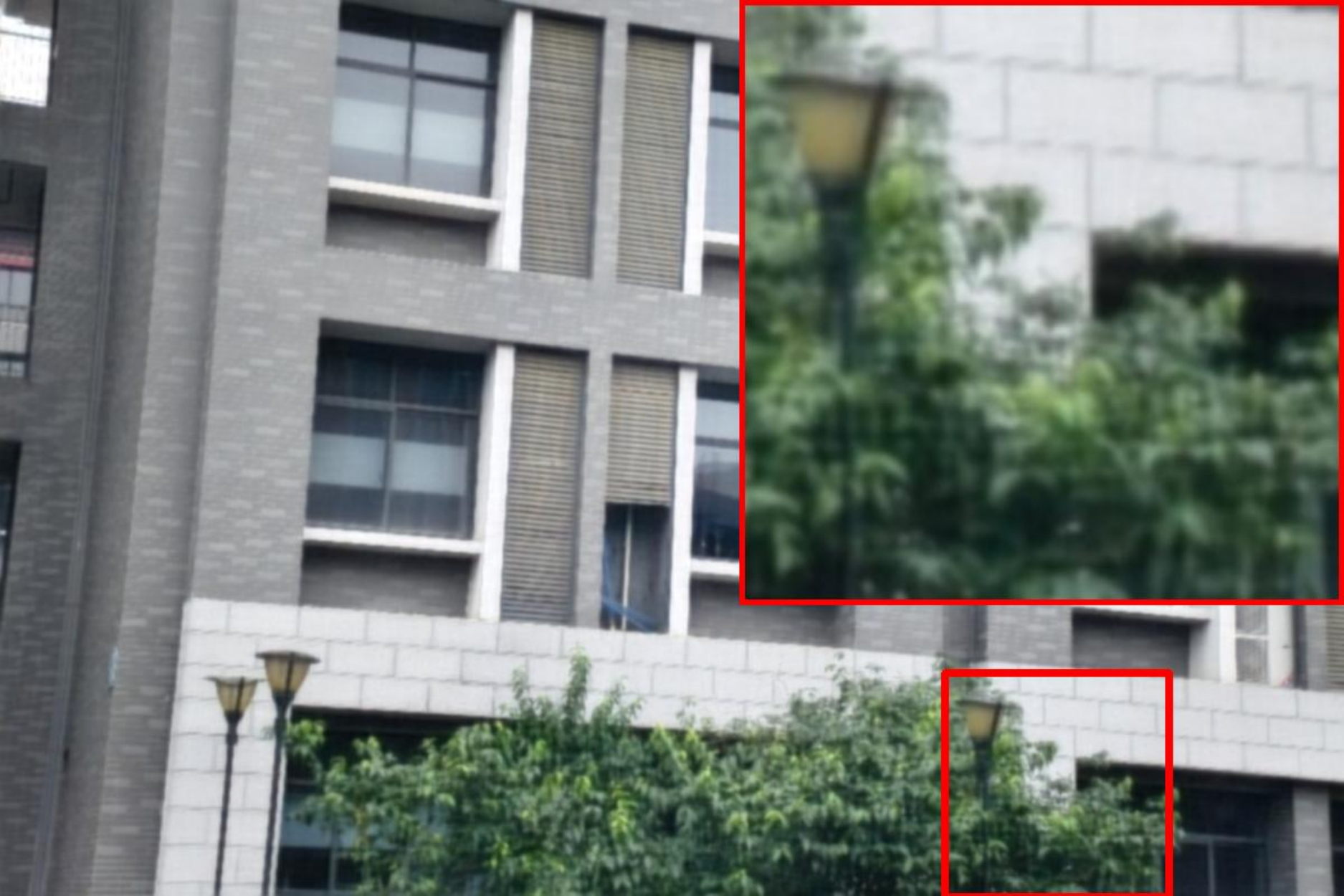}%
        \label{fig:wave}%
    }\hspace{0.5mm}%
    \subfloat[Retinexformer]{%
        \includegraphics[width=0.195\linewidth]{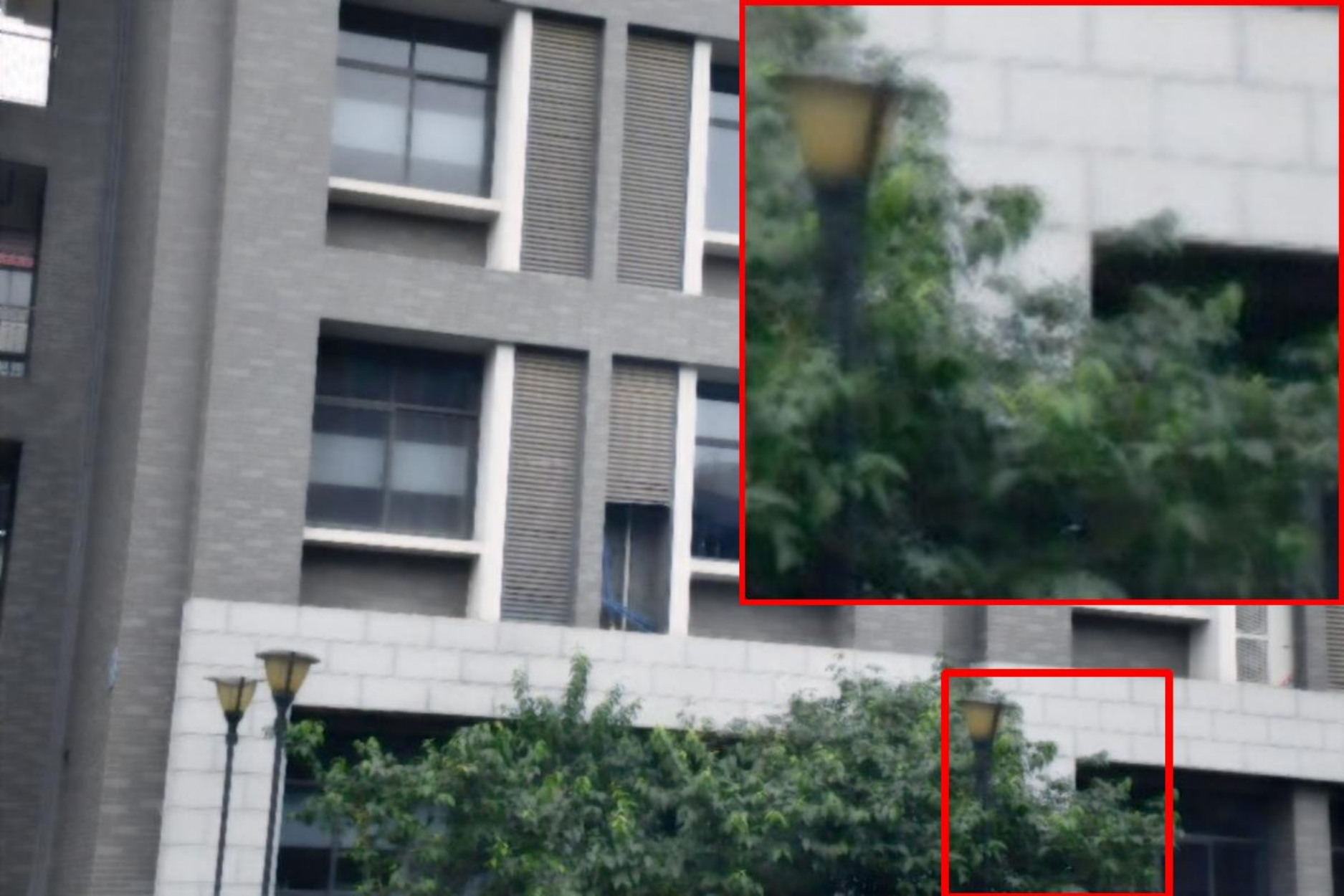}%
        \label{fig:retinex}%
    }\hspace{0.5mm}%
    \subfloat[\textbf{Ours}]{%
        \includegraphics[width=0.195\linewidth]{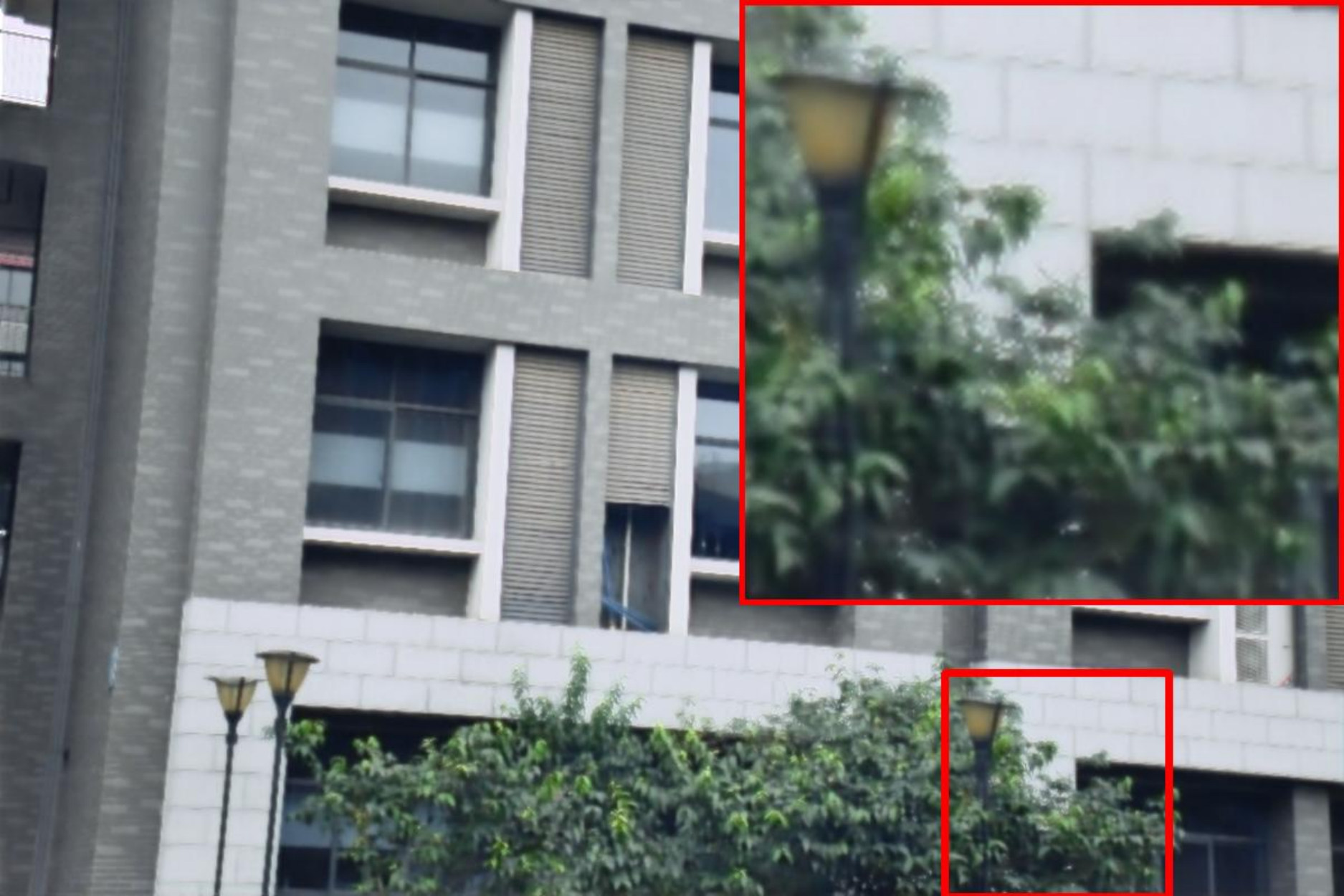}%
        \label{fig:ours}%
    }
    \vspace{-0.2cm}
    \caption{Visual comparison with state-of-the-art methods on the LSRW-Nikon dataset.}
    \vspace{-0.4cm}
    \label{fig:visual_compare_Nikon}
\end{figure*}

\begin{figure}[!t]
    \centering
    \includegraphics[width=1\linewidth]{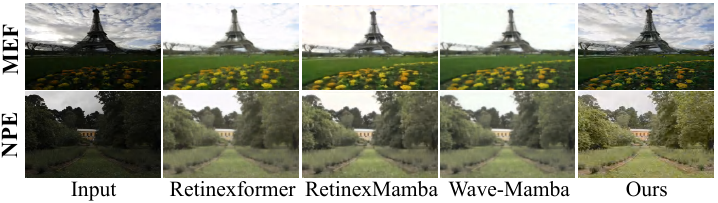}
    \vspace{-0.3cm}
    \caption{Visual comparison on MEF and NPE datasets.}
    \vspace{-0.2cm}
    \label{fig:nopir}
\end{figure}

\begin{table}[t]  
\centering  
\resizebox{0.5\textwidth}{!}{  
\begin{tabular}{l|ccccc|c}  
\toprule  
\textbf{Methods} & \textbf{LIME} & \textbf{VV} & \textbf{DICM} & \textbf{NPE} & \textbf{MEF} & \textbf{AVG} \\
\midrule  
GLARE~\cite{zhou2024glare} & 4.343 & 4.262 & 4.858 & 5.467 & 3.906 & 4.567 \\
LYTNet~\cite{brateanu2401lyt} & 4.540 & 4.639 & 4.187 & 4.372 & 3.959 & 4.339 \\
LANet~\cite{yang2023learning} & 3.680 & 4.670 & \underline{3.902} & 4.856 & 3.931 & 4.208 \\
FourLLIE~\cite{Wang2023FourLLIEBL} & 3.814 & 4.222 & 4.365 & 4.914 & 3.831 & 4.229 \\
Wave-Mamba~\cite{Zou2024WaveMambaWS} & 4.451 & 4.713 & 4.568 & 4.545 & 4.767 & 4.608 \\
Retinexformer~\cite{cai2023retinexformer} & \underline{3.442} & 3.719 & 4.013 & 3.891 & \underline{3.730} & 3.759 \\
UHDFormer~\cite{wang2024uhdformer} & 4.352 & 4.287 & 4.424 & 4.406 & 4.745 & 4.442 \\
FECNet~\cite{Huang2022DeepFE} & 6.043 & 3.751 & 4.140 & 4.508 & 4.716 & 4.632 \\
CWNet~\cite{Zhang2025CWNetCW} & 3.921 & \underline{3.583} & \textbf{3.662} & \underline{3.610} & 3.741 & \underline{3.701} \\  
\midrule  \rowcolor{gray!15}
\textbf{Ours} & \textbf{3.100} & \textbf{2.517} & 5.322 & \textbf{2.934} & \textbf{2.805} & \textbf{3.336} \\
\bottomrule   

\end{tabular}  }
\vspace{-0.2cm}
\caption{Quantitative comparison using NIQE metric (lower is better) on unpaired real-world datasets.} 
\vspace{-0.2cm}
\label{tab:niqe} 
\end{table}

\section{Experiments}

\subsection{Datasets and Experimental Settings}
\noindent \textbf{Datasets.} 
We evaluate our method on three paired low-light enhancement benchmarks: LOL-v1~\cite{lol}, LOL-v2-Syn~\cite{lol}, and LSRW~\cite{lsrw}. 
LOL-v1 contains 485 training pairs and 15 test pairs. 
LOL-v2-Syn includes 900 training pairs and 100 test pairs, covering more diverse scenes and degradation levels. 
LSRW is a large-scale real-world dataset consisting of Nikon D7500 and Huawei P40 Pro subsets, with 5,650 training pairs (3,170 Nikon and 2,480 Huawei) and 50 test pairs (20 Nikon and 30 Huawei).
To further evaluate generalization, we conduct zero-shot testing on five unpaired datasets: LIME~\cite{lime}, NPE~\cite{npe}, MEF~\cite{mefl}, DICM~\cite{dicm}, and VV. 
In addition, to assess practical utility in downstream vision tasks, we use our method as a preprocessing module for semantic segmentation on NightCity~\cite{Tan2020NightTimeSP}, object detection on ExDark~\cite{Yang2019AdvancingIU}, and face detection on DarkFace~\cite{Zamir2020LearningEF}.

\noindent \textbf{Implementation Details.} 
Our model is implemented in PyTorch 2.0.1 with CUDA 11.8, and trained on a single NVIDIA GeForce RTX 3090 GPU (24\,GB). 
The host machine is equipped with an Intel Core i9-13900K CPU, running Ubuntu 22.04.
During training, input images are randomly cropped to $256 \times 256$ patches. 
We use the AdamW optimizer with $\beta_1 = 0.9$ and $\beta_2 = 0.99$, and train for 160,000 iterations with a batch size of 8. 
The initial learning rate is set to $4 \times 10^{-4}$ and decayed using MultiStepLR. 
The total loss is a weighted sum of $L_1$, perceptual, and SSIM losses. 
For fair comparison, we measure FLOPs and inference time under a unified protocol. 
FLOPs are calculated with the THOP library using a fixed input size of $256 \times 256$. 
All compared methods are evaluated using officially provided code and weights. 
To ensure reproducibility, all methods are re-run under the identical hardware and software environment described above.

\begin{figure}[t]
  \centering
  \includegraphics[width=1.0\linewidth]{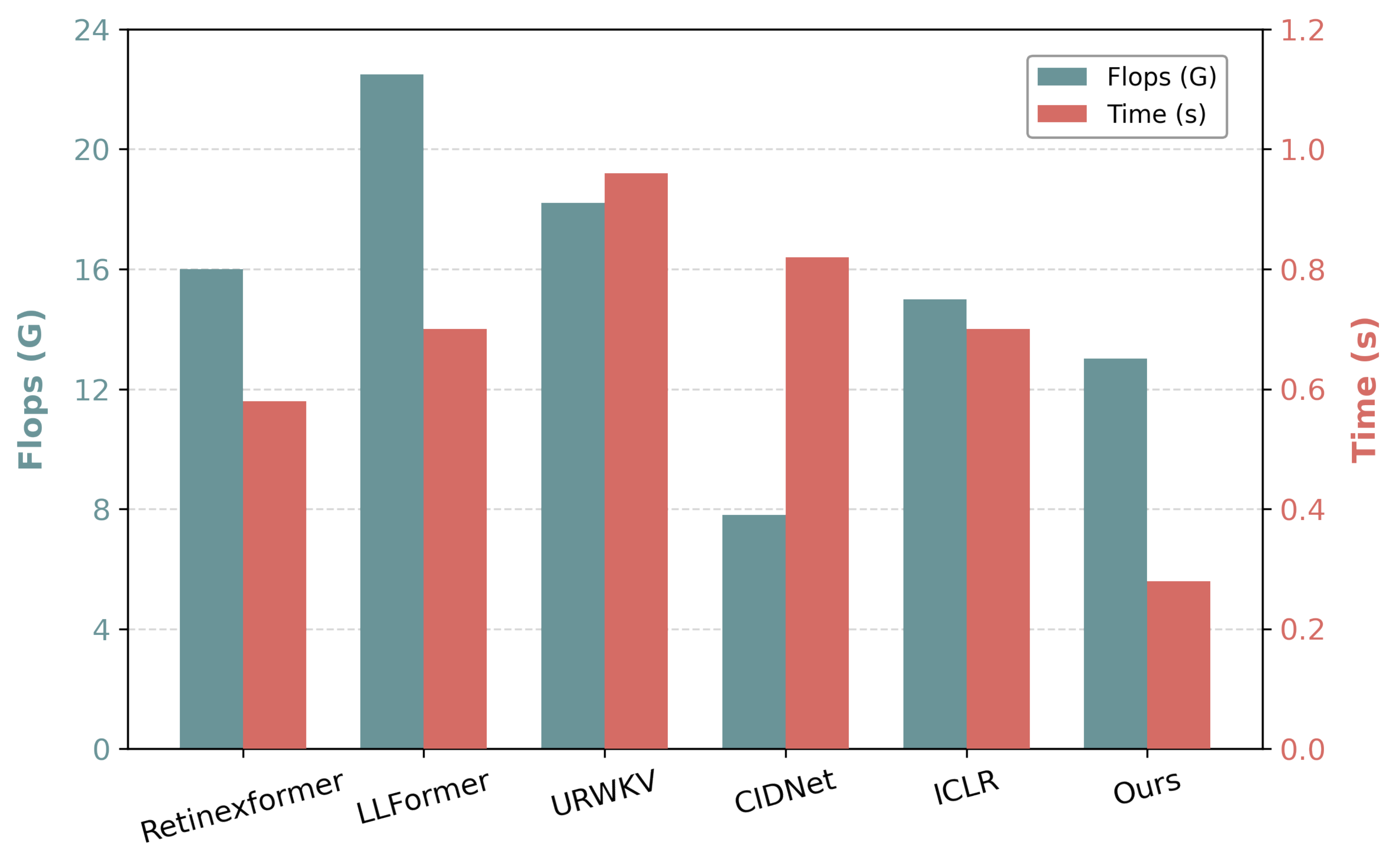}
  \vspace{-0.6cm}
  \caption{Comparison of computational efficiency in terms of FLOPs and average inference time. Our method achieves the lowest inference time while maintaining a moderate computational cost.}
  \vspace{-0.3cm}
  \label{fig:performance_comparison}
\end{figure}

\vspace{-0.2cm}
\subsection{Comparison with State-of-the-Art Methods}

To verify the effectiveness of the proposed method, we compare it with a wide range of representative low-light image enhancement approaches, including traditional methods such as LIME~\cite{lime}, NPE~\cite{npe}, and SRIE~\cite{retinex1}, as well as deep learning-based methods including Kind~\cite{kind}, Kind++~\cite{kind++}, MIRNet~\cite{lowlight9}, SNR-Aware~\cite{lowlight8}, FECNet~\cite{Huang2022DeepFE}, FourLLIE~\cite{Wang2023FourLLIEBL}, UHDFour~\cite{li2023embedding}, Retinexformer~\cite{cai2023retinexformer}, DMFourLLIE~\cite{Zhang2024DMFourLLIEDA}, RetinexMamba~\cite{Bai2024RetinexmambaRM}, CWNet~\cite{Zhang2025CWNetCW}, ICLR~\cite{Xu2025ICLRIA}, CIDNet~\cite{yan2025hvi}, URWKV~\cite{Xu2025URWKVUR}, and Wave-Mamba~\cite{Zou2024WaveMambaWS}.

\subsubsection{Quantitative Evaluation}
Table~\ref{tab:main_results} reports the quantitative comparison on LOL-v1, LOL-v2-Syn, LSRW-Huawei, and LSRW-Nikon in terms of PSNR, SSIM, and LPIPS. Overall, our method achieves highly competitive performance across all four benchmarks.
Specifically, our method achieves the best PSNR on LOL-v2-Syn and LSRW-Nikon, and ties for the best PSNR on LOL-v1. In addition, it achieves the best LPIPS on LOL-v2-Syn, the best SSIM on LSRW-Huawei, and the best SSIM on LSRW-Nikon. On the remaining metrics, our method generally ranks among the top-performing methods, indicating that it provides a favorable balance between fidelity, structural preservation, and perceptual quality.

\subsubsection{Qualitative Evaluation}
\noindent \textbf{LOL-v1.} 
As shown in Fig.~\ref{fig:visual_compare_V1}, several methods improve global visibility but still suffer from color distortion, detail loss, or over-enhancement. 
For instance, FourLLIE and UHDFour exhibit noticeable color bias, while CIDNet tends to produce overexposed regions. Our method restores the scene with more faithful colors and clearer object boundaries, yielding a result visually more consistent with the reference image.
\noindent \textbf{LOL-v2-Syn.} 
As shown in Fig.~\ref{fig:visual_compare_Syn}, existing methods struggle to jointly preserve exposure, color fidelity, and local texture. 
FourLLIE, UHDFour, and Wave-Mamba tend to produce insufficient illumination enhancement, leaving dark regions under-restored. 
By contrast, Retinexformer and CIDNet improve brightness more aggressively but may introduce color deviation or over-enhancement in highlighted regions. 
Our method generates results visually closest to the reference, with more natural exposure, better color consistency, and clearer recovery of fine textures such as water ripples.
\noindent \textbf{LSRW-Huawei.} 
As shown in Fig.~\ref{fig:visual_compare_Huawei}, FourLLIE tends to produce blurry results, whereas UHDFour and CIDNet may introduce visible color shifts. 
In comparison, our method better preserves textural details around object boundaries and recovers a more natural color distribution, yielding a more balanced enhancement.
\noindent \textbf{LSRW-Nikon.} 
As shown in Fig.~\ref{fig:visual_compare_Nikon}, this example is particularly challenging due to weak illumination and fine structural details in the vegetation region. 
Several competing methods either over-smooth local textures or introduce visible color inconsistency. 
In contrast, our method better preserves structural details in the foliage and surrounding edges while maintaining more natural tones, demonstrating stronger robustness in practical low-light scenarios.

\begin{table}[t]
    \centering
    \caption{Ablation study on the effectiveness of the two-stage architecture.}
    \vspace{-0.2cm}
    \label{tab:ablation_stage}
    \setlength{\tabcolsep}{2mm} 
    \begin{tabular}{l|cc|ccc}
    \toprule
    \textbf{Model} & \textbf{Stage-I} & \textbf{Stage-II} & \textbf{PSNR} $\uparrow$ & \textbf{SSIM} $\uparrow$ & \textbf{LPIPS} $\downarrow$ \\
    \midrule
    A & $\checkmark$ & $\times$ & 17.52 & 0.7164 & 0.1496 \\
    B & $\times$ & $\checkmark$ & 25.62 & 0.9330 & 0.0240 \\  \rowcolor{gray!15}
    \textbf{Ours} & $\checkmark$ & $\checkmark$ & \textbf{26.42} & \textbf{0.9371} & \textbf{0.0222} \\
    \bottomrule
    \end{tabular}
    \vspace{-0.2cm}
\end{table}
\begin{table}[t]
    \centering
    \caption{Ablation study of APACF.}
    \vspace{-0.2cm}
    \label{tab:ablation_apacb}
    \setlength{\tabcolsep}{3.5mm}
    \begin{tabular}{l|ccc}
    \toprule
    \textbf{Configuration} & \textbf{PSNR} $\uparrow$ & \textbf{SSIM} $\uparrow$ & \textbf{LPIPS} $\downarrow$ \\
    \midrule
    w/o Frequency Branch & 25.62 & 0.9330 & 0.0240 \\
    w/ Hard Splitting & 24.92 & 0.9311 & 0.0270 \\ \rowcolor{gray!15}
    \textbf{w/ APACF (Ours)} & \textbf{26.42} & \textbf{0.9371} & \textbf{0.0222} \\
    \bottomrule
    \end{tabular}
    \vspace{-0.2cm}
\end{table}
\begin{table}[t]
    \centering
    \caption{Ablation study on different prior sources.}
    \vspace{-0.2cm}
    \label{tab:ablation_prior_source}
    \setlength{\tabcolsep}{3mm}
    \begin{tabular}{l|ccc}
    \toprule
    \textbf{Prior Source} & \textbf{PSNR} $\uparrow$ & \textbf{SSIM} $\uparrow$ & \textbf{LPIPS} $\downarrow$ \\
    \midrule
    Raw-Prior (Input) & 23.15 & 0.9139 & 0.0418 \\
    GT-Prior (Reference) & 19.10 & 0.8664 & 0.0600 \\
    Depth Map~\cite{yang2024depth} & 24.87 & 0.9294 & 0.0342 \\
    Edge Map~\cite{xu2023low} & 23.96 & 0.9253 & 0.0384 \\\rowcolor{gray!15}
    \textbf{ERP (Ours)} & \textbf{26.42} & \textbf{0.9371} & \textbf{0.0222} \\
    \bottomrule
    \end{tabular}
    \vspace{-0.2cm}
\end{table}
\begin{table}[t]
    \centering
    \caption{Ablation study of GGEB and PIGE.}
    \vspace{-0.2cm}
    \label{tab:ablation_fusion}
    \setlength{\tabcolsep}{2mm}
    \begin{tabular}{l|cc|ccc}
    \toprule
    \textbf{Model Variant} & \textbf{GGEB} & \textbf{PIGE} & \textbf{PSNR} $\uparrow$ & \textbf{SSIM} $\uparrow$ & \textbf{LPIPS} $\downarrow$ \\
    \midrule
    Baseline & $\times$ & $\times$ & 25.51 & 0.9345 & 0.0251 \\
    w/ GGEB Only & $\checkmark$ & $\times$ & 25.44 & 0.9297 & 0.0260 \\
    w/ PIGE Only & $\times$ & $\checkmark$ & 25.31 & 0.9352 & 0.0244 \\ \rowcolor{gray!15}
    \textbf{Ours} & $\checkmark$ & $\checkmark$ & \textbf{26.42} & \textbf{0.9371} & \textbf{0.0222} \\
    \bottomrule
    \end{tabular}
    \vspace{-0.2cm}
\end{table}

\subsubsection{Generalization on Unpaired Real-World Datasets}
To further evaluate generalization beyond paired supervised benchmarks, we conduct zero-shot experiments on five widely used unpaired real-world low-light datasets: LIME~\cite{lime}, VV, DICM~\cite{dicm}, NPE~\cite{npe}, and MEF~\cite{mefl}. 
These datasets cover diverse degradation patterns commonly encountered in practice, including underexposure, mixed noise, motion blur, low contrast, compression artifacts, and sensor-specific variations.
\noindent \textbf{Quantitative evaluation.} 
We adopt the Natural Image Quality Evaluator (NIQE~\cite{6353522}) as the no-reference metric, where a lower score indicates better perceptual quality. 
As reported in Table~\ref{tab:niqe}, our method achieves the best average NIQE across the five datasets, outperforming the second-best method CWNet by a notable margin. 
Specifically, our method ranks first on four of the five datasets: LIME (3.100), VV (2.517), NPE (2.934), and MEF (2.805). 
On DICM, which contains severe compression artifacts, our method lags behind some competitors; however, no single method consistently dominates this challenging dataset, and our approach remains competitive in overall ranking.
\noindent \textbf{Qualitative evaluation.} 
Fig.~\ref{fig:nopir} presents visual comparisons on representative samples from MEF and NPE. 
On the MEF scene, competing methods produce over-enhanced and washed-out results, with noticeable color degradation in the sky and loss of detail in the foreground flowers. 
In contrast, our method preserves natural sky tones and recovers finer floral textures, demonstrating better generalization to real-world degradation. 
On the NPE scene, several methods suffer from residual noise or insufficient contrast, while our method achieves a cleaner and more visually pleasing result with well-balanced illumination.

\begin{figure}[t]
    \centering
    \subfloat[Input]{%
        \includegraphics[width=0.30\linewidth]{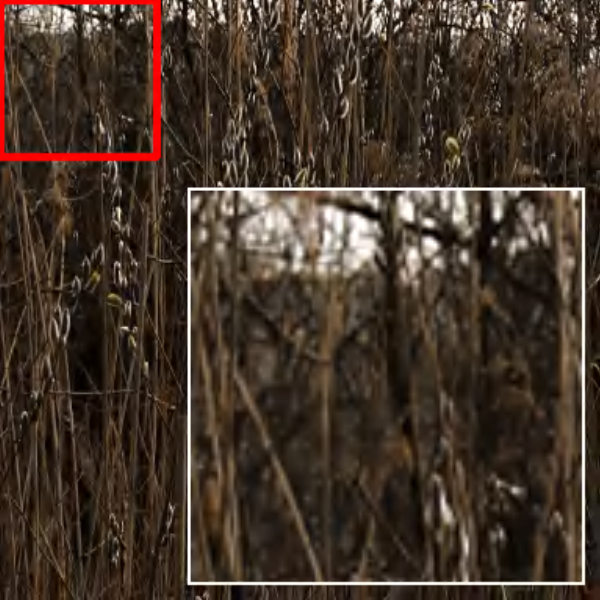}%
    }\hfil
    \subfloat[w/o GGEB/PIGE]{%
        \includegraphics[width=0.30\linewidth]{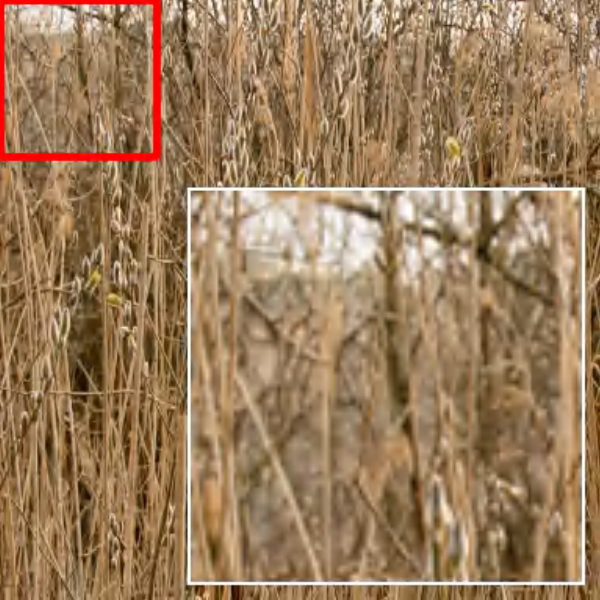}%
    }\hfil
    \subfloat[w/o GGEB]{%
        \includegraphics[width=0.30\linewidth]{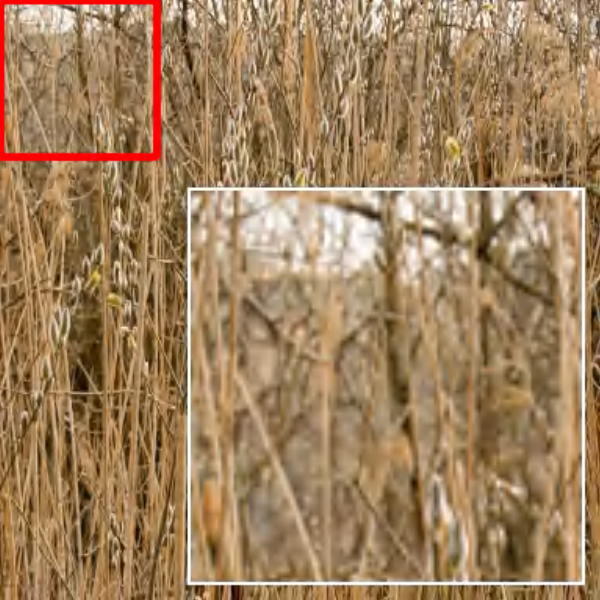}%
    }
    \vspace{-3mm} 
    \subfloat[Reference]{%
        \includegraphics[width=0.30\linewidth]{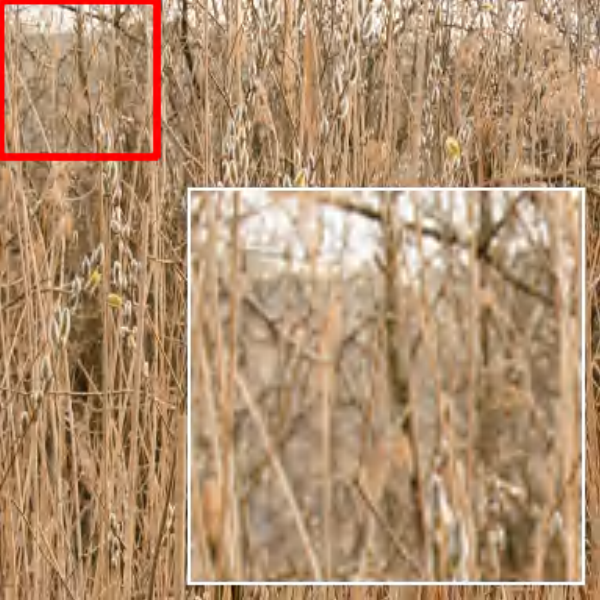}%
    }
    \subfloat[w/o PIGE]{%
        \includegraphics[width=0.30\linewidth]{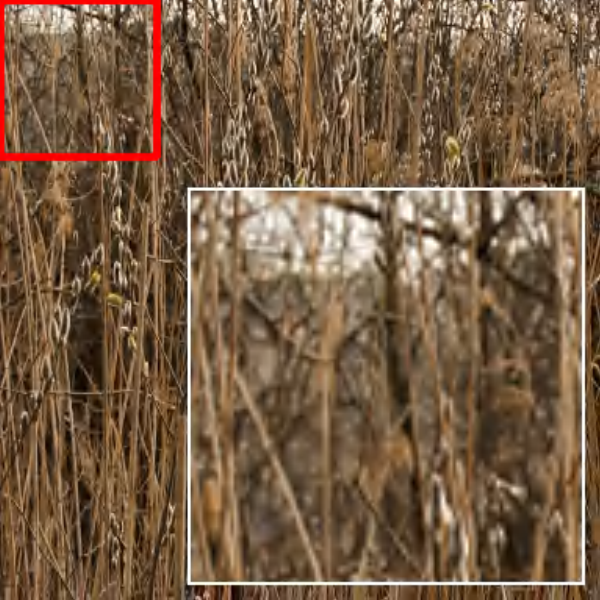}%
    }\hfil
    \subfloat[Ours]{%
        \includegraphics[width=0.30\linewidth]{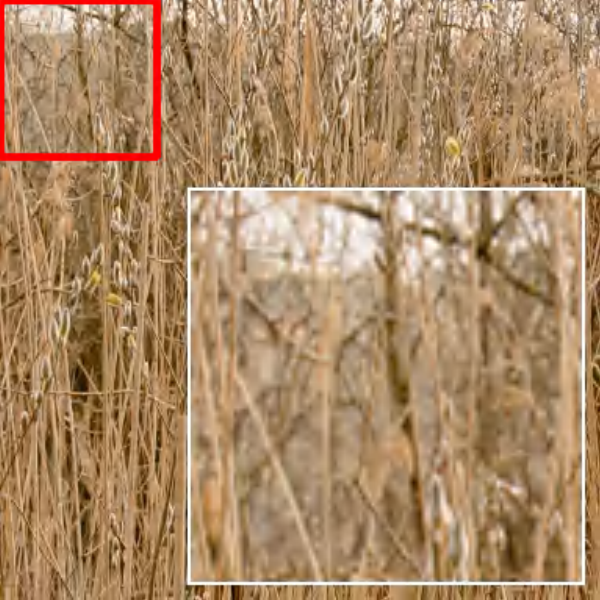}%
    }\hfil
    \vspace{-0.2cm}
    \caption{Visual ablation comparison.}
    \vspace{-0.3cm}
    \label{fig:ablation_GGEB_vis}
\end{figure}

\subsubsection{Computational Efficiency Analysis}

To assess the practical efficiency of different methods, we compare FLOPs and the average inference time on an NVIDIA RTX 3090 GPU with an input resolution of \(256 \times 256\). As shown in Fig.~\ref{fig:performance_comparison}, our method achieves the lowest inference time among all compared methods while maintaining a moderate computational cost.
Compared with transformer-based or large-capacity models such as Retinexformer, LLFormer, and URWKV, our method is substantially more efficient in both runtime and overall computation. Although CIDNet has lower FLOPs, its actual inference time is noticeably higher, indicating that lower theoretical complexity does not necessarily translate into faster execution. In contrast, our method achieves a more favorable trade-off between computational cost and latency, which highlights its potential for real-time or resource-constrained low-light enhancement applications.

\begin{figure}[!t]
  \centering
  \includegraphics[width=1.0\linewidth]{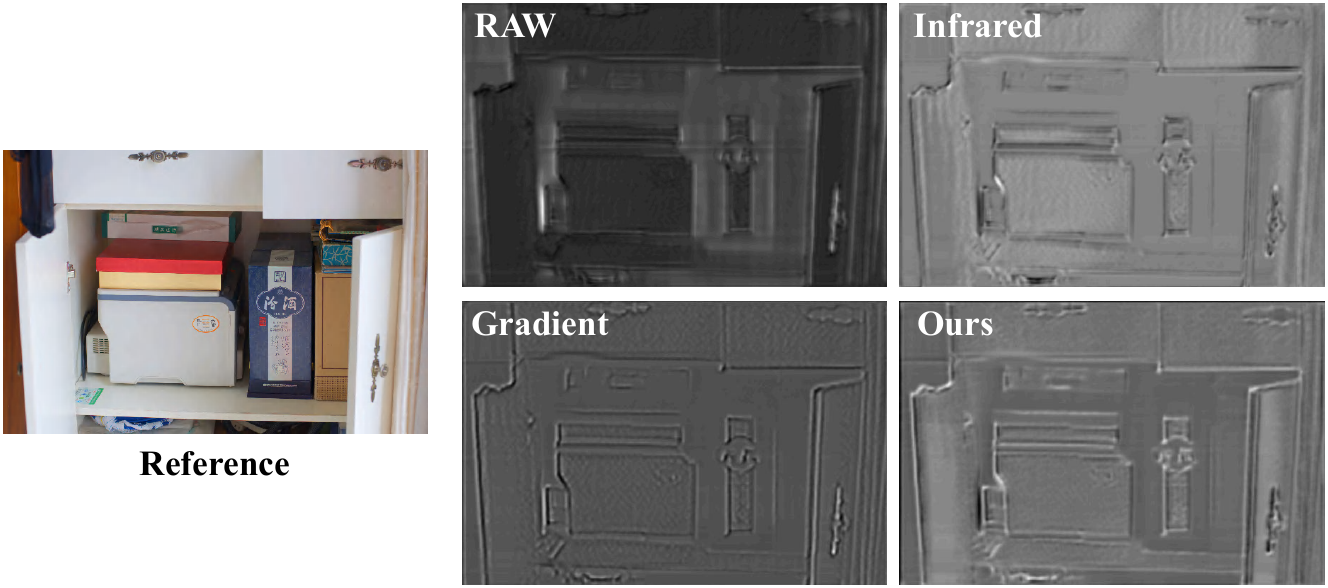}
  \vspace{-0.2cm}
  \caption{Visualization of intermediate features under different prior settings, including no prior (RAW), infrared-only prior, gradient-only prior, and the proposed ERP.}
  \vspace{-0.2cm}
  \label{fig:feature_vis}
\end{figure}

\subsection{Ablation Studies and Analysis}
We conduct ablation studies to evaluate the effectiveness of each component.
\subsubsection{Effectiveness of the Two-Stage Architecture}
Table~\ref{tab:ablation_stage} reports the contribution of the proposed two-stage design. Using Stage I alone gives limited performance, suggesting that frequency-domain rectification by itself is insufficient for high-quality enhancement. Using Stage II alone yields much better results, but still falls short of the full model. When both stages are combined, our method achieves the best performance on all metrics. This confirms that Stage I provides a reliable global restoration basis, while Stage II further refines local details for better overall enhancement.

\subsubsection{Impact of APACF}
As reported in Table \ref{tab:ablation_apacb}, APACF achieves the best performance on all three metrics. The hard-splitting variant performs the worst, indicating that independently modeling amplitude and phase weakens spectral consistency.

\subsubsection{Prior Reliability Analysis}
As shown in Table \ref{tab:ablation_prior_source}, the proposed endogenous prior achieves the best results by a clear margin on all metrics. In contrast, the raw prior performs poorly because severe degradation in the input image limits the reliability of the extracted guidance.
Interestingly, the GT prior does not improve performance and even yields the worst results. A possible reason is that, although the GT image is visually clean, its prior representation is not well aligned with the feature distribution learned by the prior-guided refinement network. Since PIGE is optimized jointly with priors derived from the Stage I intermediate result, directly replacing them with GT-based priors introduces a distribution mismatch and weakens the prior-backbone interaction.

To further examine the effectiveness of different prior sources, we additionally compare depth map~\cite{yang2024depth} and edge map~\cite{xu2023low} as alternative structural priors. However, they still lag behind the proposed ERP, which fuses pseudo-infrared and gradient priors extracted from the rectified Stage I output. This confirms that ERP offers more reliable and distribution-compatible guidance than single-modality priors.

These results reveal an important insight for prior-guided LLIE: the most visually clean prior is not necessarily the most effective one. 
Instead, a successful prior should jointly satisfy two conditions---structural reliability and distribution compatibility with the refinement network.
Our endogenous prior fulfills both: it inherits clean structures from the rectified $I_{s1}$ while maintaining distributional consistency with the features that PIGE is optimized to process.

\begin{figure}[!t]
    \centering
    \subfloat[\label{fig:mot_tsne_illum}]{\includegraphics[width=0.5\linewidth]{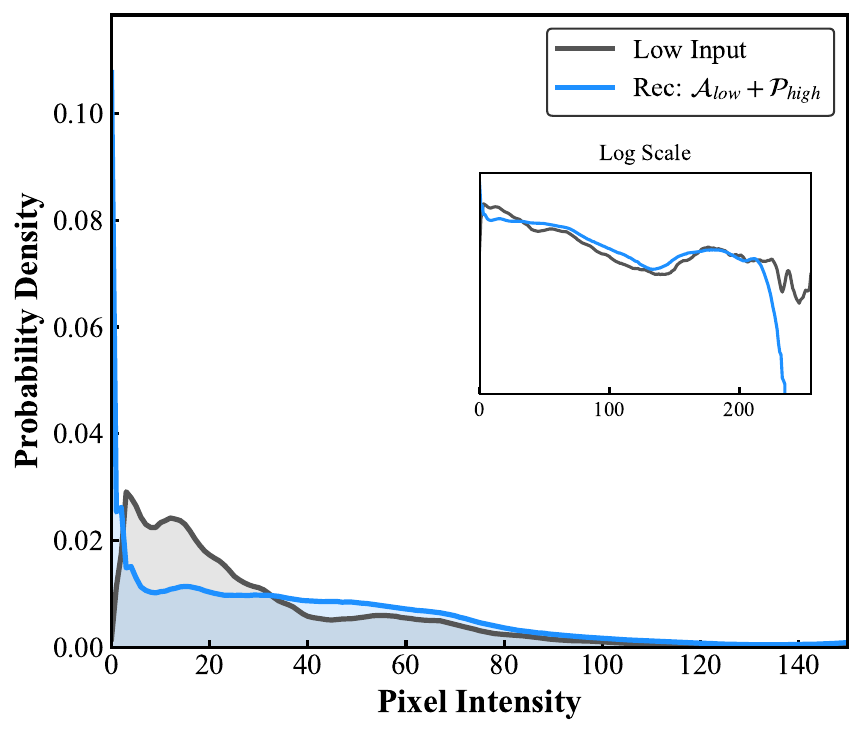}}
    \hfill
    \subfloat[\label{fig:mot_hist_illum}]{\includegraphics[width=0.5\linewidth]{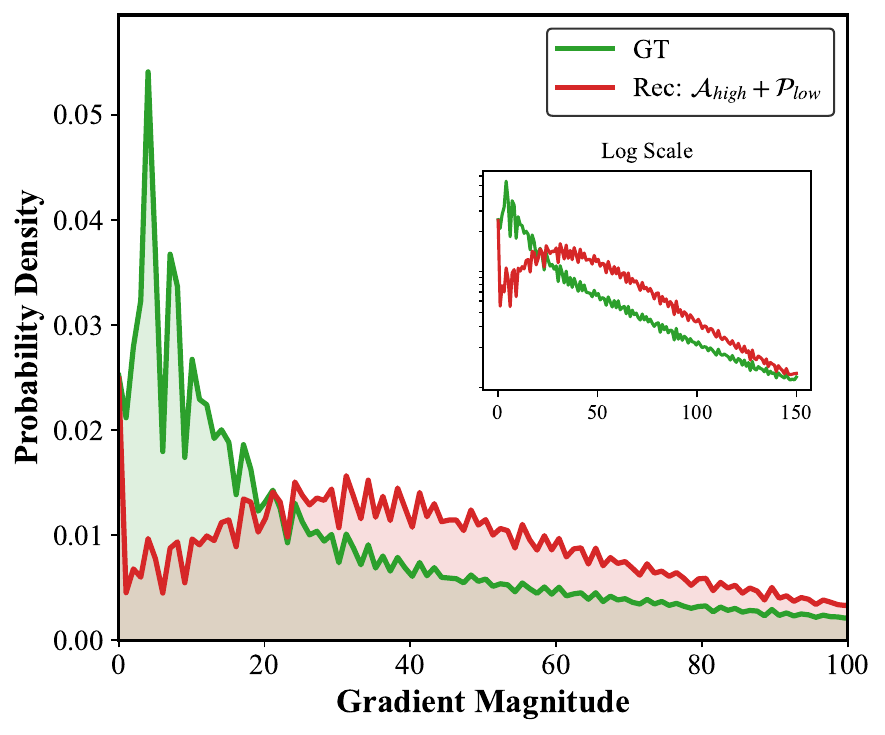}}
    \vspace{-0.2cm}
    \caption{Statistical evidence of bidirectional spectral coupling.
    (a) Replacing only the low-light phase with a normal-light one shifts the pixel intensity distribution toward brighter regions, indicating that phase also affects illumination.
    (b) Combining the ground-truth amplitude \((\mathcal{A}_{high})\) with the low-light phase still yields a gradient distribution that deviates from the GT.}
    \vspace{-0.2cm}
    \label{fig:motivation1}
\end{figure}

\begin{figure}[!t]
  \centering
  \subfloat[Reference]{%
    \includegraphics[width=0.35\linewidth]{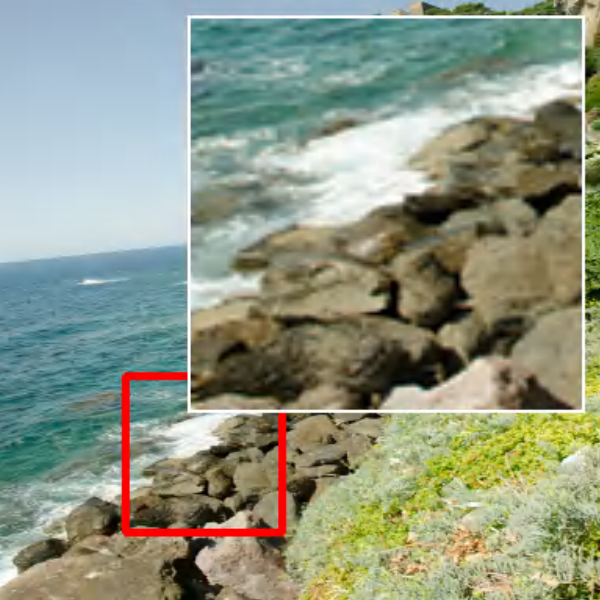}%
  }\hspace{0.1cm}
  \subfloat[w/o Frequency Branch]{%
    \includegraphics[width=0.35\linewidth]{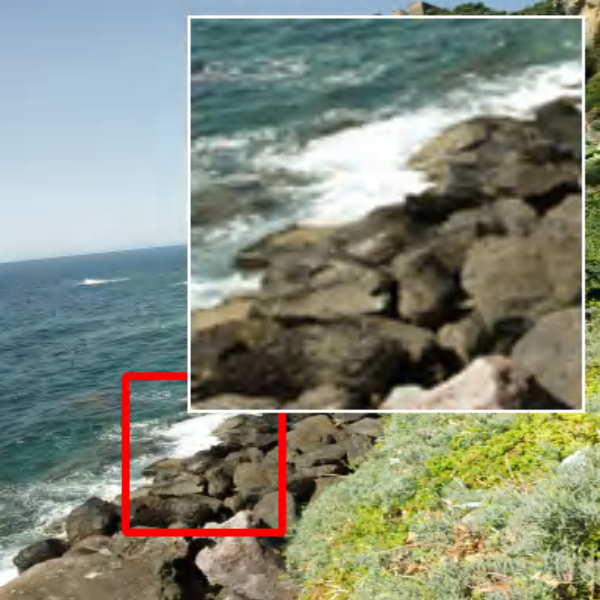}%
  } 
  \vspace{-0.01mm}
  \subfloat[w/ Hard Splitting]{%
    \includegraphics[width=0.35\linewidth]{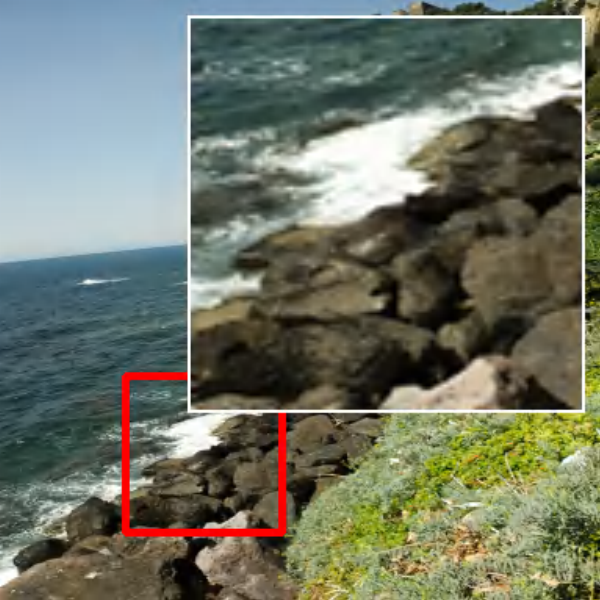}%
  }\hspace{0.1cm}
  \subfloat[\textbf{Ours (w/ APACF)}]{%
    \includegraphics[width=0.35\linewidth]{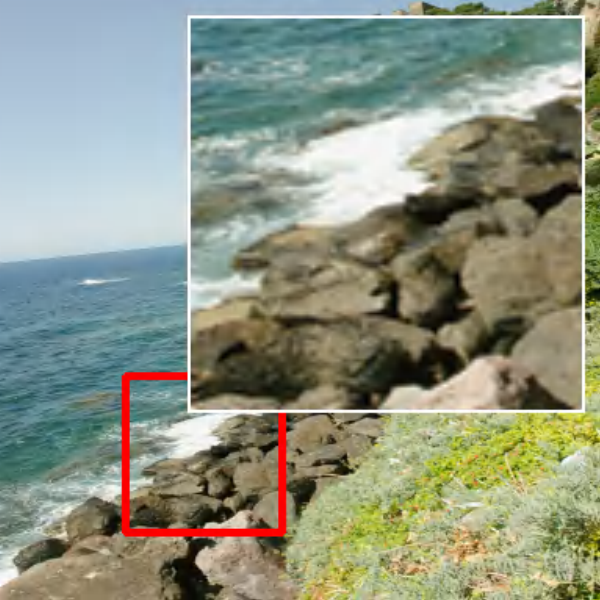}%
  }
  \vspace{-0.2cm}
  \caption{Visual ablation comparison for APACF.}
  \vspace{-0.3cm}
  \label{fig:ablation_apacf_vis}
\end{figure}

\subsubsection{Effectiveness of GGEB and PIGE}

Finally, we verify the synergy between GGEB and PIGE in Table \ref{tab:ablation_fusion}. Surprisingly, deploying either module individually degrades performance below the Baseline due to introduced feature interference or misleading attention. As shown in Fig.\ref{fig:ablation_GGEB_vis}, the lack of synergy manifests as inadequate brightness restoration and unnatural color shifts in the single-module results. In contrast, the combined integration ensures precise spectral modulation, yielding results with balanced exposure and accurate chromaticity.

\subsubsection{Visualization of Intermediate Features}
We visualize intermediate feature maps under four settings in Fig.\ref{fig:feature_vis}: no prior (RAW), infrared-only prior, gradient-only prior, and the proposed ERP. Compared with RAW, both single-prior variants produce stronger responses. However, the infrared prior mainly enhances semantic regions while lacking boundary sharpness, and the gradient prior highlights structural details but provides less complete semantic activation. By contrast, the proposed ERP generates more concentrated responses, clearer edges, and better region consistency, showing that the fused prior offers more effective guidance than either modality alone.

\subsubsection{Statistical Motivation for Bidirectional Spectral Coupling}

To verify that the interaction between amplitude and phase is an intrinsic spectral property rather than an isolated visual effect, we perform a dataset-level statistical analysis on the LSRW dataset, as shown in Fig.~\ref{fig:motivation1}.
In Fig.~\ref{fig:motivation1}(a), we fix the low-light amplitude \(\mathcal{A}_{low}\) and replace only the phase with that of a normal-light image. If illumination were determined solely by amplitude, the intensity distribution should remain unchanged. However, the histogram shifts clearly toward brighter regions, indicating that phase also contributes to illumination formation.
In Fig.~\ref{fig:motivation1}(b), we combine the ground-truth amplitude \(\mathcal{A}_{high}\) with the low-light phase and compare the resulting gradient distribution with that of the GT. Despite using the ideal amplitude, the reconstructed result still shows a clear mismatch, especially in the tail region, suggesting that amplitude correction alone cannot recover faithful structures when the phase is degraded.
These results confirm that amplitude and phase are strongly coupled in low-light image formation.

\begin{figure}[!t]
  \centering
  \subfloat[Reference]{%
    \includegraphics[width=0.35\linewidth]{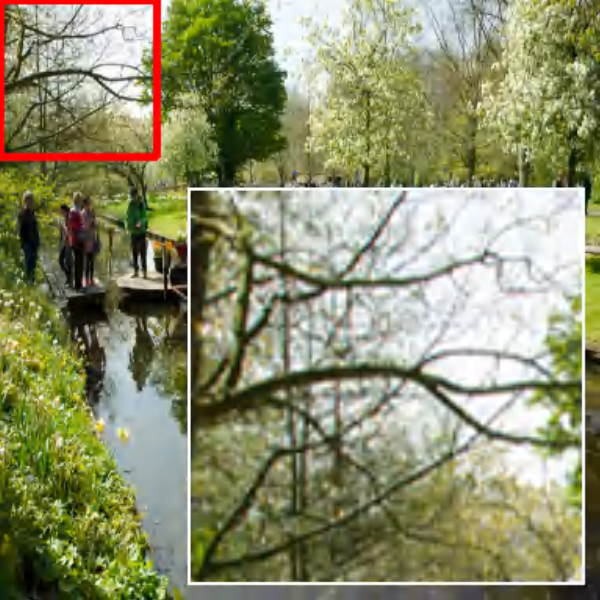}%
    \label{fig:ablation_a}%
  }\hspace{0.1cm}
  \subfloat[Raw Prior]{%
    \includegraphics[width=0.35\linewidth]{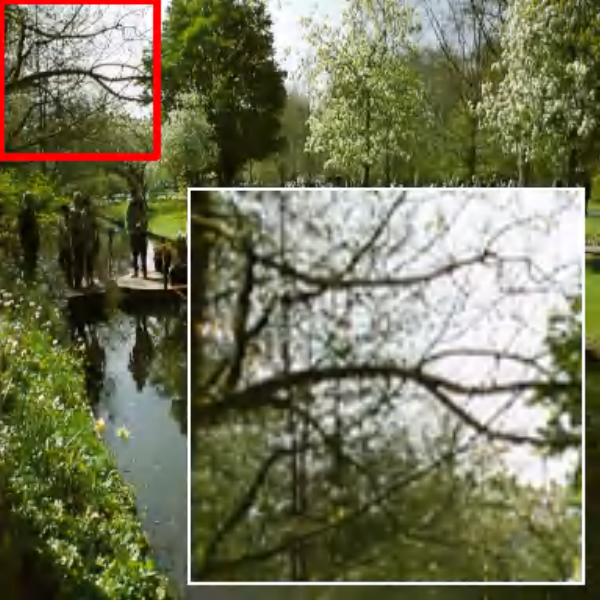}%
    \label{fig:ablation_b}%
  }  
  \vspace{-0.01mm}
  \subfloat[GT Prior]{%
    \includegraphics[width=0.35\linewidth]{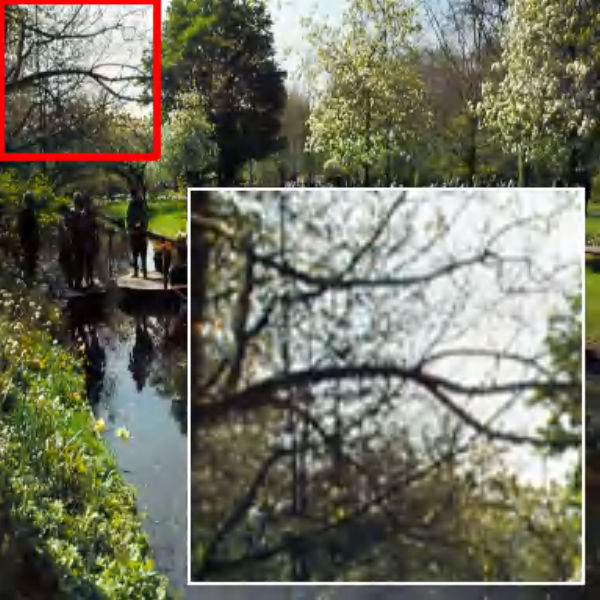}%
    \label{fig:ablation_c}%
  }\hspace{0.1cm}
  \subfloat[\textbf{Ours (ERP)}]{%
    \includegraphics[width=0.35\linewidth]{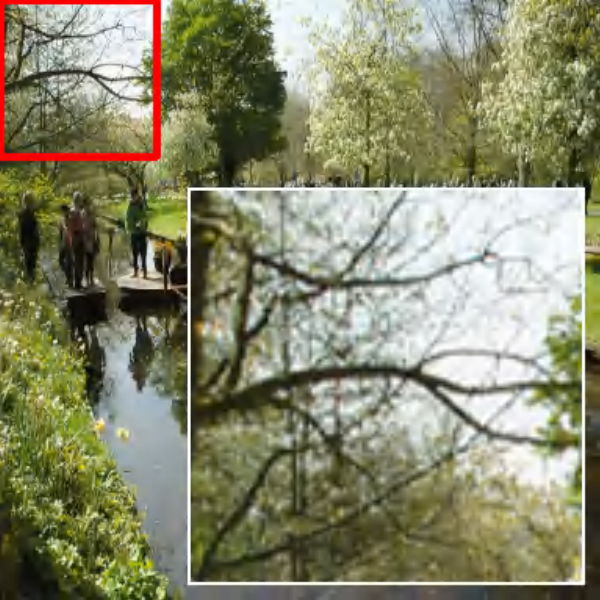}%
    \label{fig:ablation_d}%
  }
  \vspace{-0.2cm}
  \caption{Visual ablation comparison of different prior sources.}
  \vspace{-0.4cm}
  \label{fig:ablation_prior_vis}
\end{figure}

\begin{figure}[t]
    \centering
    \resizebox{\linewidth}{!}{ 
        \setlength{\tabcolsep}{0pt}    
        \renewcommand{\arraystretch}{1.2} 
        \arrayrulecolor{white}     
        \begin{tabular}{@{} c!{\vrule width 1pt}c!{\vrule width 1pt}c!{\vrule width 1pt}c!{\vrule width 1pt}c!{\vrule width 1pt}c!{\vrule width 1pt}c!{\vrule width 1pt}c!{\vrule width 1pt}c!{\vrule width 1pt}c @{}}
            \cellcolor{l_void}\parbox{2.5cm}{\centering\legfont Void} & 
            \cellcolor{l_road}\parbox{2.5cm}{\centering\legfont Road} & 
            \cellcolor{l_sidewalk}\parbox{2.5cm}{\centering\legfont Sidewalk} & 
            \cellcolor{l_building}\parbox{2.5cm}{\centering\legfont Building} & 
            \cellcolor{l_wall}\parbox{2.5cm}{\centering\legfont Wall} & 
            \cellcolor{l_fence}\parbox{2.5cm}{\centering\legfont Fence} & 
            \cellcolor{l_pole}\parbox{2.5cm}{\centering\legfont Pole} & 
            \cellcolor{l_light}\parbox{2.5cm}{\centering\legfontblk Traffic Light} & 
            \cellcolor{l_sign}\parbox{2.5cm}{\centering\legfontblk Traffic Sign} & 
            \cellcolor{l_veg}\parbox{2.5cm}{\centering\legfont Vegetation} \\
            \hline 
            \cellcolor{l_terrain}\parbox{2.5cm}{\centering\legfontblk Terrain} & 
            \cellcolor{l_sky}\parbox{2.5cm}{\centering\legfont Sky} & 
            \cellcolor{l_people}\parbox{2.5cm}{\centering\legfont People} & 
            \cellcolor{l_rider}\parbox{2.5cm}{\centering\legfont Rider} & 
            \cellcolor{l_car}\parbox{2.5cm}{\centering\legfont Car} & 
            \cellcolor{l_truck}\parbox{2.5cm}{\centering\legfont Truck} & 
            \cellcolor{l_bus}\parbox{2.5cm}{\centering\legfont Bus} & 
            \cellcolor{l_train}\parbox{2.5cm}{\centering\legfont Train} & 
            \cellcolor{l_moto}\parbox{2.5cm}{\centering\legfont Motorcycle} & 
            \cellcolor{l_bike}\parbox{2.5cm}{\centering\legfont Bicycle} \\
        \end{tabular}
    }
    \vspace{1pt} 
    \resizebox{\linewidth}{!}{
        \setlength{\tabcolsep}{0.5pt}
        \renewcommand{\arraystretch}{0.6}
        \begin{tabular}{@{} c@{\hspace{1.5mm}} cccccc @{}} 
            \sidefont Input &
            \includegraphics[width=0.155\linewidth]{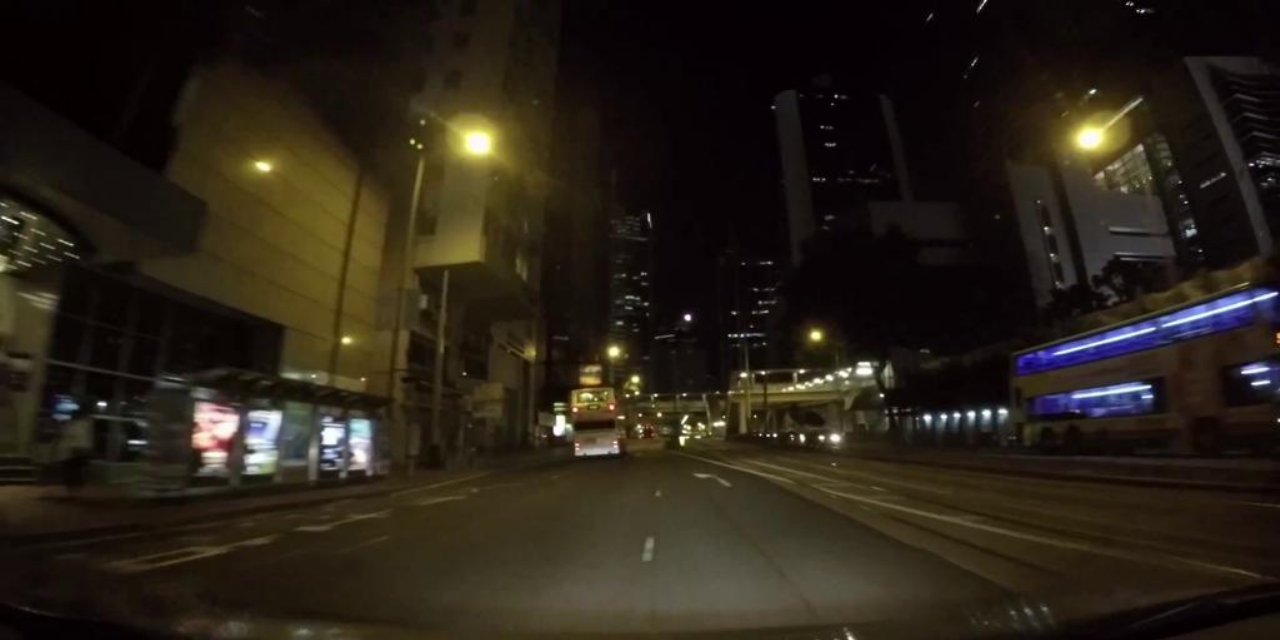} &
            \includegraphics[width=0.155\linewidth]{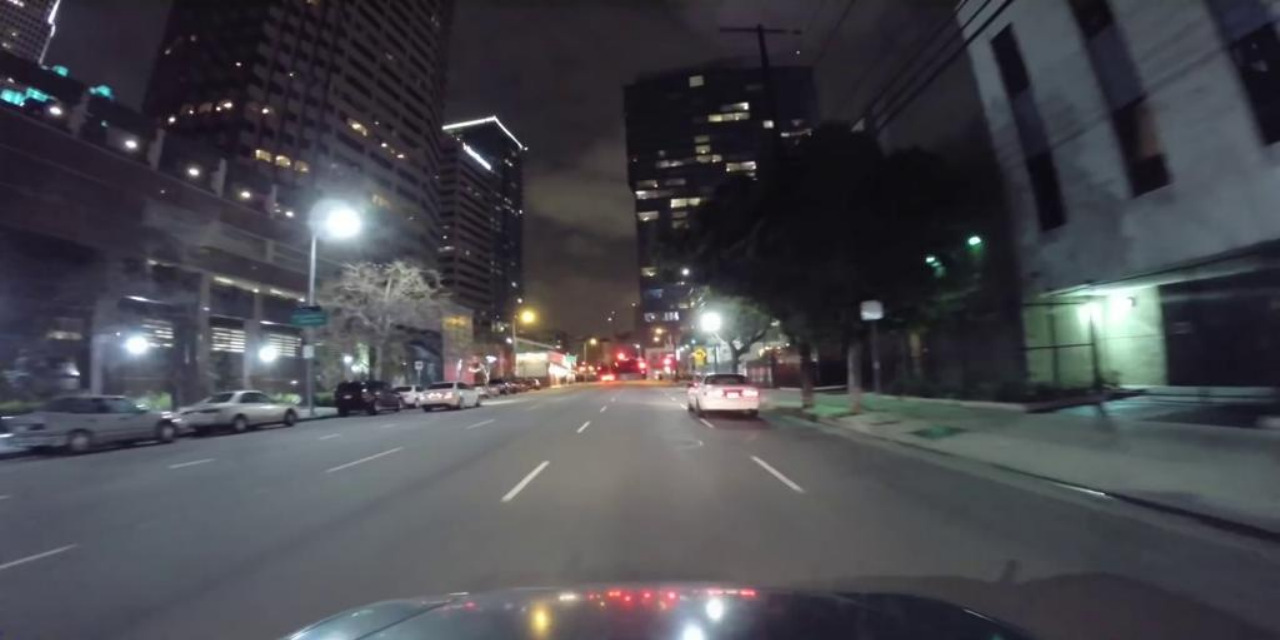} &
            \includegraphics[width=0.155\linewidth]{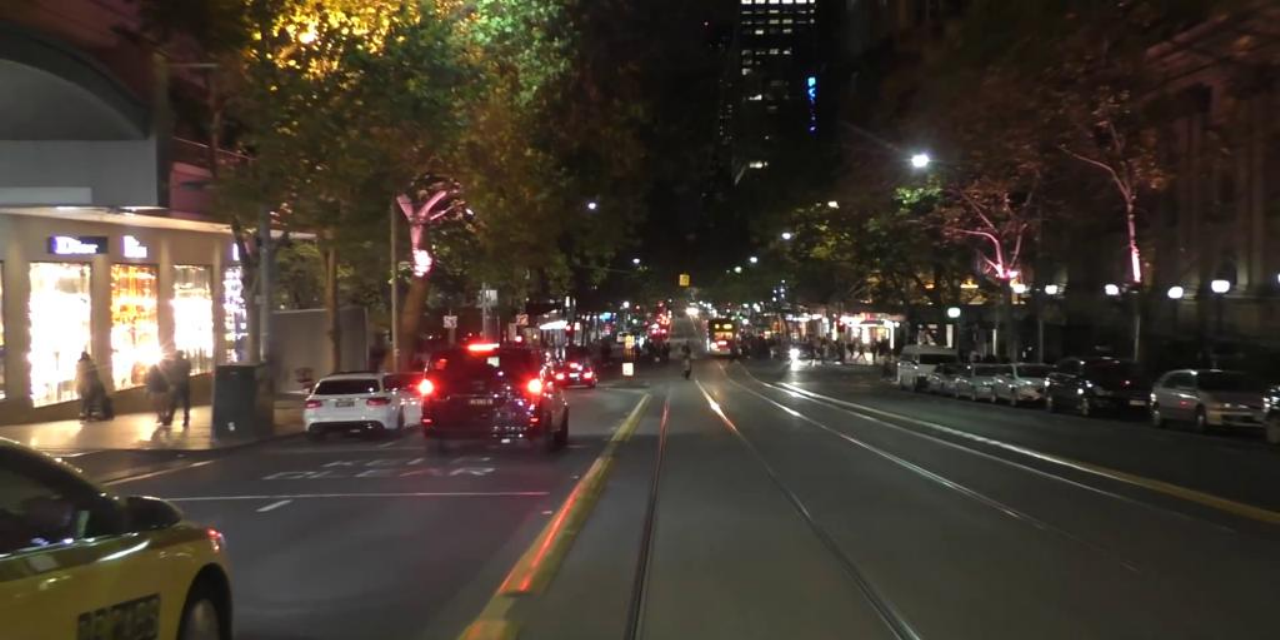} &
            \includegraphics[width=0.155\linewidth]{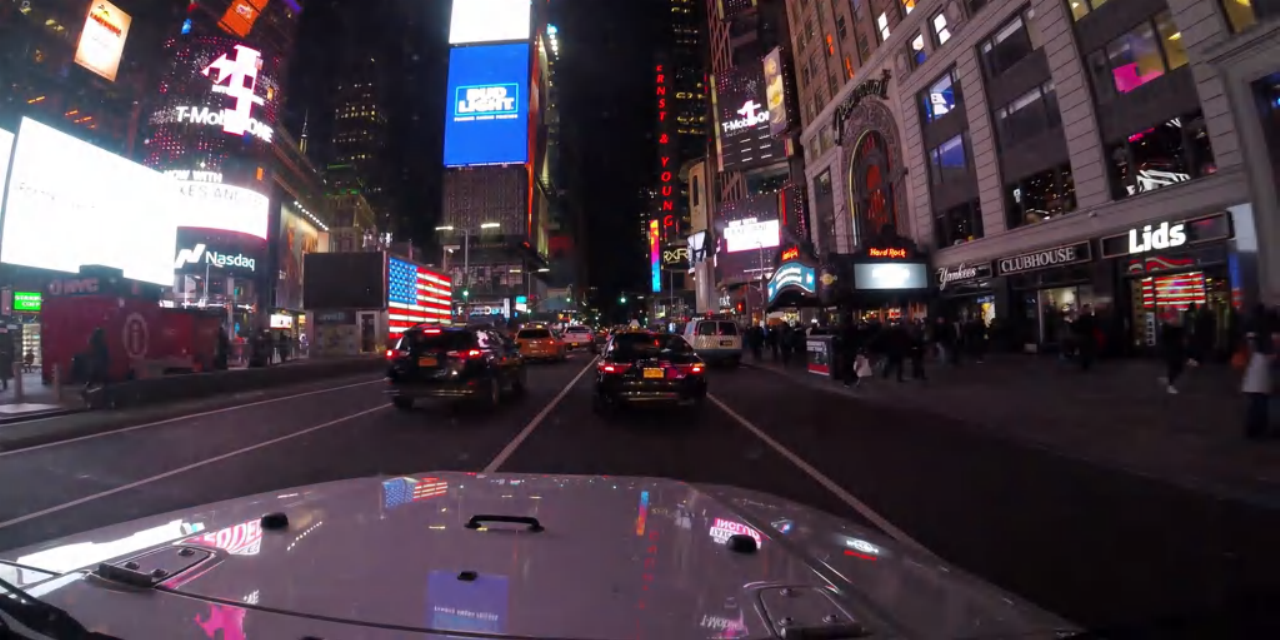} &
            \includegraphics[width=0.155\linewidth]{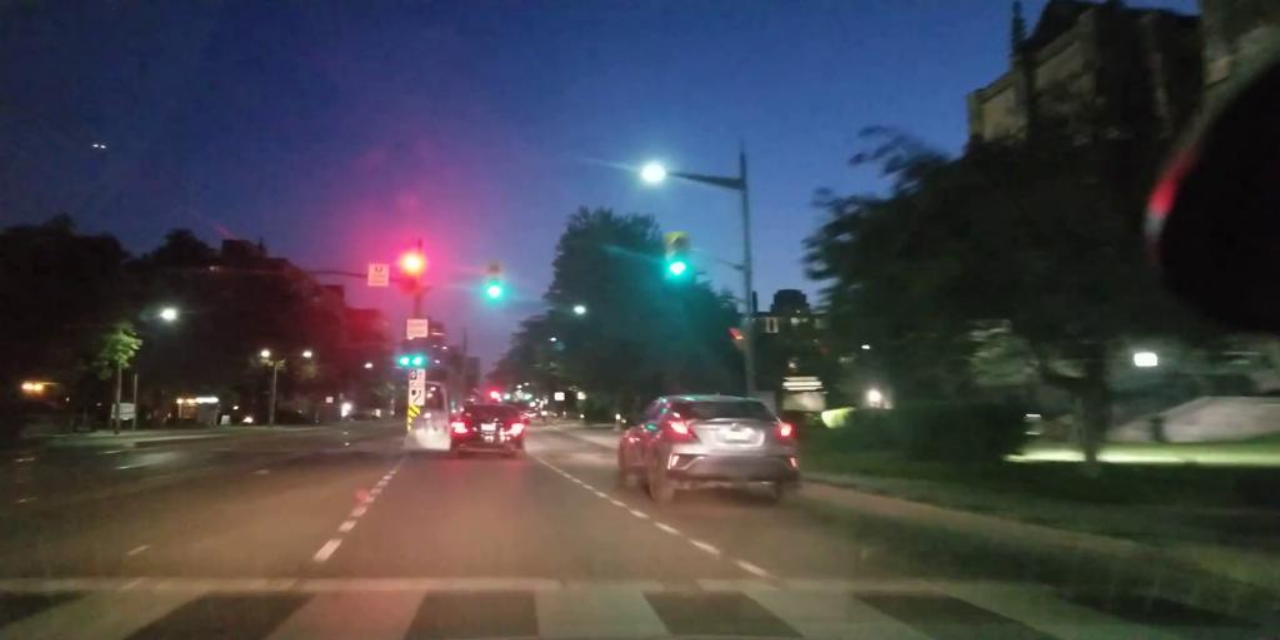} &
            \includegraphics[width=0.155\linewidth]{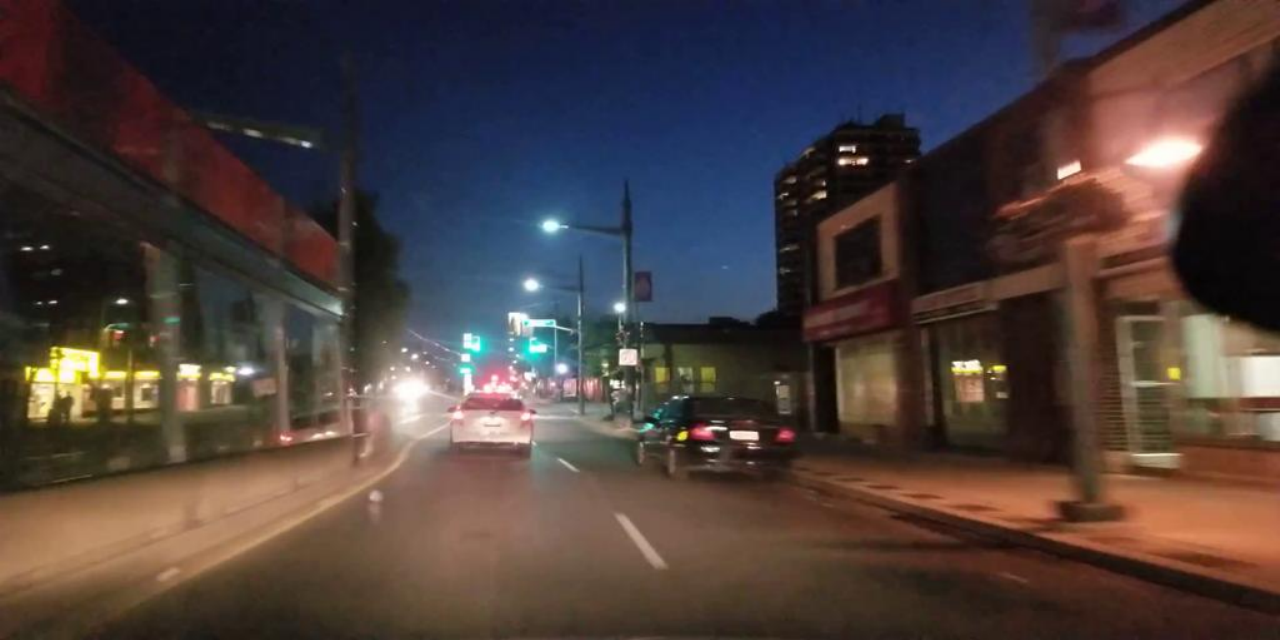} \\
            \sidefont \shortstack{Enhanced\\(Ours)} &
            \includegraphics[width=0.155\linewidth]{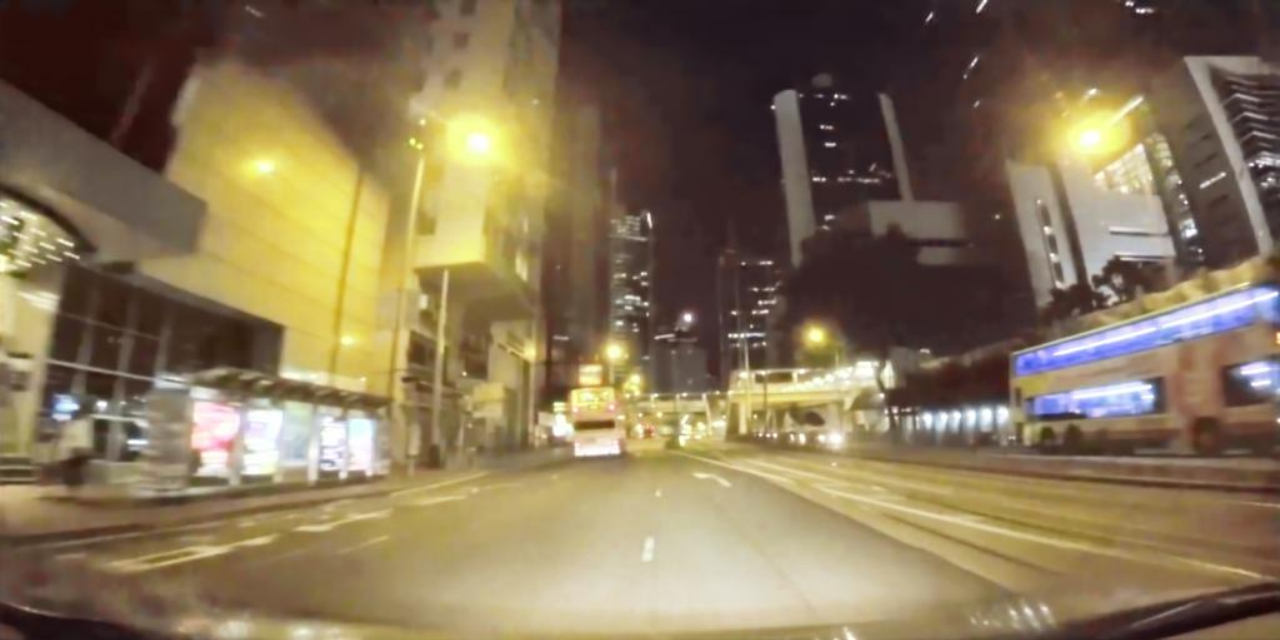} &
            \includegraphics[width=0.155\linewidth]{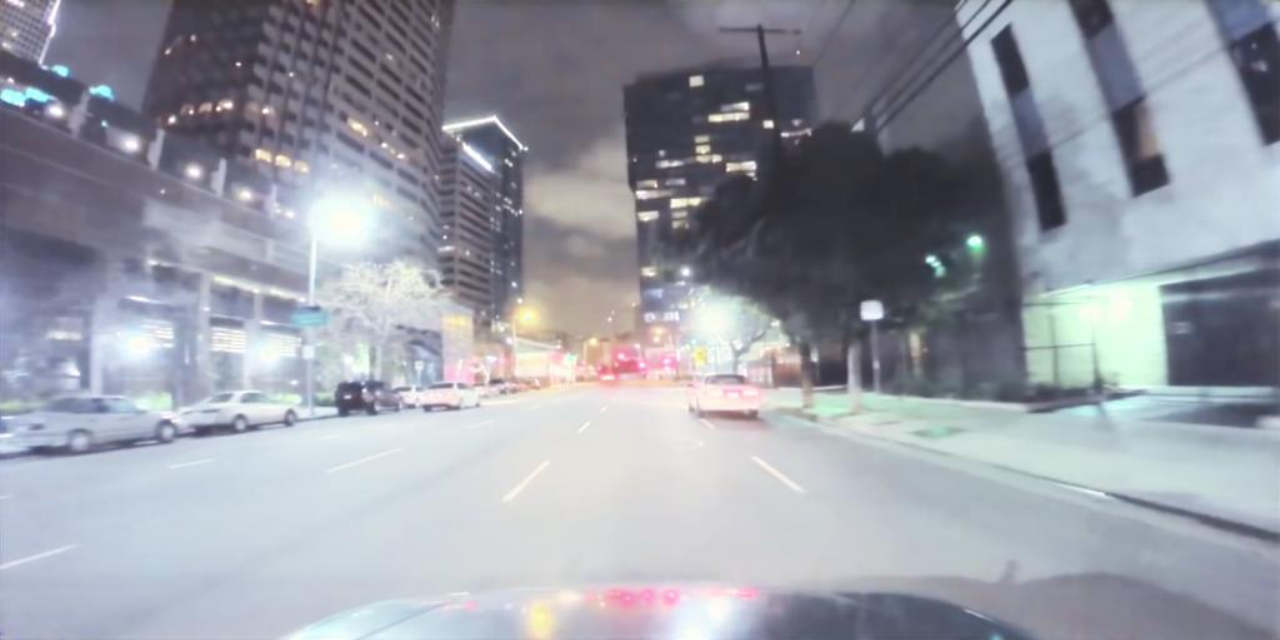} &
            \includegraphics[width=0.155\linewidth]{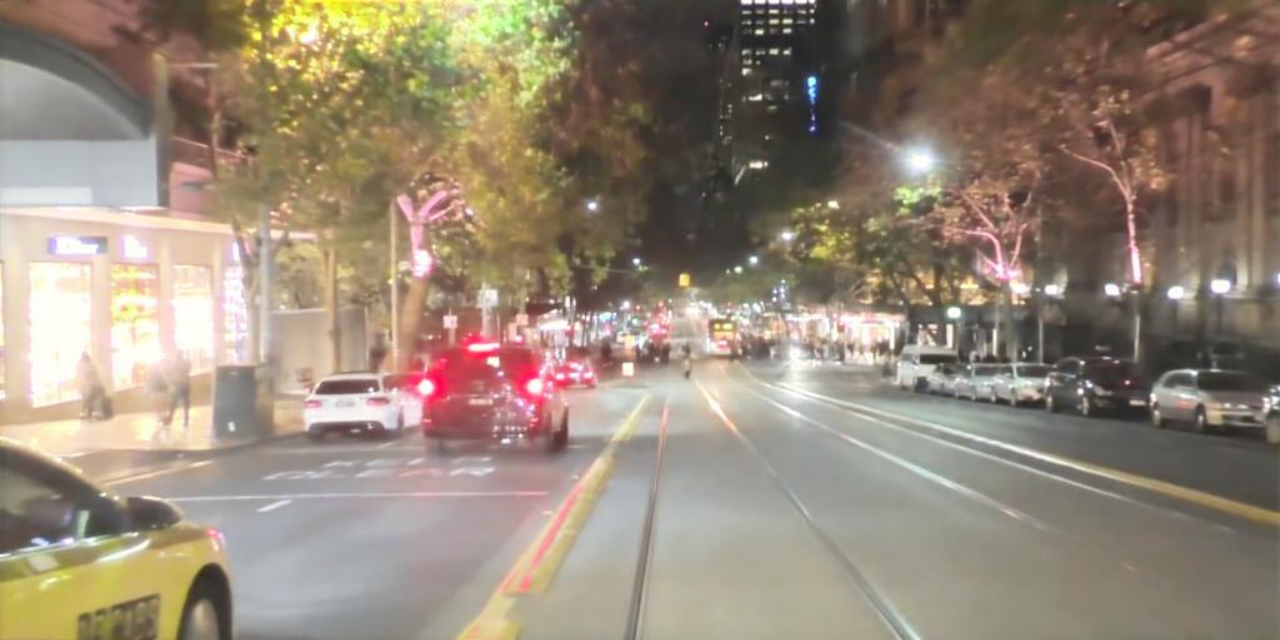} &
            \includegraphics[width=0.155\linewidth]{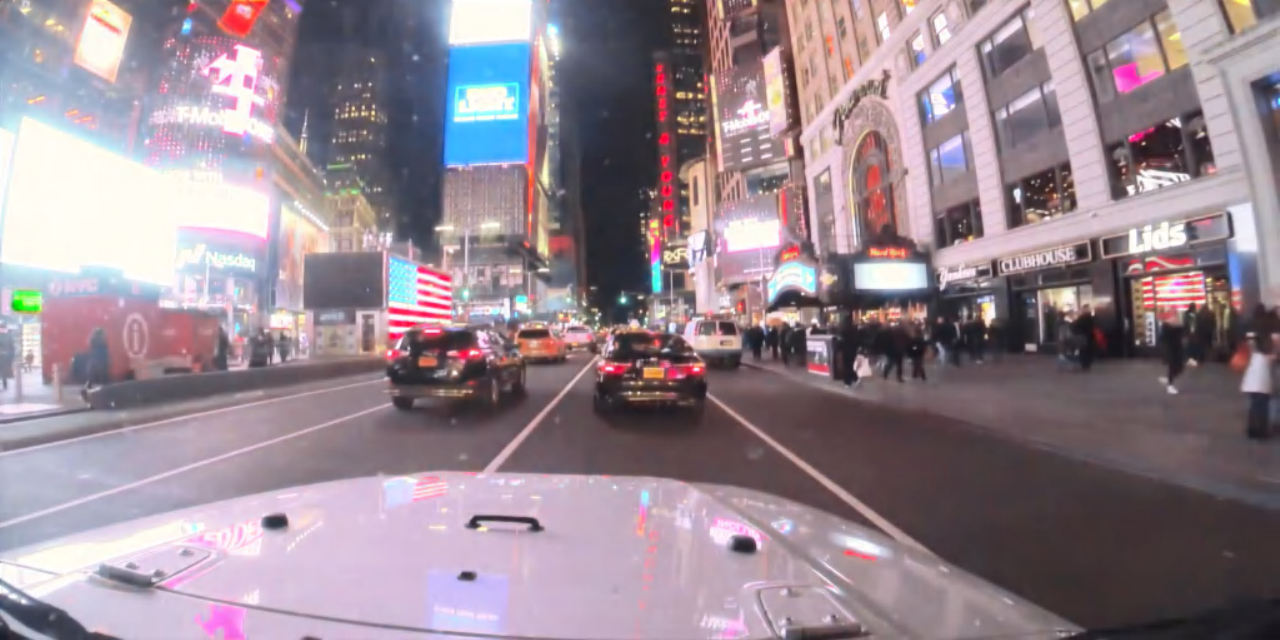} &
            \includegraphics[width=0.155\linewidth]{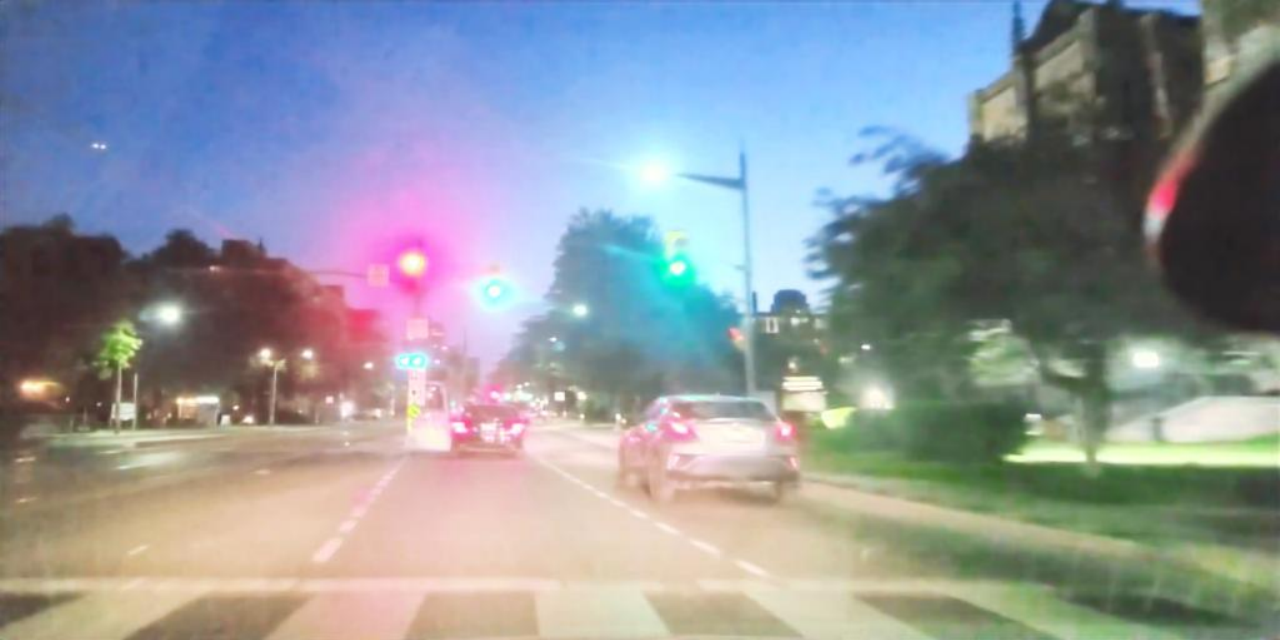} &
            \includegraphics[width=0.155\linewidth]{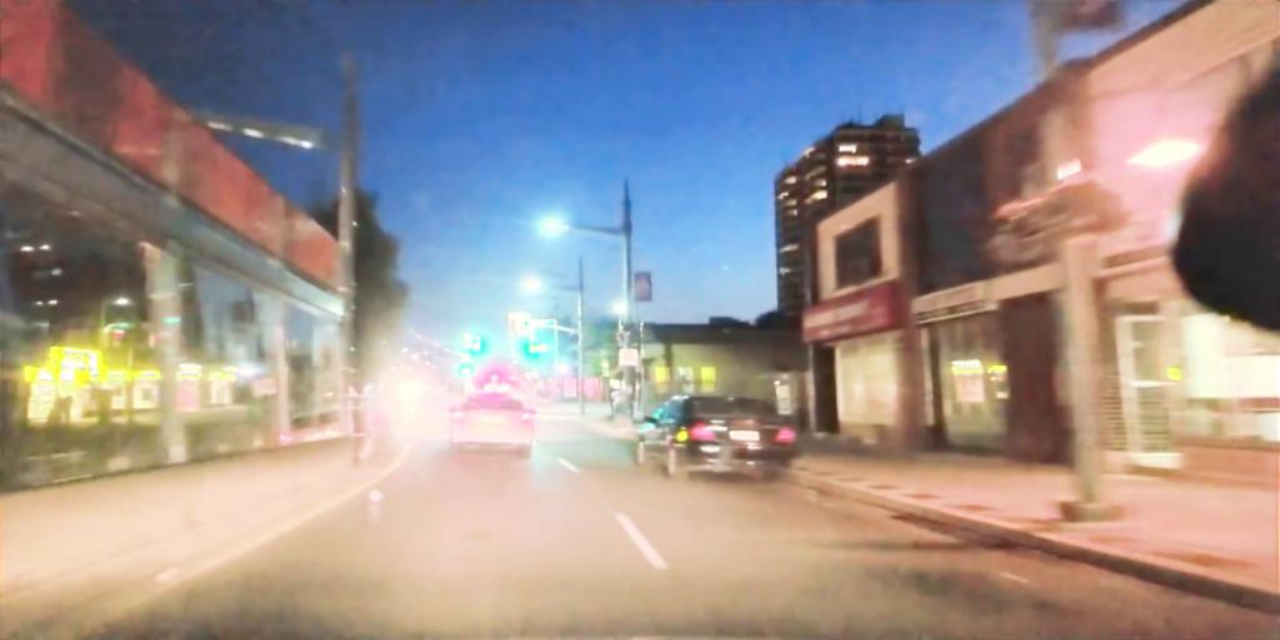} \\
            \sidefont CWNet &
            \includegraphics[width=0.155\linewidth]{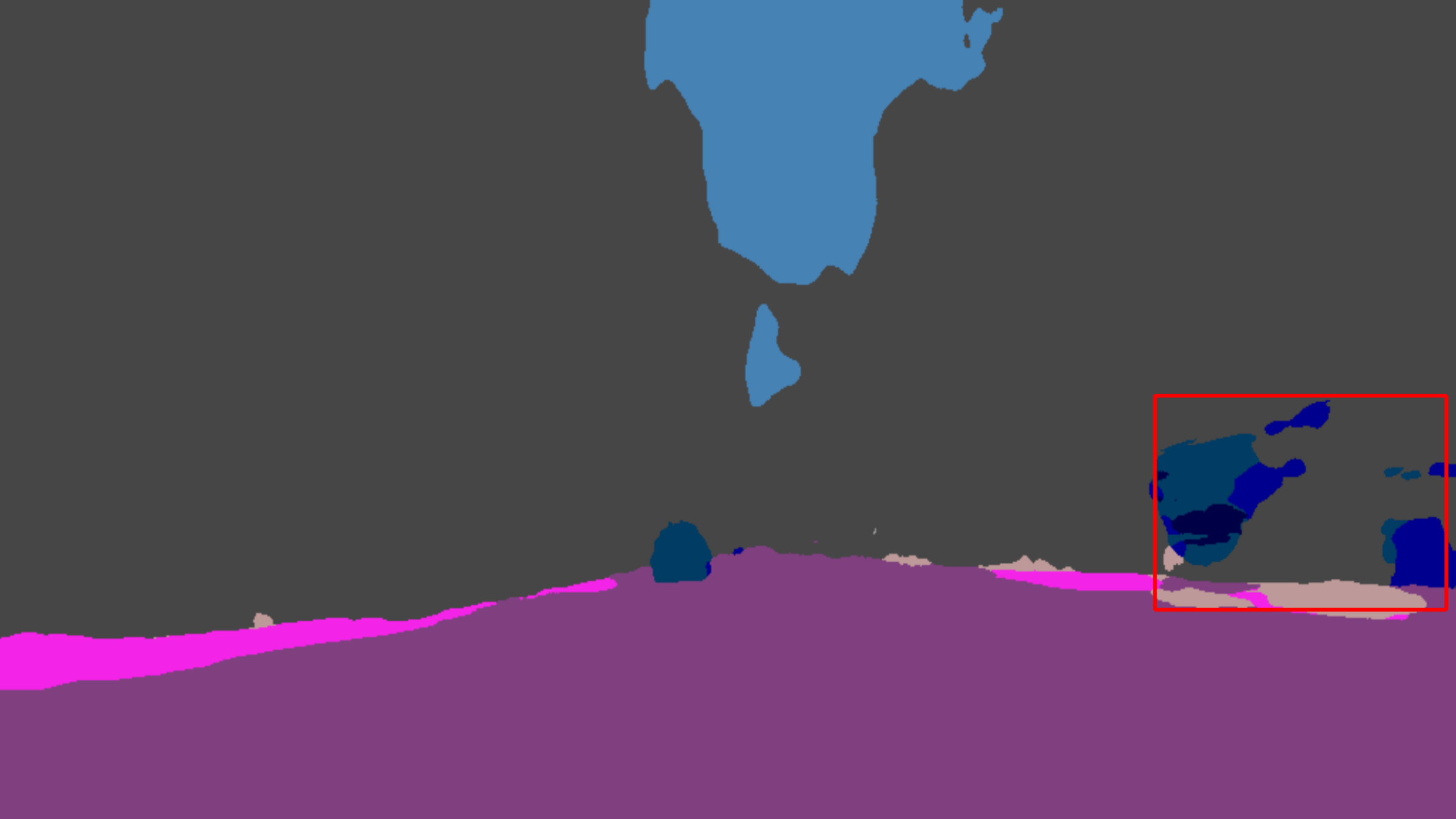} &
            \includegraphics[width=0.155\linewidth]{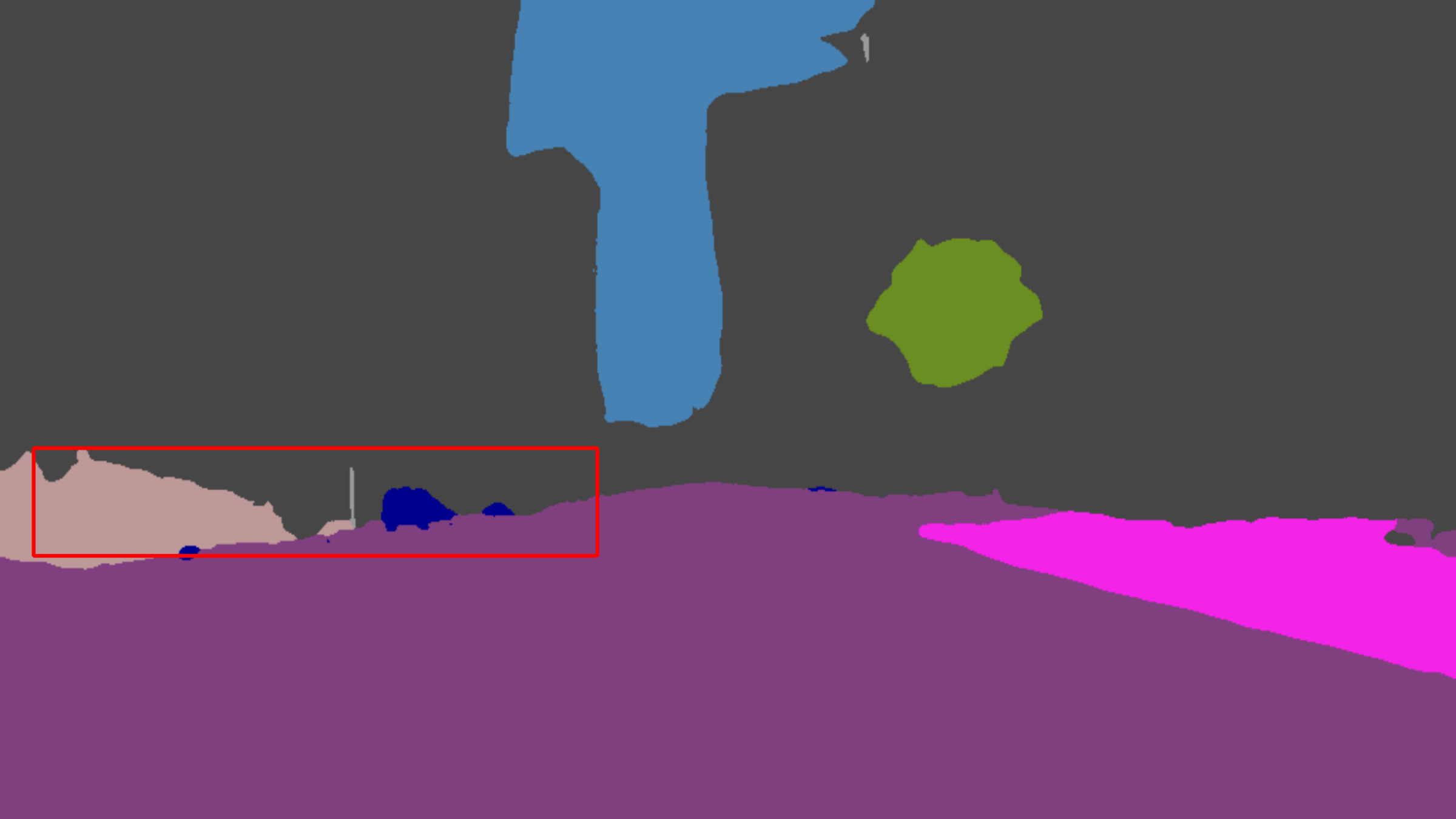} &
            \includegraphics[width=0.155\linewidth]{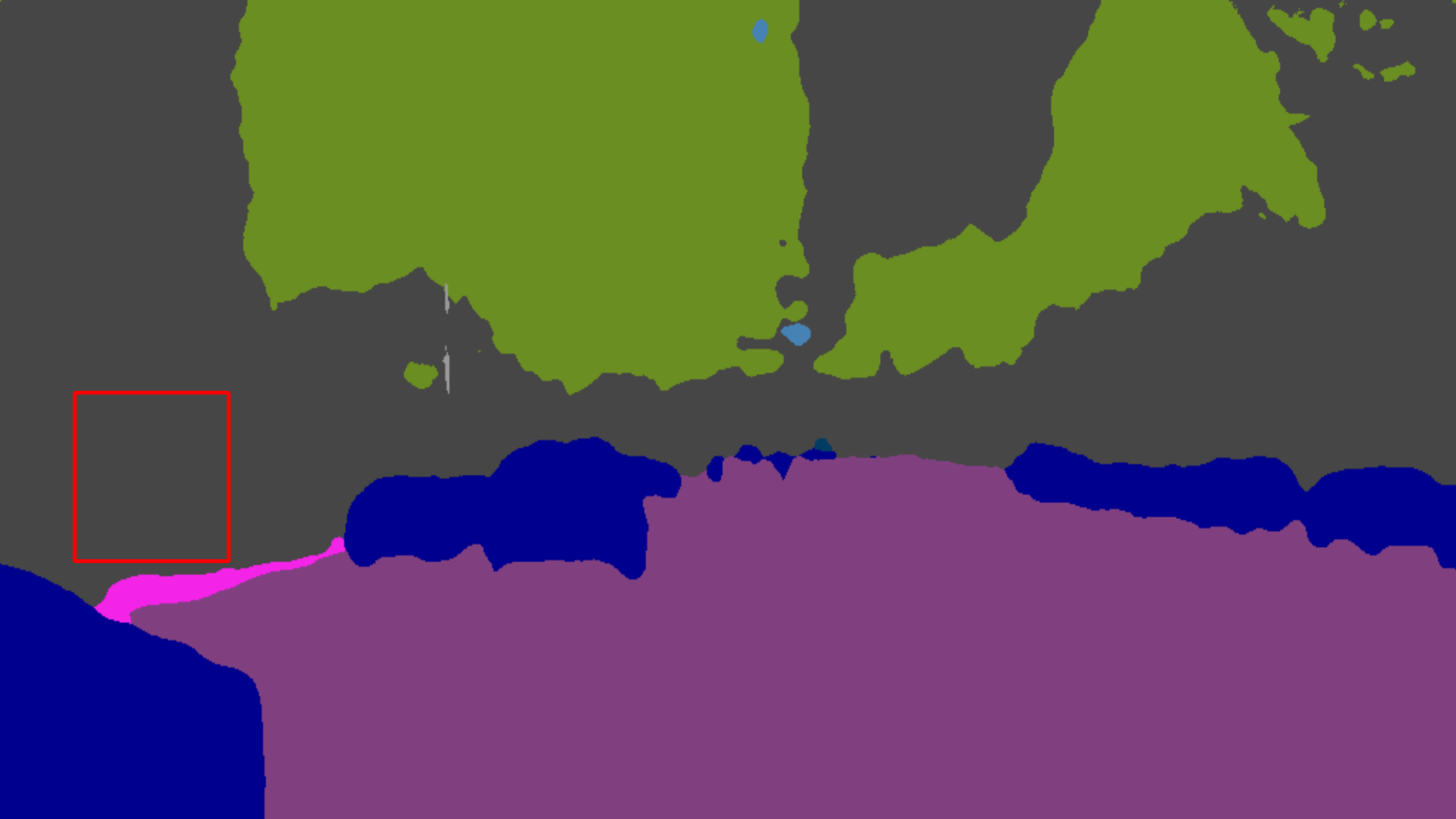} &
            \includegraphics[width=0.155\linewidth]{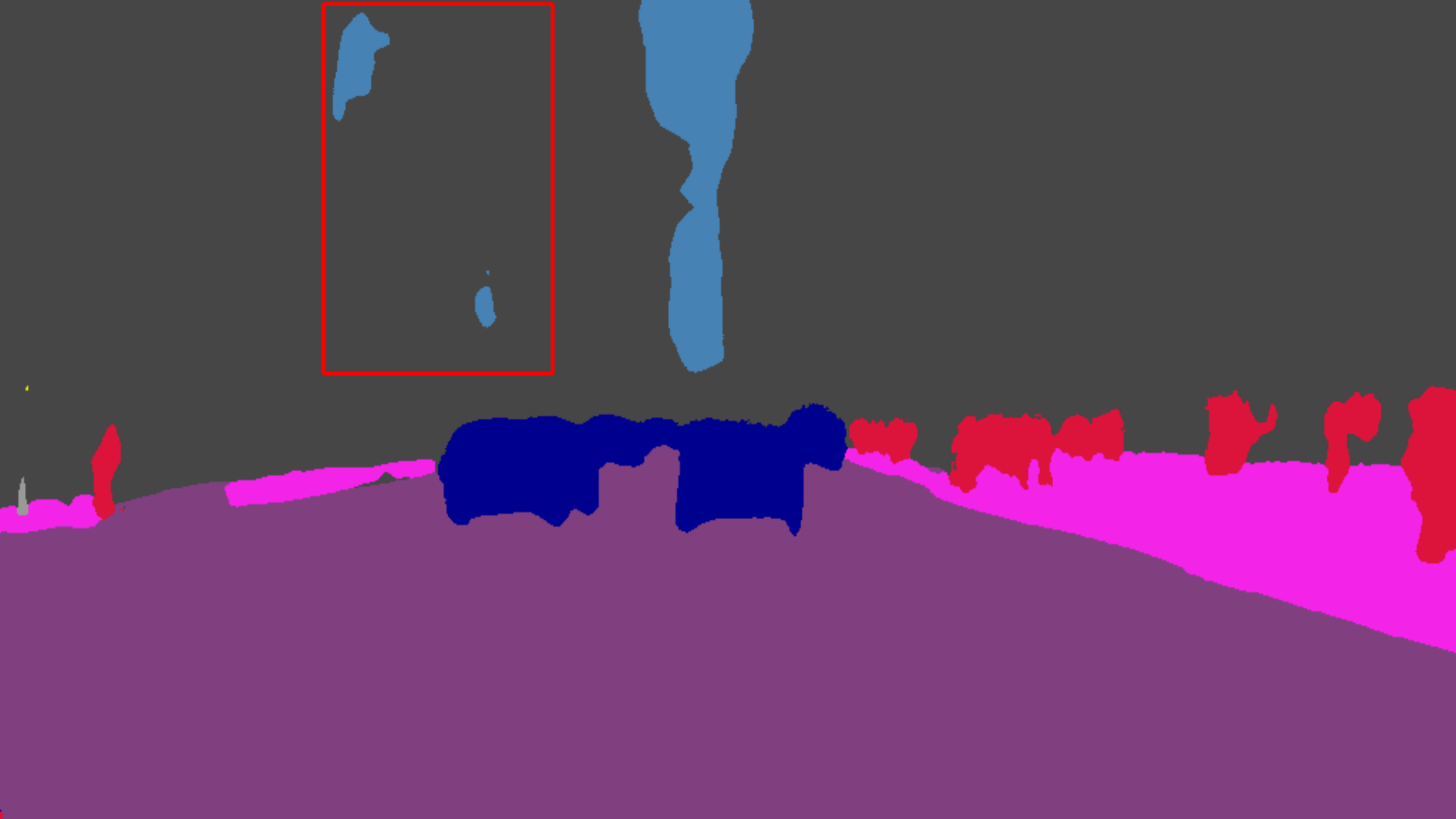} &
            \includegraphics[width=0.155\linewidth]{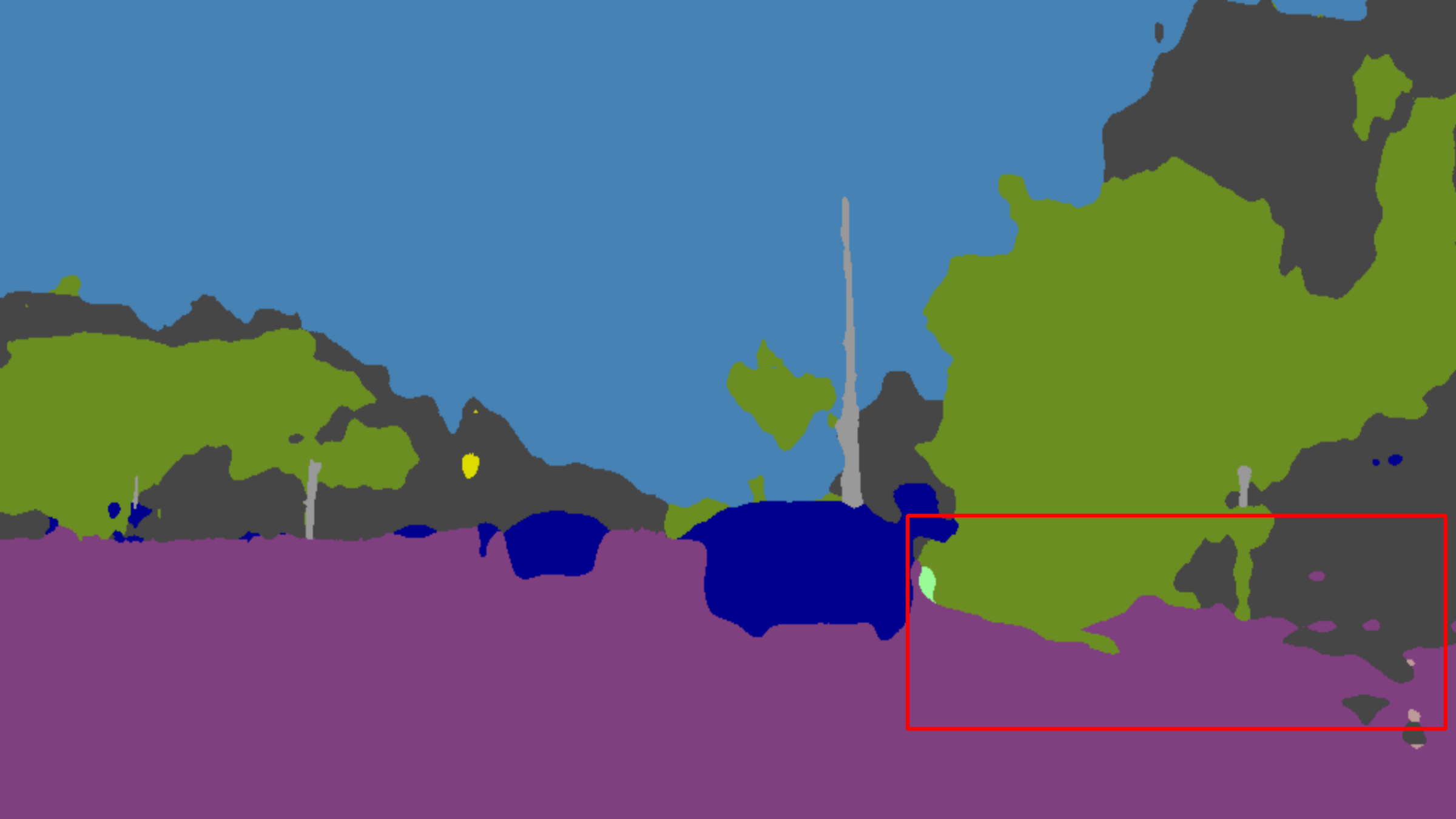} &
            \includegraphics[width=0.155\linewidth]{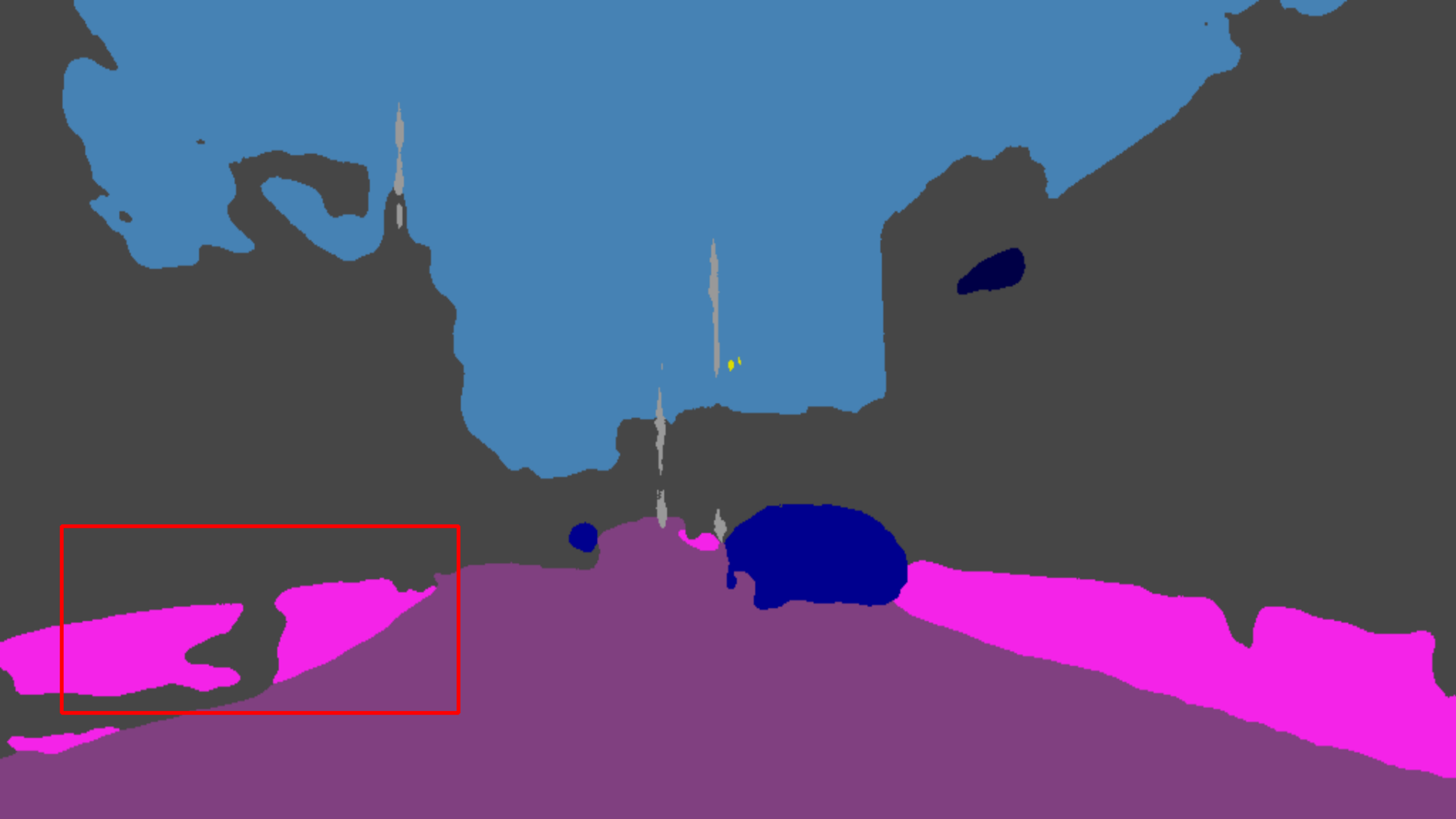} \\
            \sidefont FourLLIE &
            \includegraphics[width=0.155\linewidth]{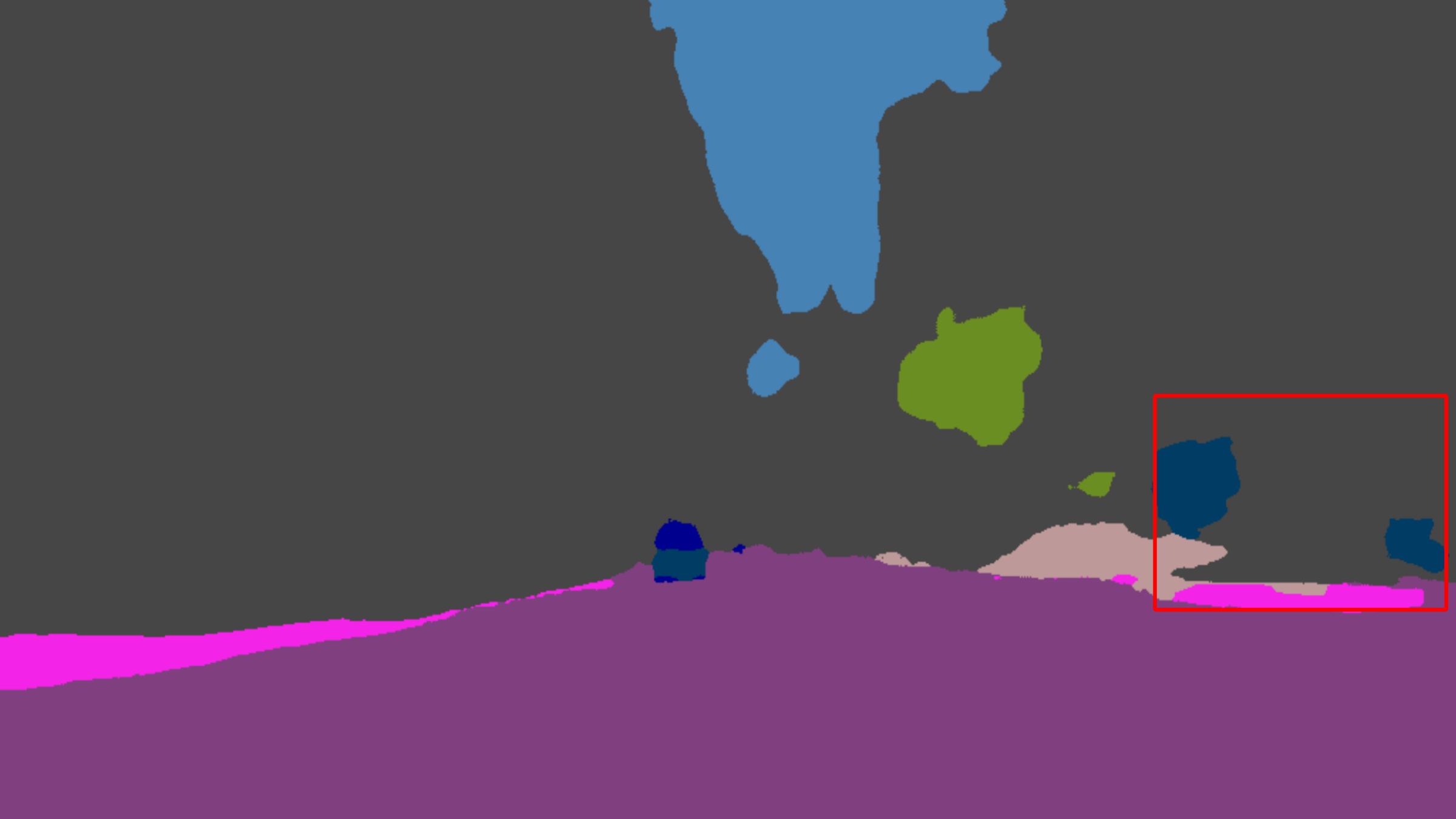} &
            \includegraphics[width=0.155\linewidth]{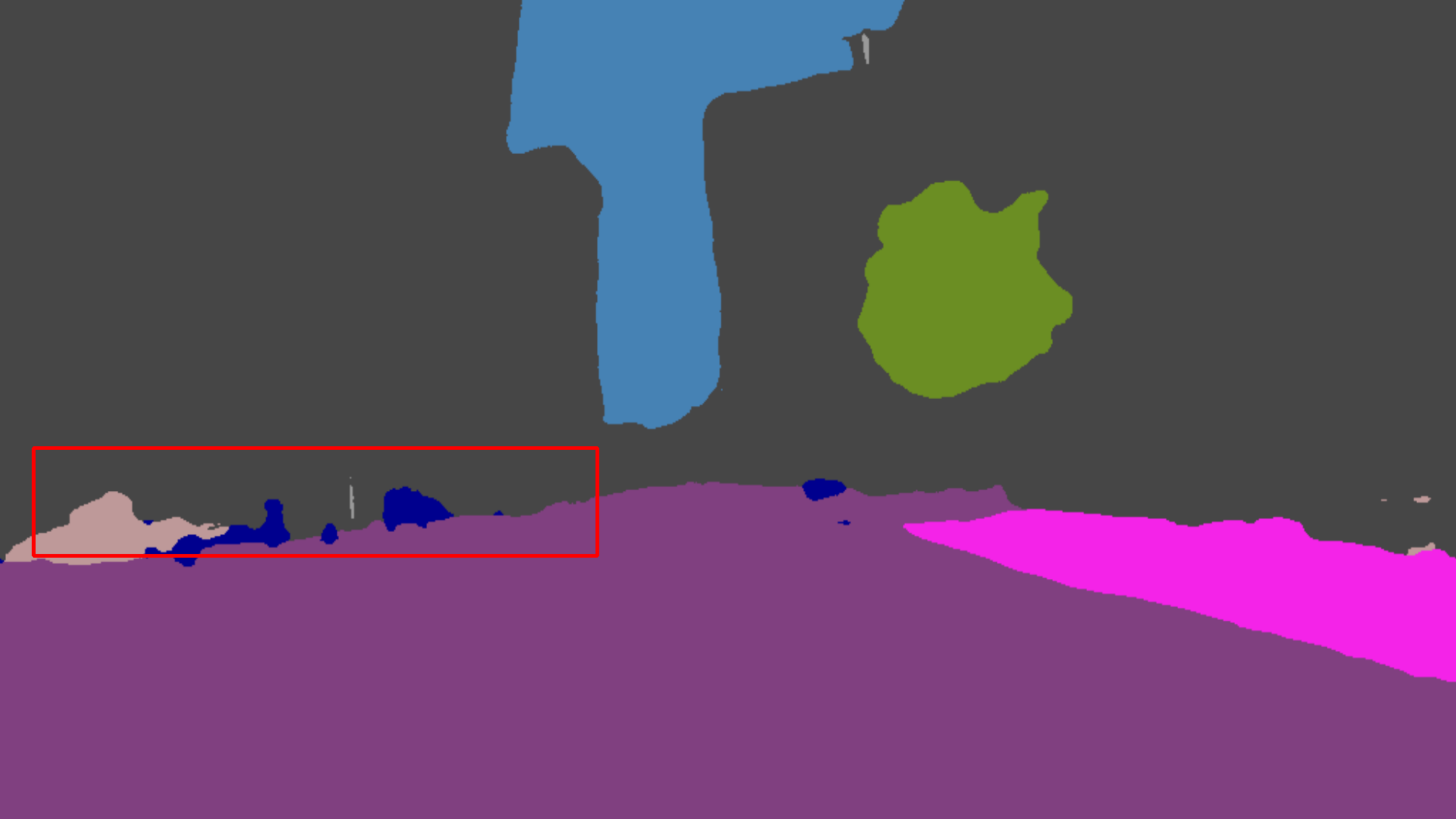} &
            \includegraphics[width=0.155\linewidth]{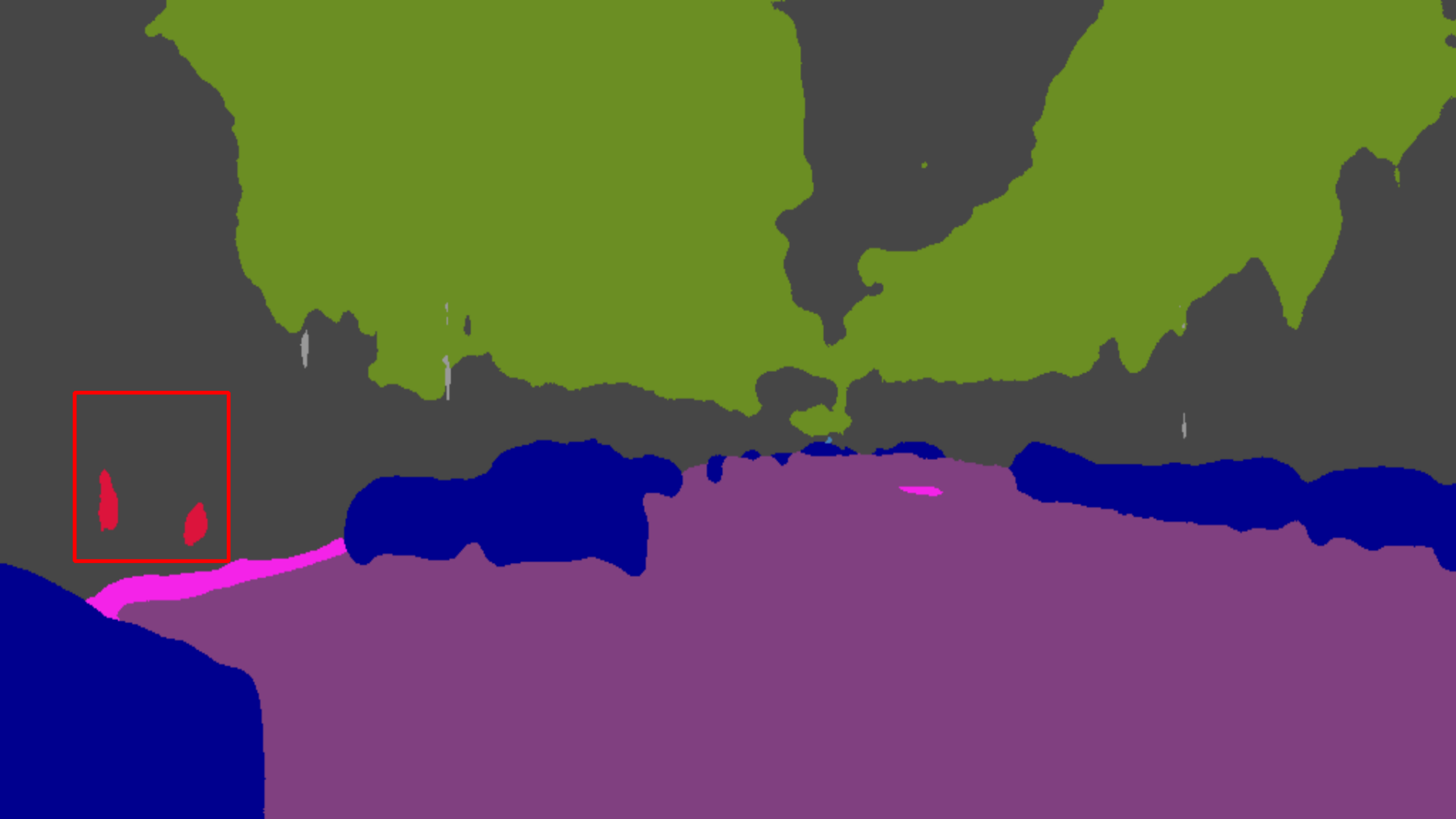} &
            \includegraphics[width=0.155\linewidth]{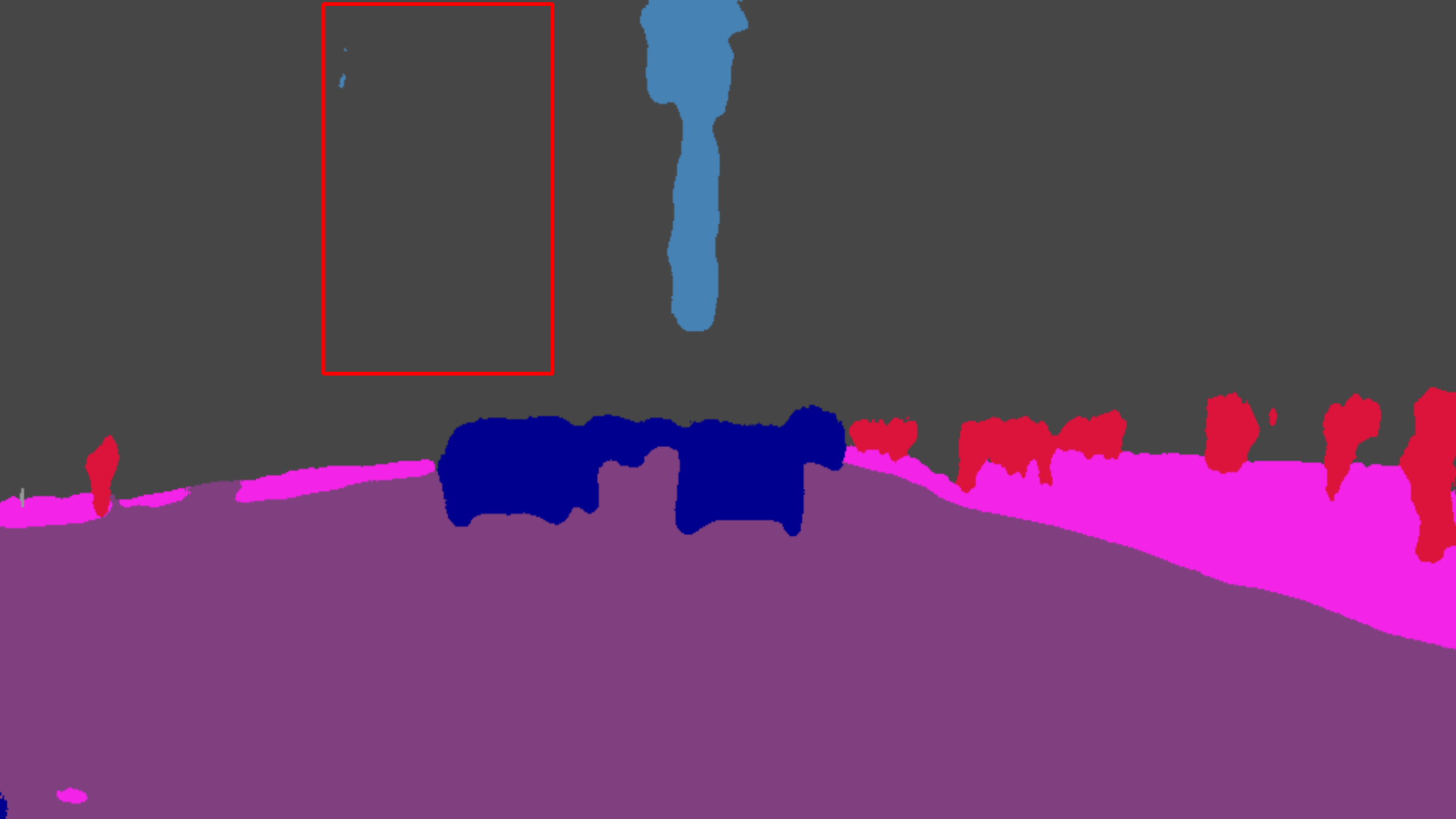} &
            \includegraphics[width=0.155\linewidth]{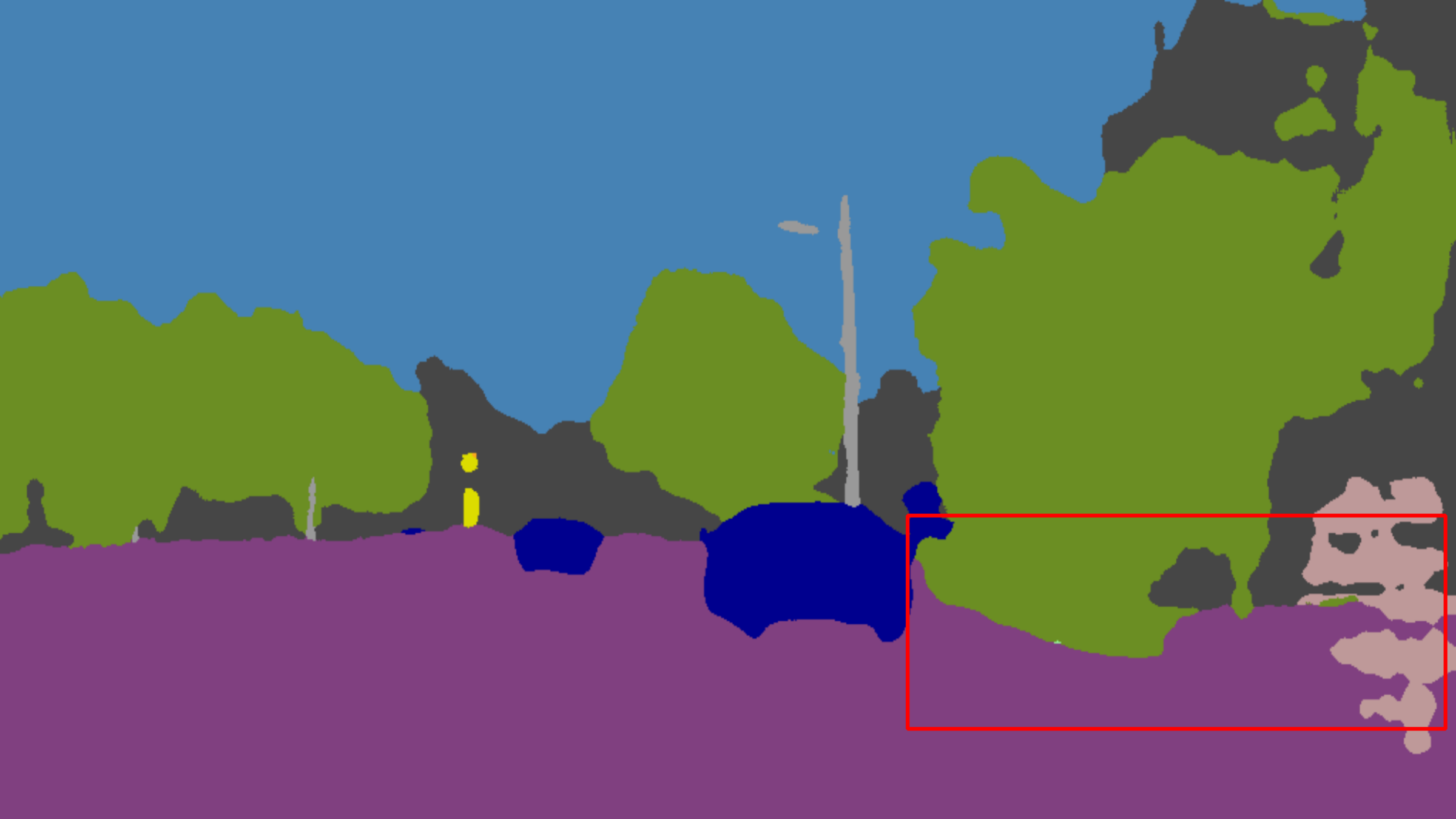} &
            \includegraphics[width=0.155\linewidth]{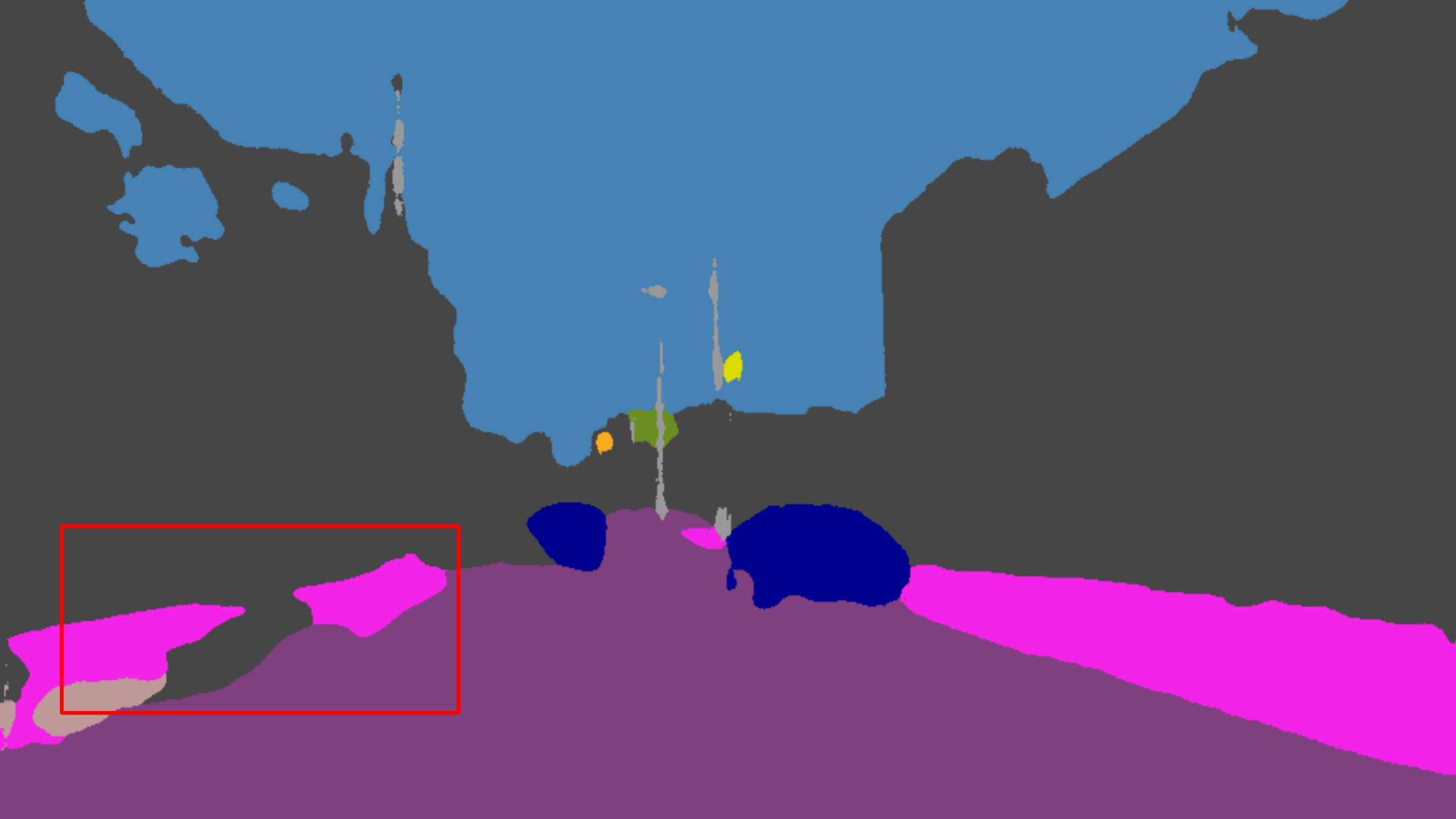} \\
            \sidefont Wave-Mam. &
            \includegraphics[width=0.155\linewidth]{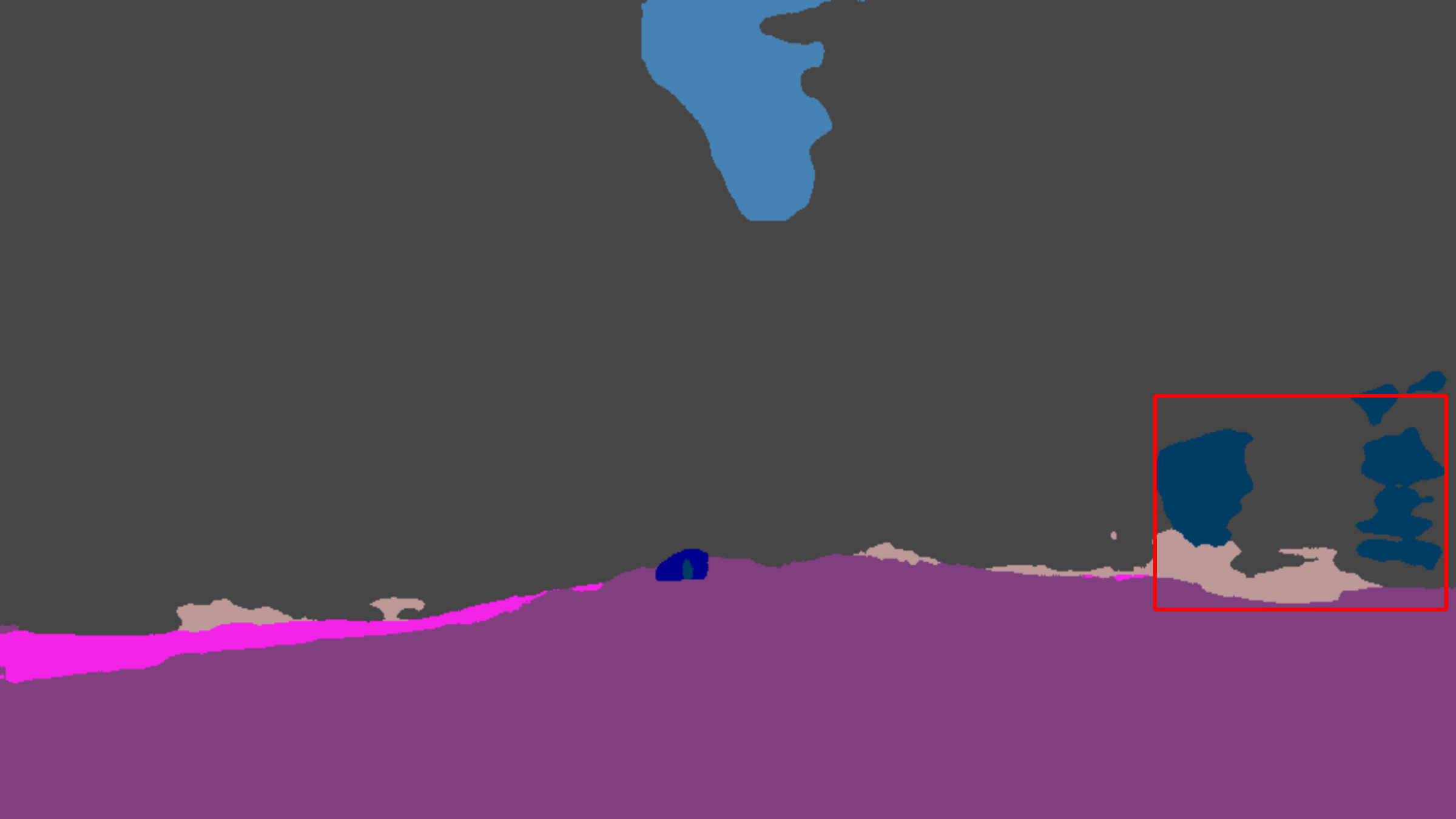} &
            \includegraphics[width=0.155\linewidth]{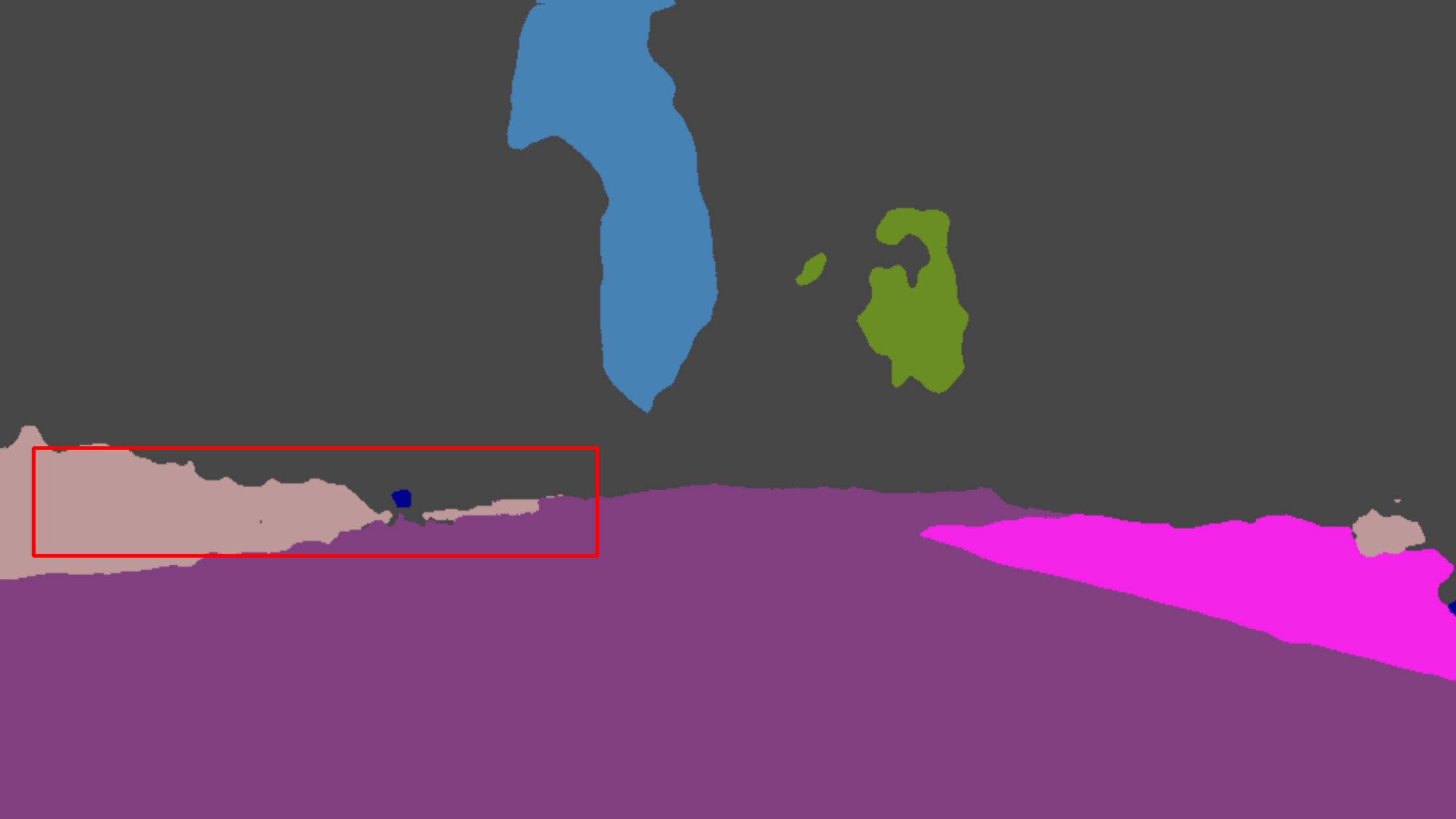} &
            \includegraphics[width=0.155\linewidth]{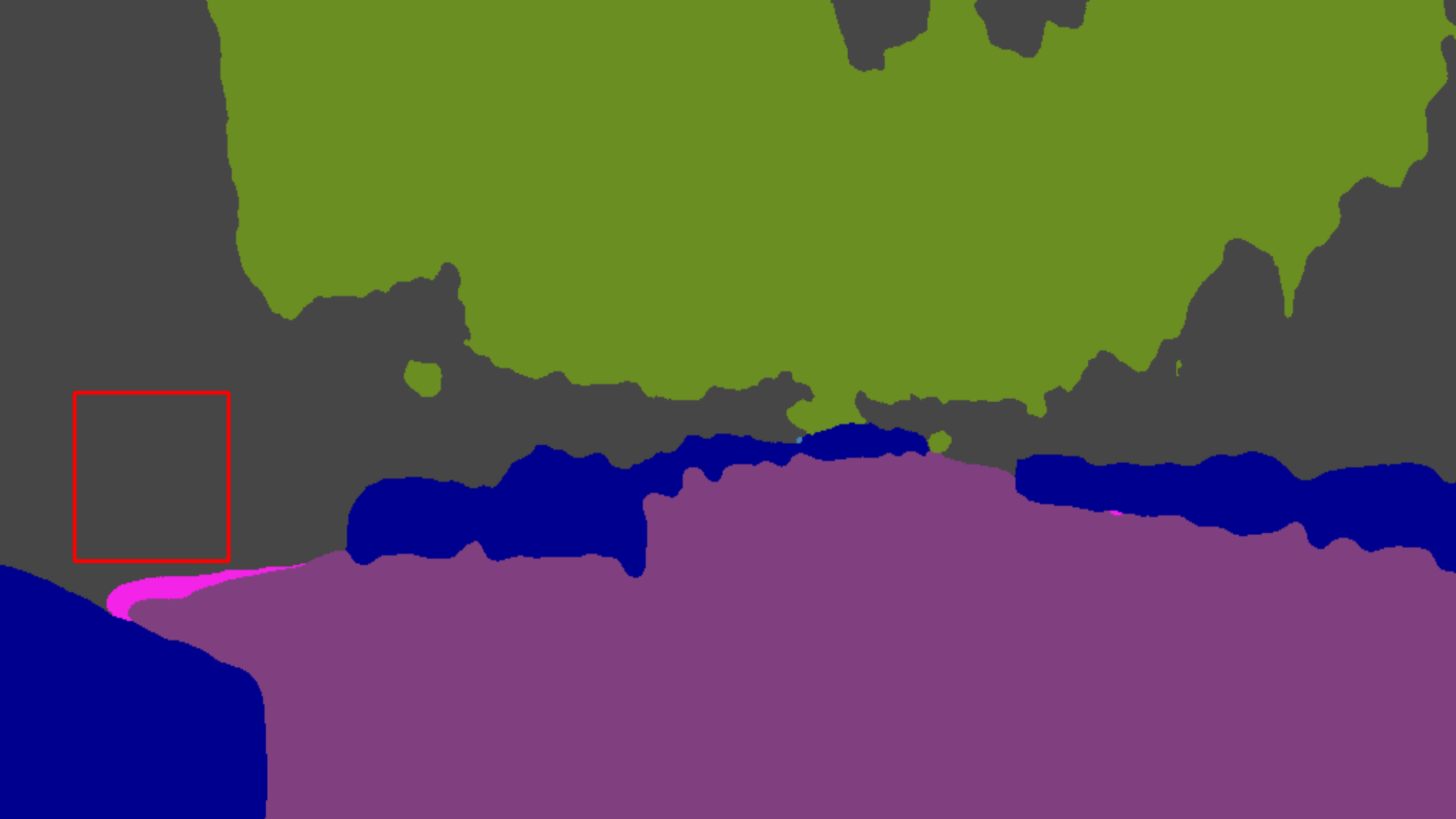} &
            \includegraphics[width=0.155\linewidth]{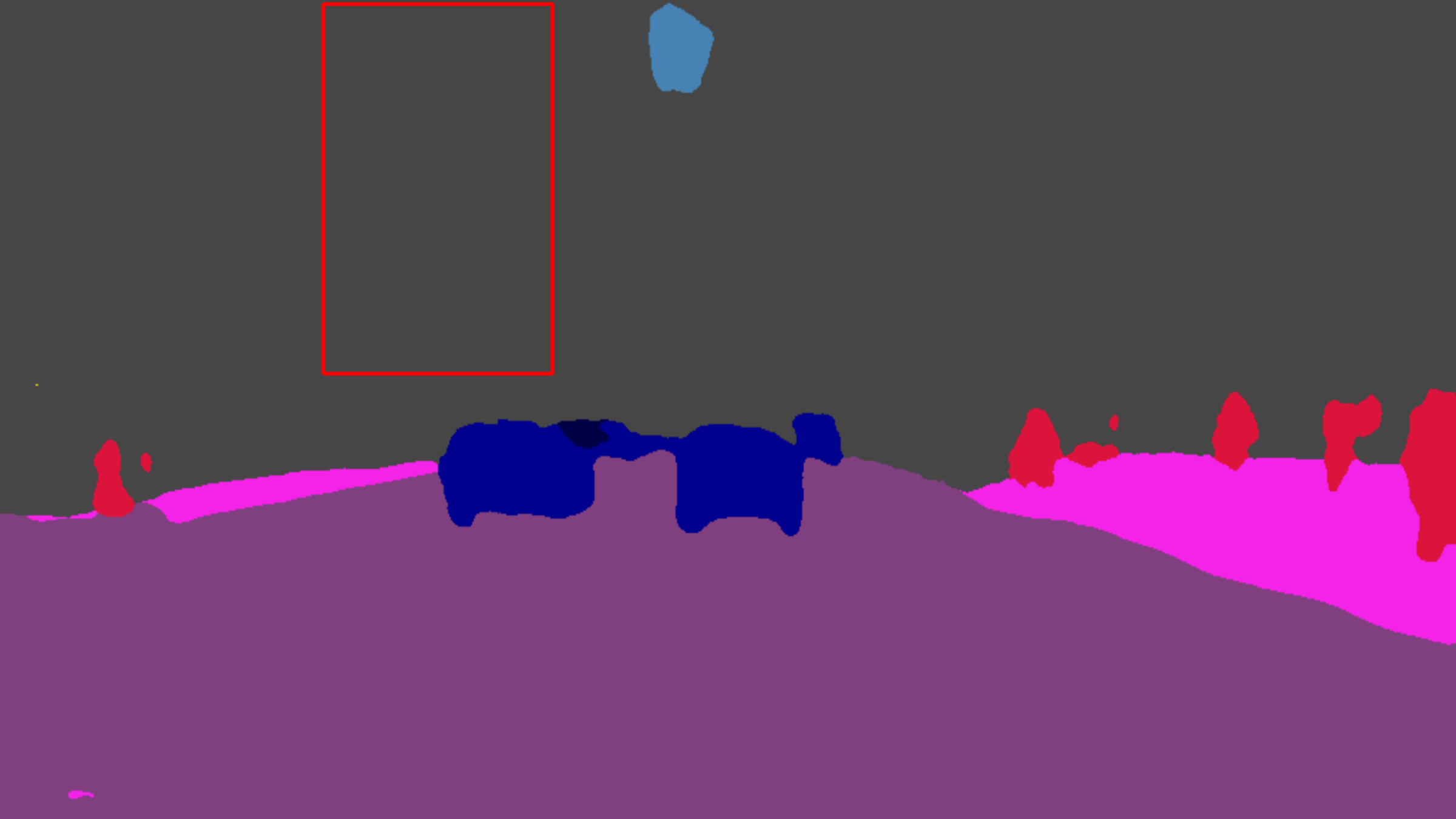} &
            \includegraphics[width=0.155\linewidth]{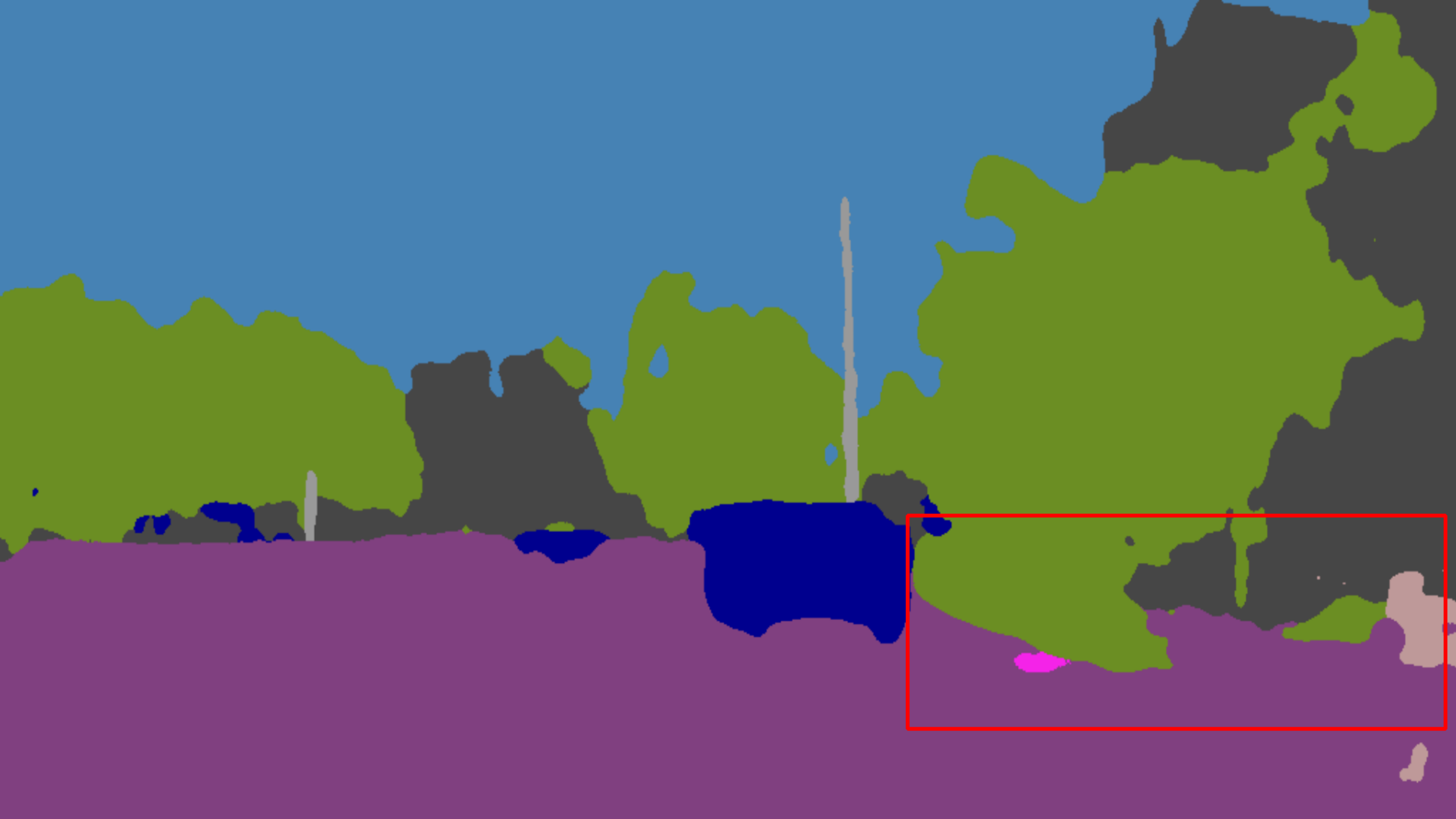} &
            \includegraphics[width=0.155\linewidth]{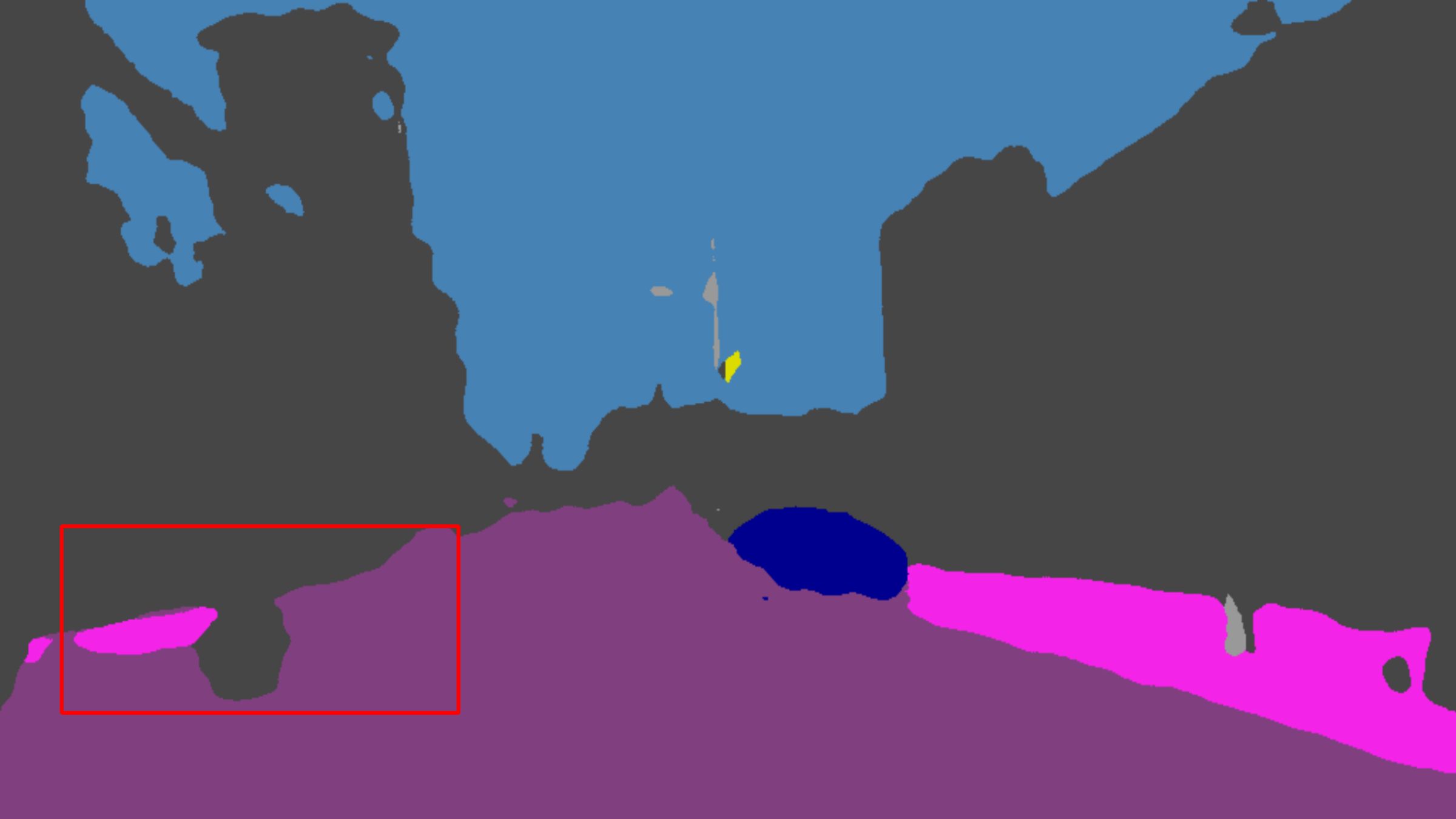} \\
            \sidefont UHDFour &
            \includegraphics[width=0.155\linewidth]{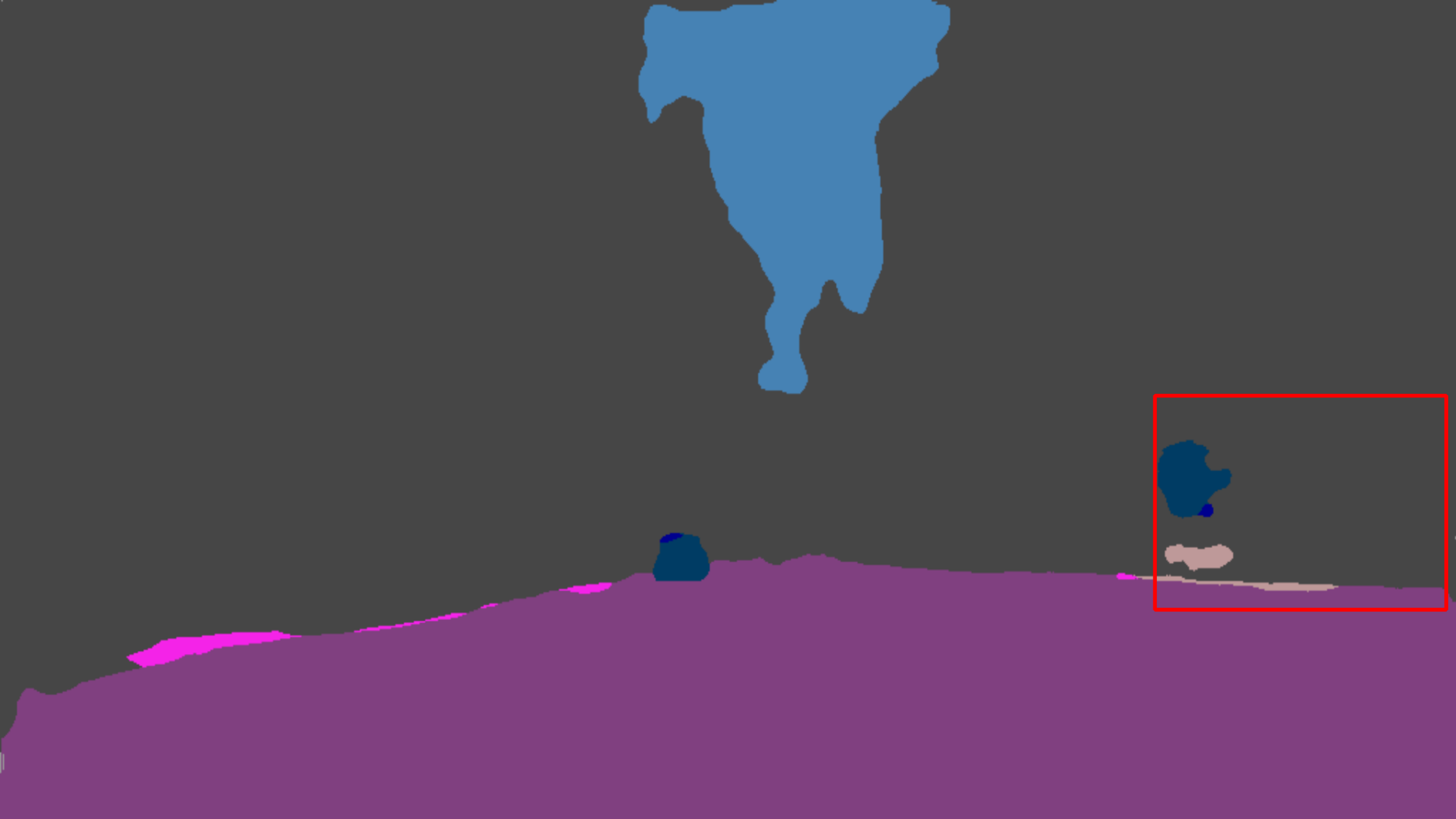} &
            \includegraphics[width=0.155\linewidth]{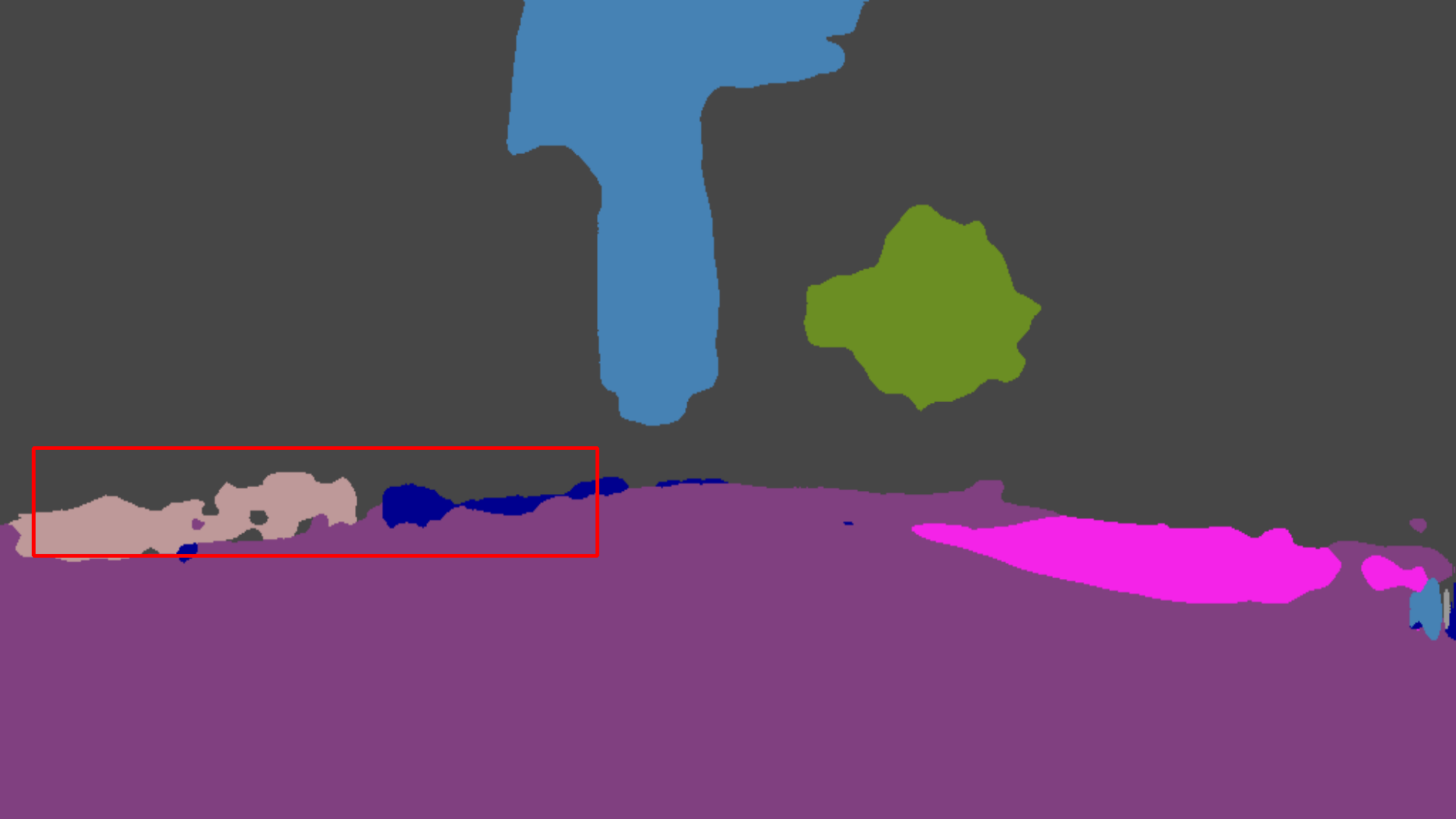} &
            \includegraphics[width=0.155\linewidth]{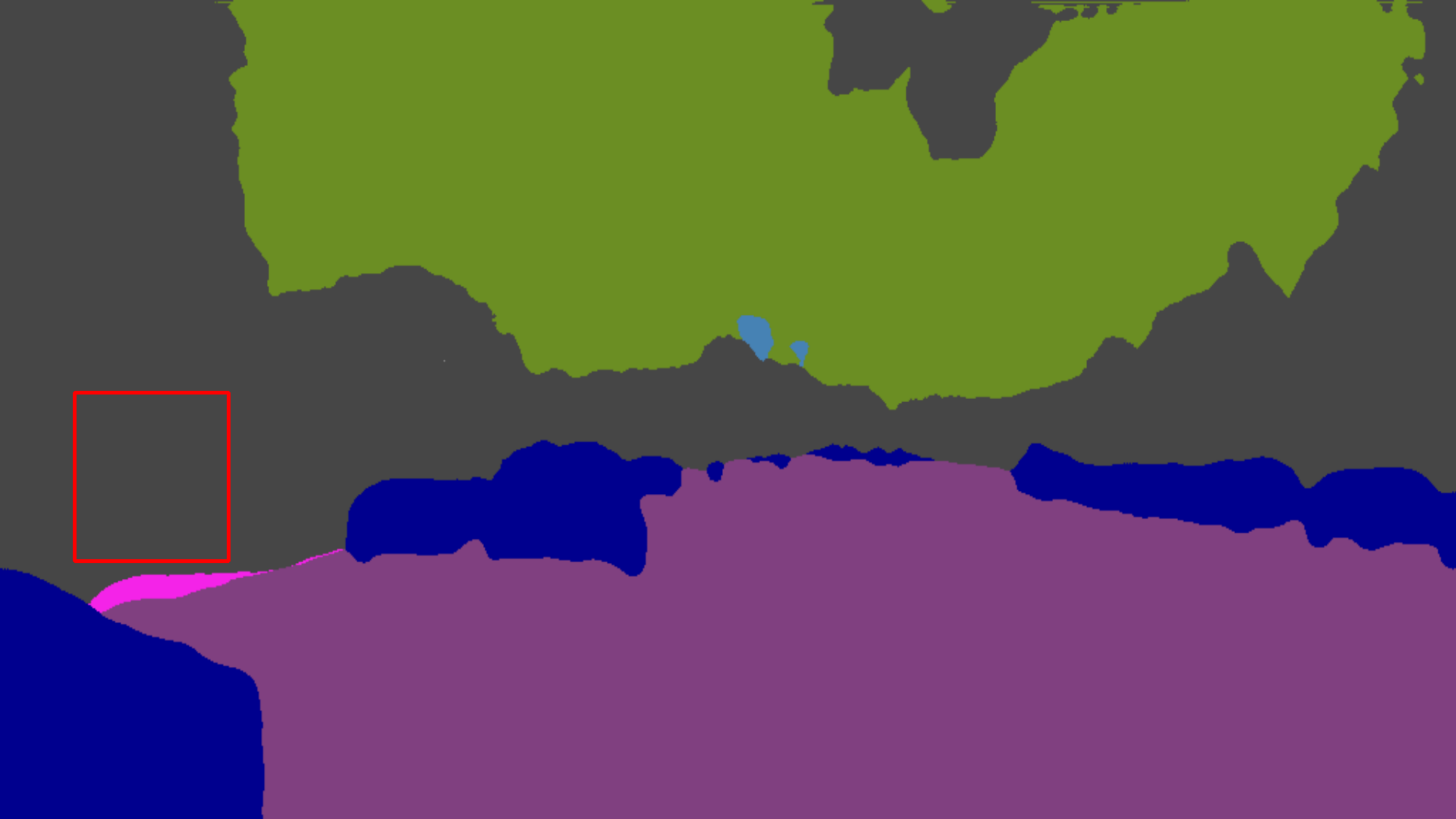} &
            \includegraphics[width=0.155\linewidth]{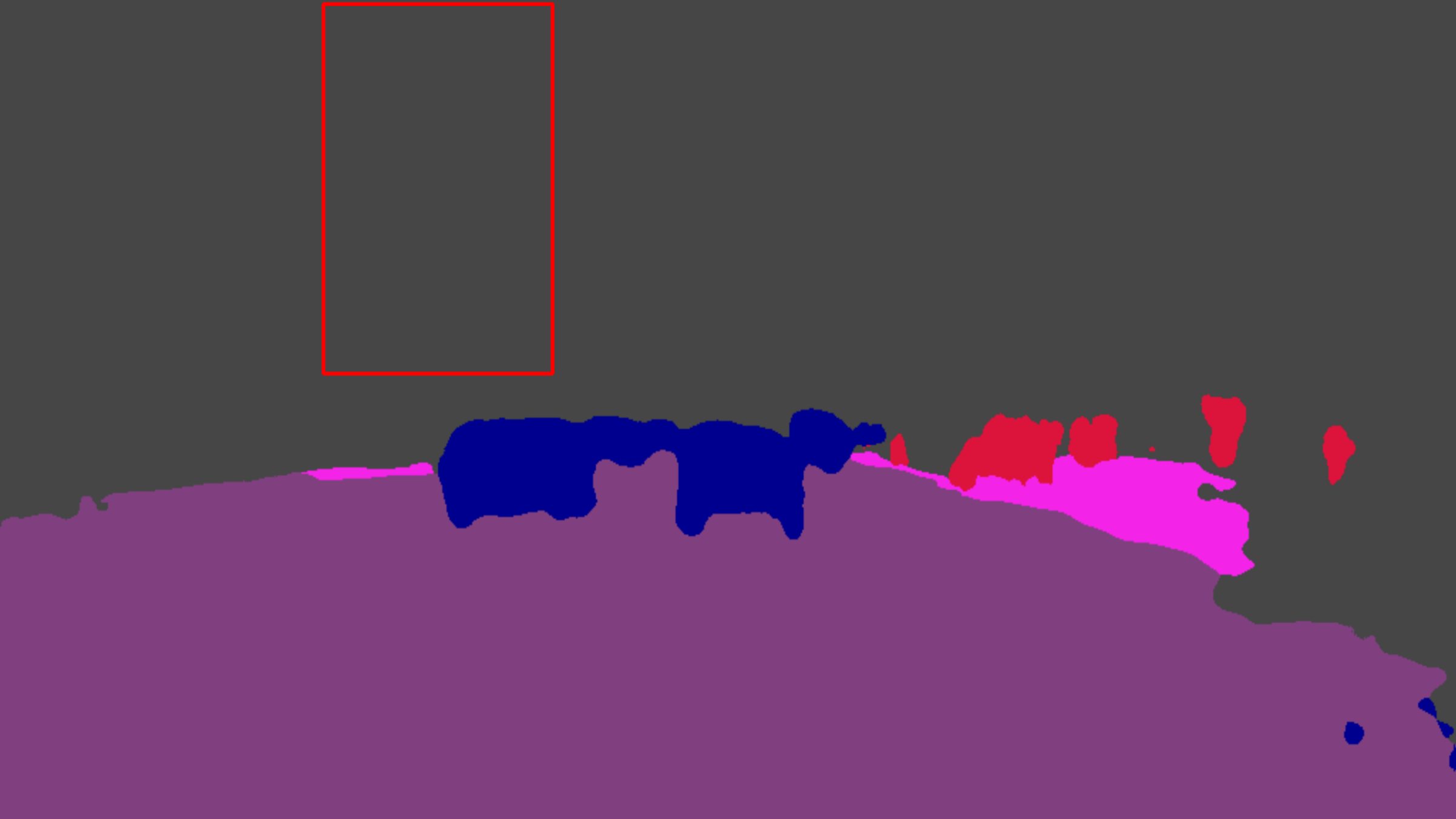} &
            \includegraphics[width=0.155\linewidth]{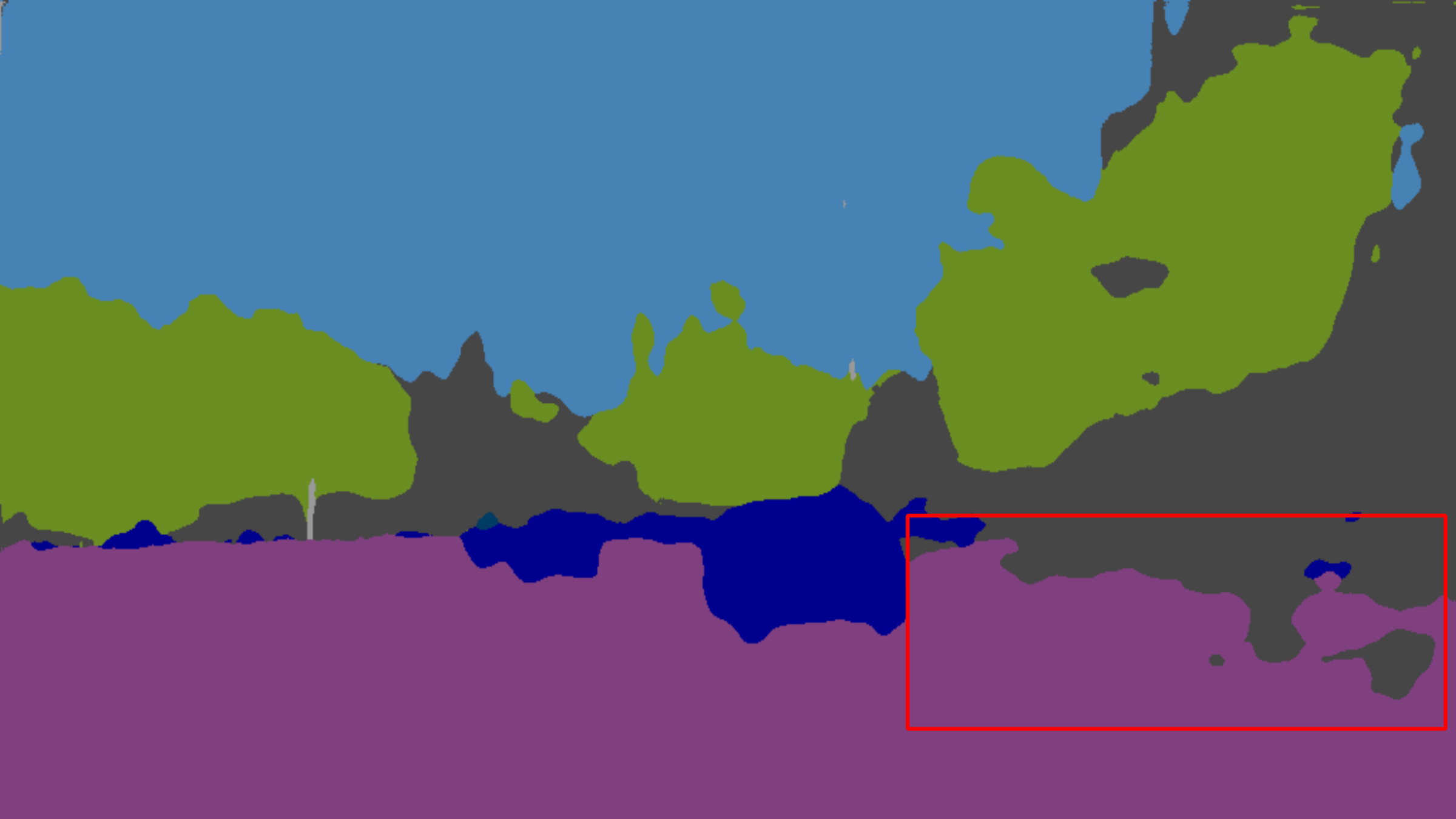} &
            \includegraphics[width=0.155\linewidth]{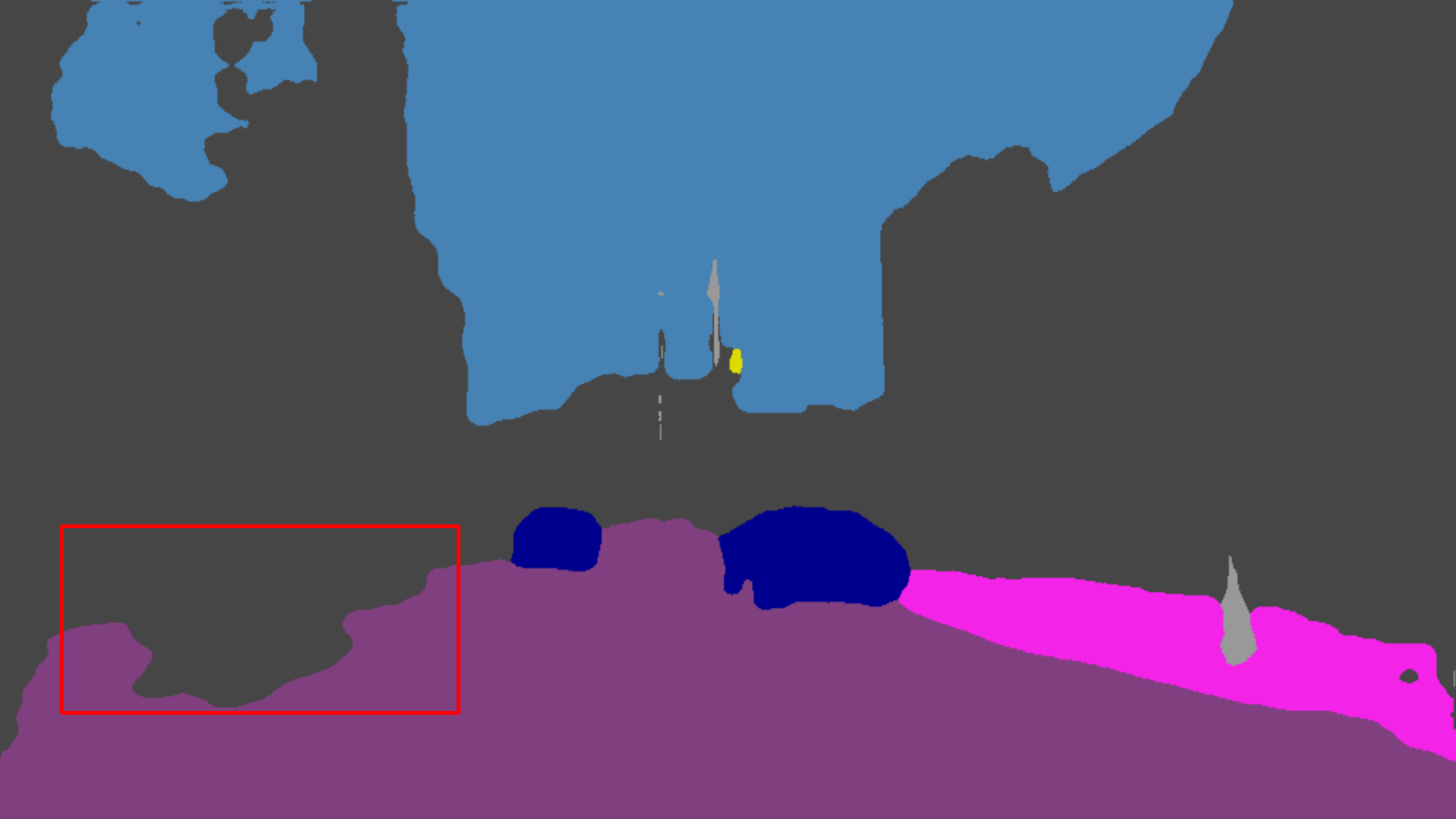} \\
            \sidefont Ours &
            \includegraphics[width=0.155\linewidth]{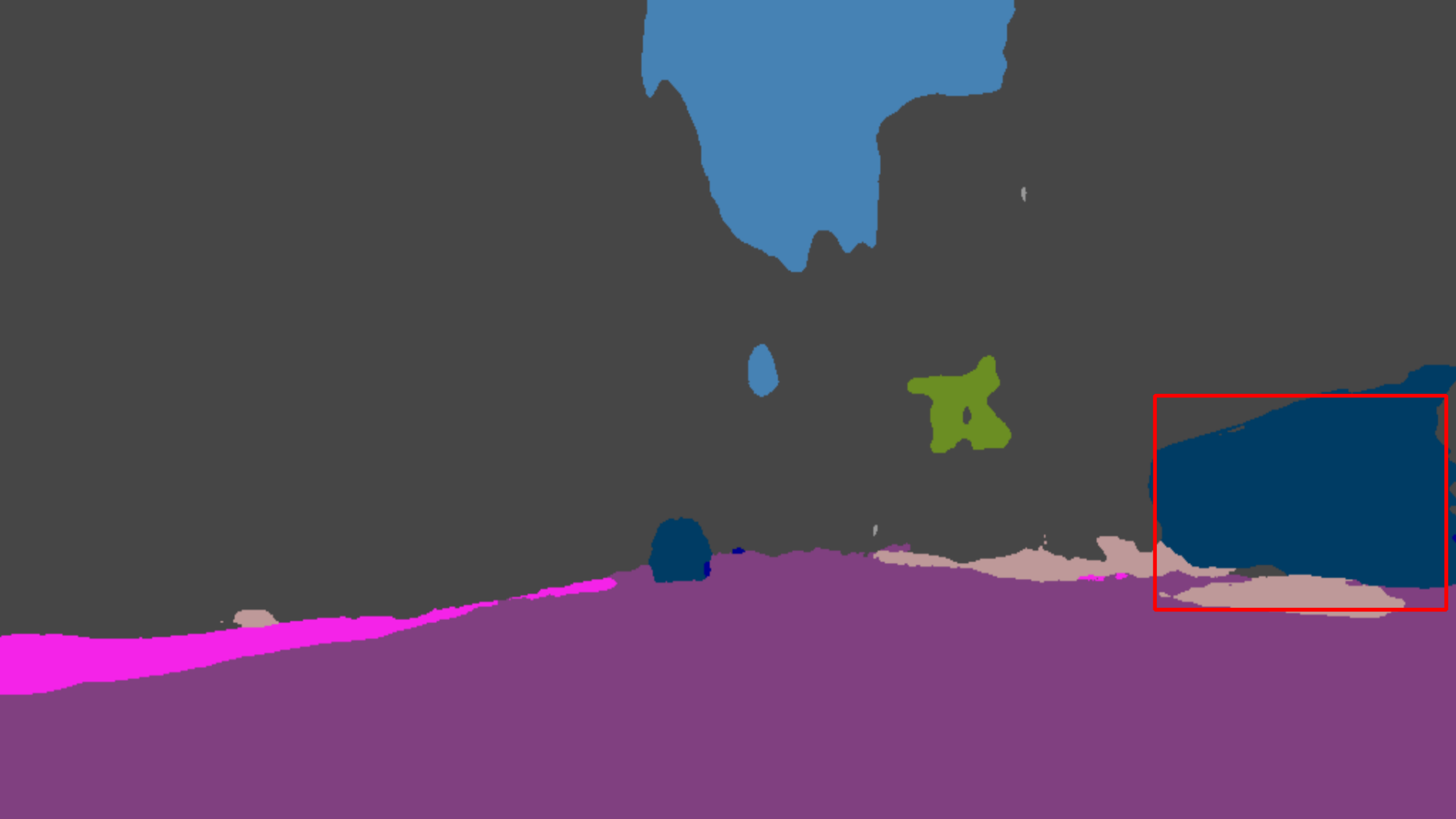} &
            \includegraphics[width=0.155\linewidth]{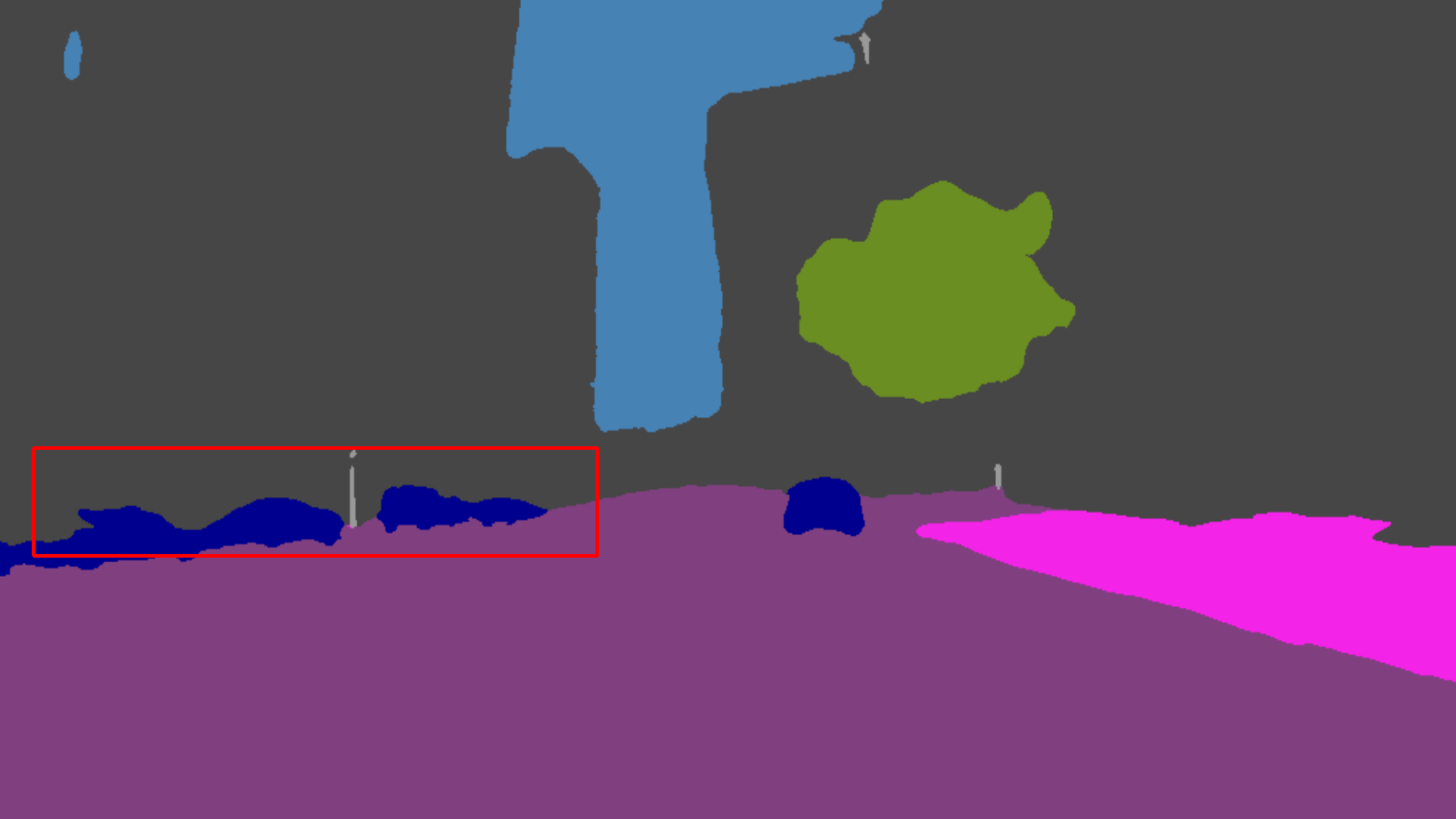} &
            \includegraphics[width=0.155\linewidth]{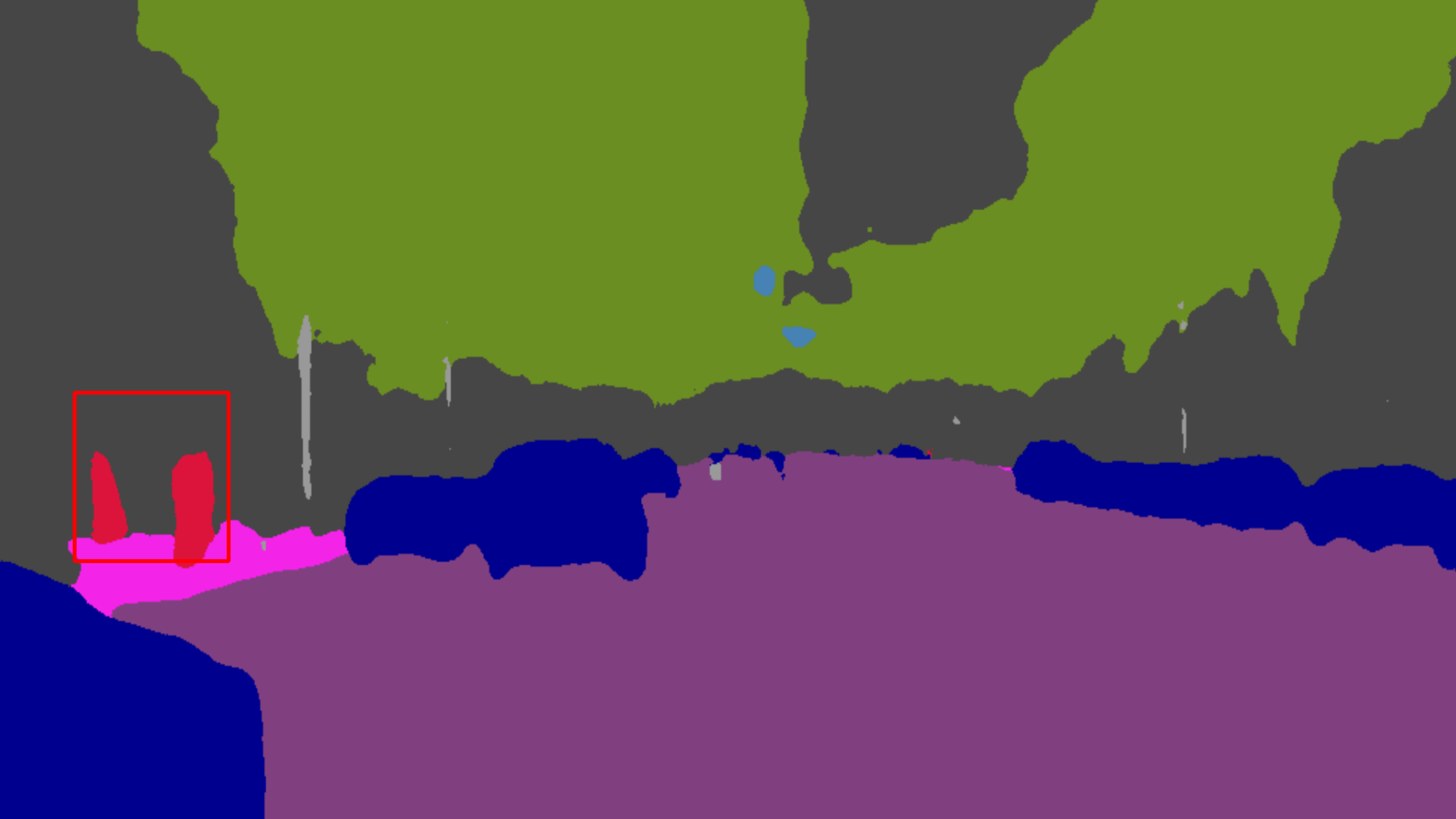} &
            \includegraphics[width=0.155\linewidth]{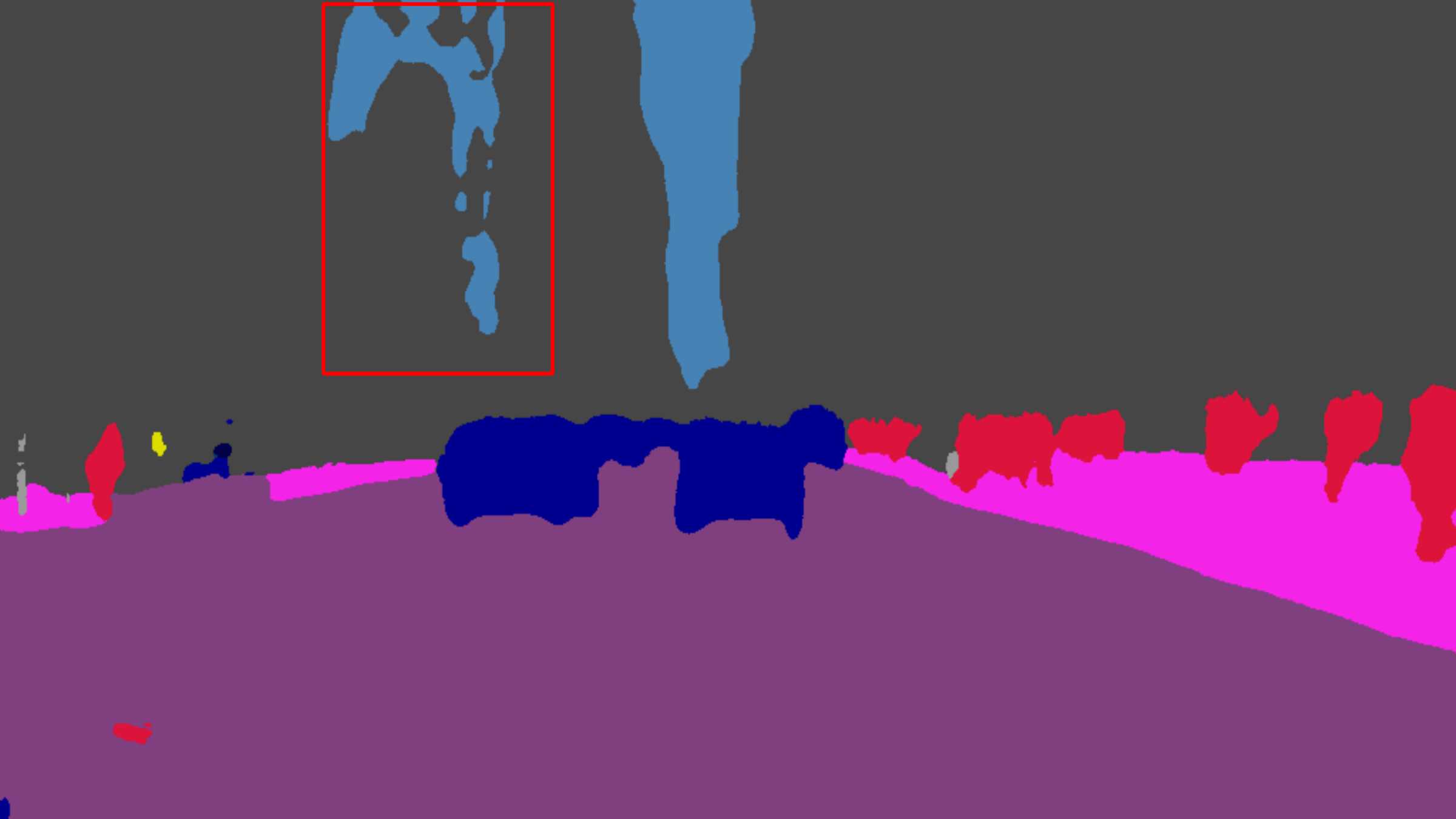} &
            \includegraphics[width=0.155\linewidth]{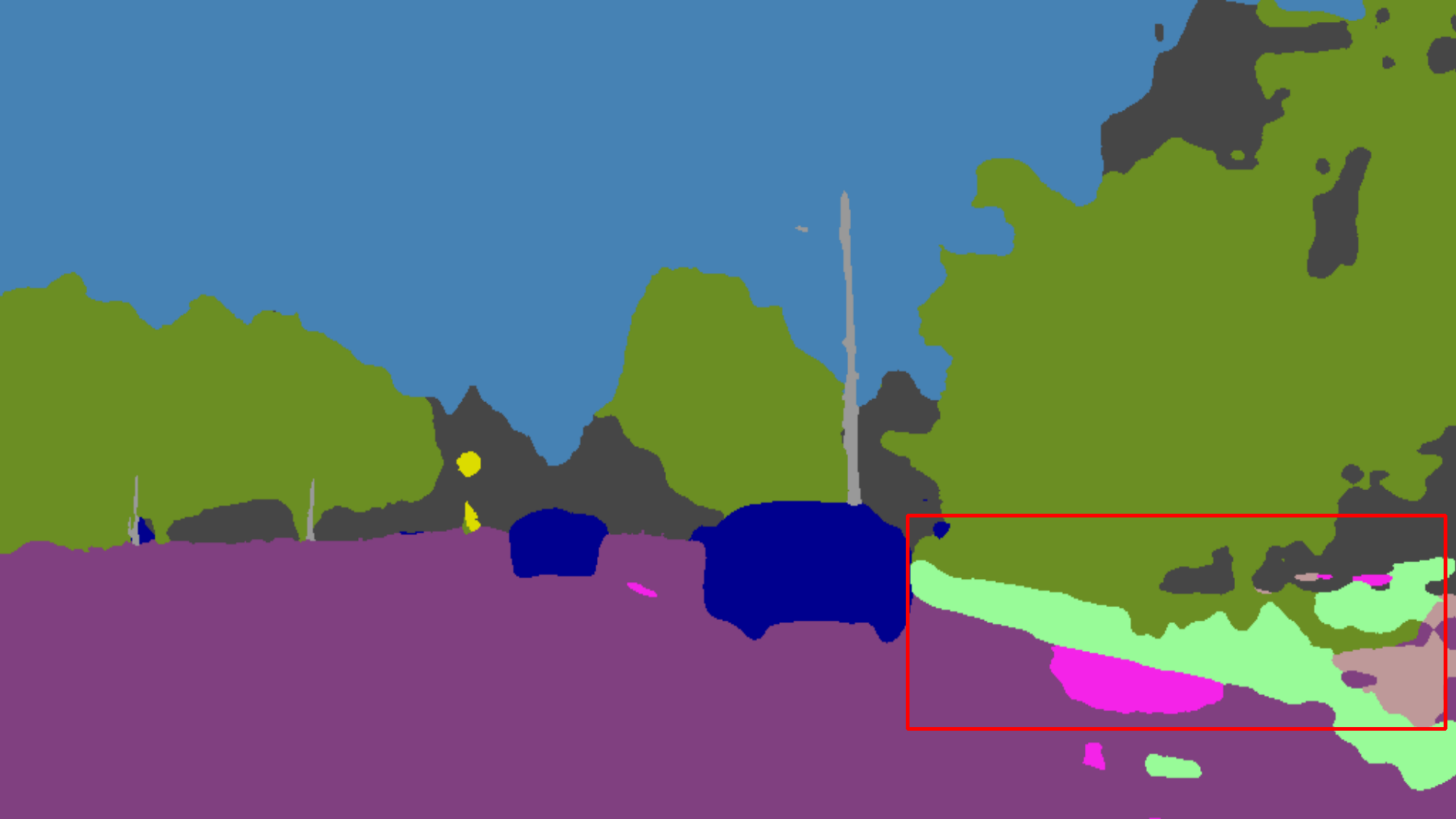} &
            \includegraphics[width=0.155\linewidth]{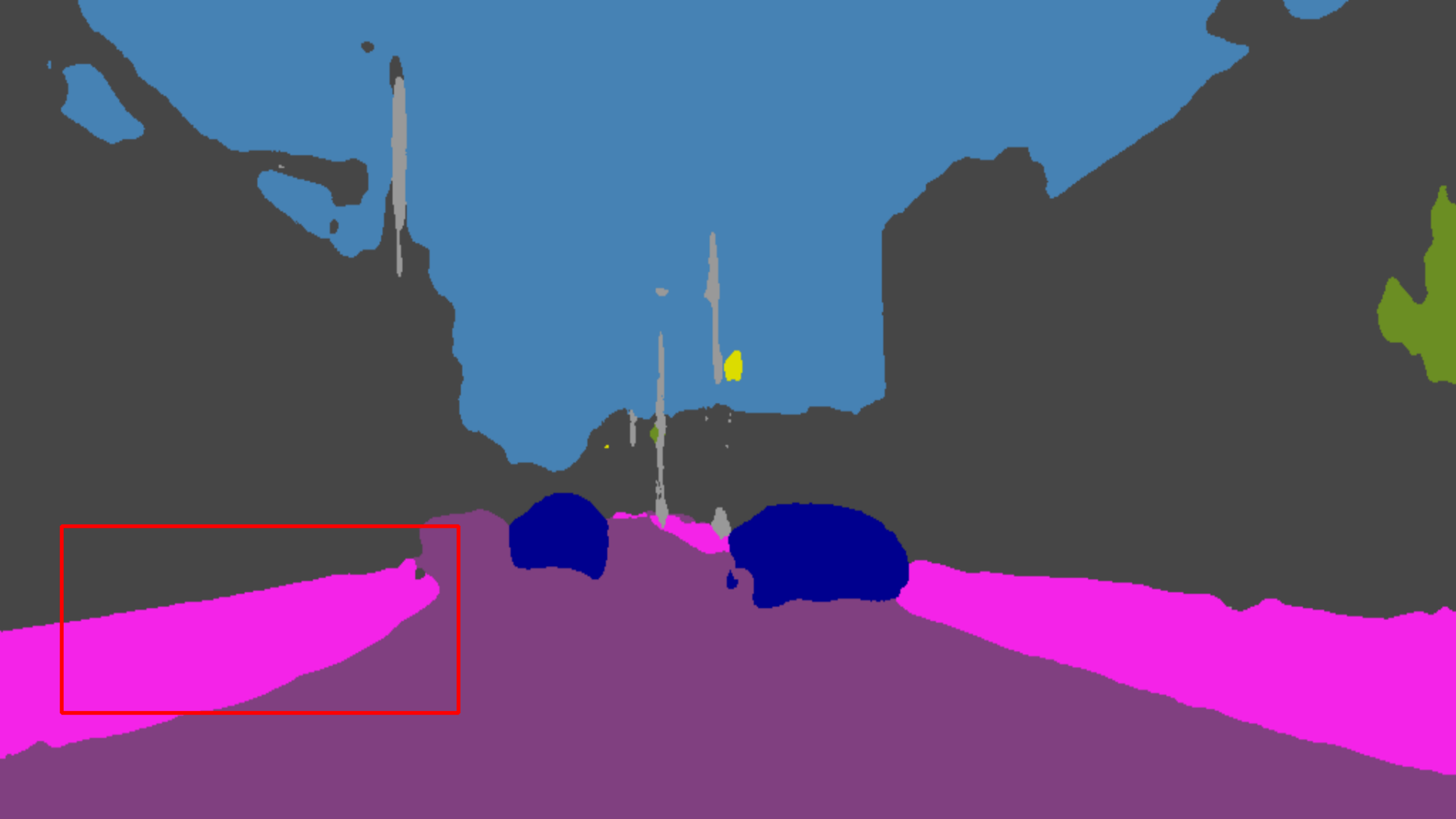} \\
            \sidefont GT &
            \includegraphics[width=0.155\linewidth]{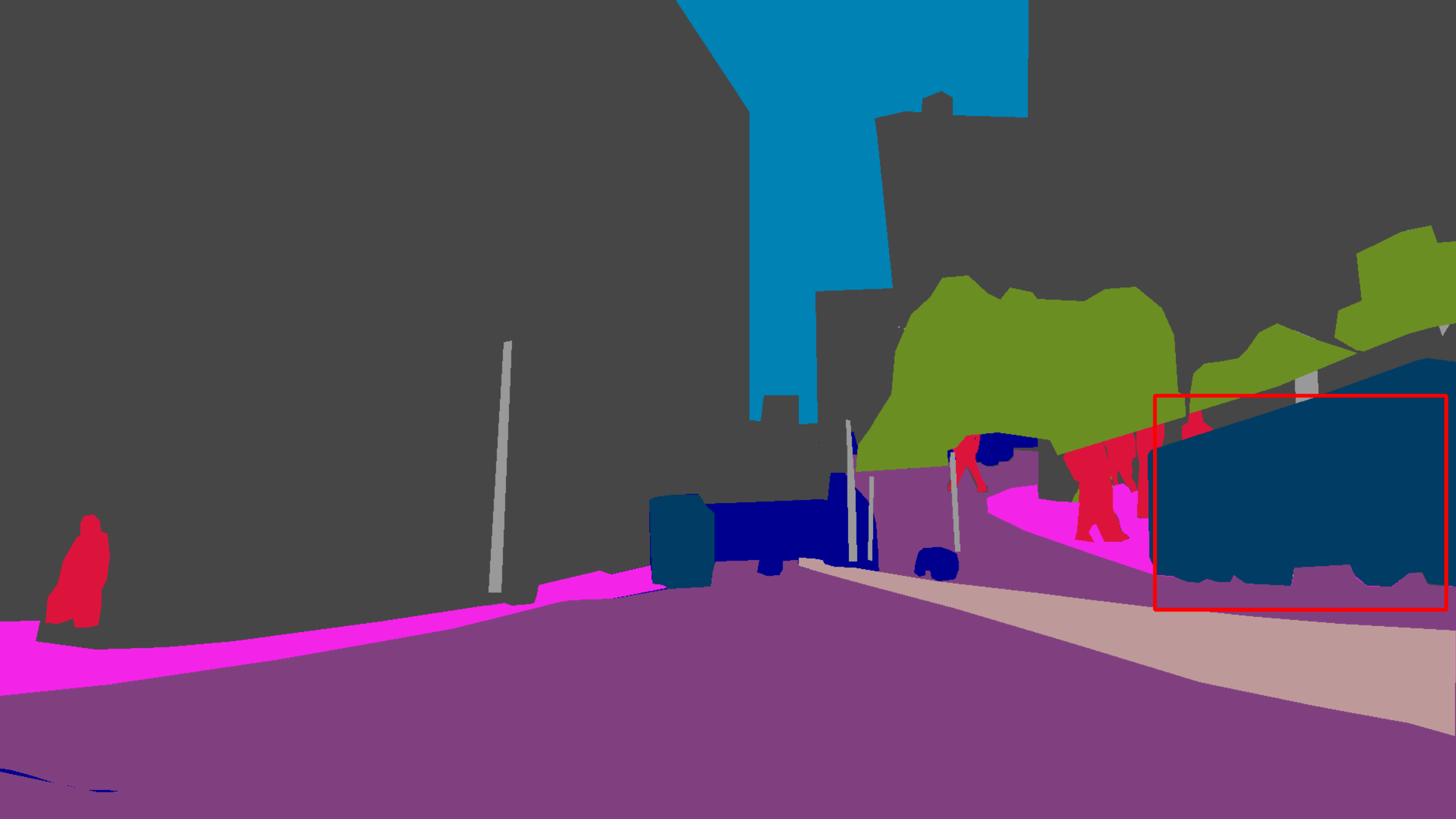} &
            \includegraphics[width=0.155\linewidth]{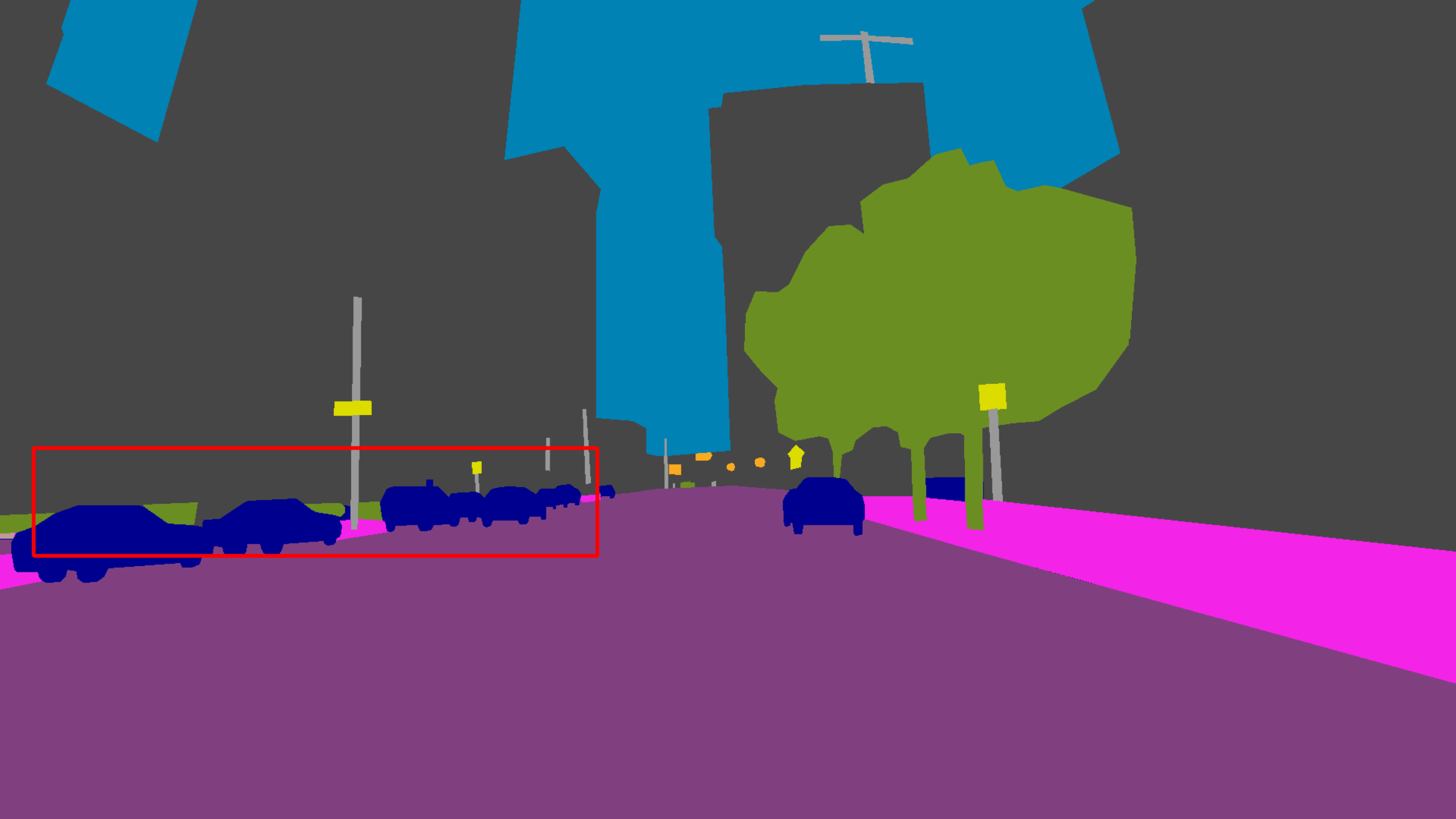} &
            \includegraphics[width=0.155\linewidth]{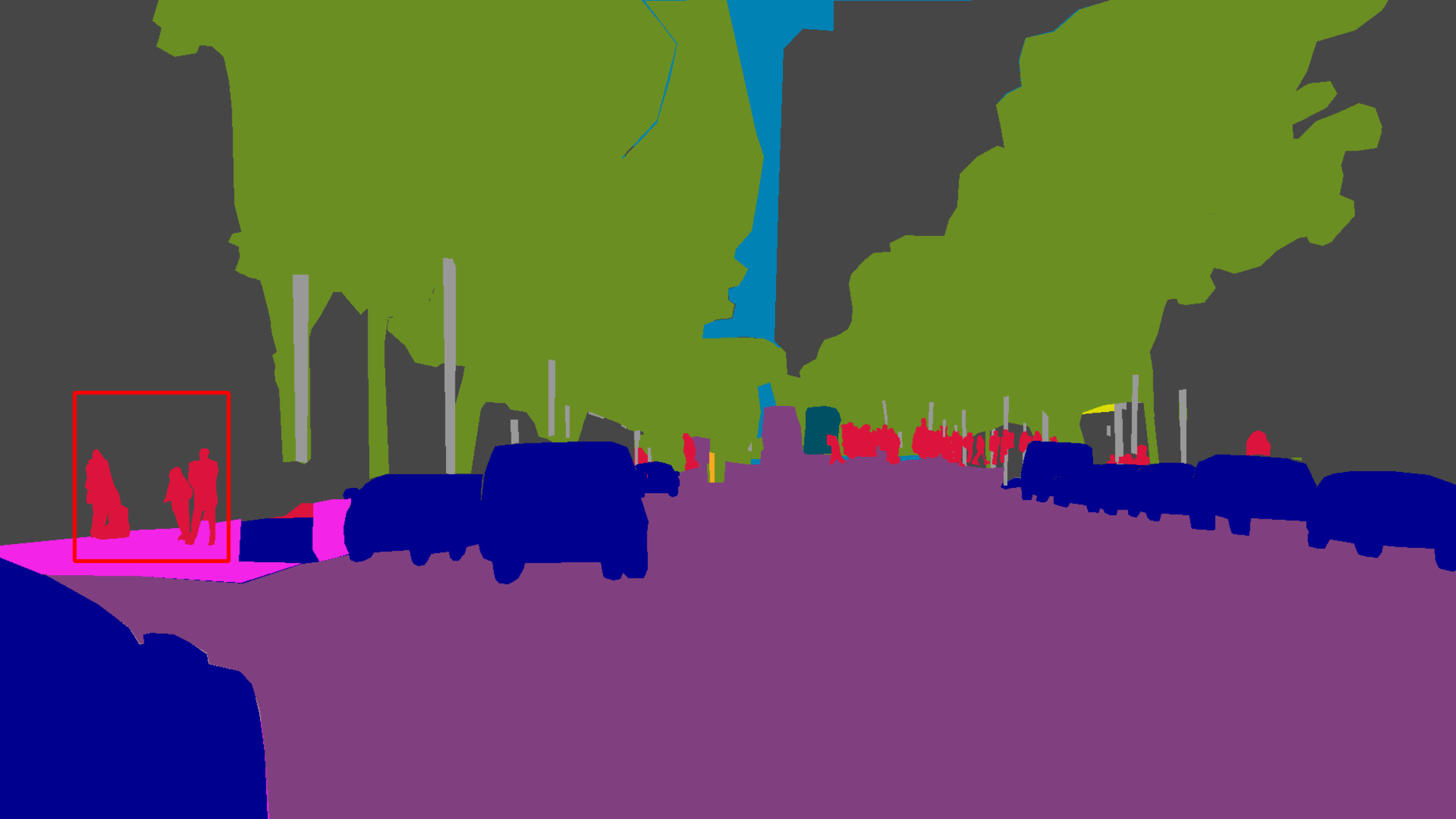} &
            \includegraphics[width=0.155\linewidth]{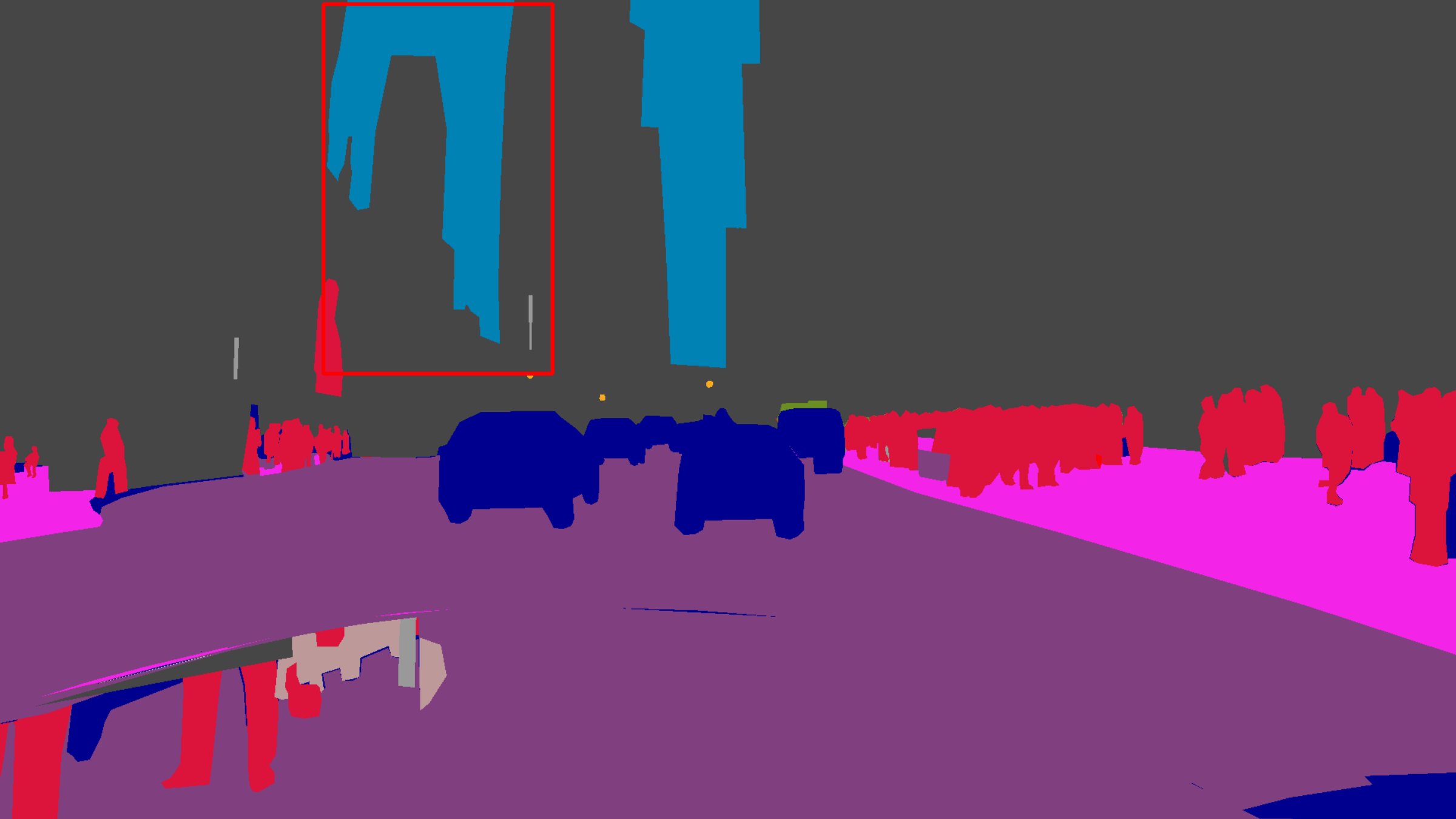} &
            \includegraphics[width=0.155\linewidth]{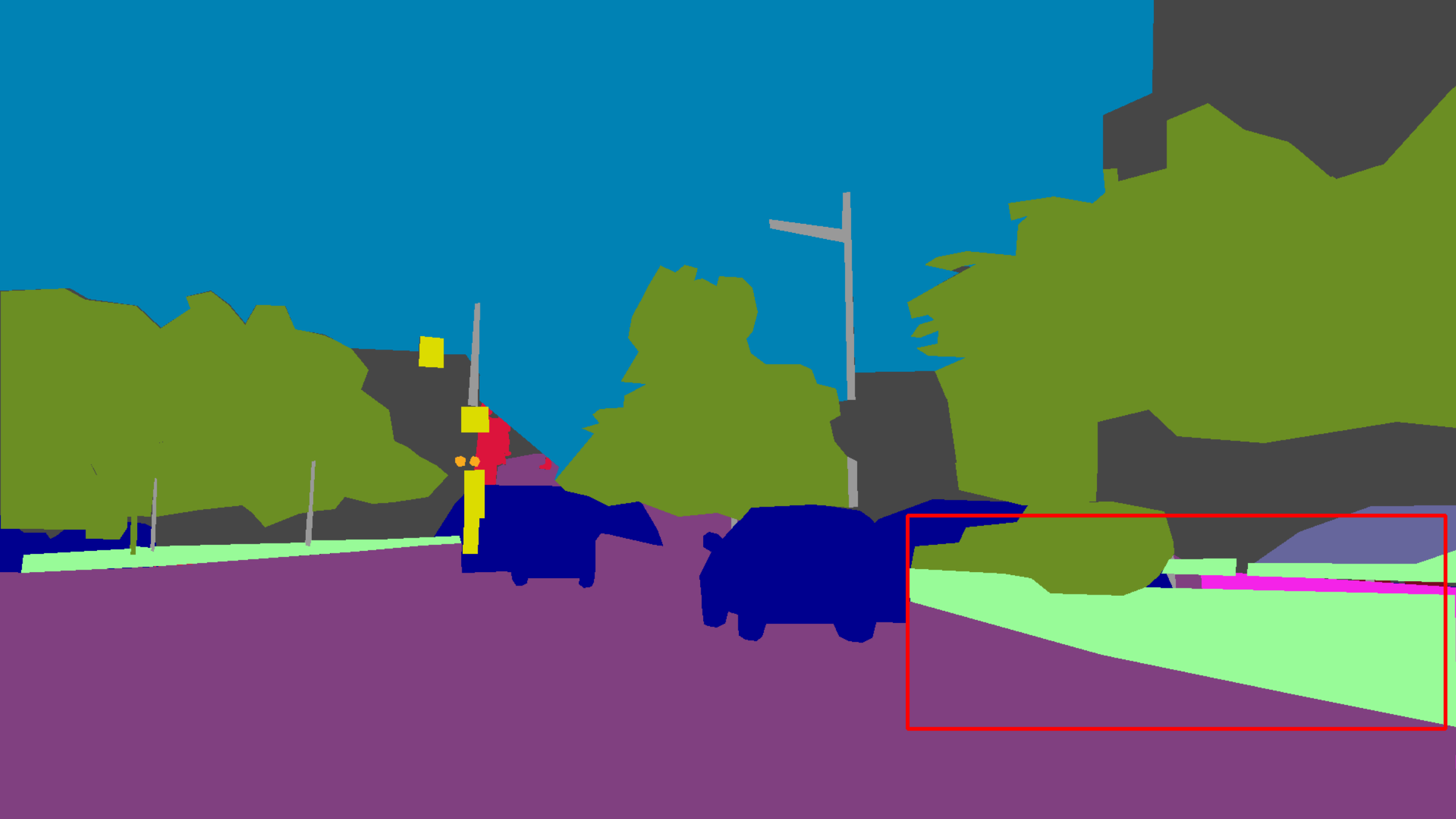} &
            \includegraphics[width=0.155\linewidth]{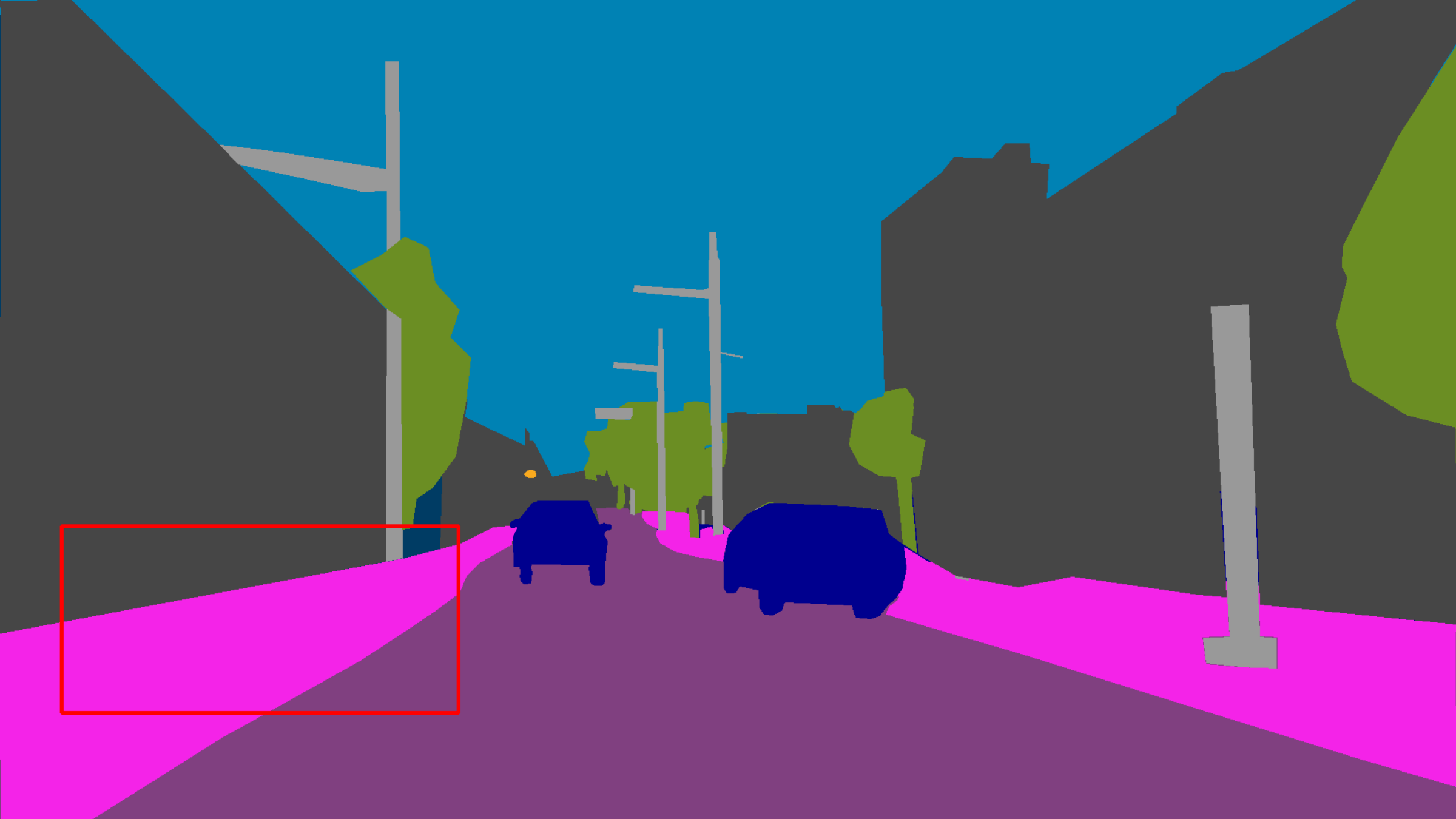} \\
        \end{tabular}
    }
    \vspace{-0.2cm}
    \caption{Qualitative comparison of semantic segmentation results on the NightCity dataset. Red boxes highlight representative local regions.}
    \vspace{-0.2cm}
    \label{fig:segmentation_vis}
\end{figure}

\begin{figure}[t]
    \centering
    \resizebox{\linewidth}{!}{
        \setlength{\tabcolsep}{0.5pt} 
        \renewcommand{\arraystretch}{0.5} 
        \begin{tabular}{cccccc}
            \includegraphics[width=0.16\linewidth]{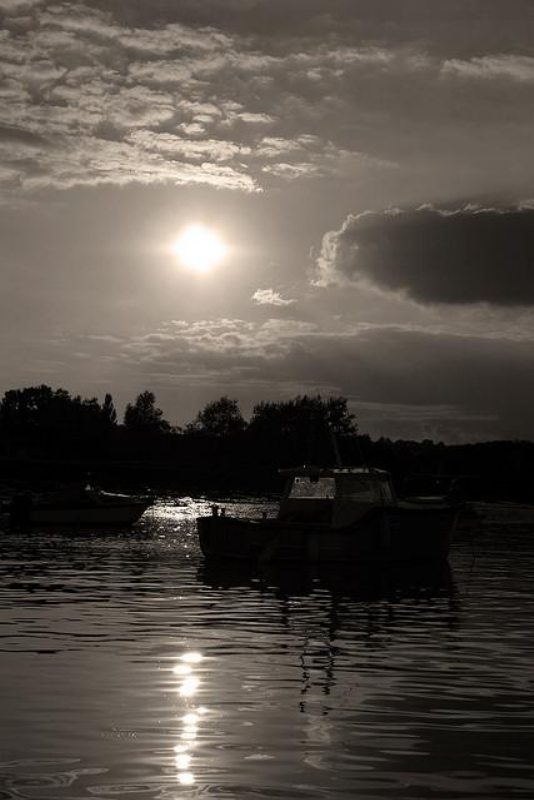} &
            \includegraphics[width=0.16\linewidth]{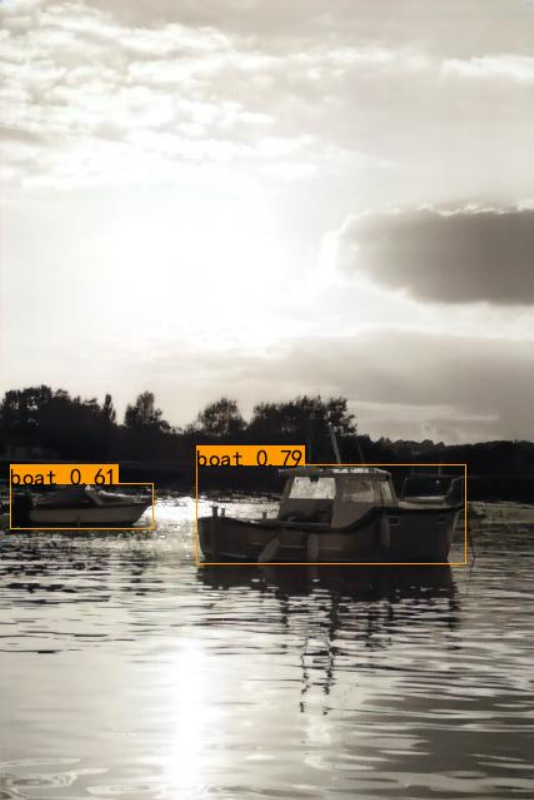} &
            \includegraphics[width=0.16\linewidth]{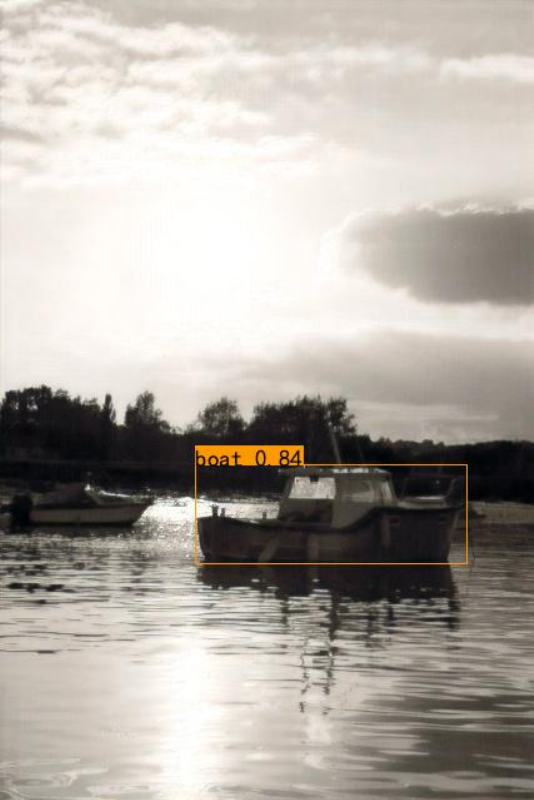} &
            \includegraphics[width=0.16\linewidth]{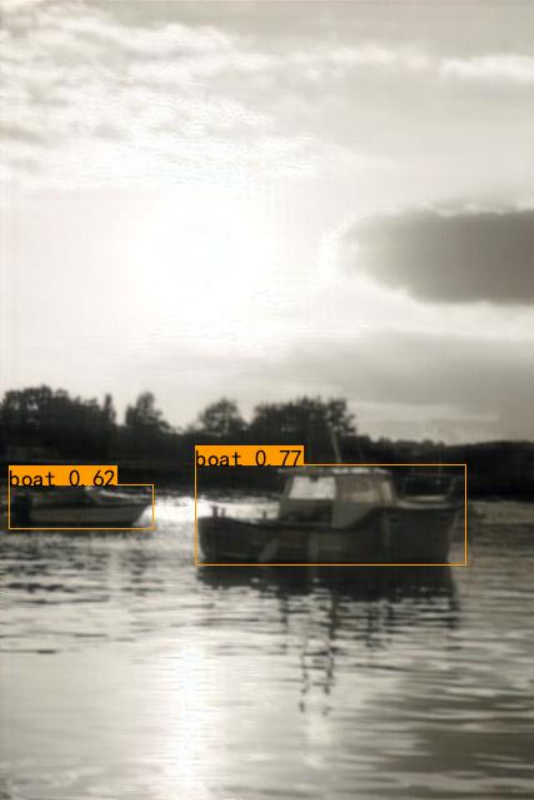} &
            \includegraphics[width=0.16\linewidth]{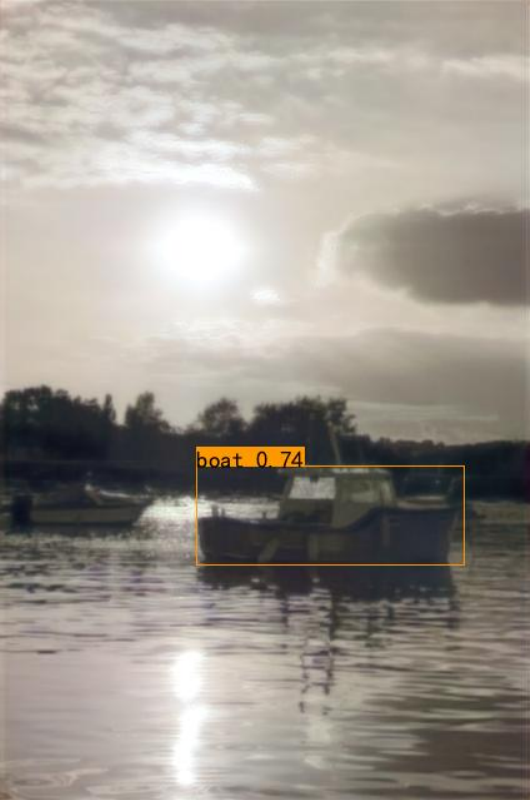} &
            \includegraphics[width=0.16\linewidth]{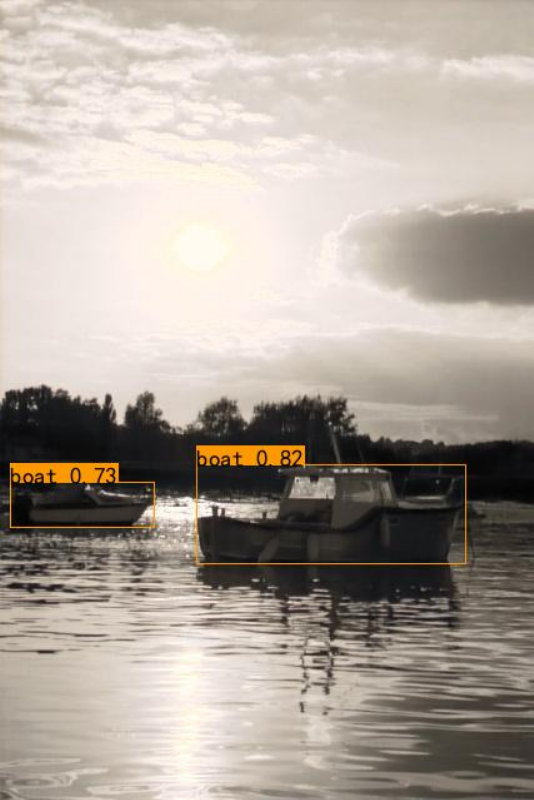} \\
            \includegraphics[width=0.16\linewidth]{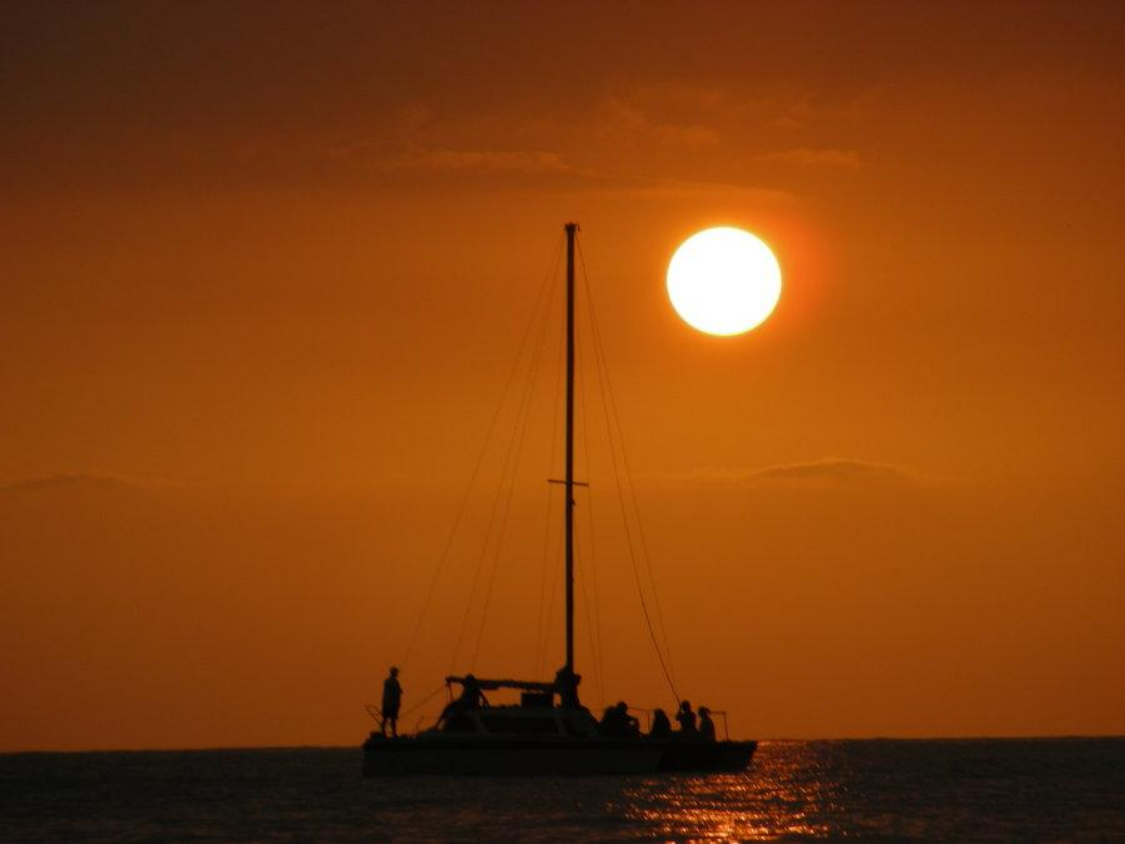} &
            \includegraphics[width=0.16\linewidth]{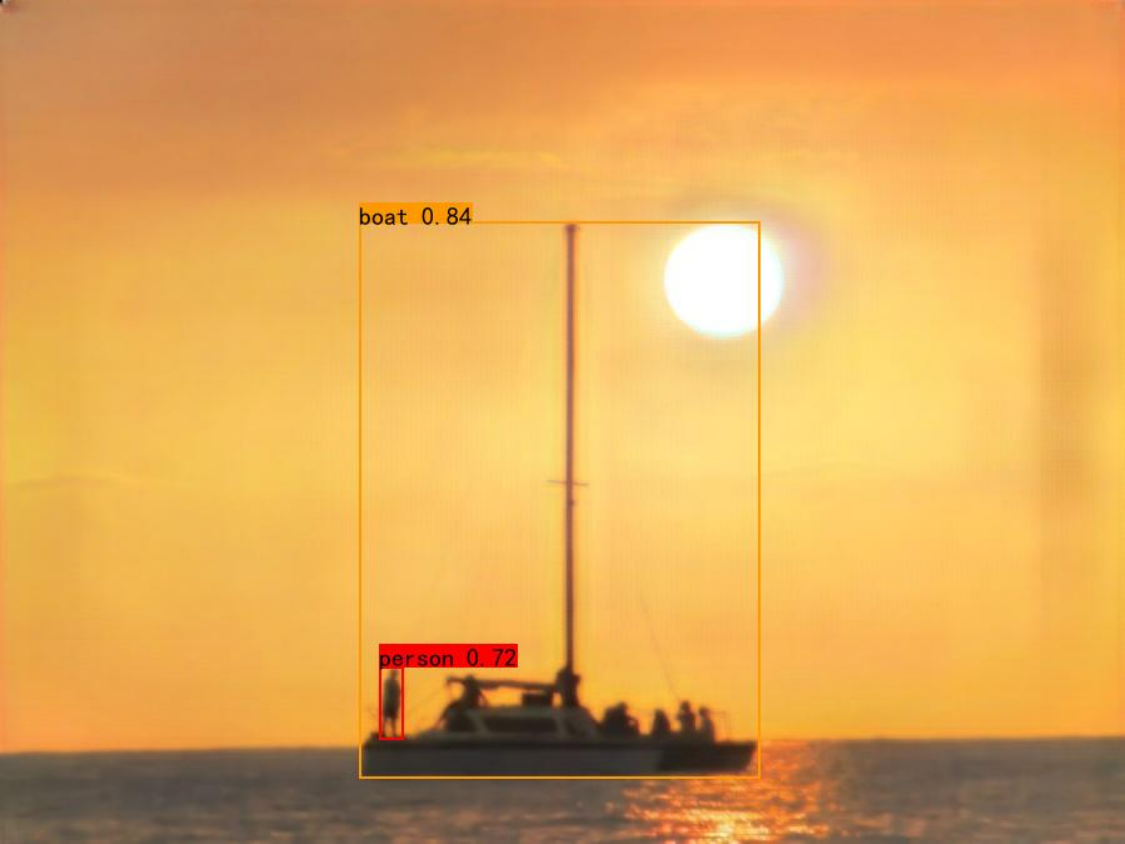} &
            \includegraphics[width=0.16\linewidth]{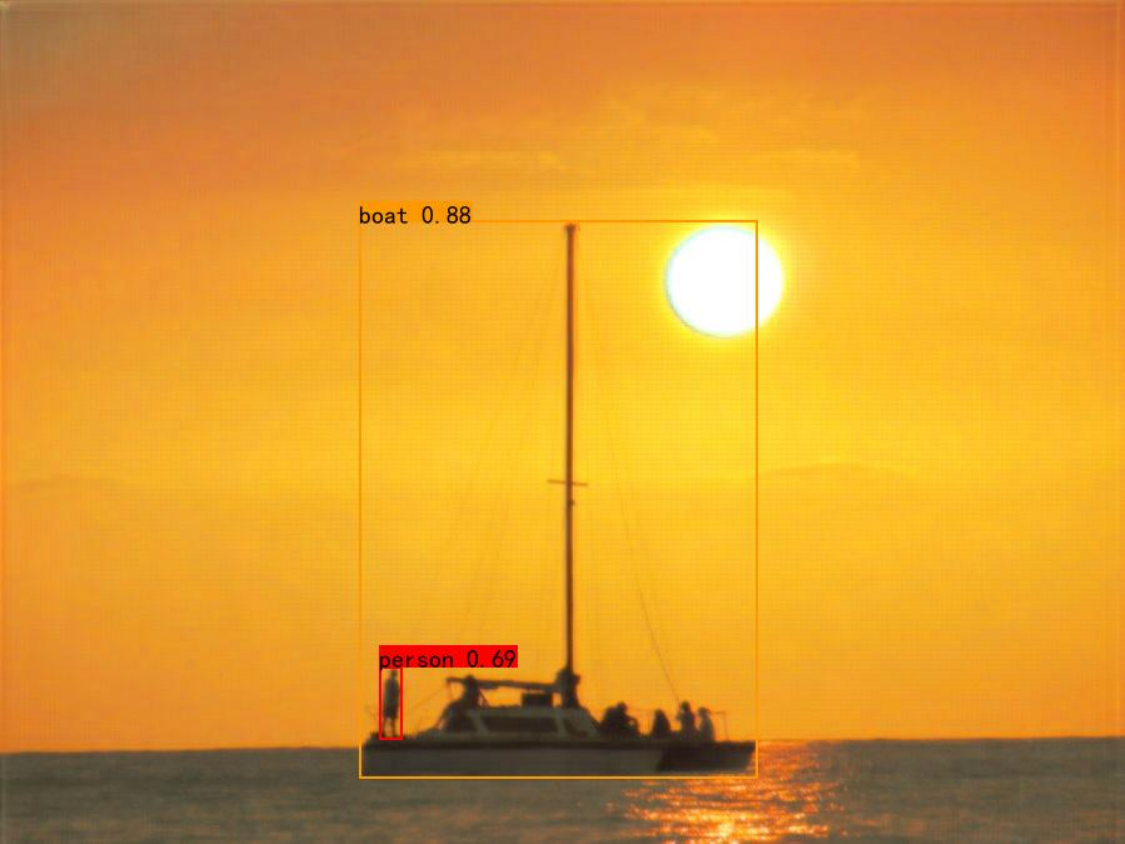} &
            \includegraphics[width=0.16\linewidth]{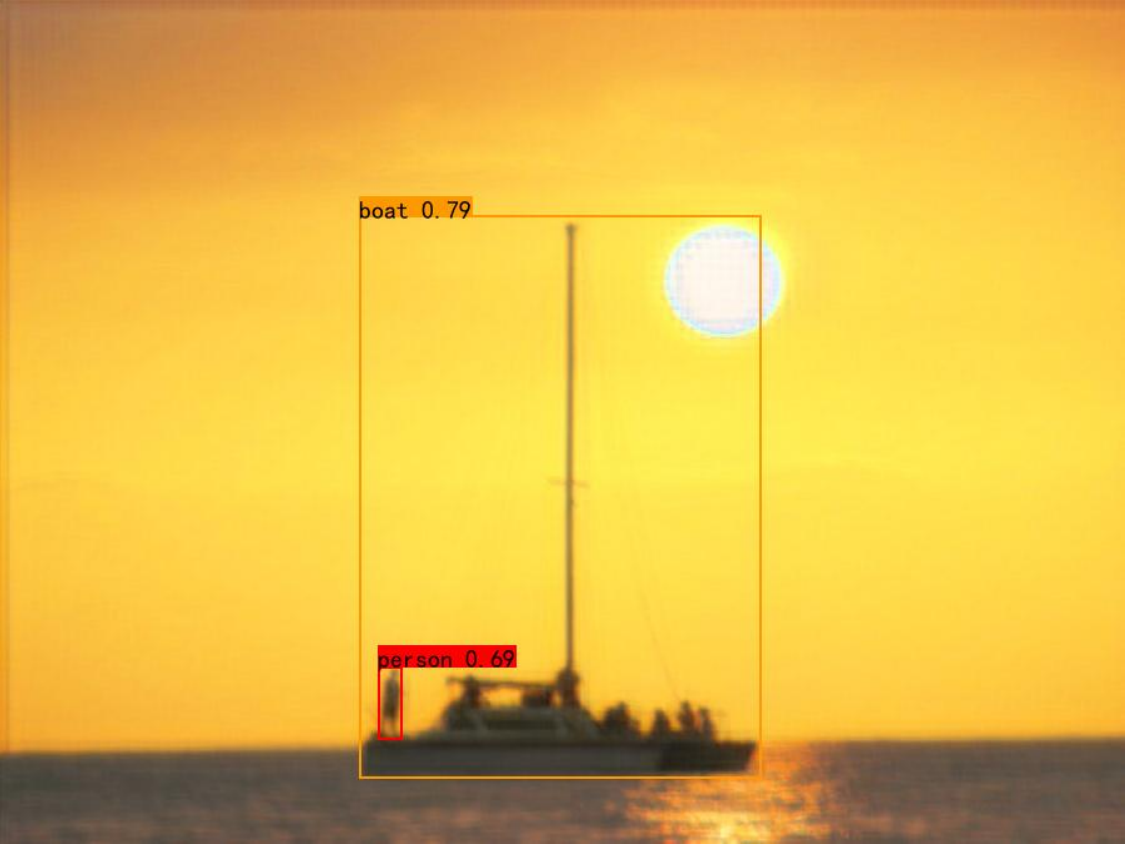} &
            \includegraphics[width=0.16\linewidth]{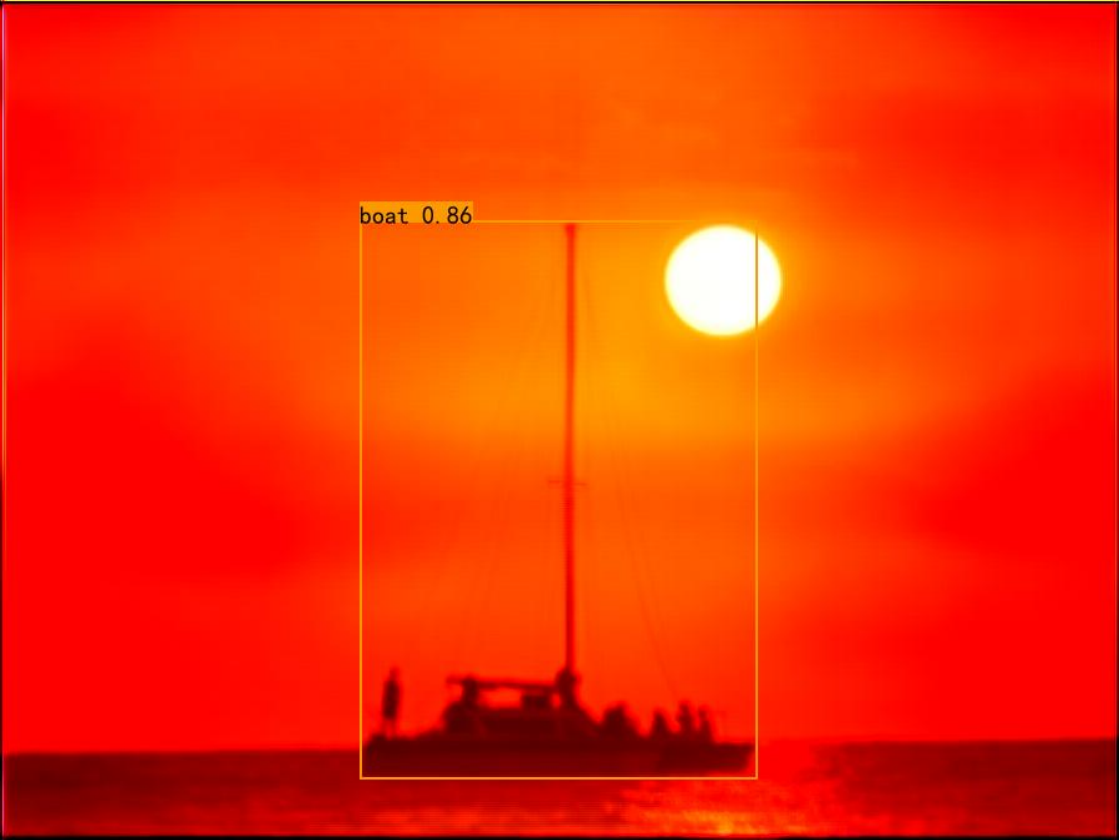} &
            \includegraphics[width=0.16\linewidth]{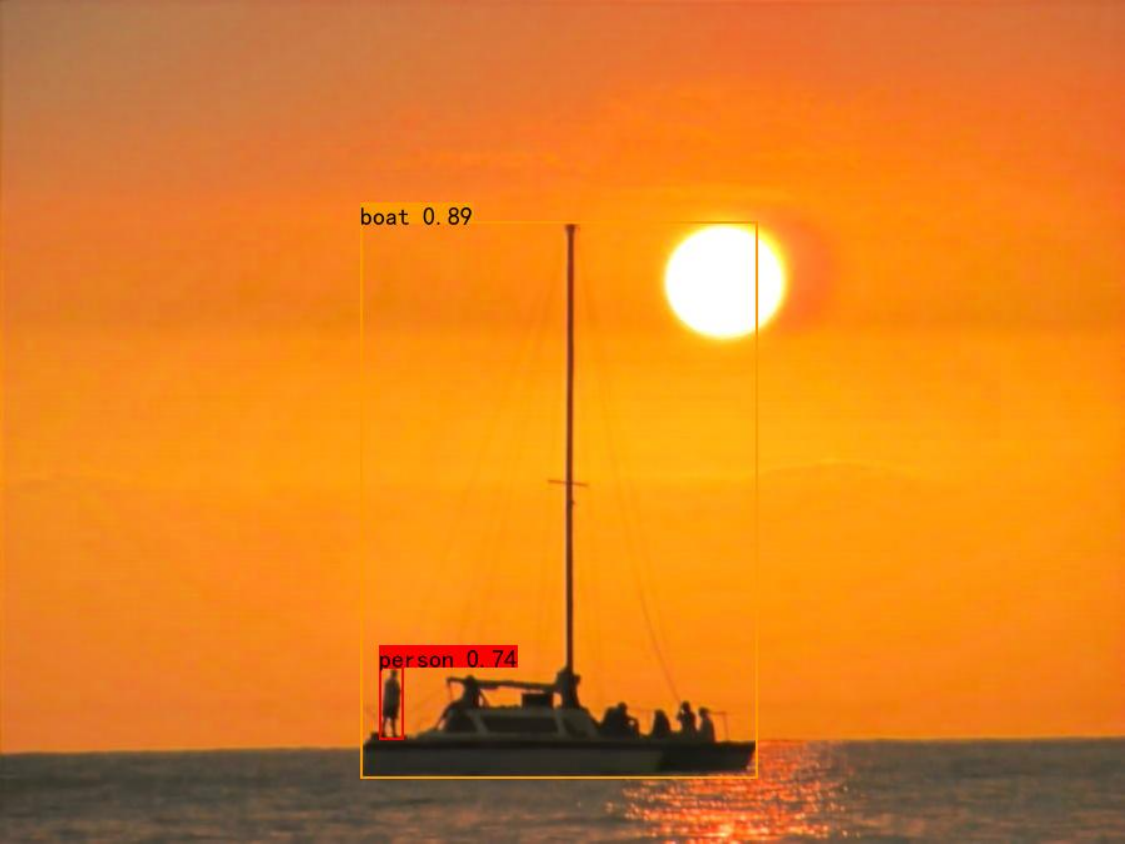} \\
            \includegraphics[width=0.16\linewidth]{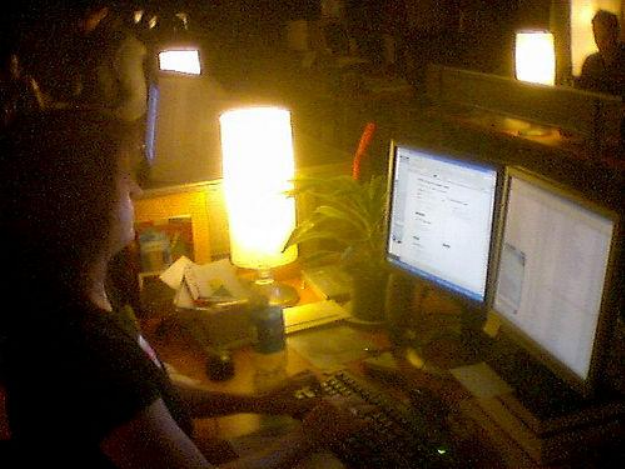} &
            \includegraphics[width=0.16\linewidth]{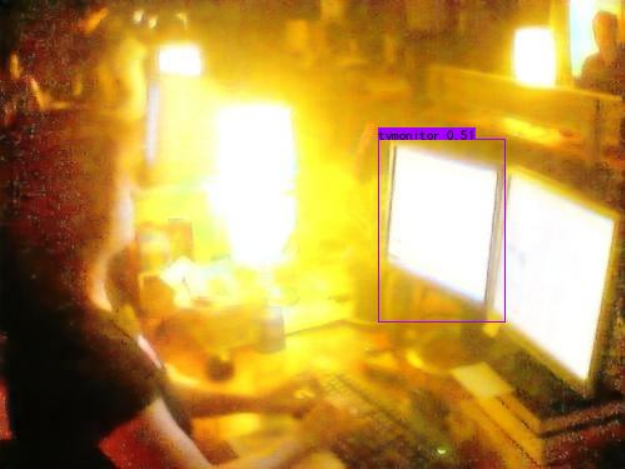} &
            \includegraphics[width=0.16\linewidth]{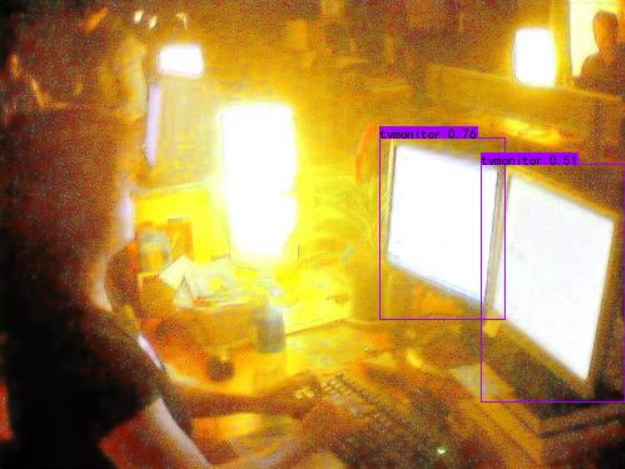} &
            \includegraphics[width=0.16\linewidth]{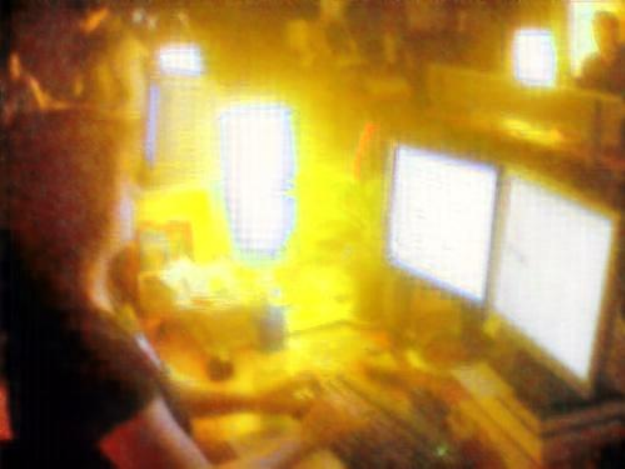} &
            \includegraphics[width=0.16\linewidth]{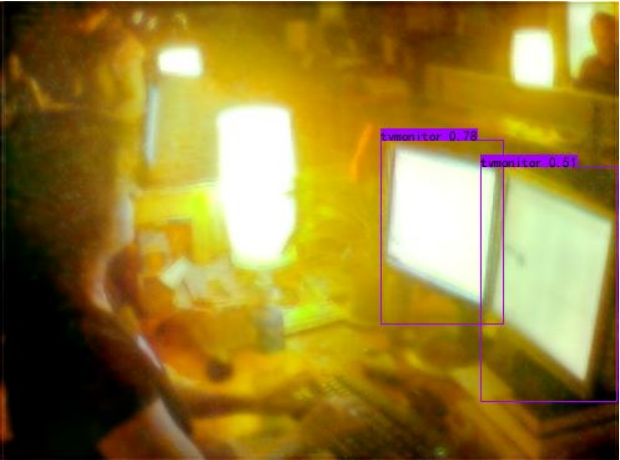} &
            \includegraphics[width=0.16\linewidth]{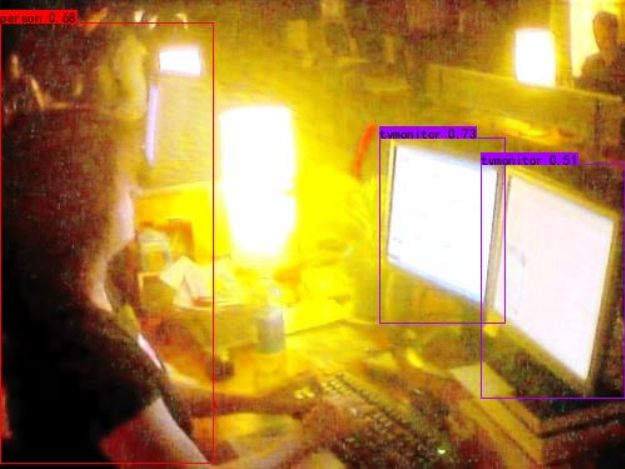} \\
            \includegraphics[width=0.16\linewidth]{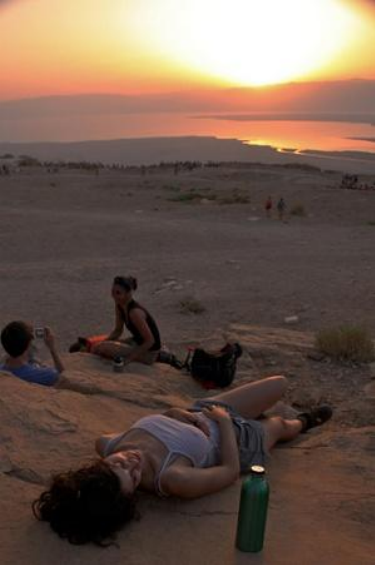} &
            \includegraphics[width=0.16\linewidth]{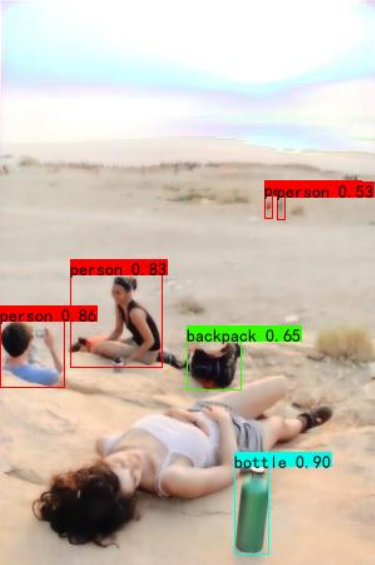} &
            \includegraphics[width=0.16\linewidth]{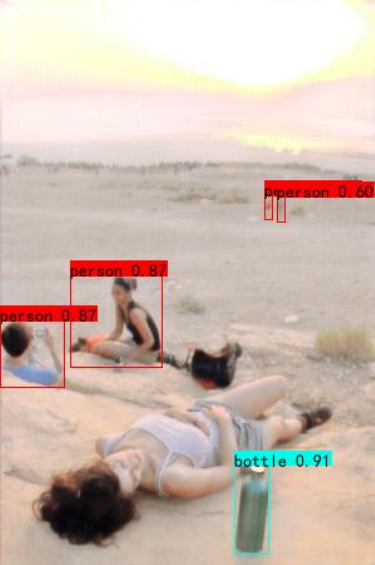} &
            \includegraphics[width=0.16\linewidth]{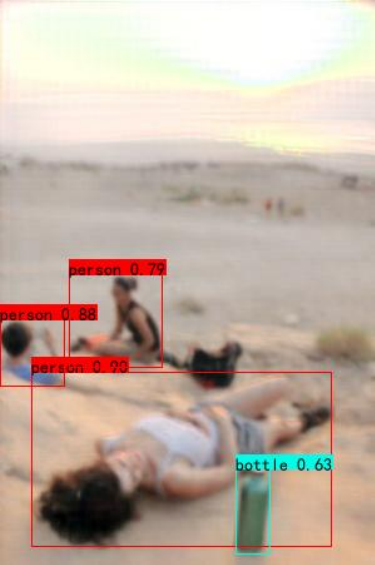} &
            \includegraphics[width=0.16\linewidth]{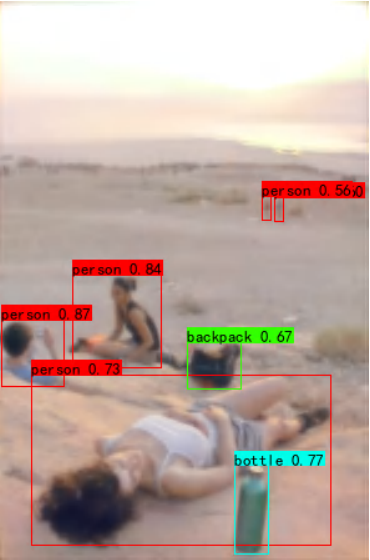} &
            \includegraphics[width=0.16\linewidth]{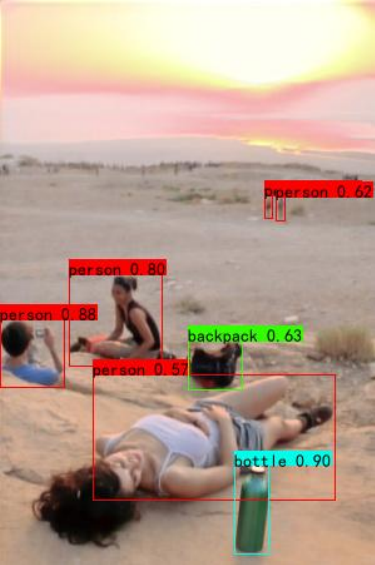} \\
            \includegraphics[width=0.16\linewidth]{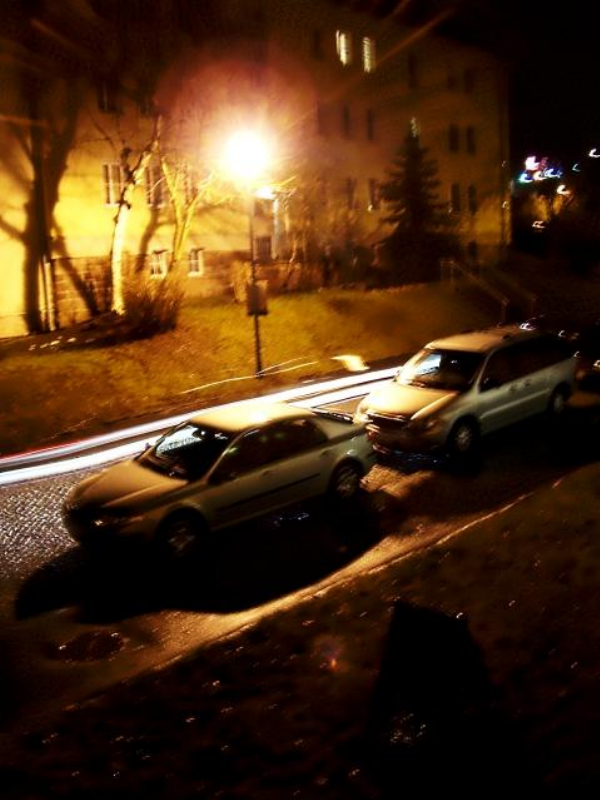} &
            \includegraphics[width=0.16\linewidth]{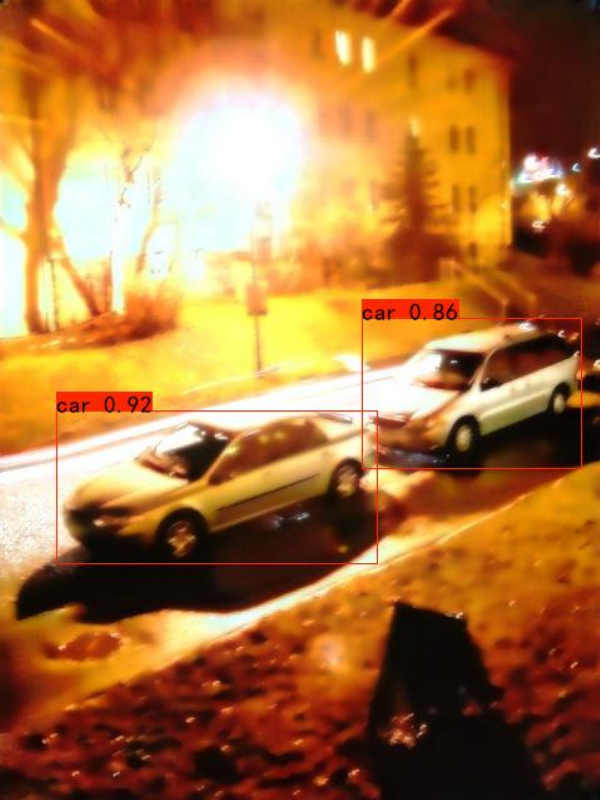} &
            \includegraphics[width=0.16\linewidth]{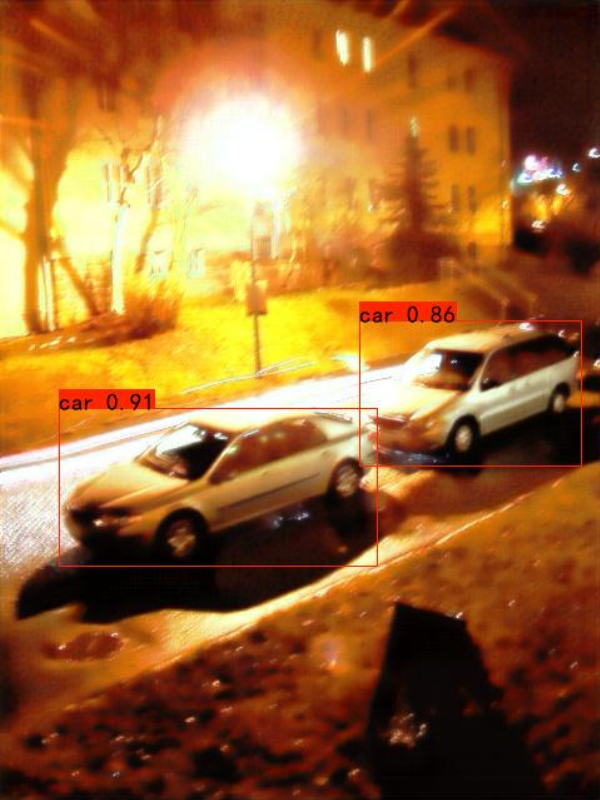} &
            \includegraphics[width=0.16\linewidth]{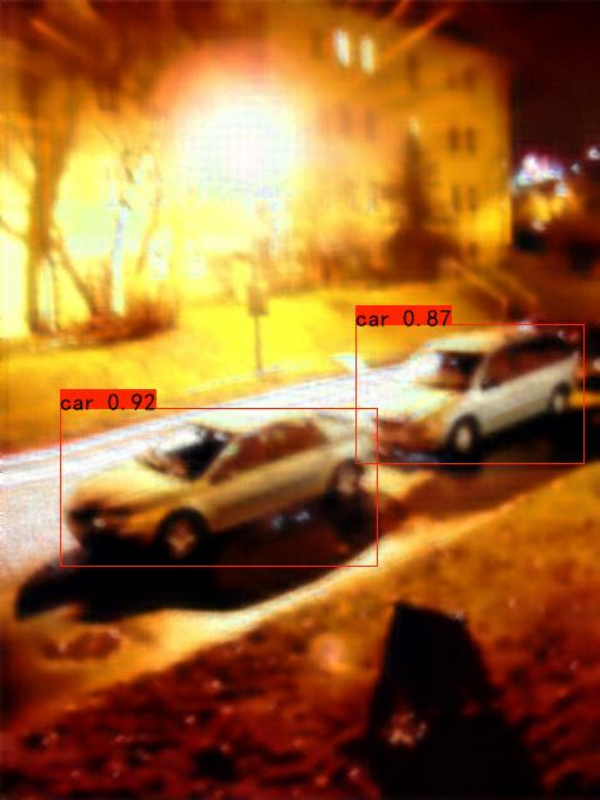} &
            \includegraphics[width=0.16\linewidth]{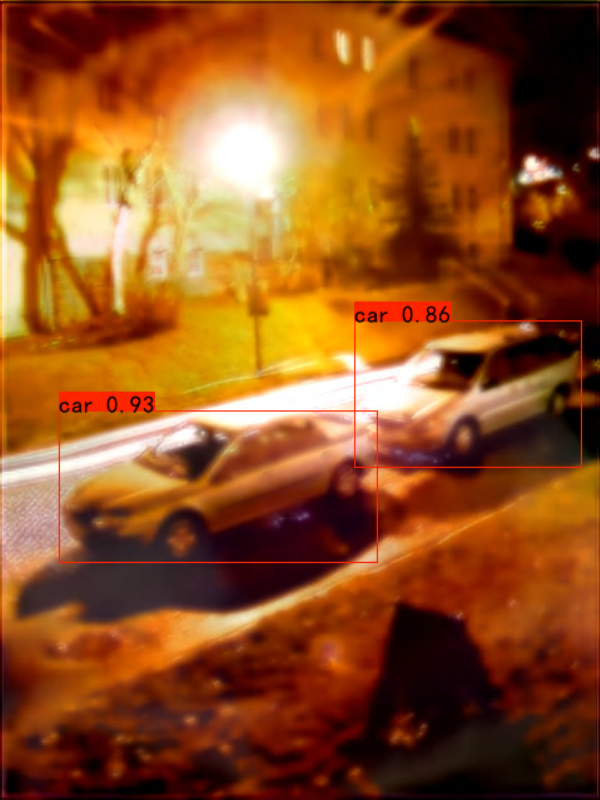} &
            \includegraphics[width=0.16\linewidth]{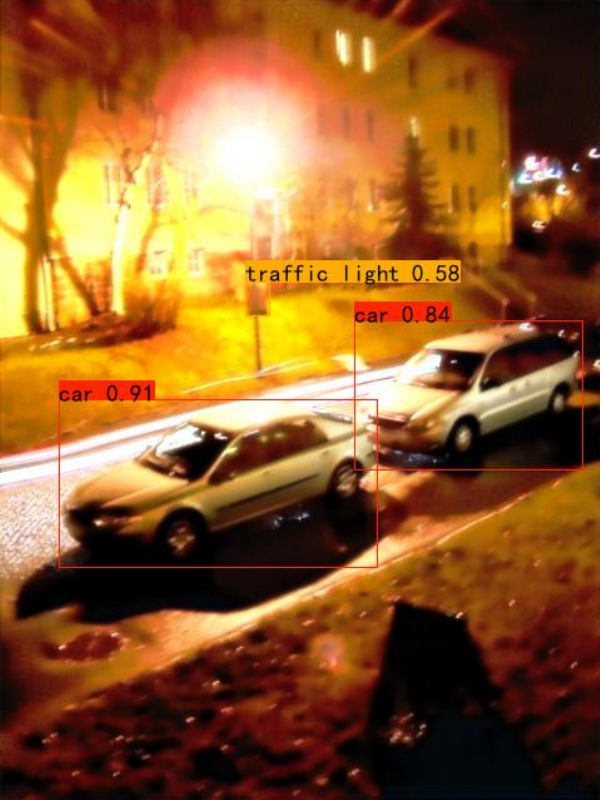} \\
            \noalign{\vspace{2pt}}
            \scriptsize \textbf{Input} &
            \scriptsize \textbf{CWNet} &
            \scriptsize \textbf{FourLLIE} &
            \scriptsize \textbf{Wave-Mamba} &
            \scriptsize \textbf{UHDFour} &
            \scriptsize \textbf{Ours} \\
        \end{tabular}
    }
    \vspace{-0.2cm}
    \caption{Qualitative comparison of object detection results in low-light scenes on the ExDark dataset.}
    \vspace{-0.2cm}
    \label{fig:detection_vis}
\end{figure}

\subsection{Visual Comparison of Ablation Results}

\begin{figure}[t]
    \centering
    \resizebox{\linewidth}{!}{
        \setlength{\tabcolsep}{0.5pt} 
        \renewcommand{\arraystretch}{0.5} 
        \begin{tabular}{cccccc}
            \includegraphics[width=0.16\linewidth]{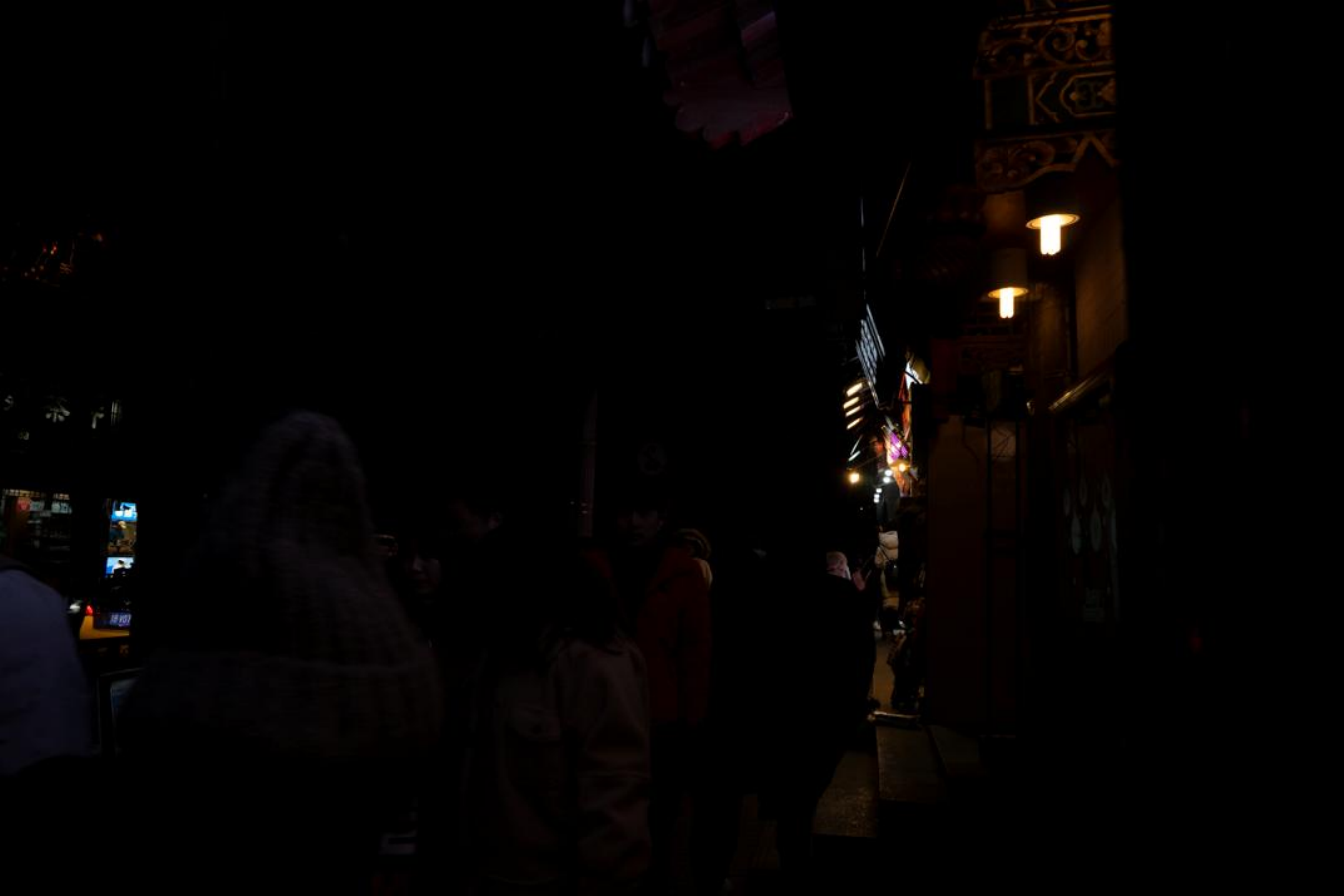} &
            \includegraphics[width=0.16\linewidth]{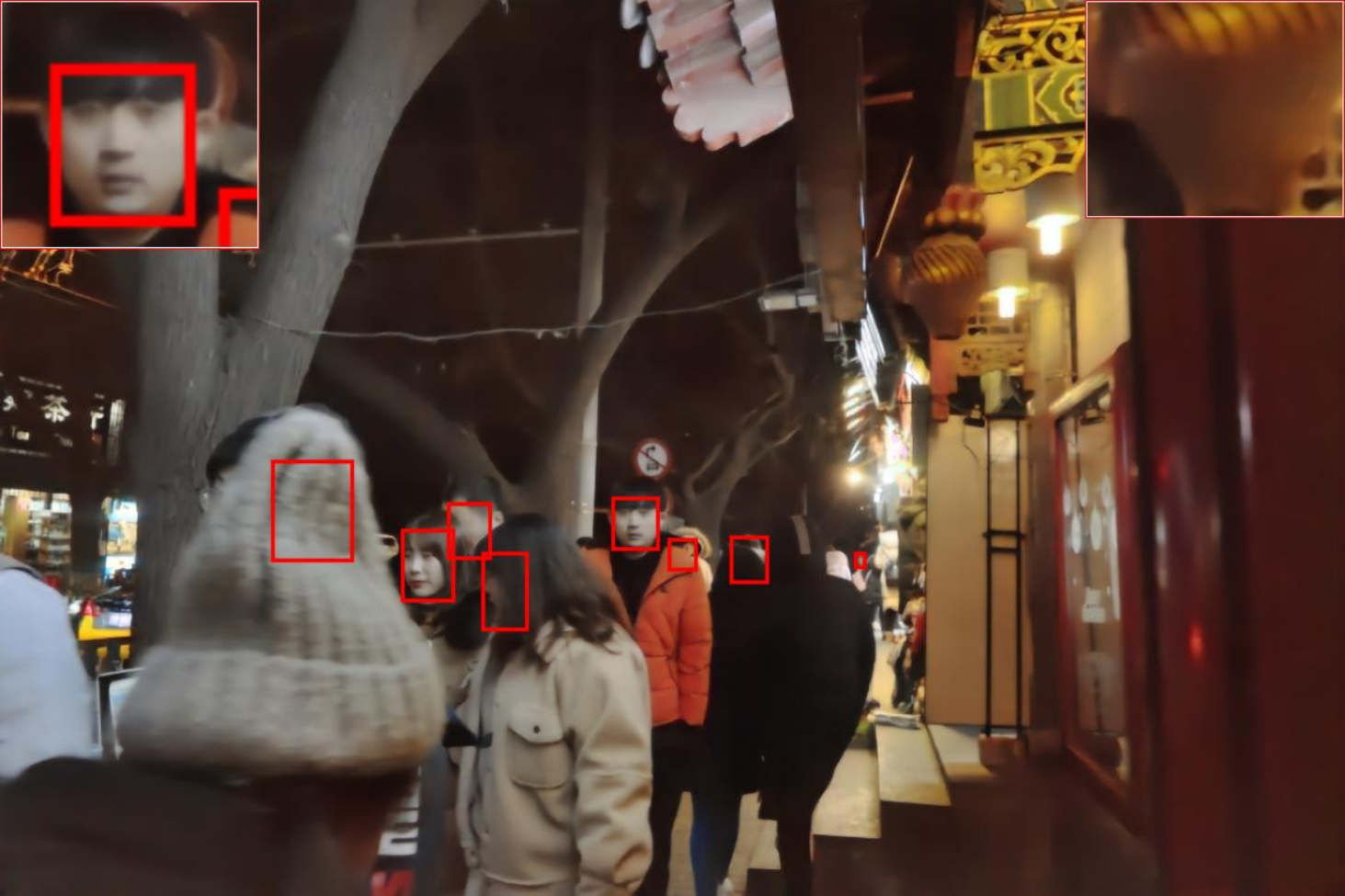} &
            \includegraphics[width=0.16\linewidth]{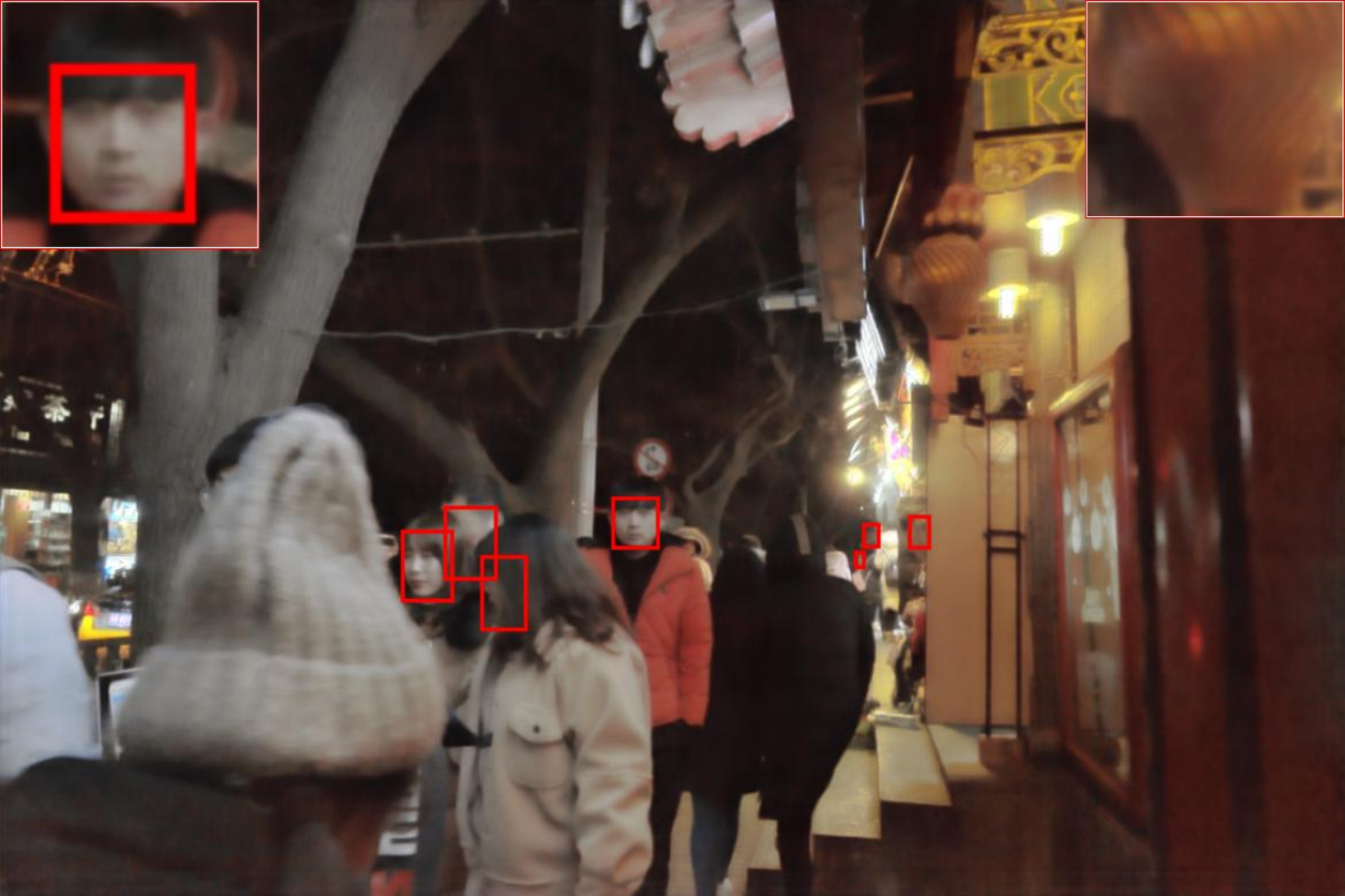} &
            \includegraphics[width=0.16\linewidth]{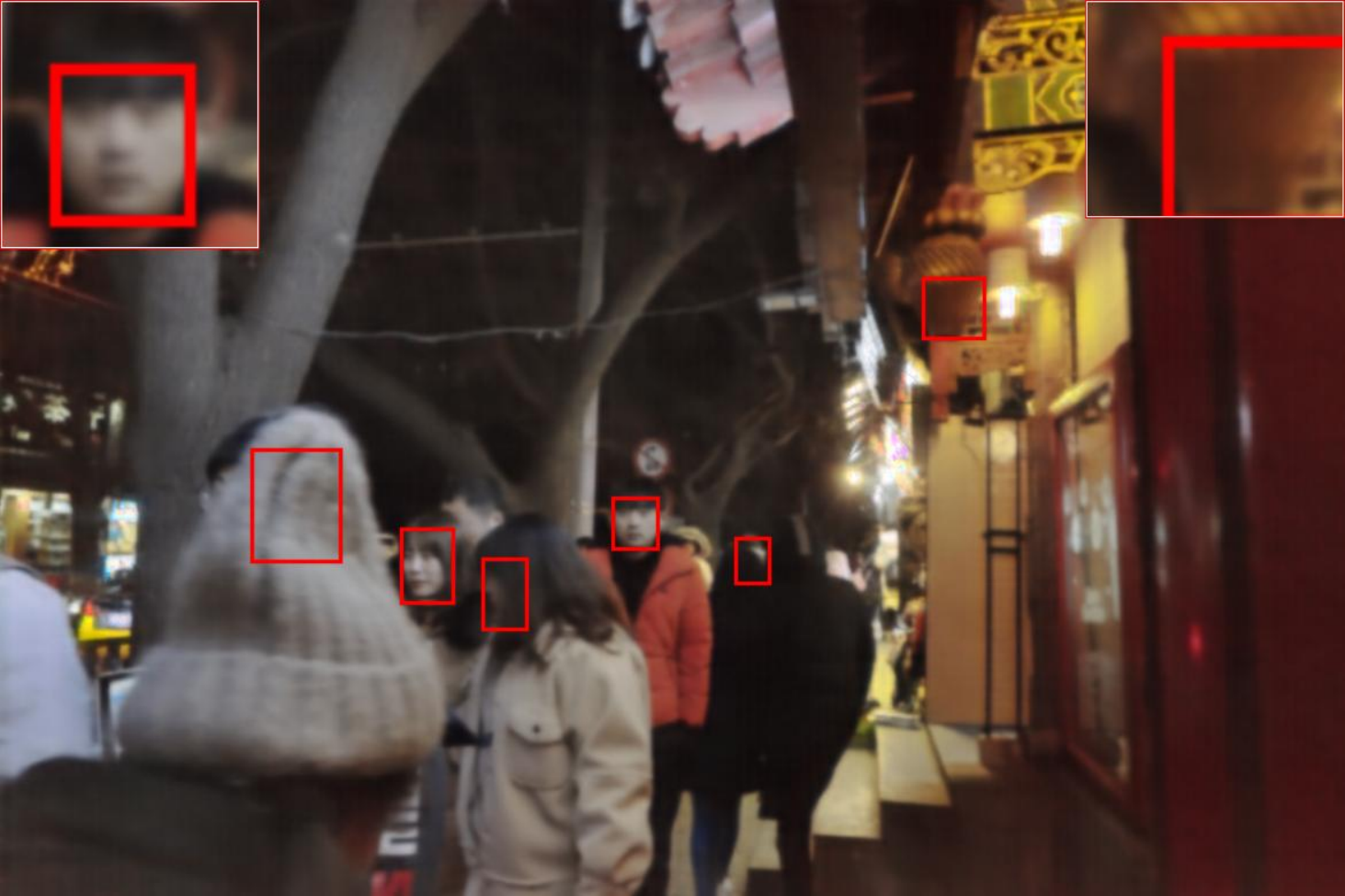} &
            \includegraphics[width=0.16\linewidth]{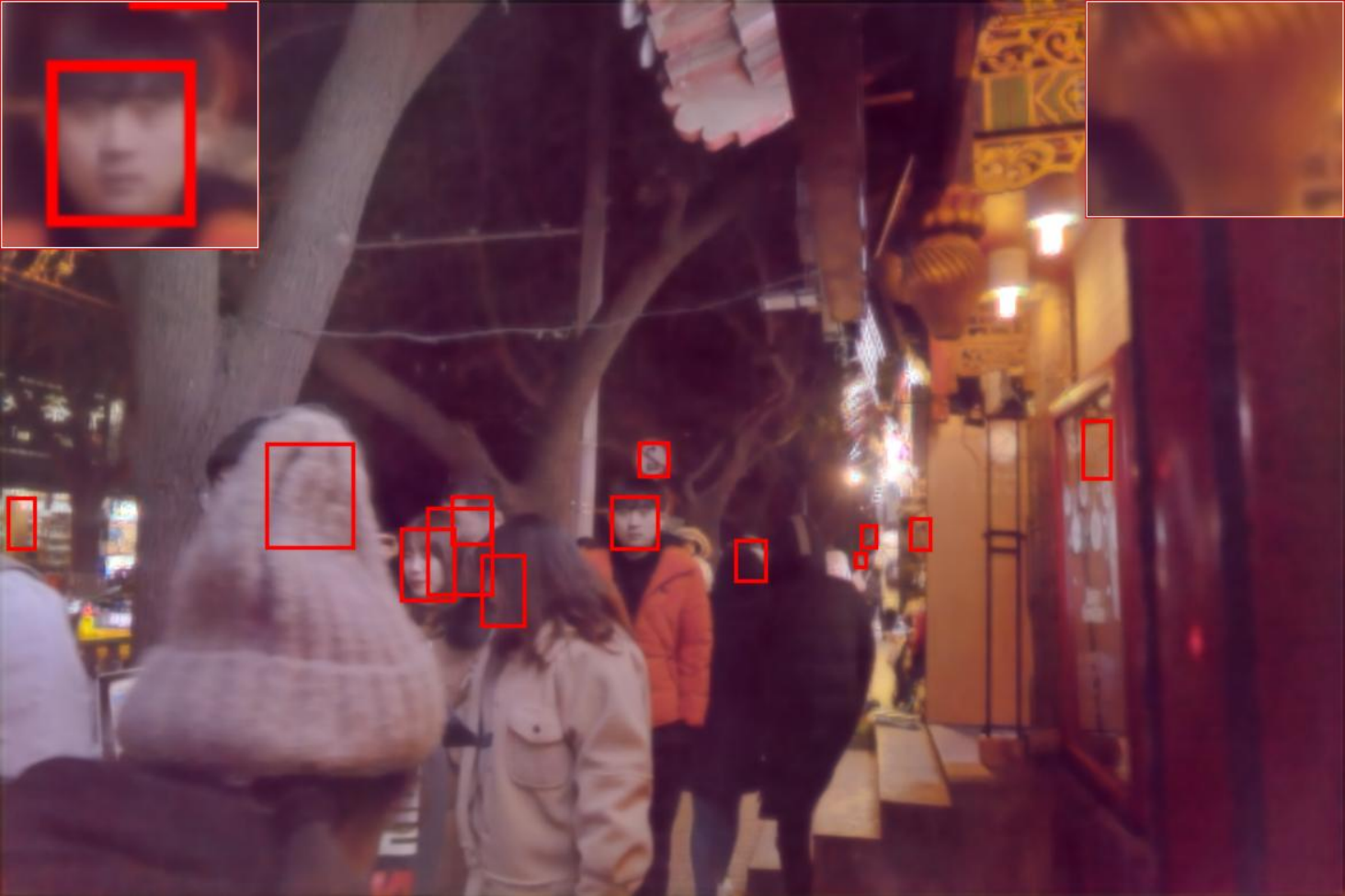} &
            \includegraphics[width=0.16\linewidth]{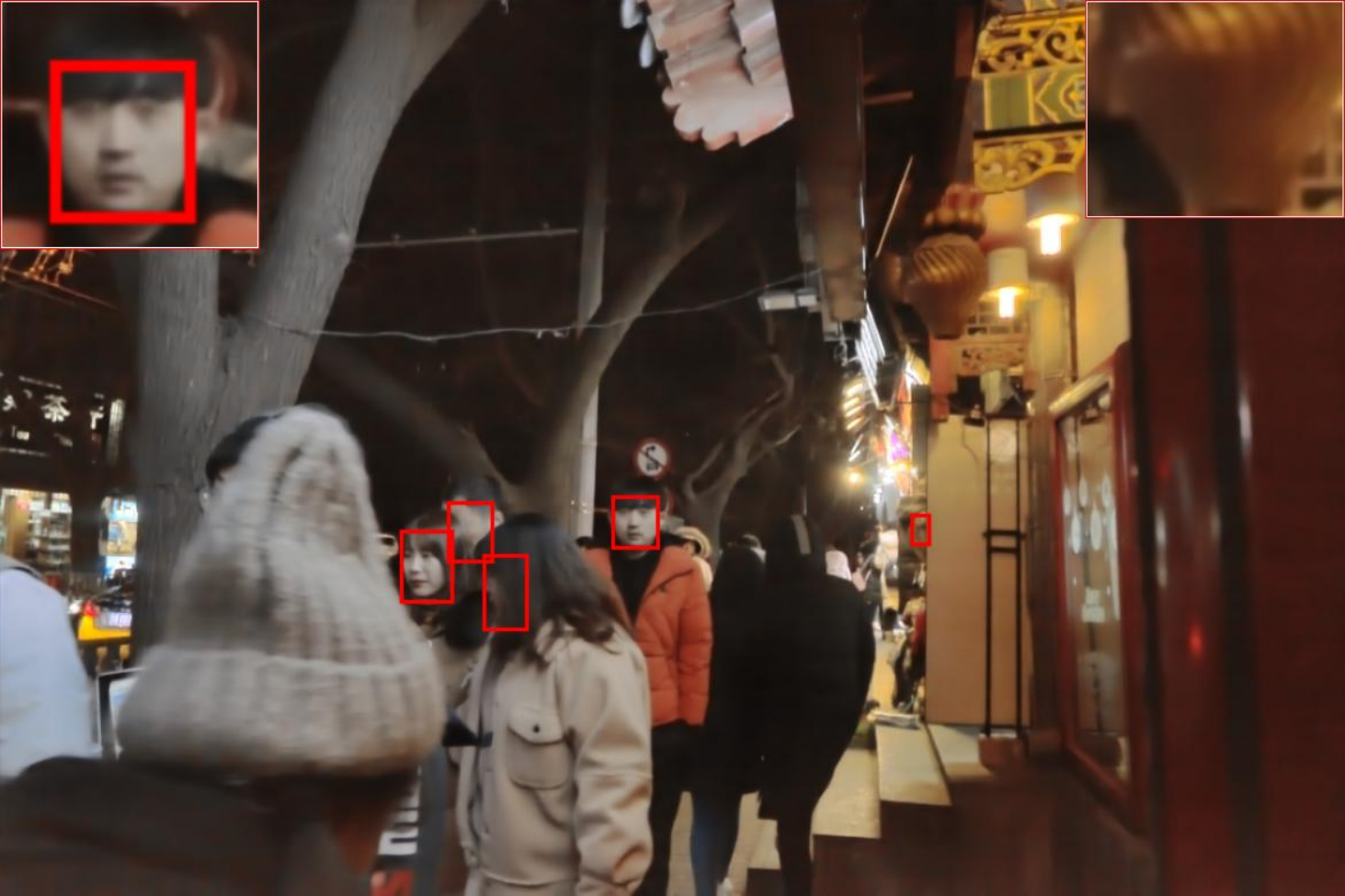} \\
            \includegraphics[width=0.16\linewidth]{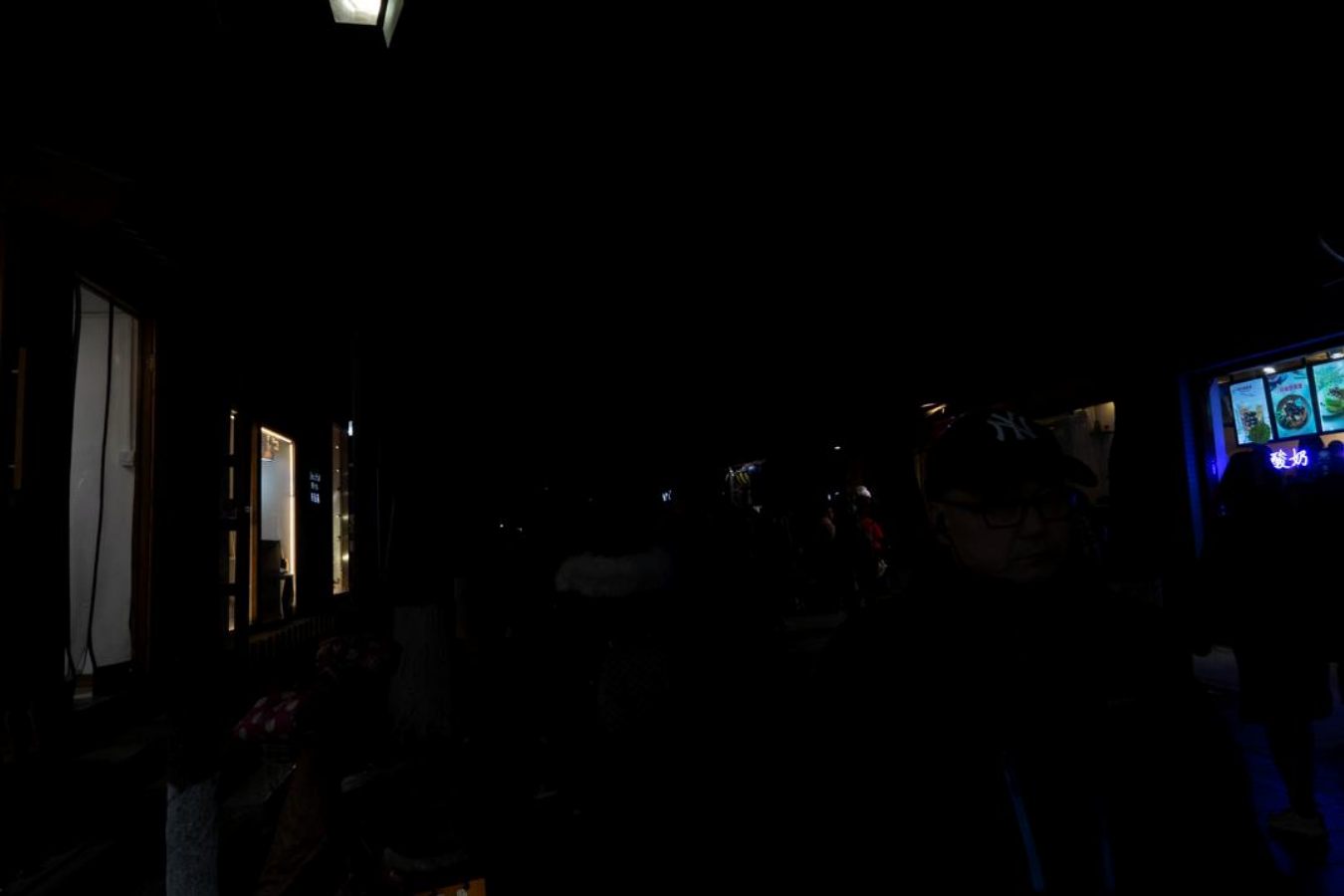} &
            \includegraphics[width=0.16\linewidth]{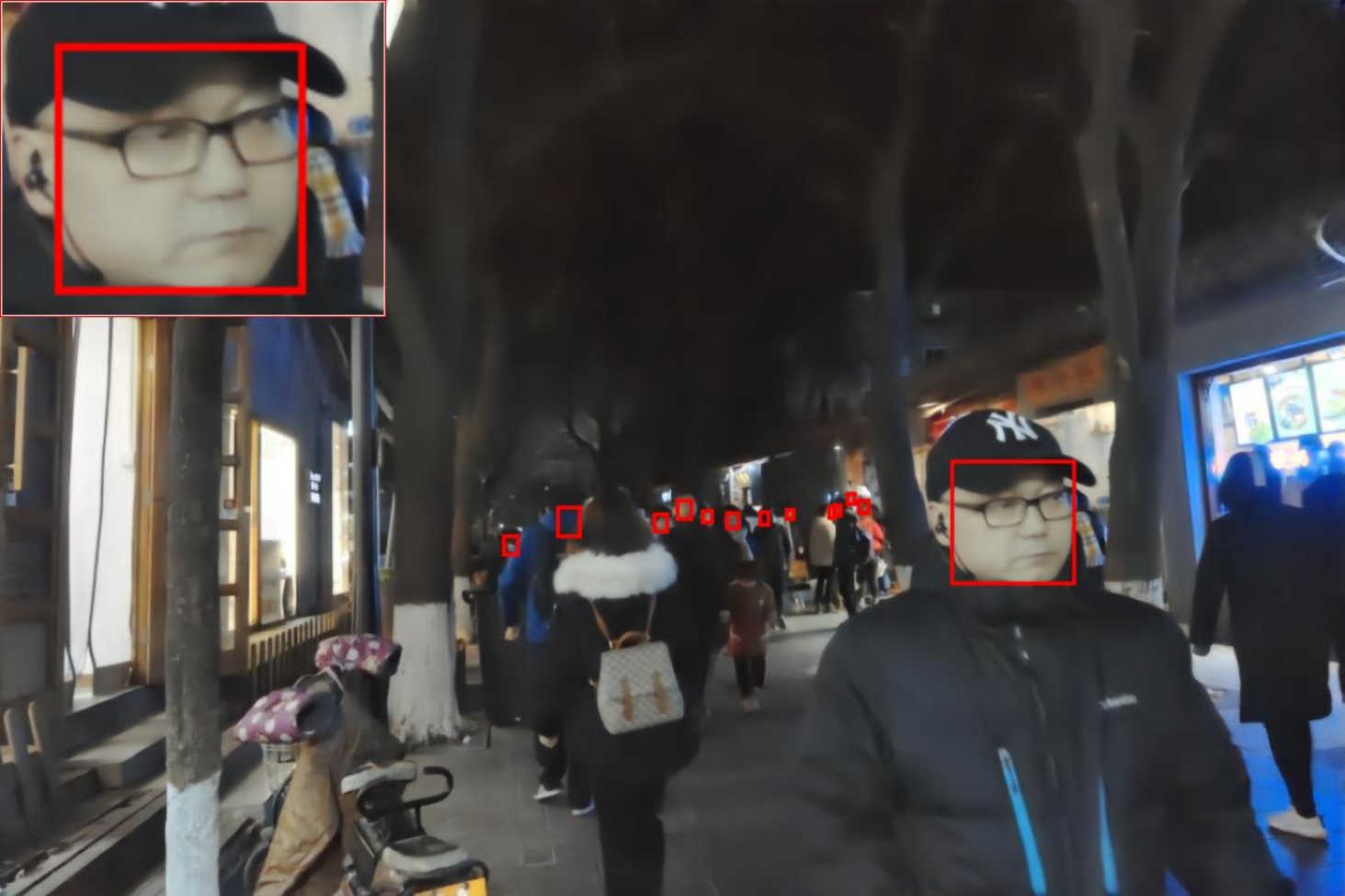} &
            \includegraphics[width=0.16\linewidth]{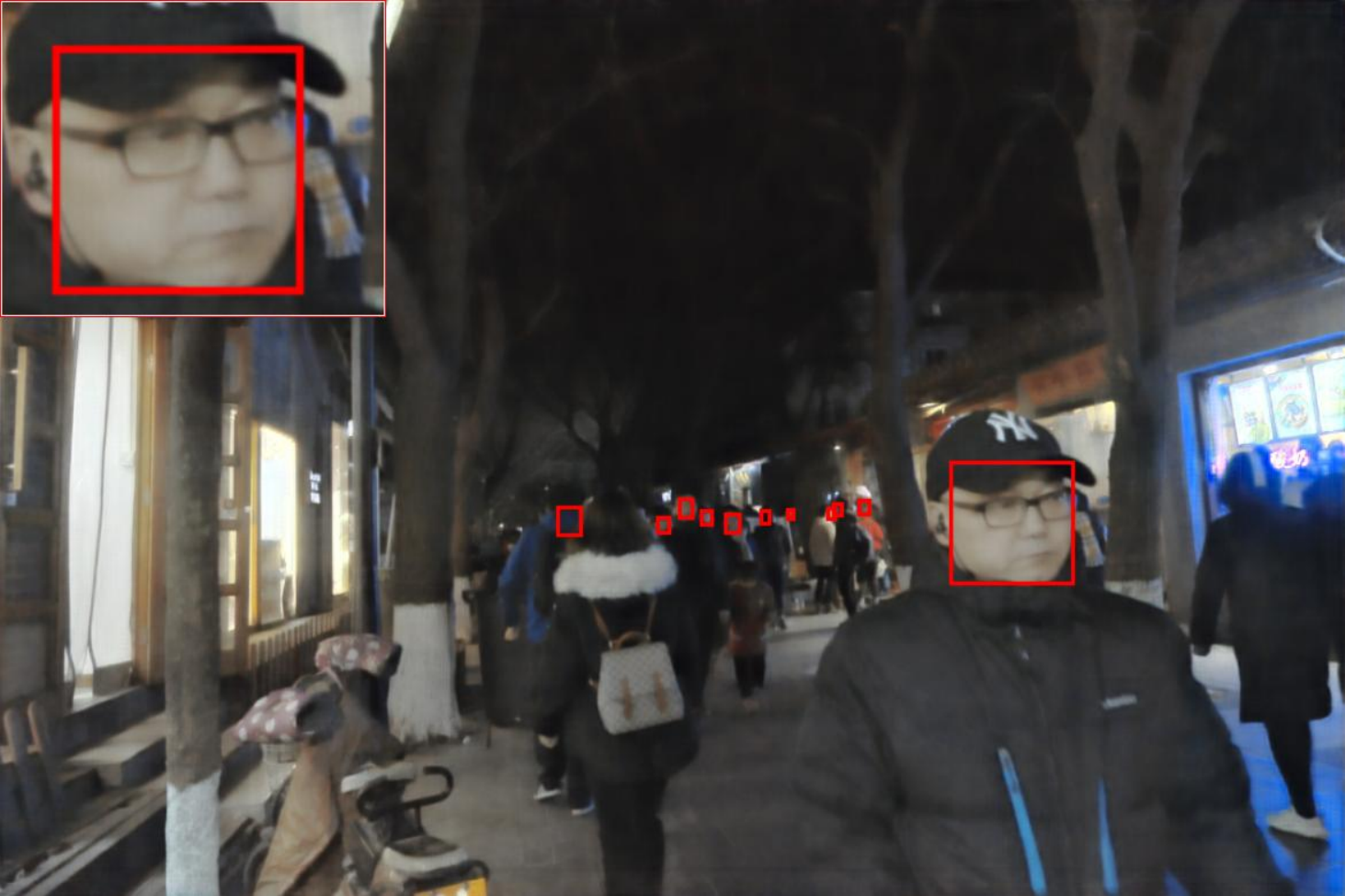} &
            \includegraphics[width=0.16\linewidth]{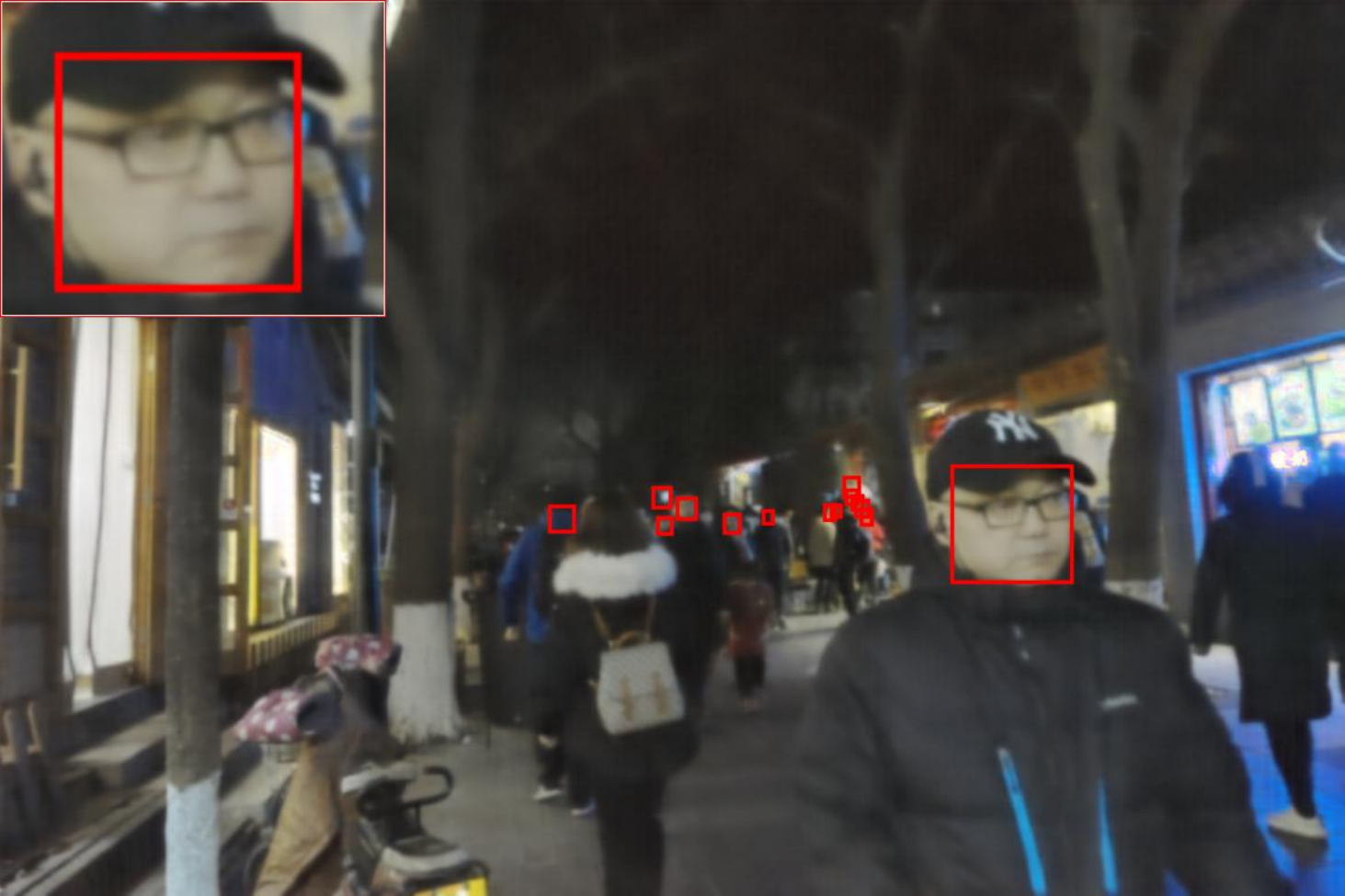} &
            \includegraphics[width=0.16\linewidth]{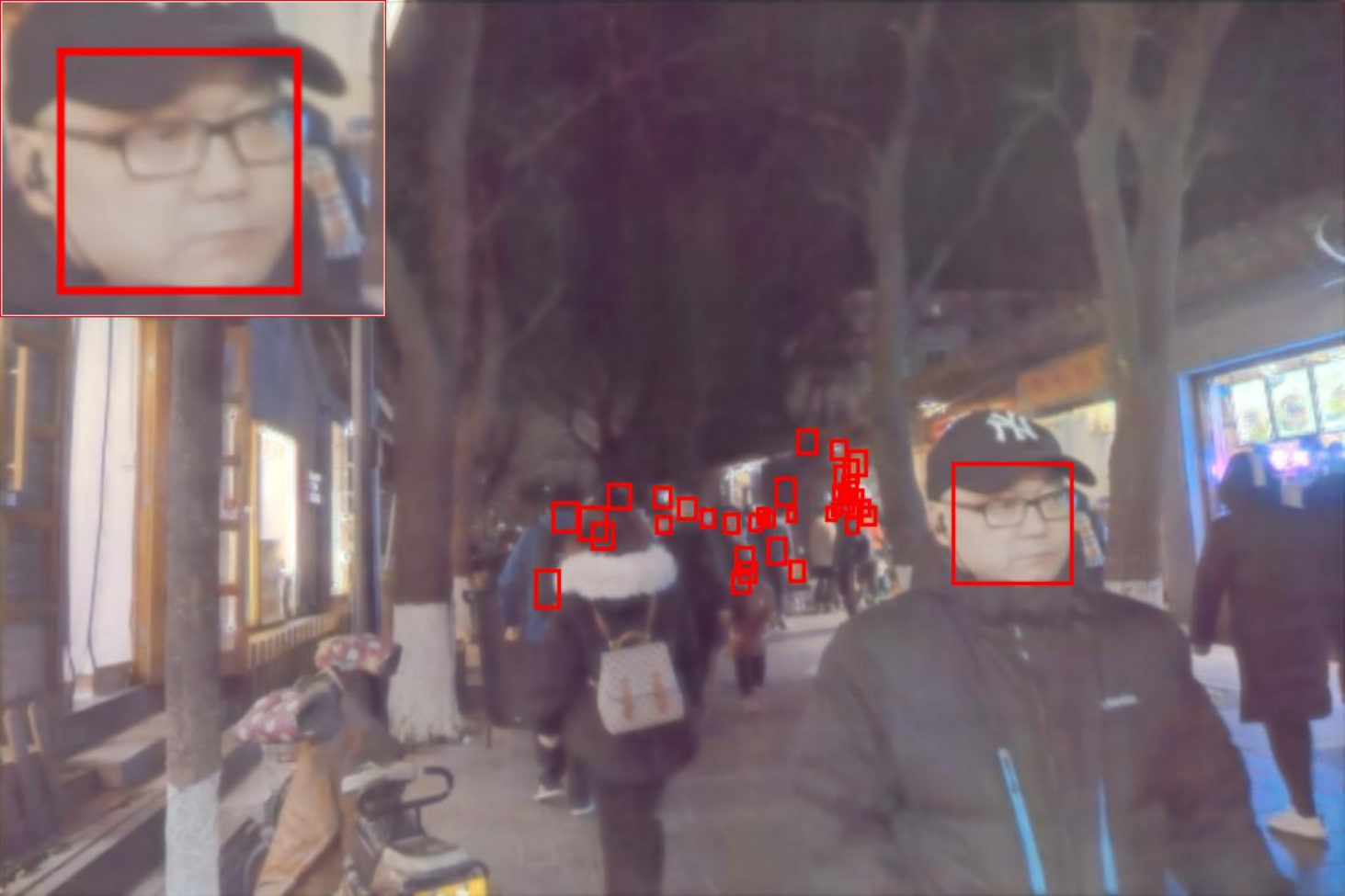} &
            \includegraphics[width=0.16\linewidth]{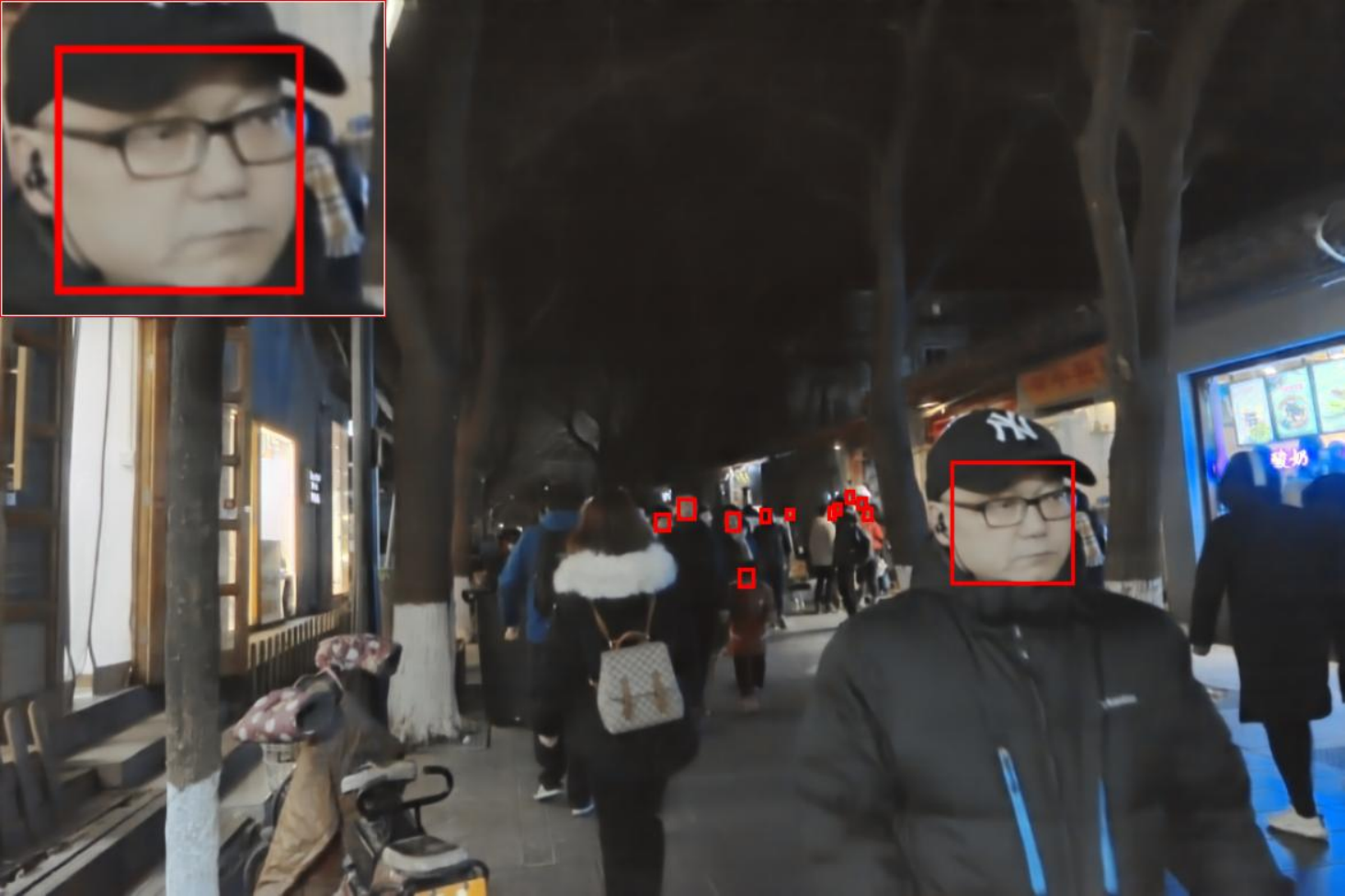} \\
            \includegraphics[width=0.16\linewidth]{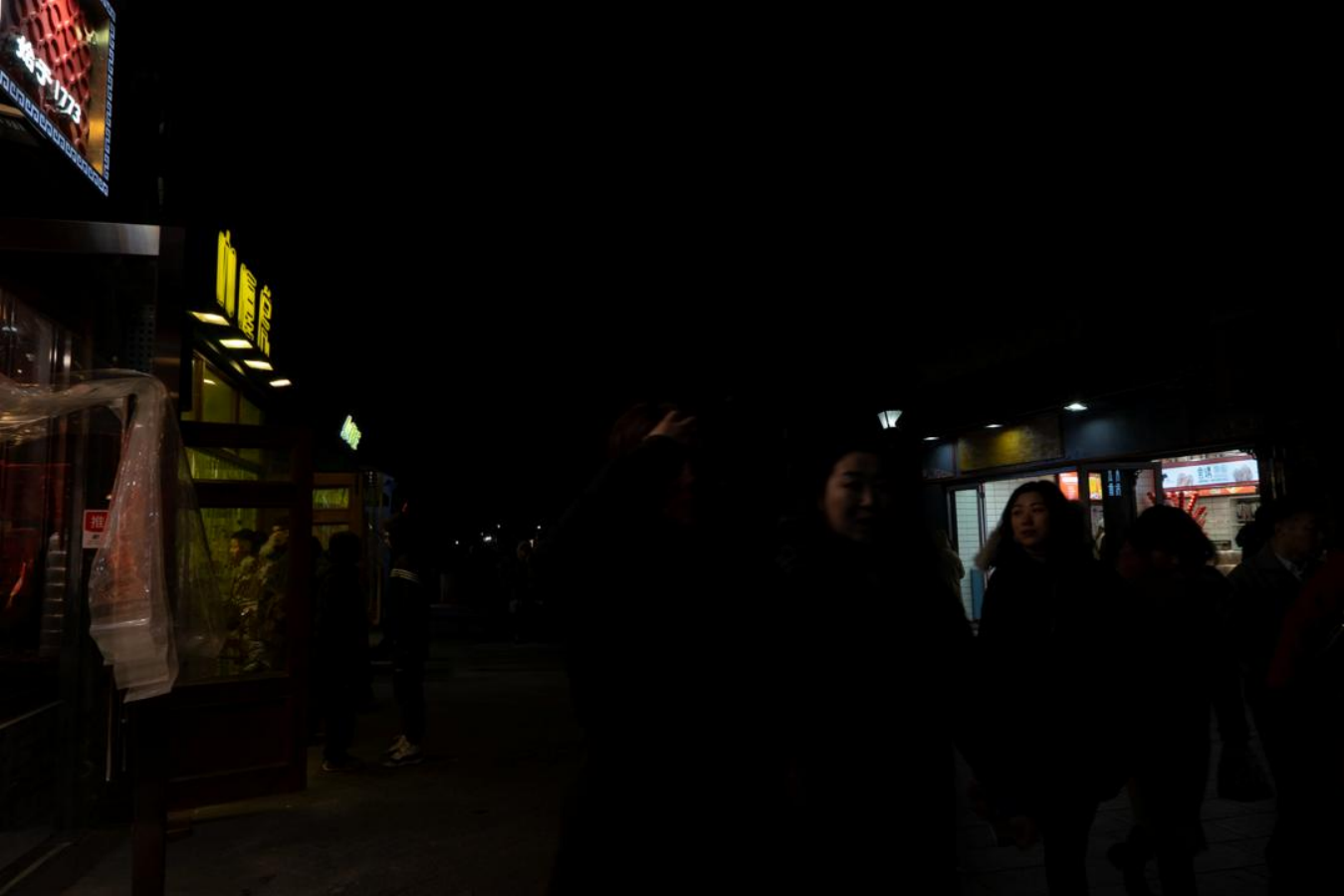} &
            \includegraphics[width=0.16\linewidth]{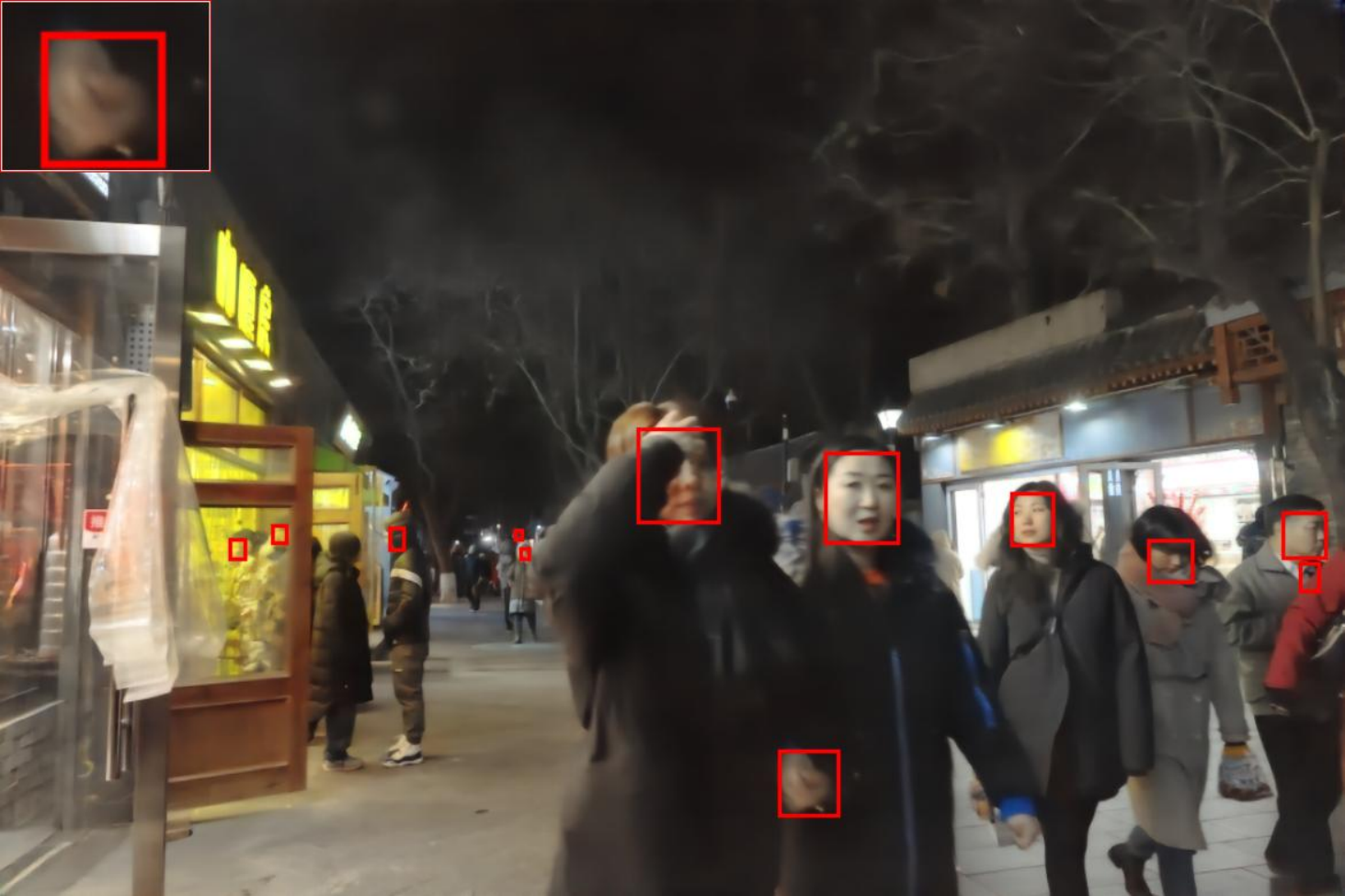} &
            \includegraphics[width=0.16\linewidth]{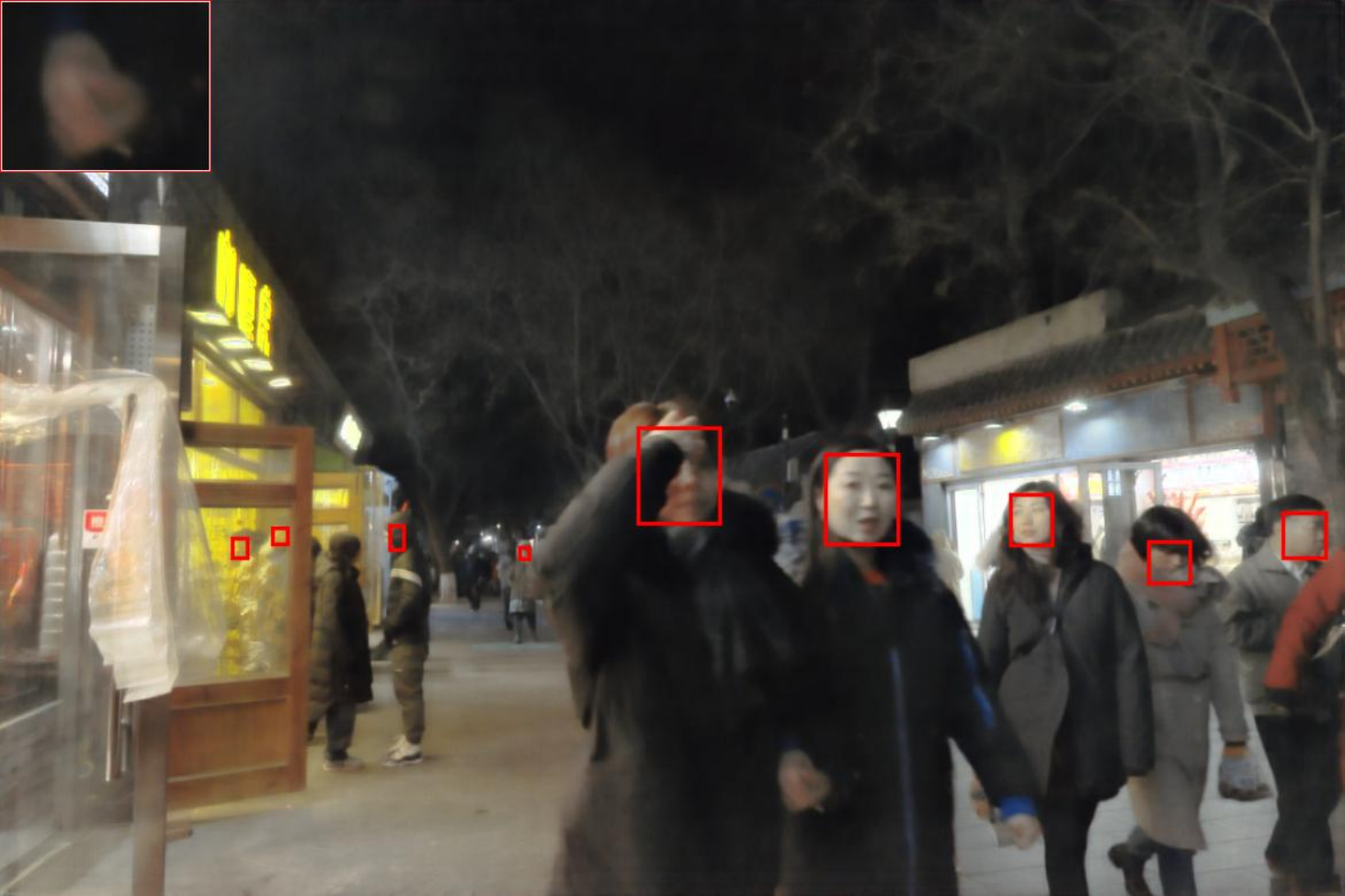} &
            \includegraphics[width=0.16\linewidth]{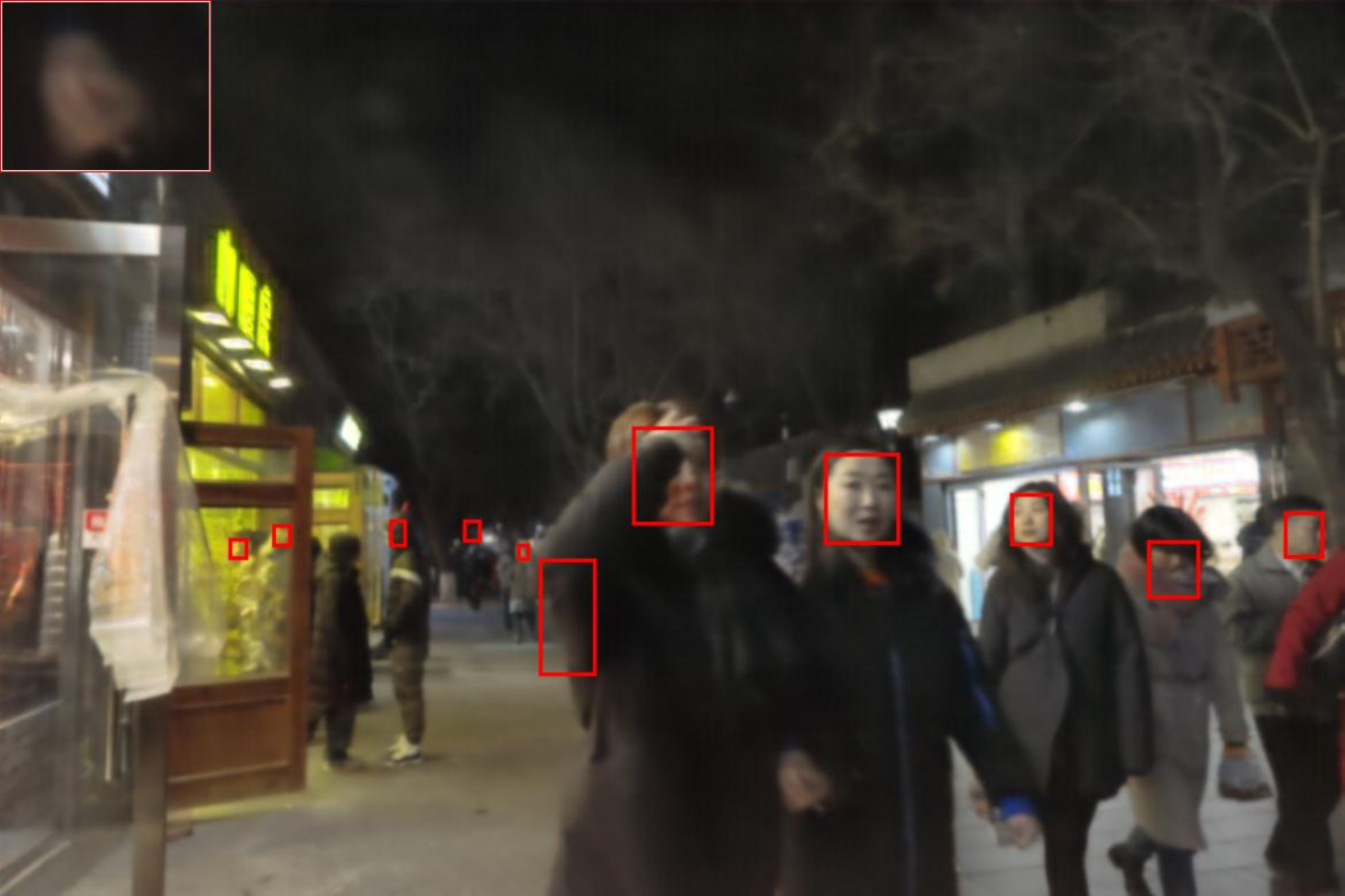} &
            \includegraphics[width=0.16\linewidth]{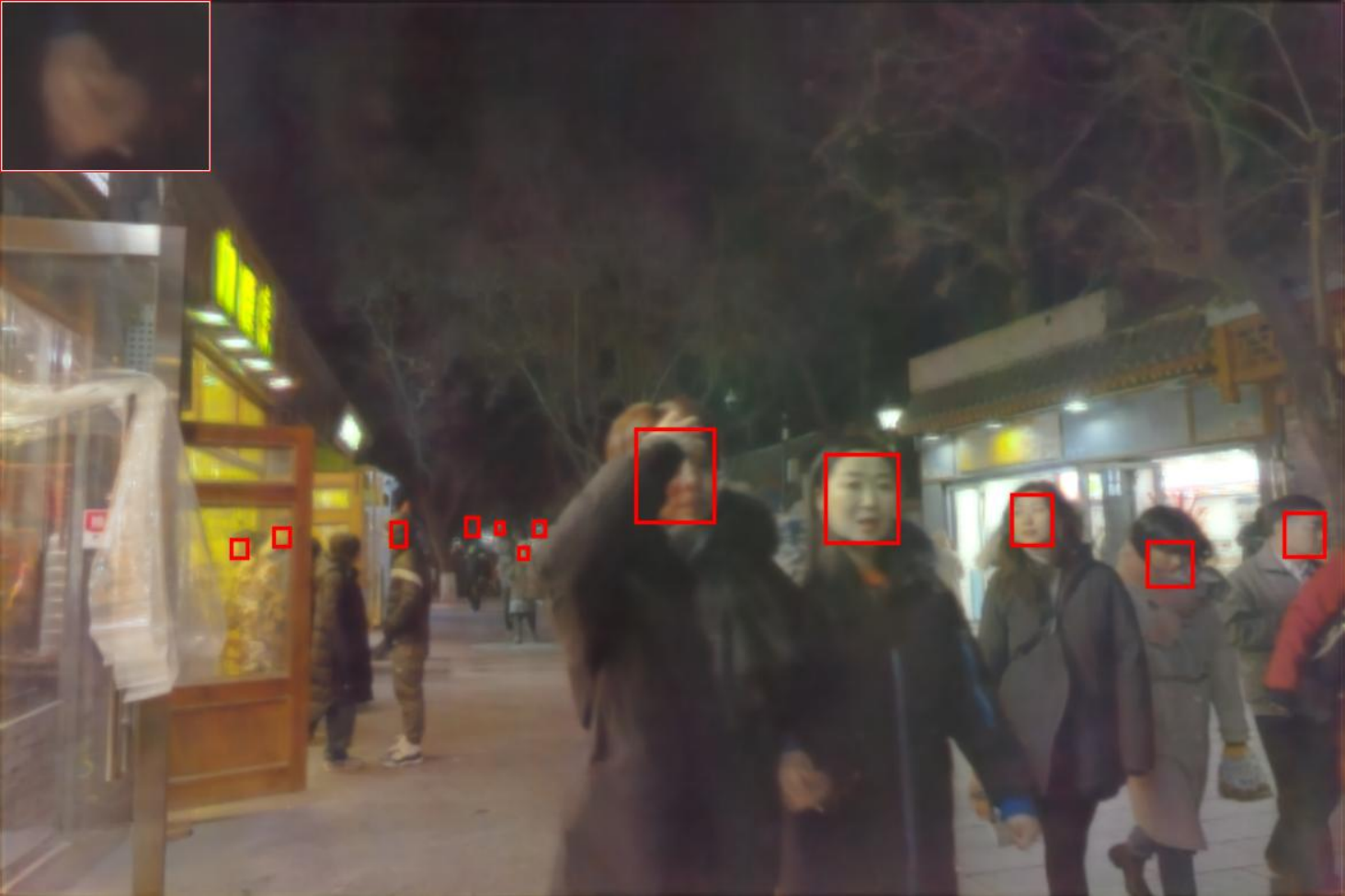} &
            \includegraphics[width=0.16\linewidth]{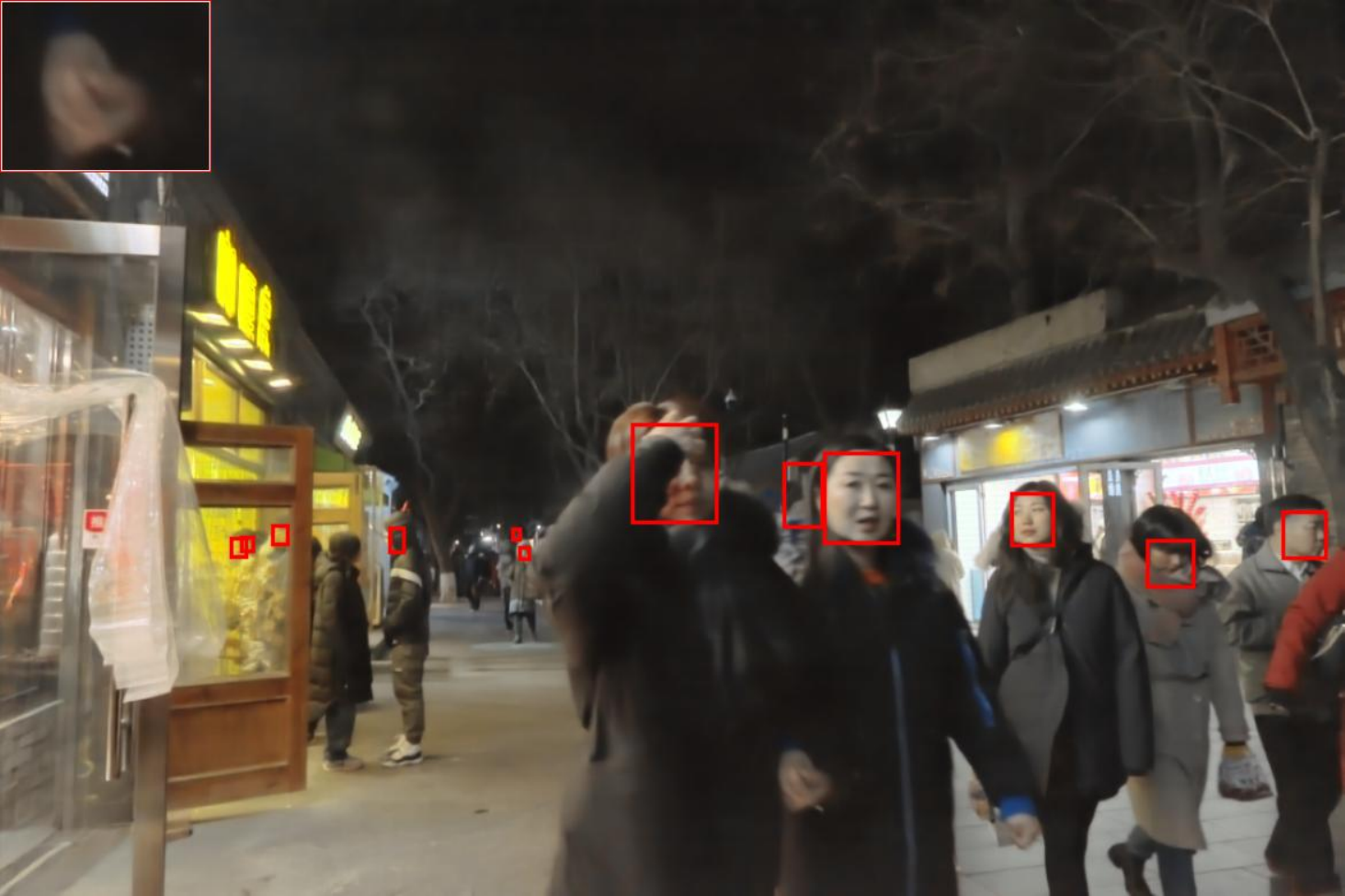} \\
            \includegraphics[width=0.16\linewidth]{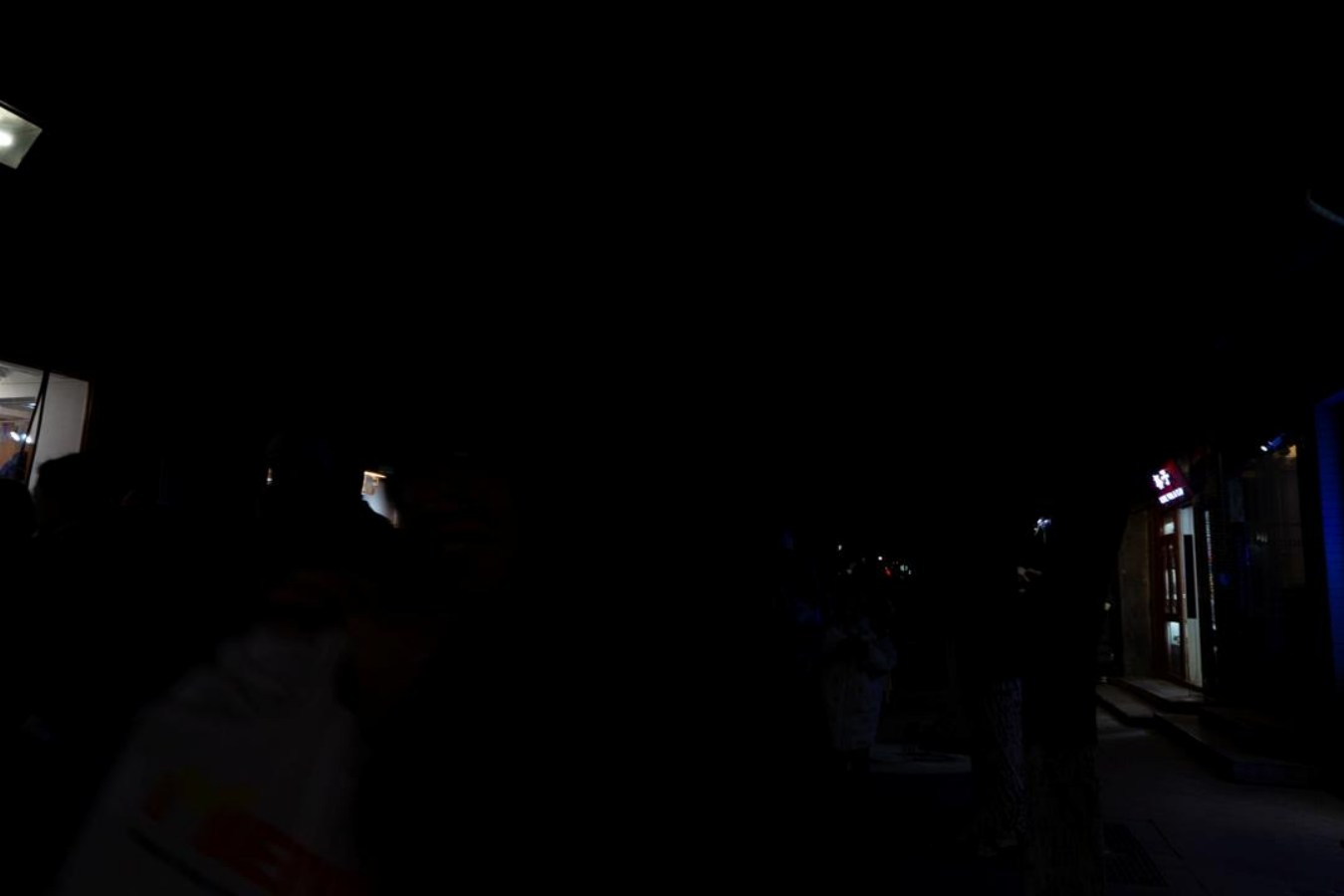} &
            \includegraphics[width=0.16\linewidth]{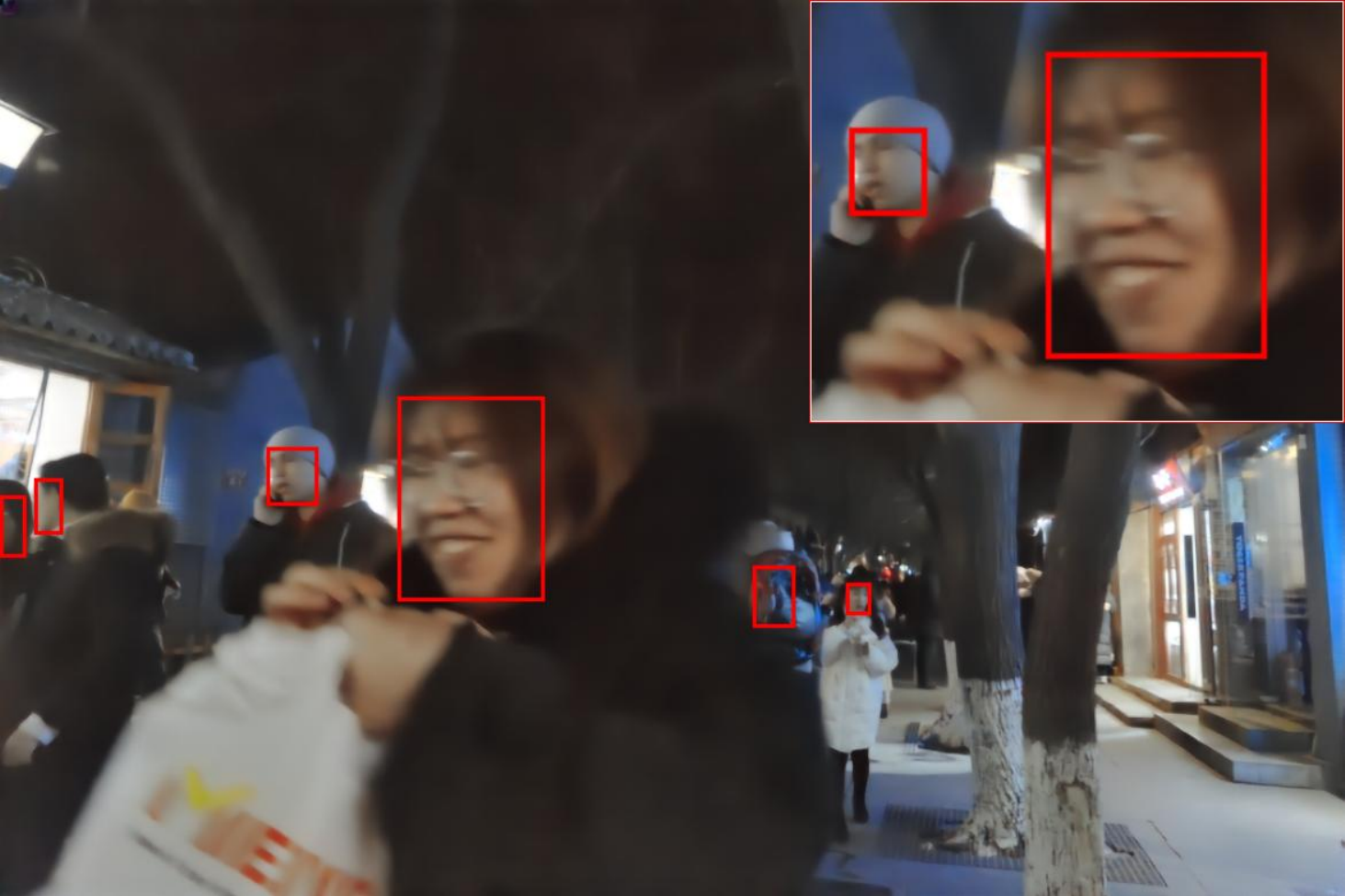} &
            \includegraphics[width=0.16\linewidth]{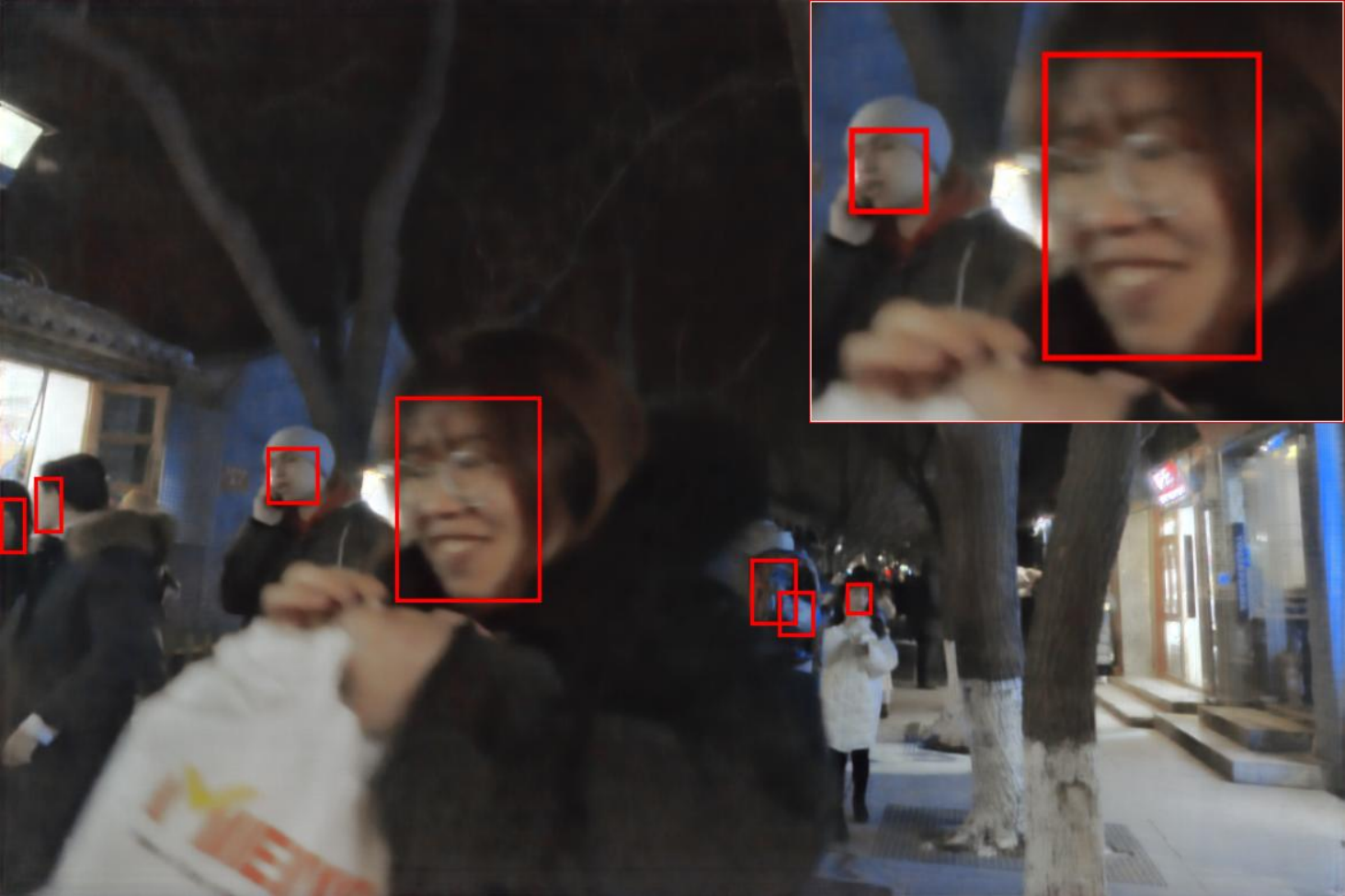} &
            \includegraphics[width=0.16\linewidth]{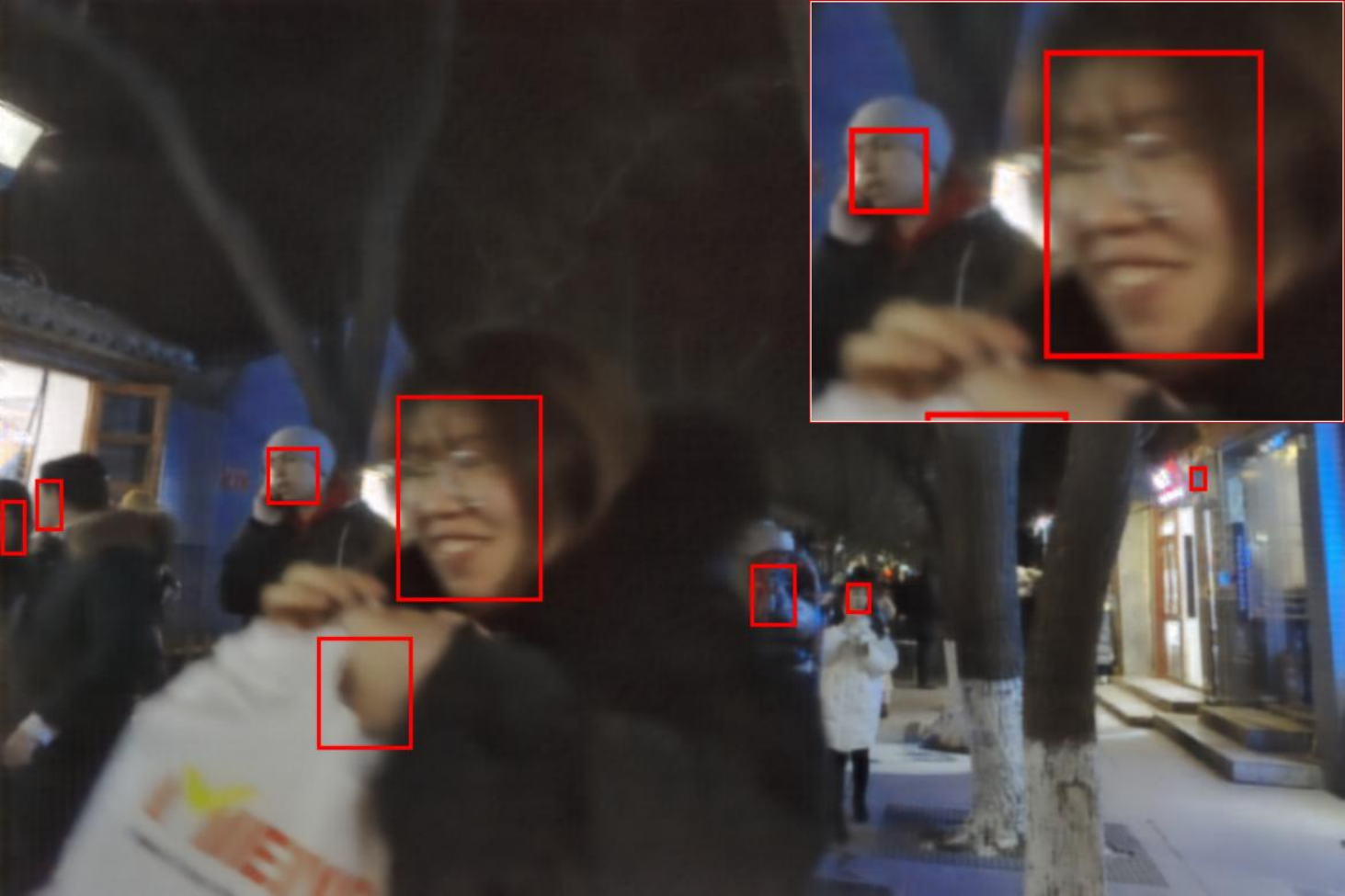} &
            \includegraphics[width=0.16\linewidth]{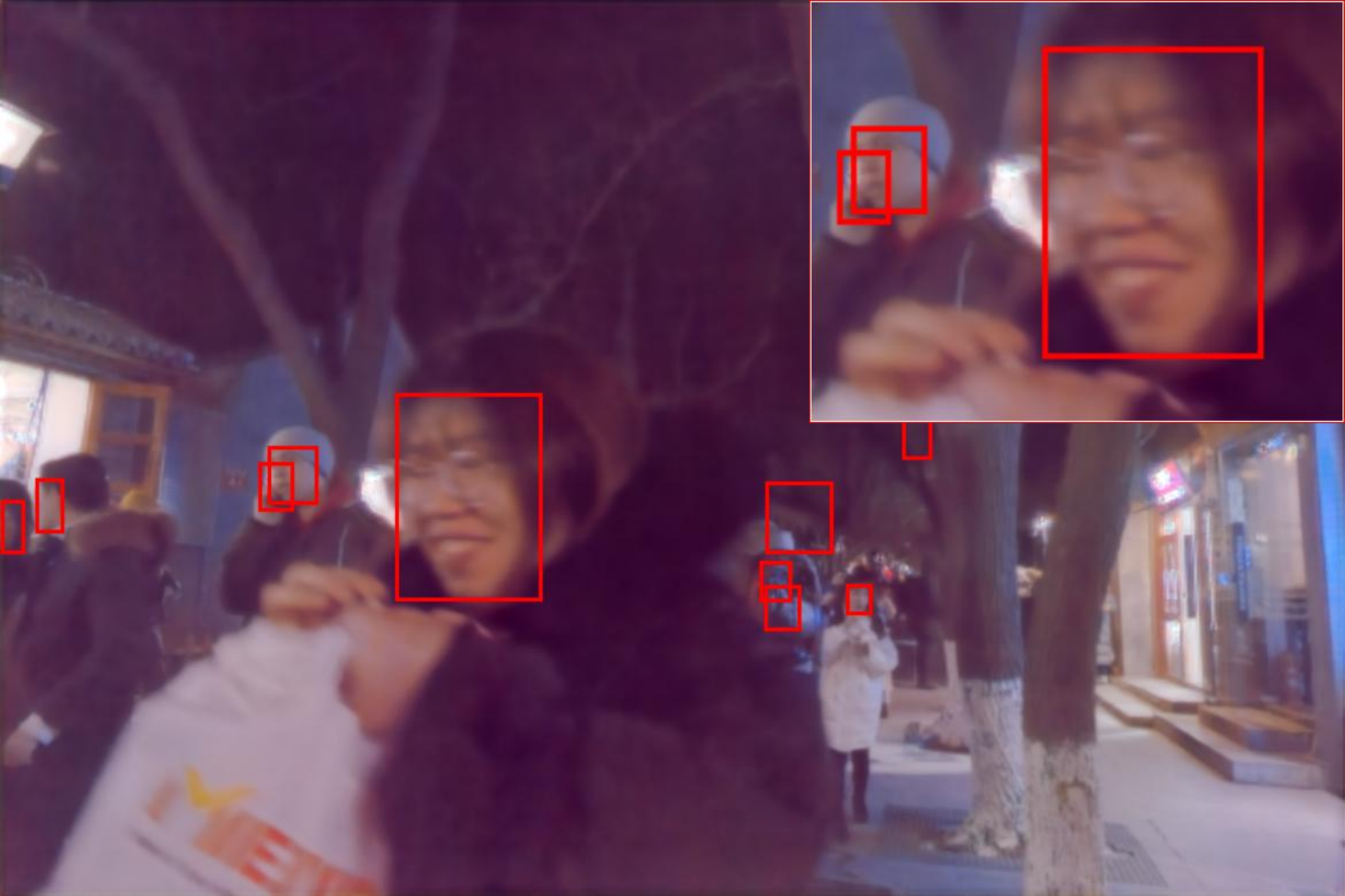} &
            \includegraphics[width=0.16\linewidth]{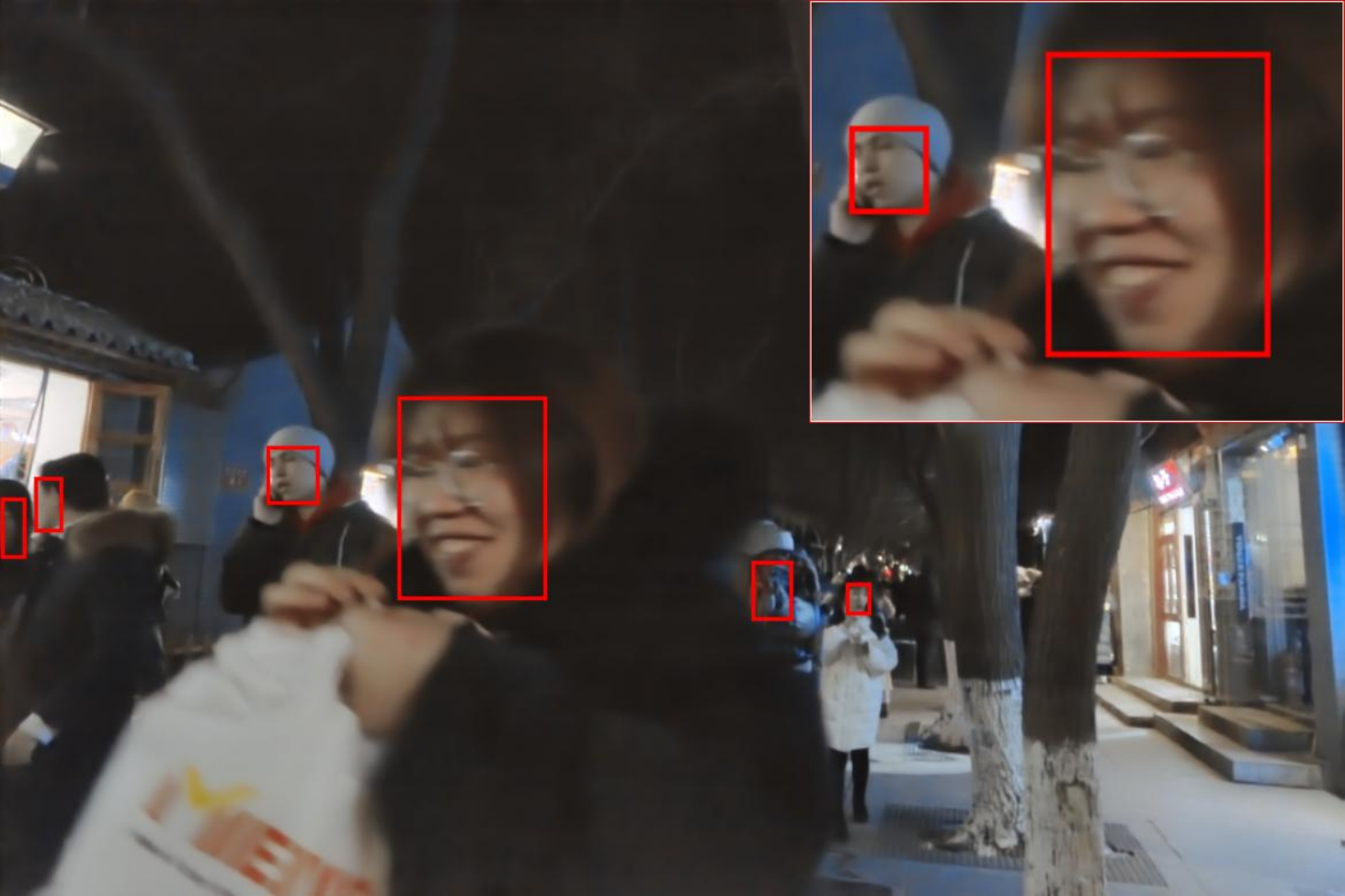} \\
            \noalign{\vspace{2pt}}
            \scriptsize \textbf{Input} &
            \scriptsize \textbf{CWNet} &
            \scriptsize \textbf{FourLLIE} &
            \scriptsize \textbf{Wave-Mamba} &
            \scriptsize \textbf{UHDFour} &
            \scriptsize \textbf{Ours} \\
            
        \end{tabular}
    }
    \vspace{-0.2cm}
    \caption{Qualitative comparison of face detection results on the DarkFace dataset.}
    \vspace{-0.2cm}
    \vspace{-0.2cm}
    \label{fig:face_detection_vis}
\end{figure}

\subsubsection{Impact of APACF}
The visual results in Fig.~\ref{fig:ablation_apacf_vis} further support this observation. Without the frequency branch, the restored image still lacks local clarity in dark regions. Hard splitting produces blurrier structures and weaker detail recovery in the highlighted rocky area. In contrast, APACF recovers clearer edges, more natural contrast, and richer textures. These results show that explicit amplitude-phase interaction is more effective than either removing frequency modeling or treating the two components independently.

\subsubsection{Prior Reliability Analysis}
As shown in Fig.~\ref{fig:ablation_prior_vis}, the raw prior fails to provide sufficient structural guidance, while the GT prior leads to obvious detail inconsistency in the enlarged branch region. By contrast, the endogenous prior produces sharper structures, better illumination, and more natural appearance. These results confirm that a reliable prior should be not only clean, but also well matched to the feature space of the restoration network.

\subsection{Applications in Vision Tasks}
Beyond improving visual quality, low-light image enhancement can also serve as a preprocessing step for downstream vision tasks. To evaluate the practical utility of our method, we further test it on three representative tasks: semantic segmentation, object detection, and face detection. By applying off-the-shelf perception models to the enhanced images, we assess whether our method can recover structures and appearance cues that are beneficial for machine perception in challenging low-light scenes.

\subsubsection{Low-Light Semantic Segmentation}
We evaluate downstream segmentation performance on the NightCity dataset. Following~\cite{Liu2022ImprovingND}, we use the enhancement model trained on LSRW-Huawei and directly feed the enhanced images into the semantic segmentation framework for inference.
As shown in Fig.~\ref{fig:segmentation_vis}, our method produces segmentation results that are visually closer to the ground truth than those of competing methods. In particular, it better preserves object boundaries and avoids over-enhancement in already bright regions, which helps distinguish adjacent categories such as pedestrians, sidewalks, and roads. In addition, clearer structural details improve the recognition of large or texture-rich regions, such as buses, vegetation, and terrain. These results indicate that our method provides more reliable visual cues for nighttime semantic understanding.

\subsubsection{Low-Light Face Detection}
To assess the effectiveness in challenging real-world scenarios with dense crowds, we conduct experiments on the DarkFace dataset. The enhanced images are fed into the DSFD face detector~\cite{Li2018DSFDDS}, pretrained on the WiderFace~\cite{Yang2015WIDERFA} dataset.
 As shown in Fig.~\ref{fig:face_detection_vis}, our method improves facial visibility while preserving local contrast and boundary details. Compared with other enhancement methods, it produces more complete and cleaner detections, especially in crowded scenes and dim regions where faces are small or partially obscured. These results further demonstrate that our method can provide effective low-light preprocessing for fine-grained visual perception tasks.

\subsubsection{Low-Light Object Detection}
We further evaluate object detection on the ExDark dataset. Following~\cite{Reis2023RealTimeFO}, we randomly sample 10\% of the images for evaluation and use a YOLOv8 detector pretrained on COCO for direct inference without fine-tuning.
As shown in Fig.~\ref{fig:detection_vis}, our method consistently improves detection quality in both extremely dark and high-contrast scenes. Compared with other enhancement methods, it better suppresses overexposure around strong light sources while preserving the visibility of target objects. This leads to higher detection confidence and more accurate localization for objects such as boats, persons, backpacks, and traffic lights. The results suggest that our method restores visual features in a way that is more compatible with generic object detectors.

\begin{figure}[!t]
    \centering
    \subfloat[Input]{%
        \includegraphics[width=0.32\linewidth]{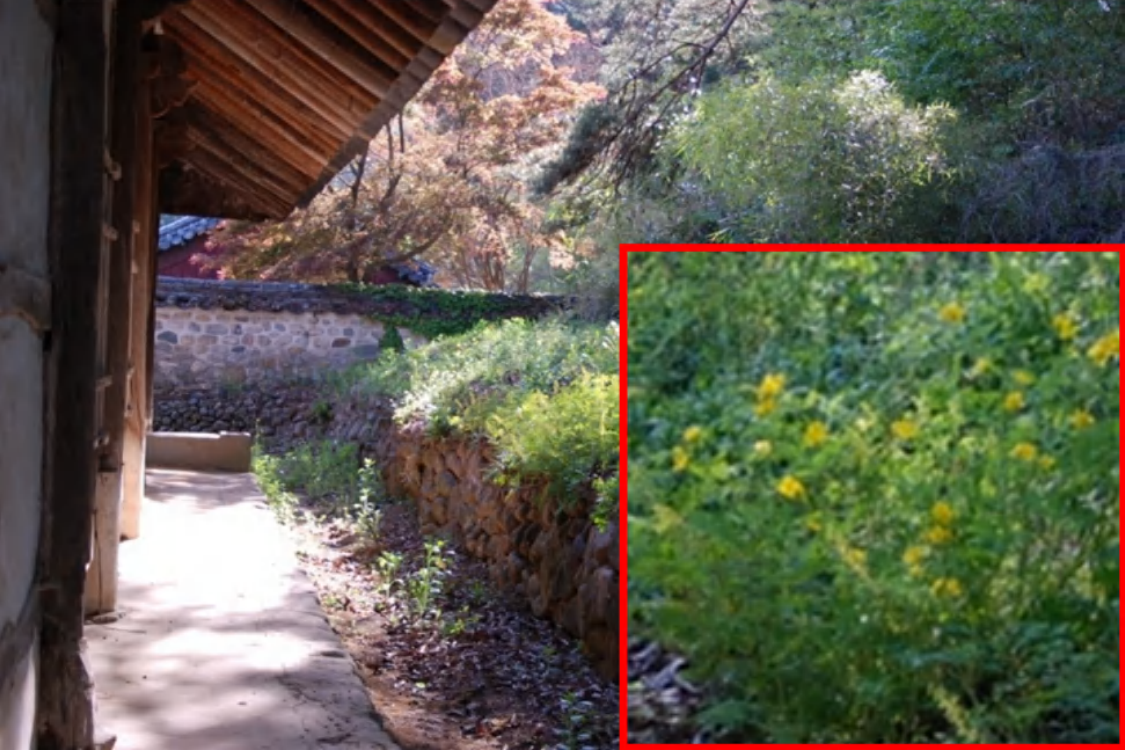}%
    }\hfil
    \subfloat[PairLIE]{%
        \includegraphics[width=0.32\linewidth]{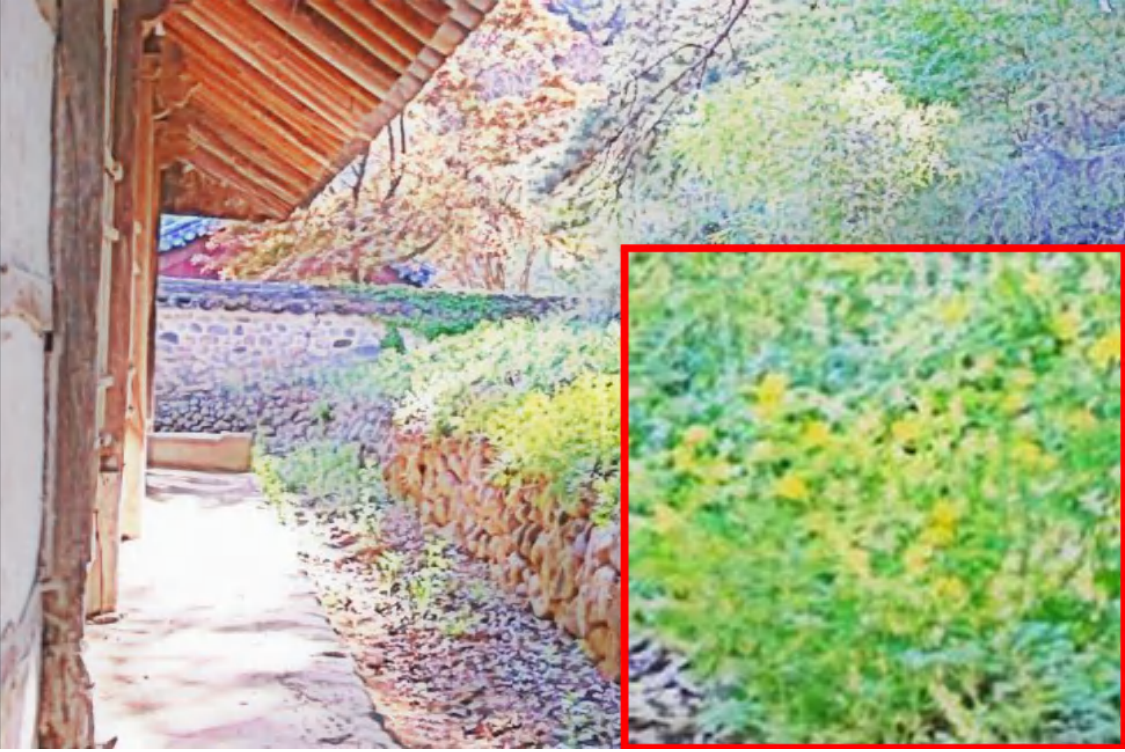}%
    }\hfil
    \subfloat[Retinexformer]{%
        \includegraphics[width=0.32\linewidth]{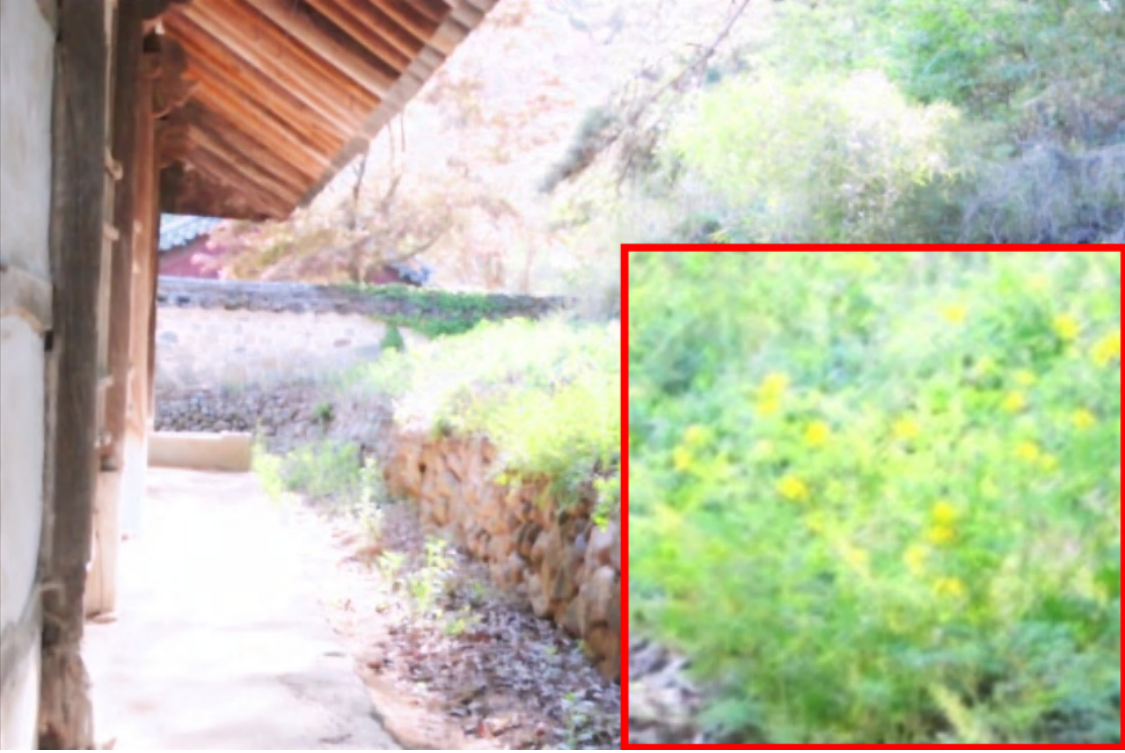}%
    }
    \vspace{1mm} 
    \subfloat[SKF]{%
        \includegraphics[width=0.32\linewidth]{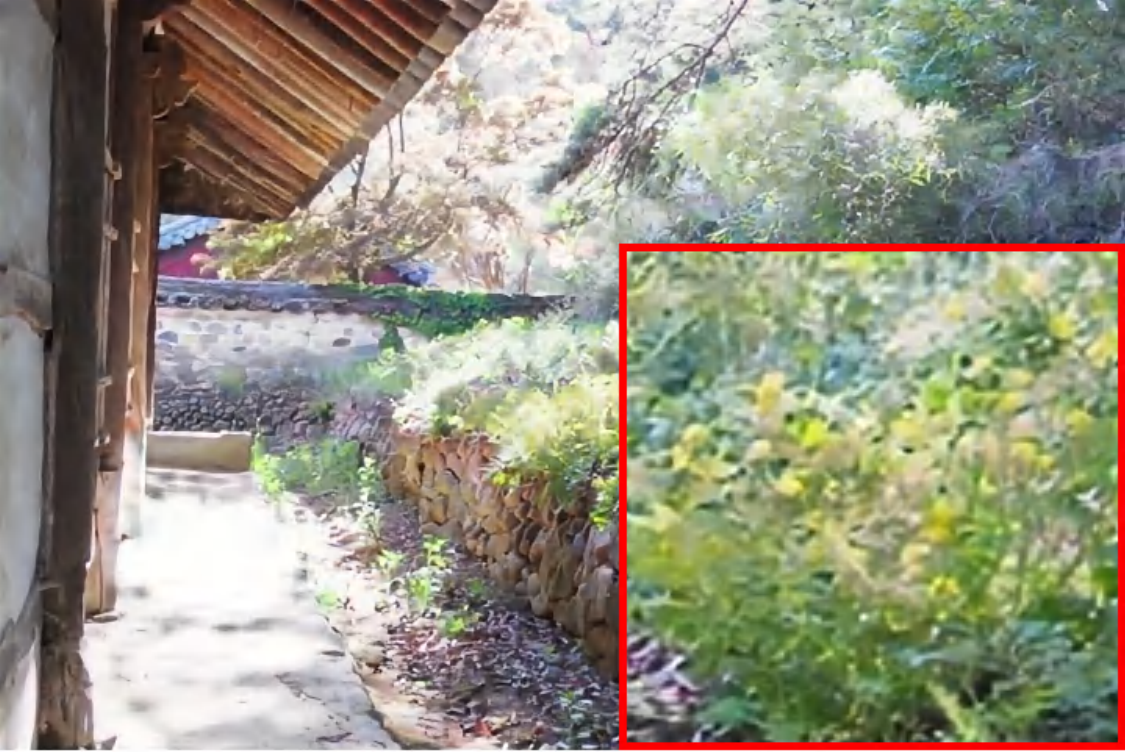}%
    }
    \subfloat[UHDFormer]{%
        \includegraphics[width=0.32\linewidth]{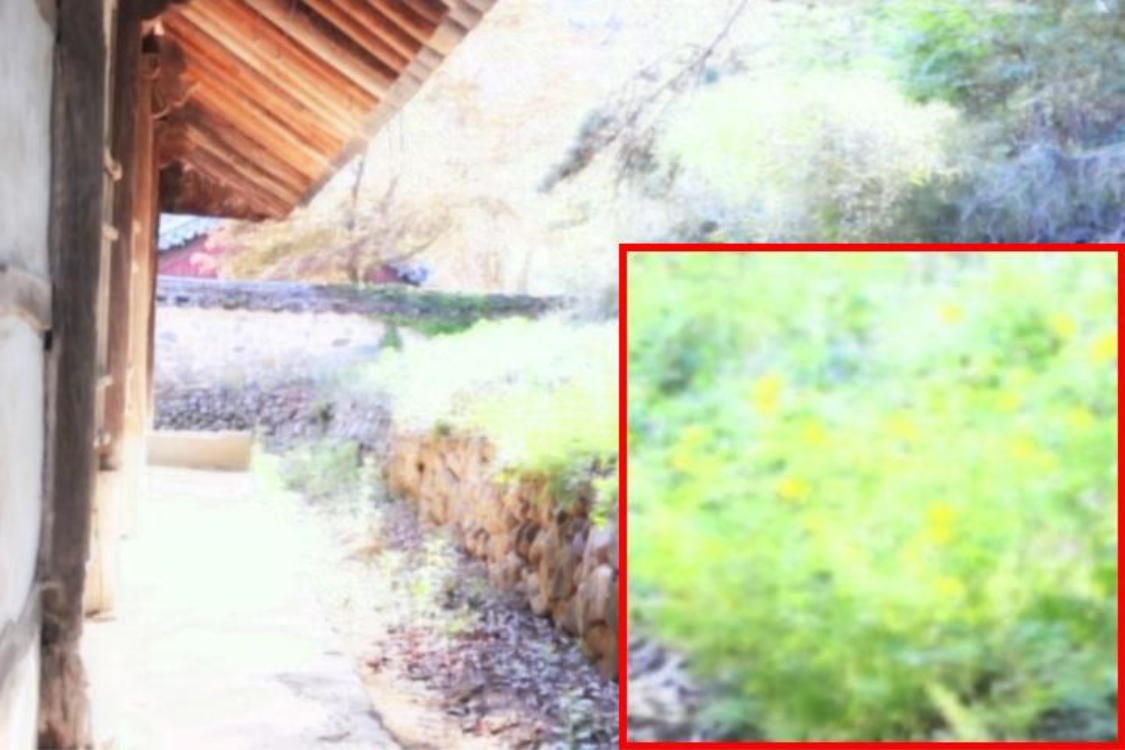}%
    }\hfil
    \subfloat[Ours]{%
        \includegraphics[width=0.32\linewidth]{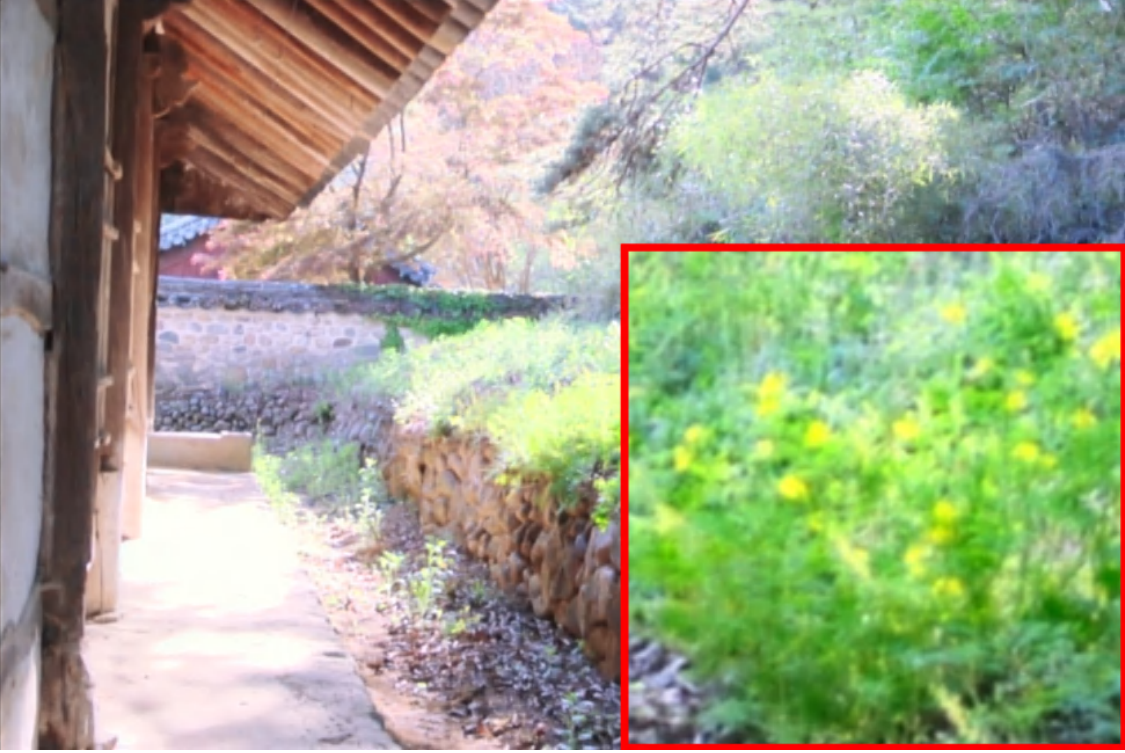}%
    }\hfil
    \vspace{-0.2cm}
    \caption{Visual limitation analysis on real-world scenes.}
    \vspace{-0.4cm}
    \label{fig:limitations_comparison}
\end{figure}
\vspace{-0.1cm}

\subsection{Limitation}
Although our method achieves competitive performance in global exposure control and color consistency and reduces artifacts common in existing methods, certain limitations remain. As illustrated in Fig.~\ref{fig:limitations_comparison}, the generated images exhibit a loss of high-frequency details compared to their natural appearance. This phenomenon primarily stems from our prior fusion guidance strategy. Since the pseudo-infrared prior inherently represents low-frequency illumination structures, guiding it with sparse gradient information can lead to an oversmoothed fused representation, resulting in a slight lack of fine-grained textures in the final output. In future work, we plan to incorporate frequency-domain attention mechanisms to better disentangle and enhance high-frequency texture components.

\section{Conclusion} 
In this work, we revisit frequency-domain learning for LLIE and tackle two key issues in previous methods: oversimplified amplitude--phase decoupling and unreliable priors from degraded inputs. We propose APACF, showing that modeling spectral coupling is crucial for artifact suppression and natural illumination recovery. We further introduce an ERP that leverages the network's intermediate high-quality states instead of external modalities or noisy inputs; together with GGEB and PIGE, it guides restoration with clean, structure-aware features. Extensive results demonstrate highly competitive performance on multiple benchmarks with strong efficiency. Future work will explore extending this endogenous prior mechanism to other degradation tasks, such as image dehazing and deraining, to further verify its generalization capabilities.
\bibliographystyle{IEEEtran}
\bibliography{IEEEabrv,reference}

\end{document}